\documentclass[12pt,oneside, a4paper]{book}
\usepackage{etex}
\usepackage[table,dvipsnames]{xcolor}
\usepackage[pdftex]{graphicx}
\usepackage{rotating}
\usepackage{epsfig}
\usepackage{epstopdf}
\usepackage{mathtools}
\usepackage{listings}
\usepackage{color}
\usepackage{svg}
\usepackage{fancyvrb}

\usepackage[T1]{fontenc}
\usepackage{cmap}
\usepackage[utf8]{inputenc}
\usepackage[english]{babel}
\usepackage{ae}
\usepackage[unicode]{hyperref}
\usepackage{mathptmx}
\usepackage{amscd}
\usepackage{amssymb}
\usepackage{xfrac}
\usepackage{amsmath}
\usepackage{amsfonts}
\usepackage{amsthm}
\usepackage{algorithmicx}
\usepackage{algpseudocode}
\usepackage{wrapfig}

\usepackage{siunitx}
\usepackage[left=2.5cm,right=2.5cm,top=2.5cm,bottom=2.5cm]{geometry}
\usepackage{setspace} 
\usepackage{fancyhdr} %
\usepackage{xspace}

\usepackage{ragged2e}

\usepackage[most]{tcolorbox}
\usepackage{tikz}
\usetikzlibrary{arrows,shapes,positioning,shadows,trees,mindmap}
\usepackage[edges]{forest}
\usetikzlibrary{arrows.meta}
\colorlet{linecol}{black!75}
\usetikzlibrary{backgrounds}
\usetikzlibrary{arrows,shapes}
\usetikzlibrary{tikzmark}
\usetikzlibrary{calc}
\usepackage{xkcdcolors} %

\usepackage{hhline}
\usepackage{caption}
\usepackage{subcaption}
\usepackage{cleveref}
\usepackage{enumerate}
\usepackage{delarray}
\usepackage{array}  %
\usepackage{tabularx} %
\usepackage{multirow}  %
\usepackage{booktabs}
\usepackage{pifont} %
\usepackage{wasysym}
\usepackage{subeqnarray}
\usepackage{pdflscape} %
\usepackage{enumitem} %
\usepackage{xspace}
\usepackage{adjustbox}
\setlist{nolistsep}   %
\usepackage{colortbl}
\usepackage{adjustbox}

\newtcolorbox{exampleblock}[1]{colback=green!10!white,colframe=green!60!black,title=#1}

\renewcommand{\arraystretch}{1.5} %

\usepackage[square, numbers, sort]{natbib} 
\let\savenumberline\numberline
\def\numberline#1{\savenumberline{#1.}}

\makeatletter
\renewcommand*\l@chapter[2]{%
  \ifnum \c@tocdepth >\m@ne
  \addpenalty{-\@highpenalty}%
  \vskip 1.0em \@plus\p@
  \setlength\@tempdima{1.5em}%
  \begingroup
  \parindent \z@ \rightskip \@pnumwidth
  \parfillskip -\@pnumwidth
  \leavevmode \bfseries
  \advance\leftskip\@tempdima
  \hskip -\leftskip
  #1\nobreak\normalfont\leaders\hbox{$\m@th
    \mkern \@dotsep mu\hbox{.}\mkern \@dotsep
    mu$}\hfill\nobreak\hb@xt@\@pnumwidth{\hss #2}\par
  \penalty\@highpenalty
  \endgroup
  \fi}
\makeatother

\makeatletter
\renewcommand*\env@matrix[1][\arraystretch]{%
  \edef\arraystretch{#1}%
  \hskip -\arraycolsep
  \let\@ifnextchar\new@ifnextchar
  \array{*\c@MaxMatrixCols c}}
\makeatother

\AtBeginDocument{\addtocontents{toc}{\protect\thispagestyle{empty}}}
\AtBeginDocument{\addtocontents{lof}{\protect\thispagestyle{empty}}}
\AtBeginDocument{\addtocontents{lot}{\protect\thispagestyle{empty}}}

\graphicspath{{./images/}}

\newcolumntype{L}[1]{>{\raggedright\arraybackslash}p{#1}}

\begin{document}

\clearpage{}%
\frontmatter

\begin{titlepage}
  \fontsize{16pt}{20pt}\selectfont
  \fontfamily{phv}\fontseries{mc}\selectfont
  \newgeometry{left=3cm,right=3cm,top=3cm,bottom=2.5cm}
  \setlength{\intextsep}{0pt plus 0pt minus 0pt}

  \begin{center}
    \begin{figure}[ht!]
      \begin{center}
        \includegraphics[height=4.1184cm, width=5.94cm]{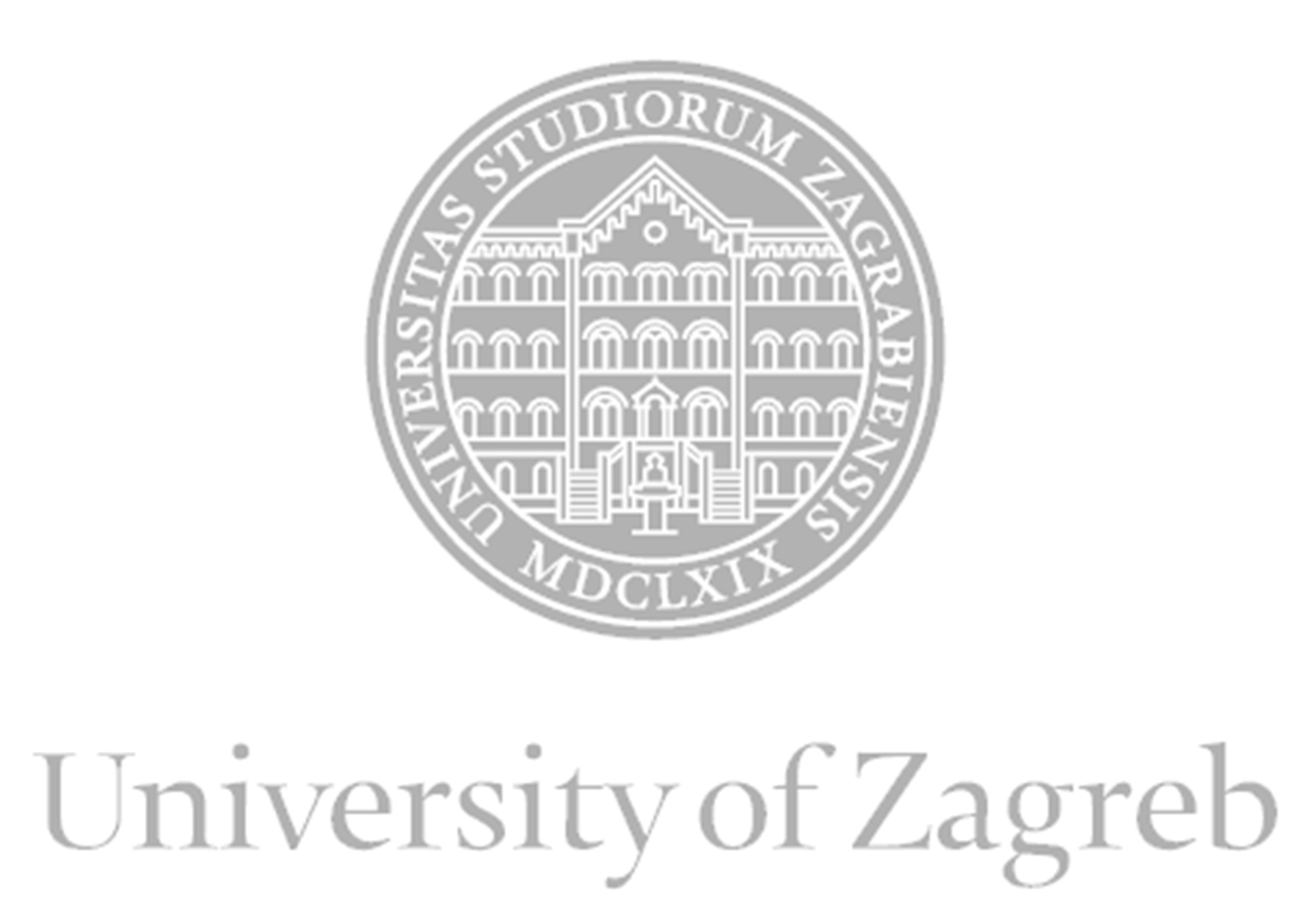}
      \end{center}
    \end{figure}
    \vspace{0cm}
    {FACULTY OF ELECTRICAL ENGINEERING AND COMPUTING} \\
    \vspace{3cm}
    David Dukić \\
    \vspace{2cm}
    {\fontsize{22pt}{22pt}\selectfont
\textbf{
IMPROVING TRANSFER LEARNING FOR SEQUENCE LABELING TASKS BY ADAPTING PRE-TRAINED NEURAL LANGUAGE MODELS}} \\
    \vspace{2cm}  
    DOCTORAL THESIS \\    
    \vfill{Zagreb, 2025}
  \end{center}
  \restoregeometry
\end{titlepage}

\begin{titlepage}
  \fontsize{16pt}{20pt}\selectfont
  \fontfamily{phv}\fontseries{mc}\selectfont
  \newgeometry{left=3cm,right=3cm,top=3cm,bottom=2.5cm}
  \setlength{\intextsep}{0pt plus 0pt minus 0pt}

  \begin{center}
    \begin{figure}[ht!]
      \begin{center}
        \includegraphics[height=4.1184cm, width=5.94cm]{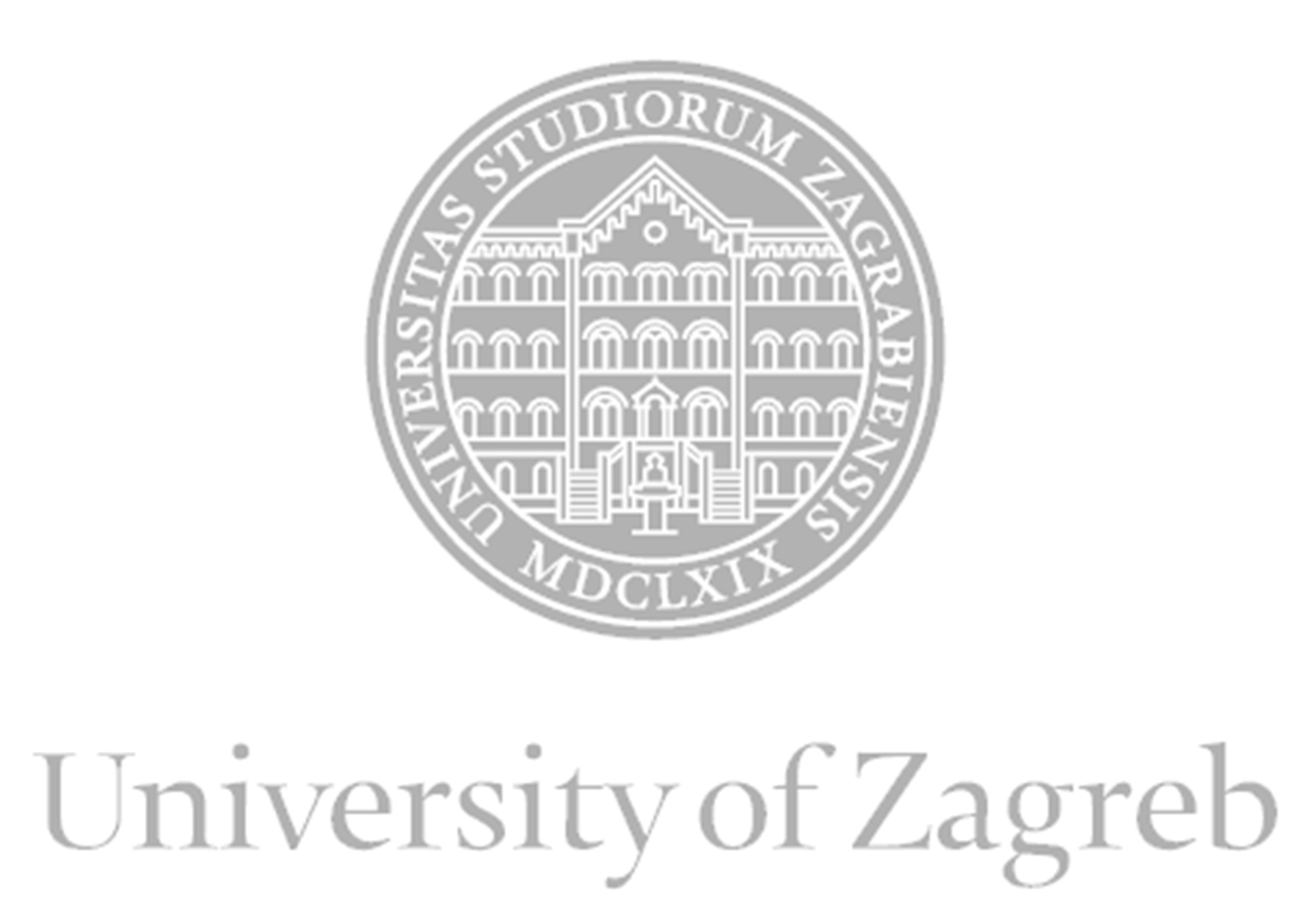}
      \end{center}
    \end{figure}		
    \vspace{0cm}
    {\fontsize{16pt}{16pt}{FACULTY OF ELECTRICAL ENGINEERING AND COMPUTING}} \\
    \vspace{3cm}
    David Dukić \\
    \vspace{2cm}
    {\fontsize{22pt}{22pt}\selectfont\textbf{
IMPROVING TRANSFER LEARNING FOR SEQUENCE LABELING TASKS BY ADAPTING PRE-TRAINED NEURAL LANGUAGE MODELS}} \\
    \vspace{2cm}   
    DOCTORAL THESIS \\  
    \vspace{5cm}   %
    Supervisor: Professor Jan Šnajder, PhD \\
    \vfill{Zagreb, 2025}
  \end{center}
  \restoregeometry
\end{titlepage}

\begin{titlepage}
  \fontsize{16pt}{20pt}\selectfont
  \fontfamily{phv}\fontseries{mc}\selectfont
  \newgeometry{left=3cm,right=3cm,top=3cm,bottom=2.5cm}
  \setlength{\intextsep}{0pt plus 0pt minus 0pt}

  \begin{center}
    \begin{figure}[ht!]
      \begin{center}
        \includegraphics[height=4.1184cm, width=5.94cm]{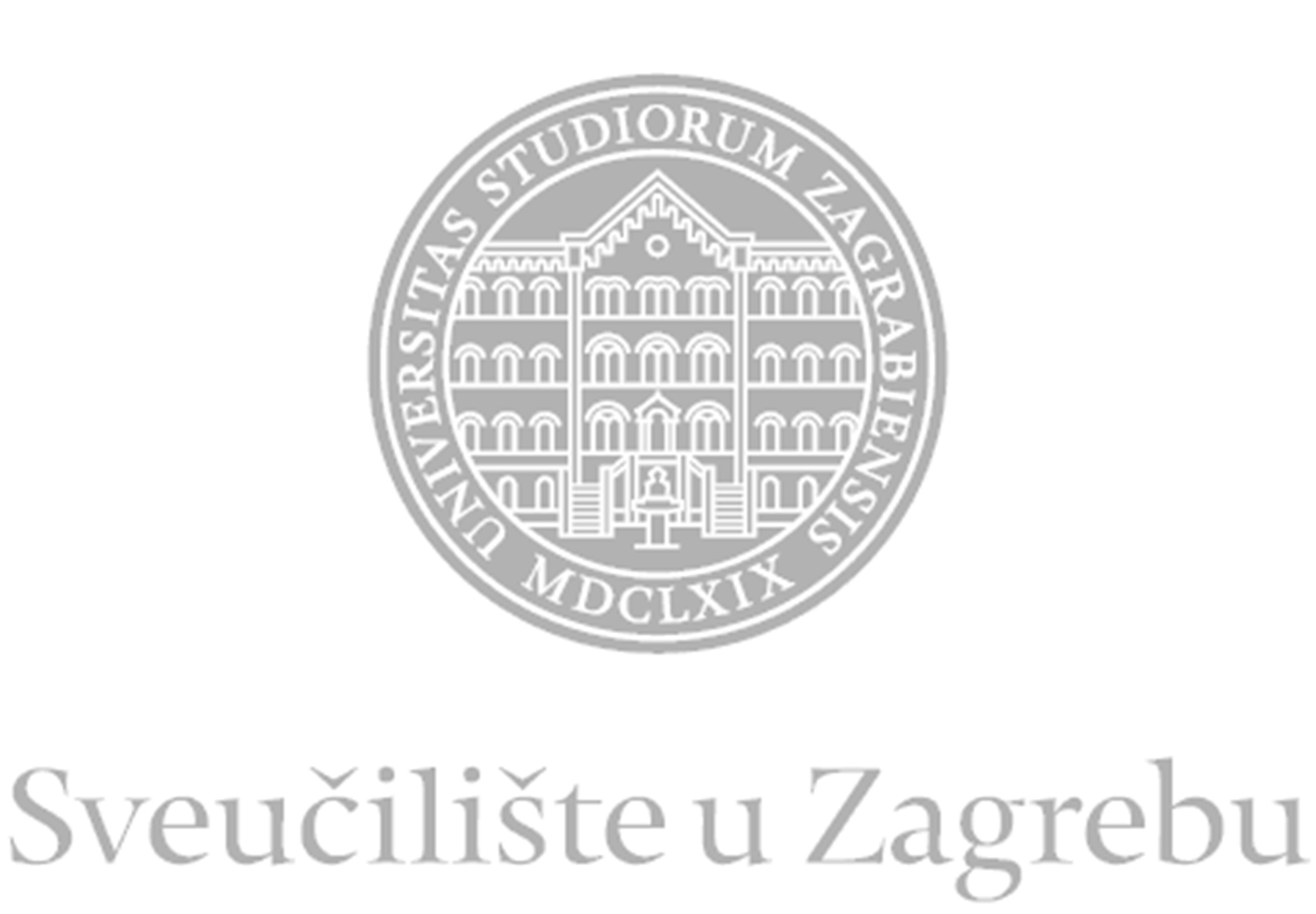}
      \end{center}
    \end{figure}		
    \vspace{0cm}
    {FAKULTET ELEKTROTEHNIKE I RAČUNARSTVA} \\
    \vspace{3cm}
    David Dukić \\
    \vspace{2cm}
    {\fontsize{22pt}{22pt}\selectfont\textbf{
POBOLJŠANJE PRIJENOSNOGA UČENJA ZA ZADATKE OZNAČAVANJA SLIJEDA PRILAGODBOM PREDTRENIRANIH NEURONSKIH JEZIČNIH MODELA
}} \\
    \vspace{2cm}    
    DOKTORSKI RAD \\
    \vspace{5cm}    %
	Mentor: Prof.~dr.~sc.~Jan Šnajder \\
    \vfill{Zagreb, 2025.}
  \end{center}
  \restoregeometry
\end{titlepage}

\begin{titlepage}
  \begin{minipage}{\dimexpr\textwidth-1cm}
    \vspace{3cm}
    The doctoral thesis has been made at the University of Zagreb,
    Faculty of Electrical Engineering and Computing, at the Department
    of Electronics, Microelectronics, Computer and Intelligent systems,
    as part of the Text Analysis and Knowledge Engineering Laboratory (TakeLab).

    \vspace{1cm}
    Supervisor: Professor Jan Šnajder, PhD

    \vspace{1cm}
    The doctoral thesis consists of: 145 pages

    \vspace{1cm}
    Doctoral thesis num.: \rule{3.5cm}{0.4pt}
  \end{minipage}
\end{titlepage}

\clearpage{}%
\clearpage{}%
\thispagestyle{empty}

\section*{About the Supervisor}

Jan Šnajder has received his BSc, MSc, and PhD degrees in Computer Science from the University of Zagreb, Faculty of Electrical Engineering and Computing (FER), Zagreb, Croatia, in 2002, 2006, and 2010, respectively. From September 2002 he was working as a research assistant, from 2011 as Assistant Professor, from 2016 as Associate Professor, and from 2021 as Full Professor at the Department of Electronics, Microelectronics, Computer and Intelligent Systems at FER. He was a visiting researcher at the Institute for Computational Linguistics at the University of Heidelberg, the Institute for Natural Language Processing at the University of Stuttgart, the National Institute of Information and Communications Technology in Kyoto, and the University of Melbourne. He participated in a number of research and industry projects in the field of natural language processing and machine learning. He has (co-) authored more than 100 papers in journals and conferences in natural language processing and information retrieval, and has been reviewing for major journals and conferences in the field. He is the lecturer in charge for six courses at FER and has supervised and co-supervised more than 100 BA and MA theses. He is a member of IEEE, ACM, ACL, the secretary of the Croatian Language Technologies Society, the co-founder and secretary of the Special Interest Group for Slavic NLP of the Association for Computational Linguistics (ACL SIGSLAV). He was a member of the Centre of Research Excellence for Data Science and Advanced Cooperative Systems and the associate editor of the Journal of Computing and Information Technology. He has been awarded the Silver Plaque ``Josip Lončar'' in 2010, the Croatian Science Foundation fellowship in 2012, the fellowship of the Japanese Society for the Promotion of Science in 2014, and the Endeavour Fellowship of the Australian Government in 2015.

\newpage
\section*{O mentoru}

Jan Šnajder diplomirao je, magistrirao i doktorirao u polju računarstva na Sveučilištu u Zagrebu, Fakultetu elektrotehnike i računarstva (FER), 2002., 2006.~odnosno 2010.~godine. Od 2002.~godine radio je kao znanstveni novak, od 2011.~godine kao docent, od 2016.~godine kao izvanredni profesor, a od 2021.~godine kao redoviti profesor na Zavodu za elektroniku, mikroelektroniku, računalne i inteligentne sustave FER-a. Usavršavao se na Institutu za računalnu lingvistiku Sveučilišta u Heidelbergu, Institutu za obradu prirodnog jezika Sveučilišta u Stuttgartu, Nacionalnom institutu za informacijske i komunikacijske tehnologije u Kyotu te Sveučilištu u Melbourneu. Sudjelovao je na nizu znanstvenih i stručnih projekata iz područja obrade prirodnog jezika i strojnog učenja. Autor je ili suautor više od 100 znanstvenih radova u časopisima i zbornicima međunarodnih konferencija u području obrade prirodnog jezika i pretraživanja informacija te je bio recenzent za veći broj časopisa i konferencija iz tog područja. Nositelj je šest predmeta na FER-u te je bio mentor ili sumentor studentima na više od 100 preddiplomskih i diplomskih radova. Član je stručnih udruga IEEE, ACM, ACL, tajnik Hrvatskoga društva za jezične tehnologije te suosnivač i tajnik posebne interesne skupine za obradu prirodnog jezika za slavenske jezike pri udruzi za računalnu lingvistiku (ACL SIGSLAV). Bio je član Znanstvenog centra izvrsnosti za znanost o podacima i kooperativne sustave te pridruženi urednik časopisa Journal of Computing and Information Technology (CIT). Dobitnik je Srebrne plakete ``Josip Lončar'' 2010. godine, stipendije Hrvatske zaklade za znanost 2012.~godine, stipendije Japanskog društva za promicanje znanosti 2014.~godine te stipendije australske vlade Endeavour 2015.~godine.\clearpage{}%
\clearpage{}%
\textit{This diamond was, indeed, made under pressure.}

\vspace{1cm}

First and foremost, I dedicate this thesis to my grandparents. Your unwavering support inspired me to embark on and survive this PhD journey. I am forever grateful for your boundless love, care, and patience. \textit{Hvala dida i nana!}

\vspace{1cm}

To my parents, thank you for your constant support, for teaching me what are the important values in life, and for giving me the strength and resilience to face whatever life brings.

\vspace{1cm}

To my sister, thank you for always being by my side and giving me the motivation to push through.

\vspace{1cm}

To my girlfriend, thank you for your endless patience, love, and understanding. You have been my anchor and my inspiration, giving me the strength to overcome every obstacle along the way.

\vspace{1cm}

To the team at Srce and the supercomputer Supek, I am deeply grateful for your prompt and reliable support, especially when my research required significant resources. Your assistance was invaluable to my work.

\vspace{1cm}

To my colleagues in the lab, thank you for your companionship along the way, unforgettable conferences together, late nights in the office, and after-work drinks.

\vspace{1cm}

Finally, to my supervisor, thank you for helping me realize my potential. Your patience, insightful advice—both academic and personal—and unwavering guidance have been pivotal in shaping this chapter of my life. I am truly grateful to have had you as my mentor.
\clearpage{}%
\clearpage{}%
\thispagestyle{empty}

\section*{Abstract}

This doctoral thesis improves the transfer learning for sequence labeling tasks by adapting pre-trained neural language models. The proposed improvements in transfer learning involve introducing a multi-task model that incorporates an additional signal, a method based on architectural modifications in autoregressive large language models, and a sequence labeling framework for autoregressive large language models utilizing supervised in-context fine-tuning combined with response-oriented adaptation strategies. Annotating domain-specific target data is expensive, which often results in domain-specific data unavailability, diminishing the effectiveness of transfer learning across various domains. The first improvement is given in the context of domain transfer for the event trigger detection task, a crucial task in the event extraction pipeline. The domain transfer of event trigger detection task in low(er)-resource transfer scenarios is suboptimal and it can be improved by incorporating an additional domain-agnostic signal into the pre-trained neural language model. The signal is incorporated using a multi-task model architecture, which leverages the additional signal from a domain-independent text processing system. The proposed model improves transfer performance and achieves the best results when combined with domain adaptation techniques using the pre-training mechanism of the underlying pre-trained neural language model. The second improvement reduces the discrepancy between pre-training and target sequence labeling tasks by modifying the model architecture. For that purpose, a method is proposed to enable bidirectional information flow across layers of autoregressive large language models. The method enables large language models to adapt directly to sequence labeling tasks, effectively utilizing the full context. This is reflected in improved model performance on various sequence labeling tasks, surpassing strong baselines. %
The third improvement puts emphasis on utilizing autoregressive large language models as text generators. With the increased scale of pre-training, large language models demonstrated surprising in-context learning abilities. To optimally leverage these abilities, a framework for fine-tuning large language models in a generative manner is devised, showing that their adaptation to sequence labeling tasks can be improved via response-oriented adaptation strategies, utilizing demonstrations in the context through supervised in-context fine-tuning. More specifically, large language model in-context learning abilities are improved by directly fine-tuning on in-context learning data, while steering the model to adapt to responses in the context using modifications of the loss function. %
The proposed model, method, and framework demonstrate that pre-trained neural language models achieve their best performance on sequence labeling tasks when adapted through targeted transfer learning paradigms.

\vspace{1cm}
\textbf{Keywords}: natural language processing, pre-trained neural language models, sequence labeling, transfer learning, multi-task model, supervised in-context fine-tuning\clearpage{}%
\clearpage{}%
\chapter*{Prošireni sažetak}

\section*{Poboljšanje prijenosnoga učenja za zadatke označavanja slijeda prilagodbom predtreniranih neuronskih jezičnih modela}

Označavanje slijeda (engl.~\textit{sequence labeling}) jedan je od ključnih nadziranih zadataka u području obrade prirodnoga jezika (OPJ). Označavanje slijeda pripada skupini zadataka razumijevanja prirodnoga jezika. Ti zadaci zahtijevaju duboko lingvističko razumijevanje konteksta i semantičkoga značenja riječi i rečenica. Za razliku od većine zadataka razumijevanja prirodnoga jezika koji se fokusiraju na klasifikaciju čitavog teksta (na primjer rečenice ili dokumenta), označavanje slijeda dodjeljuje oznake pojedinačnim riječima ili podriječima, što taj zadatak čini složenijim, ali i više primjenjivim jer rezultati zadataka označavanja slijeda služe kao temelj za mnoge primjene kao što su izgradnja baza znanja te izgradnja sustava za praćenje vijesti. Dodatno, sustavi temeljeni na označavanju slijeda nezamjenjiv su dio složenijih sustava za ekstrakciju informacija.

Kao i većina zadataka OPJ-a danas, zadaci označavanja slijeda rješavaju se korištenjem modela strojnog i dubokog učenja. U početku su se zadaci OPJ-a rješavali tradicionalnim modelima strojnoga učenja koji su se oslanjali na ručno izrađene značajke. Taj tradicionalni pristup dominirao je područjem sve do ranih 2000-tih. Ubrzo, tijekom 2010-tih, područje se počelo okretati prema neuronskim pristupima koji koriste tehnike dubokoga učenja. U posljednjem desetljeću, duboko učenje steklo je veliku popularnost zahvaljujući svojoj sposobnosti da automatski izvlači značajke iz podataka pomoću dubokih neuronskih mreža. Prijelaz s tradicionalnoga strojnog učenja na duboko učenje bio je prekretnica jer su duboki neuronski modeli pokazali nadmoć u odnosu na tradicionalne pristupe na raznim zadacima. Konačno, najznačajnija transformacija OPJ-a dogodila se 2018.~godine pojavom predtreniranih neuronskih jezičnih modela (engl.~\textit{pre-trained neural language models}) (PJM) temeljenih na arhitekturi transformatora (engl.~\textit{transformer}). 

Učinkoviti mehanizmi samonadziranog predtreniranja, poput kauzalnog jezičnog modeliranja i maskiranog jezičnog modeliranja, omogućili su rješavanje širokog spektra zadataka razumijevanja prirodnoga jezika na automatiziran način kroz nadziranu prilagodbu označenim podacima iz ciljnih domena, zadataka ili skupova podataka. Predtreniranje modela s milijunima (u novije vrijeme i milijardama) parametara na ogromnim količinama tekstnih podataka stvorilo je snažne temeljne PJM-ove. Iskorištavanjem opsežnog predtreniranja, praćenog nadziranom prilagodbom (engl.~\textit{adaptation}) na označene podatke, temeljni PJM-ovi postali su standard za postizanje vrhunskih rezultata u širokom rasponu zadataka OPJ-a. Ovaj dvostupanjski postupak predtreniranja i prilagodbe naziva se prijenosno učenje (engl.~\textit{transfer learning}) i danas je prevladavajući način razvoja modela u suvremenom OPJ-u.

U modernom OPJ-u, izgradnja robusnih i učinkovitih modela za zadatke razumijevanja pri\-ro\-dno\-ga jezika podrazumijeva nadogradnje nad temeljnim PJM-ovima. Stoga se pojavio niz metoda koje nastoje poboljšati performance zadataka OPJ-a kroz fazu prilagodbe prijenosnoga učenja. To uključuje metode poput višezadaćnog učenja, odabira podskupa parametara predtreniranog modela za prilagodbu te uključivanje dodatnih parametara uz one koji već postoje u predtreniranom modelu. Iako je u tom pogledu postignut značajan napredak za mnoge klasifikacijske zadatke razumijevanja prirodnoga jezika, zadaci označavanja slijeda dobili su usporedno manje pažnje. Iako su PJM-ovi pokazali sposobnost i učinkovitost kodiranja semantike teksta u kontekstualizirane reprezentacije, faza prilagodbe za zadatke označavanja slijeda često ne uspijeva u potpunosti iskoristiti takve reprezentacije, ostavljajući prostor za moguća poboljšanja.

U skladu s time, u okviru ove doktorske disertacije predložena su poboljšanja u postupku prijenosnoga učenja za zadatke označavanja slijeda kroz intervencije u fazi prilagodbe predtreniranih neuronskih jezičnih modela. Poboljšanja su usmjerena na moderne arhitekture, uključujući koderske PJM-ove i autoregresivne dekoderske velike jezične modele (engl.~\textit{large language models}), koji su postali standard za primjenu prijenosnoga učenja u suvremenom OPJ-u. Istražujući zadatke označavanja slijeda, arhitektonska ograničenja koja ometaju postizanje visokih performanci te modificirajući postojeće sheme optimizacije, ova disertacija predlaže nekoliko novih pristupa za poboljšanje prijenosnoga učenja za zadatke označavanja slijeda kroz fazu prilagodbe.

Detaljnije, predložen je višezadaćni model za smanjenje negativnoga domenskog prijenosa zabilježenog kod koderskih PJM-ova, metoda za omogućavanje dvosmjernog protoka informacija po slojevima dekodera te radni okvir s tehnikama nadziranog finog ugađanja u kontekstu prilagođen zadacima označavanja slijeda. Kroz opsežne eksperimente potvrđuje se učinkovitost predloženih poboljšanja i pruža dubinsko razumijevanje ostvarenih rezultata putem analize njihovih snaga i ograničenja.

\section*{1. Uvod}

U prvome poglavlju (\textit{``Introduction''}) predstavljena je motivacija za rješavanje zadataka označavanja slijeda korištenjem PJM-ova kroz intervencije tijekom faze prilagodbe prijenosnoga učenja. Također, predstavljena su znanstvena pitanja oko kojih je strukturirana doktorska disertacija. Uz znanstvena pitanja, naveden je i znanstveni doprinos doktorske disertacije te je detaljno opisana struktura disertacije.

\section*{2. Pozadina}

U drugome poglavlju (\textit{``Background''}) razlažu se povijesne i teorijske osnove za razumijevanje predloženih poboljšanja nad postojećim metodama prilagodbe neuronskih jezičnih modela. Teorijska pozadina neuronske obrade prirodnog jezika ključna je za razumijevanje PJM-ova. Počevši od distribucijske hipoteze, preko statičnih ugrađivanja (engl.~\textit{embeddings}) do metoda za obradu slijeda, početak drugog poglavlja daje temeljit pregled razvoja neuronskih modela. Opisani su neuronski modeli zasnovani na arhitekturi transformatora, koderske i de\-ko\-der\-ske varijante transformatora te metode predtreniranja istih. Dobar dio drugog poglavlja bavi se prijenosnim učenjem i njegovim paradigmama. Naglasak je stavljen na temeljne pristupe prilagodbe PJM-ova kroz: (1) uvođenje dodatnog signala svojstvenog ciljnome zadatku kroz višezadaćno učenje, slabu nadziranost i/ili ansamble, (2) arhitektonske modifikacije te (3) različite optimizacijske sheme. Na optimizacijske sheme naslanja se i dio o pristupima parametarski učinkovitoga finog ugađanja (engl.~\textit{parameter-efficient fine-tuning}). Metode parametarski učinkovitoga finog ugađanja poput adaptacije niskog ranga i kvantizirane adaptacije niskog ranga omogućuju korištenje i ugađanje malog broja parametara na učinkovit način, rezultirajući kompaktnim i proširivim modelima. S obzirom da ove metode omogućuju eksperimentiranje s velikim jezičnim modelima i korištene su u gotovo svim eksperimentima doktorske disertacije, one su u ovom poglavlju detaljno opisane. Drugo poglavlje završava formalnim uvođenjem zadatka označavanja slijeda. Istaknuta je važnost zadataka označavanja slijeda, tipični načini modeliranja zadatka unutar OPJ-a te je dana opaska o evaluaciji zadataka označavanja slijeda.

\section*{3. Zadaci, skupovi podataka i evaluacija}

U trećemu poglavlju (\textit{``Tasks, Datasets, and Evaluation''}) detaljno su opisani prototipični zadaci označavanja slijeda te skupovi podataka korišteni u eksperimentima. Opisani su zadaci i pod\-za\-da\-ci ekstrakcije događaja, zadatak ekstrakcije relacija, zadatak prepoznavanja imenovanih entiteta, zadatak cjepkanja teksta, podzadaci analize sentimenta temeljene na aspektima, zadatak rudarenja argumenata temeljenih na aspektima, zadatak označavanja slotova te zadatak označavanja semantičkih uloga. Svaki od zadataka opisan je zajedno s primjerima skupova podataka. Predstavljene su i konkretne rečenice te njihove oznake iz svakog skupa podataka korištenog u eksperimentima uz navođenje statistika za skupove podataka. U istome poglavlju detaljno je opisana i evaluacija modela za označavanje slijeda. U radu je korištena stroga evaluacija temeljena na sljedovima. Predstavljene su metrike za evaluaciju zadataka označavanja slijeda te je dan primjer evaluacije. Poglavlje završava opisom evaluacije za dekoderske modele, gdje je opisano na koji se način generirani sljedovi preslikavaju na format koji je standardiziran za evaluaciju zadataka označavanja slijeda.

\section*{4. Vanjski signal kroz višezadaćni model}

U četvrtome poglavlju (\textit{``External Signal Through Multi-task Model''}) predstavljen je nov više\-za\-da\-ćni model za smanjenje negativnoga domenskog prijenosa kod zadatka otkrivanja okidača događaja (engl.~\textit{event trigger detection}) u tekstu na temelju predikcija sustava za otvorenu ekstrakciju informacija (engl.~\textit{open information extraction}). Osim modela, predstavljena je motivacija, eksperimenti te rezultati eksperimenata. Označavanje domenski specifičnih ciljnih podataka skup je postupak, što često rezultira nedostatnom količinom podataka za učenje u ciljnoj domeni. Takav scenarij umanjuje učinkovitost prijenosnoga učenja između domena s malo označenih ciljnih podataka. Zadatak otkrivanja okidača događaja ključna je komponenta u cjevovodu ekstrakcije događaja. Domenski prijenos zadatka otkrivanja okidača događaja u scenarijima s ograničenim resursima pokazuje suboptimalne rezultate koje je moguće poboljšati ugradnjom dodatnog domenski neovisnoga signala u predtrenirani neuronski jezični model. Signal se integrira kroz višezadaćnu arhitekturu modela koja iskorištava dodatne informacije iz domenski neovisnoga sustava za obradu teksta---sustava za otvorenu ekstrakciju relacija temeljenog na pravilima. Ideja se zasniva na sličnosti relacija i okidača kao predikatnih struktura i pretpostavci da se relacije mogu iskoristiti za poboljšanje prijenosa u scenarijima kada nemamo označene okidače događaja u ciljnoj domeni ili ih imamo označeno vrlo malo. Predloženi višezadaćni model poboljšava performance prijenosa te postiže najbolje rezultate u kombinaciji s tehnikama domenske prilagodbe korištenjem mehanizama kontinuiranoga predtreniranja temeljnog PJM-a.

\section*{5. Uklanjanje kauzalne maske po slojevima}

U petome poglavlju (\textit{``Layer-wise Causal Mask Removal''}) predstavljena je metoda u svrhu poboljšanja performanci autoregresivnih velikih jezičnih modela na zadacima označavanja slijeda omogućavanjem dvosmjernoga protoka informacija po slojevima dekodera. Osim metode, predstavljena je motivacija, eksperimenti te rezultati eksperimenata. Veliki jezični modeli su uobičajeno predtrenirani korištenjem kauzalnoga jezičnog modeliranja koje uči dekoderski model da generira koherentni tekst i ne dopušta korištenje konteksta slijeda desno od trenutačno obrađivanog tokena. Takva restrikcija ostvarena je pomoću trokutaste kauzalne maske (engl.~\textit{causal mask}), koja se koristi u kombinaciji s matricom težina pozornosti. Predtrenirani dekoderski modeli su zbog toga lošiji u rješavanju zadataka označavanja slijeda od koderskih modela koji za vrijeme predtreniranja imaju pristup kako lijevom tako i desnom kontekstu. Predložena metoda koja omogućuje dvosmjeran protok informacija po slojevima omogućuje velikim jezičnim modelima izravnu prilagodbu na zadatke označavanja slijeda, učinkovito iskorištavajući cjelokupni kontekst i pretvarajući dekoderske modele u snažne kodere. Eksperimenti pokazuju da tako preinačeni modeli ostvaruju poboljšane performance na različitim zadacima označavanja slijeda, nadmašujući snažne osnovne modele poput predtreniranih koderskih modela i dekoderskih modela fino ugođenih pomoću instrukcija (engl.~\textit{instruction-tuned decoder models}). Nadalje, pokazano je i da se poboljšanja ostvarena uklanjanjem kauzalne maske ne mogu postići na nižoj skali predtreniranja i broja parametara.

\section*{6. Nadzirano fino ugađanje u kontekstu}

U šestome poglavlju (\textit{``Supervised In-context Fine-tuning''}) predstavljen je radni okvir za označavanje slijeda za autoregresivne velike jezične modele temeljen na nadziranome finom uga\-đa\-nju u kontekstu (engl.~\textit{supervised in-context fine-tuning}) u kombinaciji sa strategijama prilagodbe orijentiranima na odgovor. S povećanjem skale predtreniranja, veliki jezični modeli su pokazali iznenađujuće sposobnosti učenja u kontekstu (engl.~\textit{in-context learning}), odnosno zaključivanja iz demonstracija predstavljenih u upitu (engl.~\textit{prompt}). Kako bi se te sposobnosti optimalno iskoristile za zadatke označavanja slijeda, predstavljen je radni okvir koji pokazuje da se prilagodba velikih jezičnih modela na zadatke označavanja slijeda može poboljšati kroz strategije prilagodbe orijentirane na odgovor, iskorištavajući demonstracije u upitu putem nadziranoga finog ugađanja u kontekstu. Točnije, sposobnosti velikih jezičnih modela za učenje u kontekstu poboljšavaju se izravnim finim ugađanjem na podacima za učenje u kontekstu, uz orijentiranje modela k prilagodbi na odgovore u kontekstu kroz modifikacije funkcije gubitka za kauzalno jezično modeliranje. Demonstrirana poboljšanja vrijede za različite zadatke označavanja slijeda. Modeli orijentirani na više odgovora u kontekstu postižu bolje rezultate učenja u kontekstu od onih koji su orijentirani samo na posljednji odgovor ili tretiraju sve dijelove upita za treniranje na jednak način. Također je analiziran i učinak treniranja bez instrukcije u kontekstu te kasnije uključenje varijacija instrukcije prilikom učenja u kontekstu. Rezultati su pokazali da instrukcija nije ključna za vrijeme treniranja za ostvarenje dobrih rezultata na različitim zadacima označavanja slijeda.

\section*{7. Rasprava}

U sedmome poglavlju (\textit{``Discussion''}) proširuju se zaključci opisani u poglavljima koja iznose predložena poboljšanja. Sažeto se iznose pronalasci te se povezuju s postavljenim znanstvenim pitanjima. Također, iznose se implikacije te ograničenja predloženih poboljšanja u postupku prijenosnog učenja za zadatke označavanja slijeda kroz intervencije u fazi prilagodbe predtreniranih neuronskih jezičnih modela.

\section*{8. Zaključak}

U osmome poglavlju (\textit{``Zaključak''}) sažimaju se poboljšanja predložena u okviru doktorske disertacije te se predlažu smjernice za buduća istraživanja u području prilagodbe PJM-ova za zadatke označavanja slijeda. Zaključeno je da su predloženi model, metoda i radni okvir razvijeni na čvrstim temeljima tehnika prilagodbe PJM-ova. Unaprjeđenje prijenosnoga učenja za zadatke označavanja slijeda uz moderne PJM-ove zahtijeva inovativne i učinkovite pristupe prilagođene modelu, zadatku i dostupnim računalnim resursima. Budući da je predtreniranje velikih jezičnih modela još uvijek nedostupno većini istraživača, očekuje se da će napredak u fazi prilagodbe prijenosnoga učenja u doglednoj budućnosti i dalje biti u središtu interesa istraživača u području OPJ-a.

\vspace{1cm}
\textbf{Ključne riječi}: obrada prirodnog jezika, predtrenirani neuronski jezični modeli, označavanje slijeda, prijenosno učenje, višezadaćni model, nadzirano fino ugađanje u kontekstu\clearpage{}%
\clearpage
\pagestyle{empty} %
\tableofcontents
\cleardoublepage %

\chapter*{List of Abbreviations}

\begin{description}
\item[AAC] Argument aspect corpus
\item[ABAM] Aspect-based argument mining
\item[ACE] Automatic content extraction
\item[ATE+ATP] Aspect term extraction and polarity
\item[BiLSTM] Bidirectional long short-term memory
\item[BIO] Begin, inside, outside
\item[CDEE] Closed-domain event extraction
\item[CLM] Causal language modeling
\item[CM] Causal mask
\item[CRF] Conditional random field
\item[DAPT] Domain-adaptive pre-training
\item[DPO] Direct preference optimization
\item[EAC] Event argument classification
\item[EAD] Event argument detection
\item[EE] Event extraction
\item[ETC] Event trigger classification
\item[ETD] Event trigger detection
\item[GPT] Generative pre-trained transformer
\item[GRU] Gated recurrent unit
\item[HMM] Hidden Markov model
\item[ICL] In-context learning
\item[IE] Information extraction
\item[IOB] Inside, outside, begin
\item[IT] Instruction tuning
\item[LLM] Large language model
\item[LoRA] Low-rank adaptation
\item[LSTM] Long short-term memory
\item[MEMM] Maximum entropy Markov model
\item[MLM] Masked language modeling
\item[MRC] Multi-response completion
\item[NER] Named entity recognition
\item[NLP] Natural language processing
\item[NLU] Natural language understanding
\item[OIE] Open Information extraction
\item[PCA] Principal component analysis
\item[PEFT] Parameter-efficient fine-tuning
\item[PLM] Pre-trained language model
\item[PoS] Part of speech
\item[QLoRA] Quantized low-rank adaptation
\item[QRC] Query response completion
\item[RC] Response completion
\item[RDRR] Right-side dependency relations ratio
\item[RE] Relation extraction
\item[RLHF] Reinforcement learning with human feedback
\item[RNN] Recurrent neural network
\item[SC] Span classification
\item[SD] Span detection
\item[SFT] Supervised fine-tuning
\item[SIFT] Supervised in-context fine-tuning
\item[SOTA] State of the art
\item[SRC] Single-response completion
\item[SRL] Semantic role labeling
\item[SVM] Support vector machine
\item[TAPT] Task-adaptive pre-training
\end{description}

\pagestyle{fancyplain} %

\mainmatter
\clearpage{}%
\chapter{Introduction}
\label{ch:intro}

Sequence labeling is one of the essential supervised tasks in the field of natural language processing (NLP). It belongs to a class of natural language understanding (NLU) tasks in NLP. NLU tasks demand a profound linguistic understanding of natural language, carefully considering the context and nuanced semantic meaning of words and sentences. Unlike most NLU tasks that focus on classifying entire text sequences or documents, sequence labeling stands out as a more fine-grained task, assigning labels to individual words or subwords within a text. This granularity makes sequence labeling more complex but, at the same time, more applicable and versatile, as its outputs serve as a basis for numerous downstream applications. For instance, sequence labeling systems are irreplaceable in the information extraction (IE) pipeline since many tasks in the pipeline can be framed as sequence labeling. Within the IE pipeline, sequence labeling tasks operate on unstructured text data to discover and systematize structured semantic representations of sequences of words or subwords.

Prototypical sequence labeling tasks include named entity recognition (NER), part-of-speech (PoS) tagging, text chunking, and semantic role labeling (SRL) \citep{tjong-kim-sang-de-meulder-2003-introduction, tjong-kim-sang-buchholz-2000-introduction, 10.1162/089120102760275983}. These sequence labeling tasks are useful on their own. Moreover, they are even more valuable when applied across various disciplines. For example, sequence labeling models are pivotal in constructing knowledge bases \citep{hoffart-etal-2011-robust}, powering news and social media monitoring systems \citep{dukic2024closed}, and facilitating research in social sciences by automating large-scale text analysis. Their capacity to extract and quantify structured information from unstructured text has made them critical for advancing interdisciplinary collaboration, lowering the gap between computer science and social sciences \citep{dukic2024takelab}. 

Developing robust and efficient models for sequence labeling is essential for building accurate and high-performing downstream applications, which help us make sense of the immense volume of data generated in the digital age. For example, news event monitoring systems enable us to differentiate important and urgent news from insignificant and deferrable news \citep{dukic2024closed}. These downstream applications rely on the outputs of sequence labeling models for event extraction. Thus, improving models for sequence labeling not only improves the accuracy of the specific task but also enhances the performance of downstream applications.

Like most NLP tasks today, sequence labeling tasks are carried out using machine learning and deep learning models. Machine learning is a subfield of artificial intelligence that involves developing statistical models that learn from data using features and generalize to new, unseen data instances.
The way NLP tasks are solved has changed over the years. Initially, NLP tasks were addressed using traditional machine learning models that relied on hand-crafted features, requiring manual effort. This traditional approach dominated the field until the early 2000s. Soon, in the 2010s, the field began to sway toward neural approaches, leveraging deep learning techniques. In the last decade, deep learning gained much traction due to its ability to automatically extract features from the data with deep neural networks, adapting to the underlying data distribution in an end-to-end manner. The transition from traditional machine learning to deep learning was a turning point, as deep neural models demonstrated superior performance over traditional approaches across various tasks. Finally, NLP has witnessed its most significant transformation in 2018 with the emergence of pre-trained language models (PLMs) based on the Transformer architecture \citep{vaswani2017attention}. Effective self-supervised pre-training mechanisms, such as causal language modeling \citep{radford2018improving} and masked language modeling \citep{devlin-etal-2019-bert}, have enabled solving a wide range of NLU tasks in an automated manner through supervised adaptation to labeled data from target domains, tasks, or datasets. Pre-training models with millions (more recently billions) of parameters on huge amounts of text data created strong foundation PLMs. By leveraging large-scale pre-training followed by supervised adaptation to task-specific labeled data, foundation PLMs have become the standard for reaching state-of-the-art (SOTA) performance across a wide range of NLP tasks. This two-step process of pre-training and adaptation is referred to as transfer learning and has established itself as a prevailing model development process in modern NLP.

In today's modern NLP, building robust and efficient models for NLU tasks implies starting from a foundation PLM. Consequently, a range of methods emerged that strive to improve the performance of NLP tasks through the adaptation phase of transfer learning. These include methods such as multi-task learning \citep{crawshaw2020multi}, selecting a subset of pre-trained model parameters for adaptation, and including additional parameters alongside pre-trained ones \citep{houlsby2019parameter}. While significant progress has been made in this regard for many NLU tasks, mostly sequence classification ones, sequence labeling tasks have received comparatively less attention. Although PLMs have demonstrated the ability to encode text semantics into contextualized representations, the adaptation phase for sequence labeling tasks often fails to fully exploit these representations, leaving room for potential improvements.

To address the aforementioned gap, this thesis proposes improvements in the transfer learning process for sequence labeling tasks through interventions in the adaptation phase of pre-trained neural language models. Our work targets modern architectures, including encoder-only PLMs and autoregressive decoder-only large language models (LLMs), which have become the de facto standard for applying transfer learning in modern NLP. By exploring the nuances of sequence labeling tasks, investigating the architectural limitations that hinder optimal performance, and modifying existing optimization schemes, we introduce several novel approaches to improve transfer learning for sequence labeling tasks through the adaptation phase.

More specifically, we propose a model to mitigate challenges such as negative domain transfer registered with encoder-only PLMs, a method to bypass limitations in information flow within autoregressive PLMs, and a framework with novel in-context fine-tuning techniques tailored for sequence labeling tasks. Through extensive experimentation, we validate the effectiveness of the proposed improvements and provide a deep understanding of the obtained findings via an analysis of their strengths and limitations.

This thesis is structured around the following research questions:

\begin{enumerate}
    \item[\textbf{RQ1}:] How to improve the transfer with little (few-shot) to none (zero-shot) domain-specific data for sequence labeling?
    \item[\textbf{RQ2}:] How to narrow the pre-train--fine-tune discrepancy effectively and efficiently during the adaptation phase of decoder-only LLMs for improved downstream sequence labeling performance?
    \item[\textbf{RQ3}:] How to utilize decoder-only LLMs for improved sequence labeling performance in a parameter-efficient manner under a unified adaptation framework?
\end{enumerate}

Addressing these research questions has the potential to advance transfer learning for sequence labeling tasks significantly. By tackling RQ1, the development of more robust few-shot and zero-shot domain transfer methods could result in high-performing sequence labeling models in specialized domains, where annotated data is expensive to obtain. Addressing RQ2 would reduce the pre-train--fine-tune discrepancy within the scope of sequence labeling for decoder-only LLMs, which in turn could unlock their underlying potential for sequence labeling tasks and bridge the performance gap with encoder-based models. The existence of a parameter-efficient framework proposed in RQ3 could enable the natural utilization of LLMs for sequence labeling, preserving their generative capabilities and making their deployment practical in compute-constrained scenarios. Collectively, these advancements could improve transfer learning for sequence labeling tasks with PLMs, leveraging both their discriminative and generative capabilities, and consequently reducing reliance on task-specific architectures.

\section{Contributions}

The doctoral thesis topic is the improvement of transfer learning in natural language processing for sequence labeling tasks through the development of new approaches for adapting pre-trained neural language models. The proposed research improves the transfer learning of encoder and autoregressive decoder models for sequence labeling tasks through interventions in the adaptation phase of pre-trained neural language models. The original scientific contribution of the doctoral thesis consists of the following:

\begin{enumerate}
    \item  A multi-task model for mitigating negative domain transfer of sequence labeling tasks based on information produced by a domain-independent text processing system;
    \item  A method for improving the performance of autoregressive large language models on sequence labeling tasks by enabling bidirectional information flow across decoder layers;
    \item A sequence labeling framework for autoregressive large language models based on supervised in-context fine-tuning combined with response-oriented adaptation strategies.
\end{enumerate}

\section{Thesis Structure}

This thesis is organized into eight chapters. %
Chapter \ref{ch:background} goes into the background and lays the historical and theoretical foundations for understanding the proposed improvements over existing adaptation methods. Further, Chapter \ref{ch:tde} details the prototypical sequence labeling tasks and datasets, giving an overview of the ones used for experiments in this thesis. Chapter \ref{ch:tde} ends with an exhaustive description of the evaluation setup for sequence labeling tasks. Next, we introduce three chapters delivering the overall contribution of this doctoral thesis. Chapter \ref{ch:method1} introduces novel multi-task models, experimental setup, experiments, and results for mitigating negative domain transfer in zero- and few-shot transfer scenarios. Next, Chapter \ref{ch:method2} presents a method operating across decoder layers for improved autoregressive LLM adaptation to sequence labeling tasks and the corresponding experimental setup, experiments, and results. Finally, to wrap up the contributions, Chapter \ref{ch:method3} proposes a fine-tuning framework for autoregressive LLMs, an experimental setup specifying how the framework is utilized, experiments, and results. The contribution chapters provide comments on the obtained results and draw first-order conclusions. In Chapter \ref{ch:discussion}, we extend our conclusions to the implications and limitations of our proposed improvements. Finally, we summarize our contributions and propose follow-up research that we leave for future work in Chapter \ref{ch:conclusion}.\clearpage{}%
\clearpage{}%
\chapter{Background}
\label{ch:background}

As stated in the previous chapter, modern NLP relies on neural approaches. Therefore, we lay the theoretical background for neural NLP in Section \ref{sec:neural_nlp}. Starting from the idea of the distributional hypothesis, through its efficient implementations to introducing pre-training methods and foundational pre-trained neural language models, Section \ref{sec:neural_nlp} prepares the reader for the transfer learning paradigms and the pillars of the pre-trained neural language model adaptation, which follow in Section \ref{sec:tl}. We conclude this chapter with Section \ref{sec:sl} on sequence labeling, which introduces the task, its importance in NLP, and typical modeling approaches.

\section{Neural Natural Language Processing} \label{sec:neural_nlp}

Although it was not recognized at the time, the foundations of neural NLP were laid in 1954 with the introduction of the distributional hypothesis, which states that words that appear in similar contexts tend to have similar meanings \citep{distrib}. Building on the idea of the distributional hypothesis, the area of distributional semantics in NLP gradually emerged and evolved. Distributional semantics creates methods that map text (words, subwords, or even smaller lexical units) into a shared semantic space. Essentially, the goal is to represent text as an array of numbers---a vector---that encodes the semantic meaning. In practice, we want the semantic space to correspond to the vector space where units of text are represented with vectors and text units with similar meanings have vectors that are closer to each other in the vector space. Methods from distributional semantics leverage linear algebra to operate on vectorized text, enabling efficient and simple manipulation through vector operations. 

Representing text as vectors offers several key advantages. First, it simplifies the application of mathematical operations to text data. Further, vectors encode both syntax and semantics, unlike raw character-based text representations that encode only syntax. Additionally, vector representations are robust to certain perturbations, such as noise or minor variations in the text, preserving semantic meaning, whereas character-based representations lack this robustness. Another significant benefit is computational efficiency. Stacking multiple text vectors into matrices allows operations to be parallelized, taking full advantage of linear algebra optimizations. These advantages have driven extensive research and development in distributional semantic models, creating various types of distributional representations.

\subsection{Static Embeddings}

\begin{figure*}
    \centering
    \includegraphics[width=0.7\linewidth]{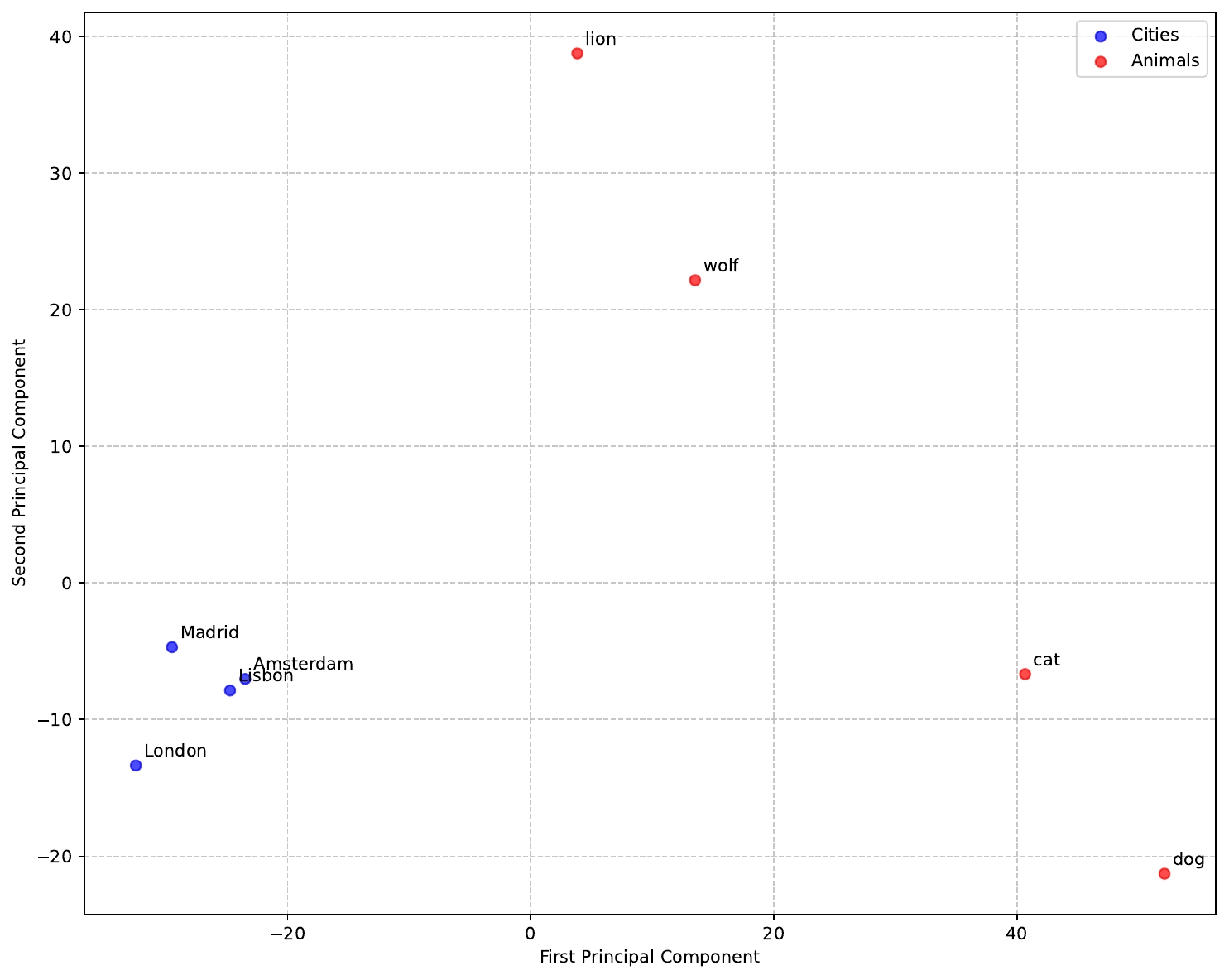}
    \caption[Visualization of embeddings in 2D-reduced vector space]{Embeddings of names of cities and animals projected into the same 2D-reduced vector space with principal component analysis. Names of cities form one group, while names of animals form another group.}
    \label{fig:word2vec_pca}
\end{figure*}

The first models that trained distributional representations of words in a neural language modeling fashion were introduced in 2000 \citep{bengio2000neural}. Soon after, the term \textit{embeddings} became widely used in the field to describe distributional representations of characters, subwords, words, sentences, or even whole documents. However, many years had to pass for researchers to witness the broader implementation of the distributional hypothesis. The neural revolution started with the introduction of \textit{Word2Vec} embeddings \cite{mikolov2013efficient}. These were the first pre-trained embeddings that captured the semantics of the words from the vocabulary of the natural language text and enabled a boom in downstream task performance. Embeddings also exhibit some favorable properties, such as the linearity of word embeddings \citep{allen2019analogies} and superposition \citep{drozd-etal-2016-word}. For example, mapping words in a vector space preserves linearity in the sense that an embedding of a word \textit{man} is related to the word \textit{king} in the same manner as an embedding of a word \textit{woman} is related to the word \textit{queen}, approximately forming a parallelogram in a shared vector space. The superposition of embeddings also holds. For example, the following operations on word embeddings are valid: $\text{king}
- \text{man} + \text{woman} \approx \text{queen}$. An interesting property of embeddings is that the relations between words are preserved even when we map the embeddings into a 2D-reduced vector space with, e.g., principal component analysis (PCA) as in Figure \ref{fig:word2vec_pca}. In this example, the names of cities are grouped together and easily separated from the animals. Furthermore, the animals naturally separate into \textit{domestic} and \textit{wild} animal subgroups with a visible distinction between the two.

The \textit{Word2Vec} embeddings were learned through either a continuous skip-gram model or a continuous bag-of-words model. The former model learns the distributional representation of the current word by predicting the context (surrounding words), while the latter learns the context (surrounding words) by predicting the word in the middle. The models that learn these representations are neural in that we train a neural network with a backpropagation algorithm and encode the representations of words from a fixed vocabulary in the network's hidden layer. After the training, we are left with embeddings of words from the vocabulary. We refer the reader to \cite{mikolov2013efficient} for more details on the exact implementation and training of the models. 

The obtained embeddings are static and do not change for the same word when used in a different context. This is a problem since it is often the case that the same words in different contexts convey different meanings. For example, in sentences \textit{He lies in bed.}~and \textit{He lies to her.}~the word \textit{lies} has the same spelling, but different meanings in different contexts. This is an example of a homonym. In the former case, the word \textit{lies} is used to describe that a person is resting on top of the bed, while the latter usage of the word \textit{lies} describes that a person is making false statements to deceive someone. Static embeddings fail to reflect that difference, having the same embedding for each word occurrence in the text, irrespective of the surrounding context. Static embeddings also struggle with polysemous words such as \textit{run}, which can mean physical movement in \textit{He wants to run in the park.}~or executing a program in \textit{He wants to run the code.} Because static embeddings assign a single representation to each word, they miss these contextual differences, resulting in worse performance on tasks such as word sense disambiguation and machine translation.

\subsection{Sequence Processing} \label{subsec:seq_proc}

Static word embeddings initiated a revolution in modern NLP, increasing the quality of features extracted from text data. However, another significant drawback of static embeddings---in addition to not modeling different meanings of a word in context---is that they were not meant to represent sequences of words, i.e., sentences and documents. One of the approaches to represent the sequence of words with static word embeddings was to average the representations of individual words that constitute a sequence. However, this simple approach often failed to capture the complexity and nuances of natural language encoded in the sequence. Therefore, researchers directed their attention to methods that would allow for a more natural processing of sequences. The most successful ones leveraged the recurrent structure of neural networks to allow for processing words in a sequence one step at a time. These methods are called recurrent neural networks (RNNs) \citep{ELMAN1990179} and were the go-to approach for modeling the semantic composition of individual words for most of the 2010s. RNNs witnessed many upgrades over the basic recurrent model, such as the introduction of gated recurrent unit (GRU) \citep{cho-etal-2014-learning}, long short-term memory network (LSTM) \citep{10.1162/neco.1997.9.8.1735} or bidirectional long short-term memory network (BiLSTM). Consequently, word and sequence representations improved with each new model. However, a key change was the introduction of the \textit{attention mechanism}. The idea was to calculate the weighted semantic composition over representations of words from the perspective of each processed word in a sequence. The attention mechanism was first combined with RNNs, which improved word representations, contextualizing each word and giving it a unique representation based on its surrounding context \citep{bahdanau2014neural}. This approach outperformed previous models for the machine translation task and paved the way for the arrival of the most used model in modern NLP---the \textit{Transformer} model \citep{vaswani2017attention}. Transformer models simplified the attention calculation and dropped the recurrence, allowing for faster and more parallelized processing of sequences.

\subsubsection{The Transformer}

\begin{figure*}
    \centering
    \hspace*{\fill}%
    \begin{subfigure}{0.5\textwidth}
        \centering
        \includegraphics[width=0.65\linewidth]{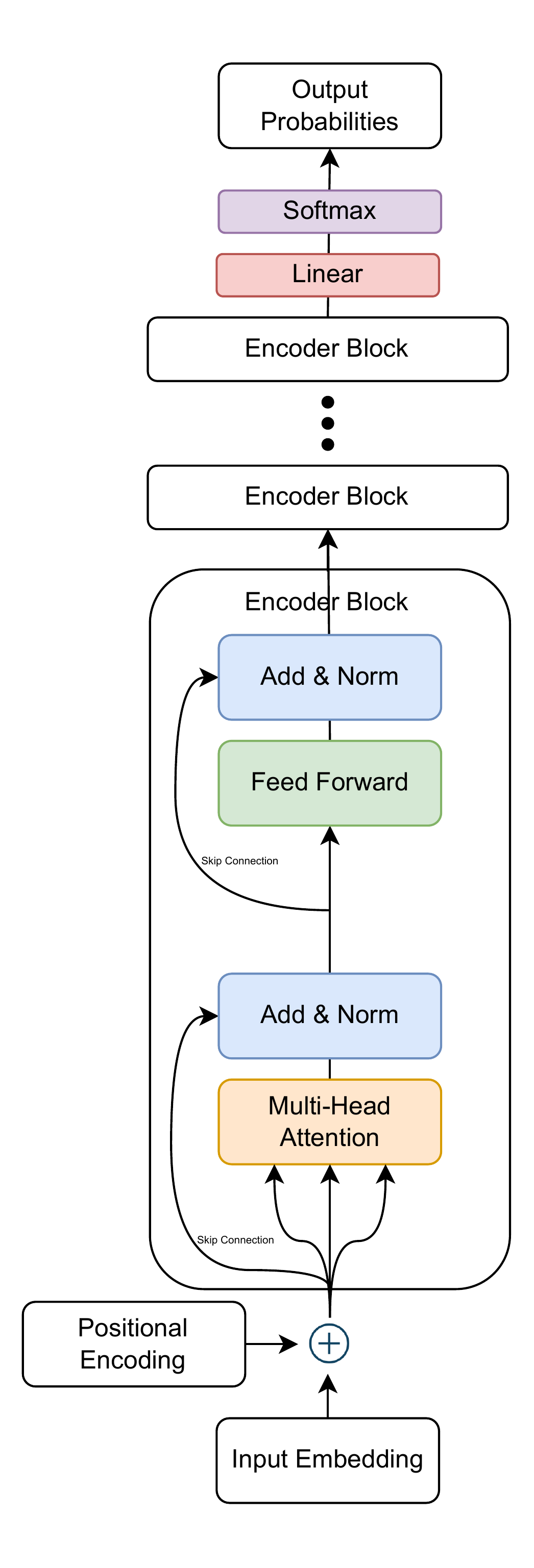}
        \vspace{0.17cm}
        \caption{}
        \label{fig:enc}
    \end{subfigure}\hfill%
    \begin{subfigure}{0.5\textwidth}
        \centering
        \includegraphics[width=0.65\linewidth]{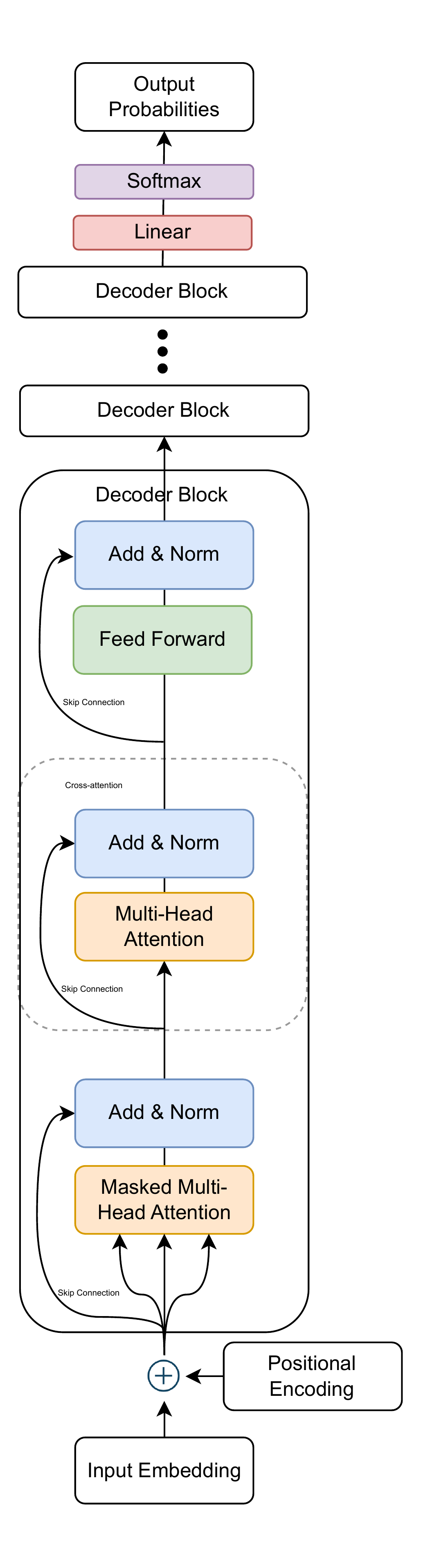}
        \caption{}
        \label{fig:dec}
    \end{subfigure}
    \hspace*{\fill}%
    \caption[The (a) encoder-only and (b) decoder-only architectures of the Transformer model]{The (a) encoder-only and (b) decoder-only architectures of the Transformer model. The cross-attention part of each decoder block is used only in the encoder-decoder architecture of the original Transformer model. For simplicity, we do not show the encoder-decoder architecture. In the original encoder-decoder Transformer model, the output of the last encoder block (a) is passed to the input of the cross-attention part of the first decoder block (b).}
    \label{fig:enc_dec}
\end{figure*}

The Transformer model is a neural sequence processing model consisting of several encoder and decoder blocks (cf.~Figure~\ref{fig:enc_dec}), where the encoder part maps inputs from the vocabulary into distributional representations, and the decoder part reconstructs the representations back into the vocabulary space. The Transformer model owes its success in modern NLP to several innovative mechanisms, including: query-key-value formulation of attention, scaled dot-product attention, self-attention, skip connections, positional encoding, and multi-head attention.

The Transformer model operates on subwords, i.e., tokens, rather than on words when processing a sequence of text. A token is a sequence of characters corresponding to a wordform or a part of a wordform, which may not necessarily correspond to a sequence of morphemes. The main motivation for splitting words into smaller units is to maintain a fixed-size vocabulary for the model, minimizing the total number of unique tokens required. For instance, we could use separate tokens for verbs such as \textit{train} and \textit{training} or \textit{swim} and \textit{swimming}. Alternatively, we might use individual tokens for \textit{train} and \textit{swim}, and a separate token for the common suffix \textit{ing}, which frequently appears in English verb inflections. By recognizing that most English verbs can occur with the suffix \textit{ing}, we significantly reduce the total number of words in the vocabulary. 

The self-attention mechanism in Transformers operates on tokens, allowing each processed token to have different embeddings based on either left, right, or both contexts. At the same time, the positional encodings account for the order of the words in the sequence. The attention mechanism used by \citep{bahdanau2014neural} is called \textit{additive}, leveraging a feed-forward network with one hidden layer to calculate the attention scores. On the other hand, the Transformer model relies on the \textit{scaled dot-product (multiplicative) attention}, which delivers better performance in terms of space and time complexity \citep{vaswani2017attention}. The scaled dot-product attention is calculated using the query $\mathbf{W}^q$, key $\mathbf{W}^k$, and value $\mathbf{W}^v$ weight matrices inspired by retrieval systems. Each word from the input sequence is first embedded with an embedding matrix $\mathbf{E}$ to form input matrix $\mathbf{X}$ and then translated to its query, key, and value representations with the following matrix multiplication operations $\mathbf{Q}=\mathbf{X}\mathbf{W}^q$, $\mathbf{K}=\mathbf{X}\mathbf{W}^k$, and $\mathbf{V}=\mathbf{X}\mathbf{W}^v$. The dot-product attention is calculated as $\mathit{softmax}\left(\mathbf{Q}\mathbf{K}^T\right)\mathbf{V}$, where $\mathit{softmax}_i(\mathbf{z}) = \frac{e^{z_i}}{\sum_{j=1}^K e^{z_j}}$ for a vector $\mathbf{z}$ with $K$ elements. To obtain the scaled dot-product attention, the dot-product attention is scaled with a scaling factor $\frac{1}{\sqrt{d_k}}$, where $d_k$ is the dimensionality of queries and keys. This scaling factor was introduced to counter the problem of small gradients that surface with dot-product attention \citep{vaswani2017attention}. The final formulation of scaled dot-product attention is given with:

\[
\mathrm{Attn}(\mathbf{Q},\mathbf{K},\mathbf{V})=\mathit{softmax}\left(\frac{\mathbf{Q}\mathbf{K}^T}{\sqrt{d_k}}\right)\mathbf{V}.
\]

The Transformer model leverages multiple query, key, and value weight matrices to form multiple attention heads, which operate in parallel and allow for joint processing of input sequences by calculating attention scores at different positions in the sequence from the perspective of different representation subspaces \citep{vaswani2017attention}. This mechanism is called \textit{multi-head attention}. For $h$ attention heads, it is given with:

\[
\mathrm{MultiHead}(\mathbf{Q}, \mathbf{K}, \mathbf{V}) = \mathrm{concat}(\mathbf{head}_1, \dots, \mathbf{head}_h)\mathbf{W}^o,
\]
\[
\mathbf{head}_i = \mathrm{Attn}(\mathbf{X}\mathbf{W}_{i}^{q},\mathbf{X}\mathbf{W}_{i}^{k},\mathbf{X}\mathbf{W}_{i}^{v})
\]

\noindent where $\mathbf{W}^o$ is the output weight matrix with learnable parameters. Multi-head attention can also be masked to endow the decoders with text-generation capabilities. This masking is performed in the decoder block in the \textit{masked multi-head attention} part of the block (cf.~Figure \ref{fig:dec}) using a triangular, causal mask, which pulls the softmax attention scores to zero for words to the right of the word we are calculating the attention scores for. 

The Transformer model was created to have both encoder and decoder parts because the model was meant to be used for machine translation, where the inputs and outputs are both sequences of words. In the encoder-decoder Transformer architecture, the output of the last encoder block from Figure \ref{fig:enc} is passed into the cross-attention layer of the first decoder block from Figure \ref{fig:dec}. Then, the output of the last block in the decoder is passed to the combination of \textit{linear} and \textit{softmax} layers to obtain output probabilities. These output layers can also be stacked on the last encoder block. This way, the decoder model can generate words from the vocabulary by mapping the output representations from the last layer to the vocabulary space, and the encoder can solve NLU tasks by classifying the whole input sequence or labeling the entire sequence. For example, the linear layer with weight matrix $\mathbf{W}$ and a bias $\mathbf{b}$ performs sequence labeling using a representation for each word from the last block $\mathbf{x}_i^{\mathrm{last}}$. The process can be described with the expression:

\[\mathit{softmax} (\mathbf{W}_\mathrm{linear}^\mathrm{T}\mathbf{x}_i^{\mathrm{last}} + \mathbf{b}_\mathrm{linear})
\]

\noindent where the dimensionality of the weight matrix $\mathbf{W}$ from the linear layer determines the size of the label space we are projecting the representation of each word $\mathbf{x}_i$ to. %

While the authors of \citep{vaswani2017attention} trained their encoder-decoder Transformer model from scratch for a specific task in a supervised manner leveraging labeled data, the research community shifted to \textit{pre-training} on unlabeled corpora using self-supervised training objectives combined with either encoder-only or decoder-only Transformer blocks.\footnote{\footnotesize{There also exist pre-training objectives for the encoder-decoder architecture, which resulted in models such as T5 \citep{raffel2020exploring}, but are not explored further in this thesis.}} Supervised training relies on labeled data, where each input is paired with a corresponding target output to teach the model a specific task. This approach is also known as \textit{learning with a teacher}. In contrast, self-supervised training eliminates the need for manual labeling by leveraging the inherent structure of unlabeled data. The model learns by discovering and exploiting hidden patterns within the data itself, using raw data as labels. When training in a self-supervised manner, the model learns linguistic structures and acquires world knowledge. Training neural language models is typically performed in batches, by grouping multiple (e.g., $32$) sequences, as empirical studies demonstrate that this approach both stabilizes gradient-based optimization and accelerates training.

The approaches for pre-training language models are usually \textit{self-supervised} because there are no assigned labels to the corpora used for pre-training. Consequently, the sequences of words in natural language become labels. Two prevailing architectures emerged, each coupled with a compatible pre-training paradigm: (1) the encoder-only architecture, with the masked language modeling (MLM) pre-training objective, and (2) the decoder-only architecture, utilizing causal language modeling (CLM) for pre-training. There were also other methods proposed for pre-training, such as replaced token detection \citep{clark2020electra} and XLNet, a generalized language model pre-training method \citep{yang2019xlnet}, but are not dominant in the field today. %

The models trained using self-supervision on large corpora are called pre-trained language models (PLMs). The effect of self-supervised pre-training objectives using unlabeled corpora was that the models acquired linguistic competence and world knowledge. Each PLM is equipped with its pre-trained tokenizer, a tool for splitting natural language text into tokens.

\subsubsection{Encoder-only and Decoder-only Models} \label{subsec:enc_dec}

The first encoder-only (henceforth \textit{encoder}) PLM that gained traction was BERT \citep{devlin-etal-2019-bert}. The BERT model used $12$ encoder blocks in a standard (BERT-base) configuration. Each block consists of multiple layers that perform transformations on input embeddings (cf.~Figure \ref{fig:enc}). These layers include performing multi-head attention, normalization, addition with input embeddings, and passing through the feed-forward neural network layer. The output of each encoder block is passed to the succeeding encoder block as its input. These blocks are architecturally identical but have separate parameters and are stacked on top of each other to form the BERT model.

\begin{figure*}
    \centering
    \begin{subfigure}{1.0\linewidth}
        \includegraphics[width=\linewidth]{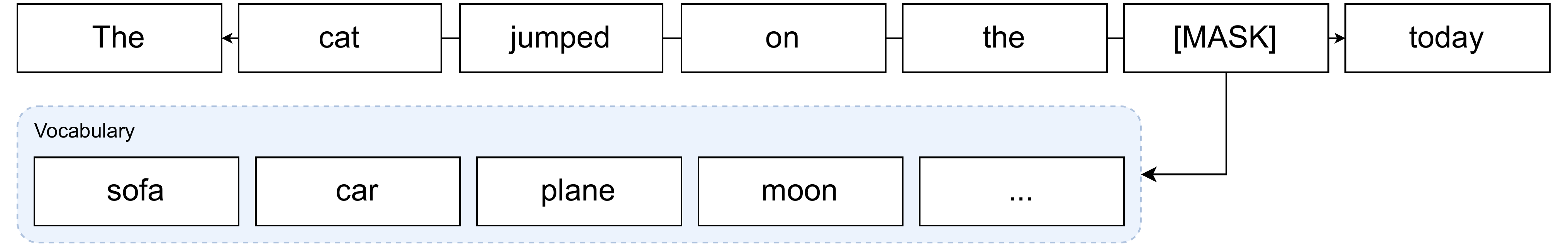}
        \caption{}
        \label{fig:mlm}
    \end{subfigure}
    
    \vspace{0.3cm} %
    
    \begin{subfigure}{1.0\linewidth}
        \includegraphics[width=\linewidth]{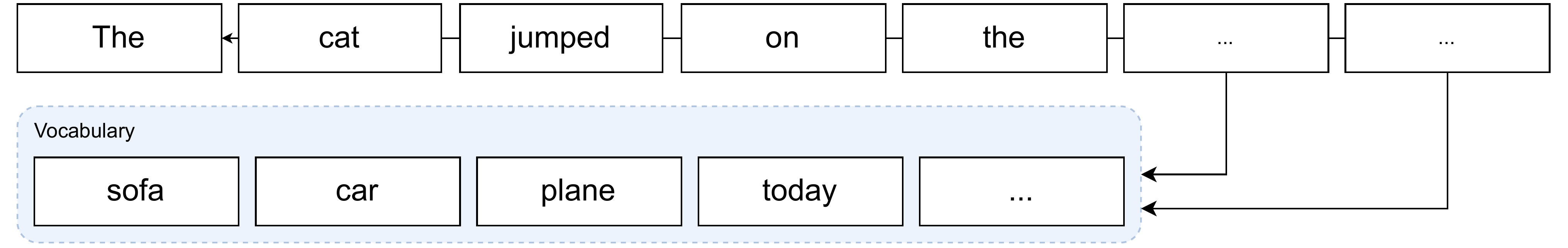}
        \caption{}
        \label{fig:clm}
    \end{subfigure}
    
    \caption[Comparison of pre-training objectives]{Comparison of pre-training objectives: (a) MLM randomly masks input tokens (shown as [MASK]) and predicts the original tokens using bidirectional context; (b) CLM predicts each token sequentially using only the left context. For simplicity, tokens are shown as whole words rather than subwords.}
    \label{fig:pretrain_objectives}
\end{figure*}

BERT was pre-trained using the MLM pre-training and next-sentence prediction objectives (next-sentence prediction was proved redundant by authors of \citep{liu2019roberta}). MLM pre-training works by masking a fixed proportion of tokens in each sequence (usually only 15\% of the tokens are masked) and training the model to predict which token from the vocabulary should be in place of each masked token. As shown in Figure \ref{fig:mlm}, this involves (1) replacing selected tokens with a special [MASK] token and (2) training the model to predict the original tokens from the vocabulary for the masked positions. For instance, given the masked input \textit{The cat jumped on the [MASK] today}, the model learns to predict \textit{sofa} rather than irrelevant alternatives such as \textit{plane} or \textit{moon}. If the token to predict is in the middle of the input sequence, the model utilizes both the left and the right context, enabling the model to excel at NLU tasks. Some examples of NLU tasks are sentiment classification, paraphrase identification, semantic textual similarity, recognizing textual entailment, question answering, and named entity recognition. MLM is a bidirectional pre-training objective, which means that the model can utilize both the left and right context through attention scores during the pre-training process. Pre-training encoders with bidirectional pre-training objectives, such as MLM, encourages them to develop robust contextual representations, ensuring better performance on NLU tasks. The model trained with MLM uses a variation of the cross-entropy loss, standardly used in deep learning models, defined as:
\[
L_{\text{MLM}} = - \frac{1}{|M|} \sum_{i \in M} \log P(x_i \mid \mathbf{h}_i^L),
\]
\noindent where the model is penalized only for wrong predictions of the tokens from the list of masked tokens $M$. The embedding from the last transformer layer $L$ for each masked token $i$ is denoted as $\mathbf{h}_i^L$. The probability of the masked token $x_i$ is conditioned on this output vector. 

Although the MLM pre-training lays strong foundations for solving NLU tasks, the downside is that only a small fixed percentage of inputs (masked ones) in the training corpora affect the loss and contribute to weight updates of the model. Furthermore, the [MASK] tokens are used only during pre-training and dropped afterward, introducing a \textit{pre-train--fine-tune discrepancy} when adapting to the target task \citep{meng2023representation}. Pre-train--fine-tune discrepancy is not limited to models pre-trained with MLM, but is a common problem in NLP \citep{khanehzar-etal-2019-modeling, clark2020electra}.

The first decoder-only (henceforth \textit{decoder}) PLM with CLM pre-training success was the \textit{generative pre-trained Transformer} (GPT) \citep{radford2018improving}. Its standard configuration was made of $12$ decoder blocks, the same as BERT-base. In GPT's decoder architecture, the cross-attention part (cf.~Figure \ref{fig:dec}) was removed from each decoder block. Thus, the only difference between the encoder and decoder architecture is that the former relies on \textit{multi-head attention}, while the latter uses \textit{masked multi-head attention}, preventing the model from utilizing the right-side context with the causal mask. As was the case for encoders, the output of each decoder block is passed to the succeeding decoder block as its input. These blocks are architecturally identical but have separate parameters and are stacked on top of each other to form the GPT model.

CLM pre-training objective is crucial to ensure that the model can be used to generate text in a natural language fashion at inference time by generating one token at a time from left to right. Formally, for a sequence of tokens $T=(t_1,\dots,t_T)$, the model is trained with a cross-entropy CLM loss defined as:
\[
L_{CLM} = - \sum_{i=1}^{|T|}\log P(t_i | t_{<i})
\]
\noindent where the prediction of the next token $t_i$ in the input sequence $T$ depends on the correct sequence of the previous tokens $t_{<i}$. Based on this loss, for the token at position $i-1$ in $T$, the model learns the probability distribution of the token at position $i$ in $T$, i.e., the next token in the sequence.

During pre-training of GPT, the model's task was to predict the next token in the sequence (choosing from the available tokens in the vocabulary), conditioning the prediction solely on the left (so far generated) context. At inference time, this translates to iterative text generation: the model samples from its predicted token distribution over the vocabulary, appends the predicted next token to the end of the input, and repeats the process \textit{autoregressively}, by feeding the input again through the model for obtaining the following token in the sequence. As shown in Figure \ref{fig:clm}, when processing the partial sequence \textit{The cat jumped on the}, the model would first assign the highest probability to the word \textit{sofa} from the vocabulary. Upon extending the context to \textit{The cat jumped on the sofa}, it would then generate the subsequent token (e.g., \textit{today}) conditioned on this expanded sequence. 

\begin{figure*}
    \centering
    \includegraphics[width=1.0\linewidth]{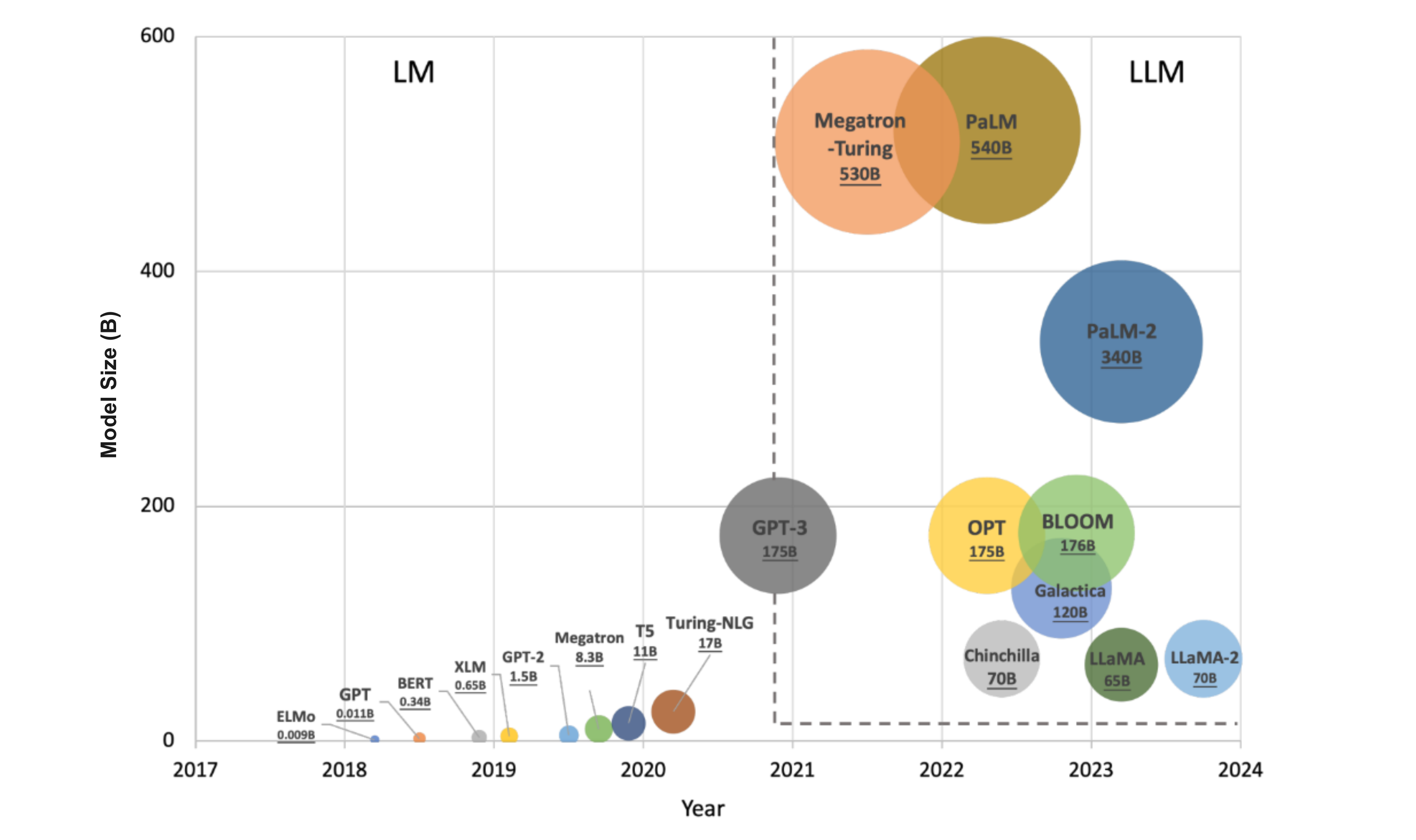}
    \caption{The scale of pre-training for neural language models \cite{scaling_lms_src}}
    \label{fig:scaling}
\end{figure*}

Both encoder and decoder architectures were explored equally in the early stages of pre-training research. In experiments that juxtaposed encoders and decoders of a comparable number of parameters (up to a couple hundred million parameters), MLM-based encoders consistently outperformed CLM-based decoders on NLU tasks \citep{devlin-etal-2019-bert}. Interestingly, the authors of \citep{shoeybi2019megatron} showed that scaling the BERT model to four billion parameters with MLM and the GPT-2 model \citep{radford2019language} to eight billion parameters with CLM brings gains on NLU tasks. Nevertheless, the community stopped scaling up encoders and continued scaling decoders to multiple billion parameters, leading to a proliferation of LLMs. The scaling of pre-trained neural language models is shown in Figure~\ref{fig:scaling}. 

Between 2017 and 2021, PLMs expanded from millions to billions of parameters. The 2021 release of GPT-3, with its $175$ billion parameters, marked a definitive transition into the LLM era. Since then, both proprietary and open-weight LLMs have been released with a higher and lower number of parameters than GPT-3. Notably, starting with 2023, more compact open-weight models (ranging from one to ten billion parameters) began outperforming their larger predecessors, such as GPT-3, despite the latter's advantage in scale. This progress has redefined what is considered an LLM. Models exceeding ten billion parameters are now typically recognized as LLMs, while the term small LLMs has emerged to describe compact, efficient models operating at the single-digit billion scale.

Scaling up the number of LLM parameters and making pre-training more efficient has given rise to notable properties in decoder LLMs, including enhanced zero- and few-shot learning capabilities (learning from zero to a few annotated examples) \citep{wei2021finetuned, brown2020language}, the ability for in-context learning (ICL) \citep{brown2020language, liu2023pre}, and chain-of-thought prompting \citep{NEURIPS2022_9d560961}. Modern LLMs such as GPT-3 \citep{brown2020language}, PaLM \citep{chowdhery2023palm}, and ChatGPT demonstrate these emerging capabilities. ICL is a method that is characteristic for LLMs as is stems from scaling up to billions of parameters. It is performed by prompting LLMs with task demonstrations. ICL typically involves giving the LLM a description of the task (instruction) and demonstrations (examples and expected responses) of how to solve the task. Using ICL, the model can solve the task better than without any demonstrations in the context. To improve the model's ability to solve tasks, a popular paradigm of instruction tuning (IT) emerged \citep{mishra-etal-2022-cross}. IT concerns training the model with CLM loss to address specific tasks in a supervised, generative manner. The model is given instructions for each data point and trained to solve the task, relying on its text generation capabilities acquired during pre-training. IT belongs to a broader family of supervised fine-tuning (SFT) techniques. This thesis leverages both variants of the PLMs---encoders and decoders in the context of pre-trained model adaptation to a domain, task, or dataset.

\section{Transfer Learning} \label{sec:tl}

Achieving SOTA results on NLU tasks typically follows a two-step approach: first, pre-training on large unlabeled corpora, and then further adapting (fine-tuning)\footnote{\footnotesize{We use the terms \textit{adaptation} and \textit{fine-tuning} interchangeably throughout the thesis, referring to the process of changing PLM weights to achieve better performance on the target task.}} the model weights on task- or domain-specific data---a process referred to as \textit{transfer learning} \citep{ruder-etal-2019-transfer, raffel2020exploring}. Pre-training enables a model to acquire foundational linguistic competence and knowledge of the world, encoded in its weights before the adaptation begins. This contrasts with training from random initialization, where no prior knowledge is present, so transfer learning is infeasible. The transfer learning process makes model training easier, reaching better generalization than training from random initialization. 

\begin{figure*}
    \centering
    \includegraphics[width=1.0\linewidth]{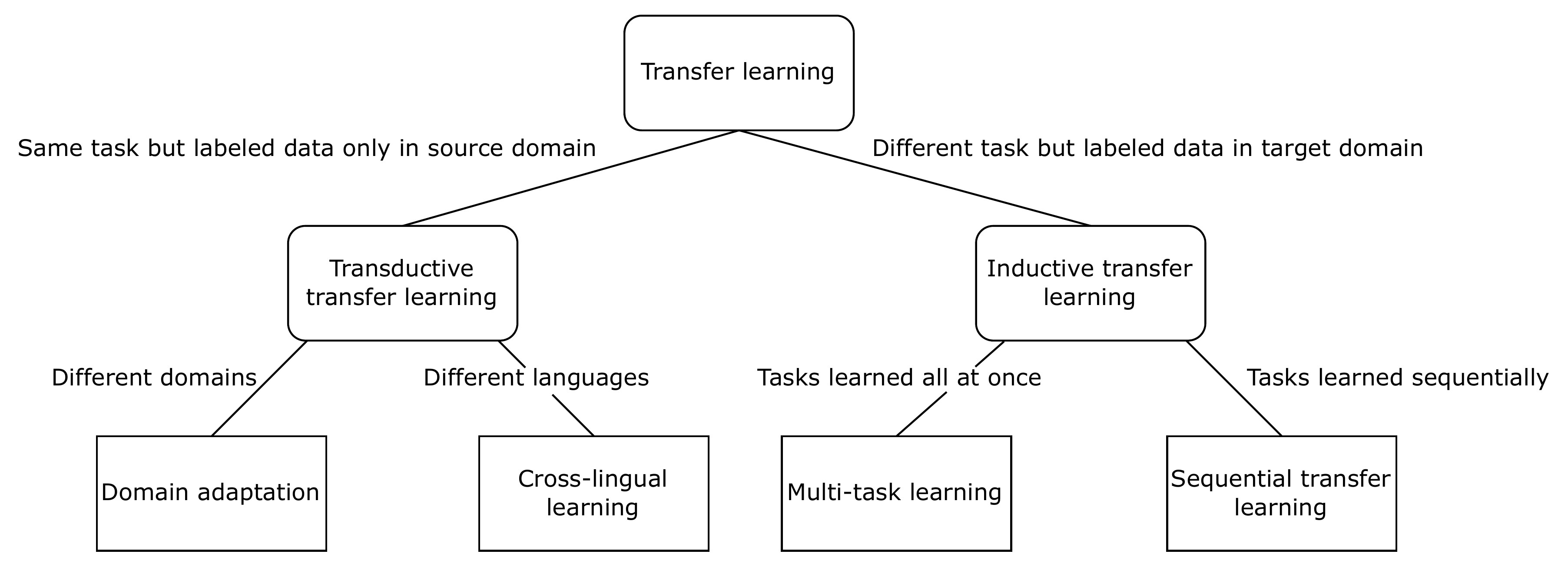}
    \caption{Transfer learning taxonomy for NLP as introduced in \cite{ruder2019neural}}
    \label{fig:tl_tax}
\end{figure*}

Transfer learning in modern NLP de facto implies starting from a pre-trained language model where the transfer is performed from the \textit{source} task and/or domain to the \textit{target} task and/or domain. The transfer is typically performed from a high-resource source task and/or domain to the low(er)-resource target task and/or domain. The transfer occurs by updating the parameters of the starting model to achieve better performance on the target task and/or domain. This starting model is commonly a pre-trained language model or a pre-trained language model already fine-tuned on a source task or domain. Depending on whether the transfer applies to the same task or a different one and the nature of the source and target domains, the author of \citep{ruder2019neural} distinguishes between \textit{transductive} and \textit{inductive} transfer learning\footnote{\footnotesize{In machine learning, there is a difference between \textit{inductive} and \textit{transductive} learning. Inductive learning (e.g., supervised) generalizes from labeled data to unseen data points, while transductive learning (e.g., self-supervised) leverages intrinsic structure in unlabeled data, uncovering patterns without explicit generalization.
}} (cf.~Figure~\ref{fig:tl_tax}). Transductive transfer learning occurs when the source and target tasks are the same, but the labeled data is available only in the source domain. When applied across different domains, this process is known as \textit{domain adaptation} \citep{gururangan-etal-2020-dont}. If transductive transfer learning is performed across different languages, it falls under \textit{cross-lingual learning} \citep{ruder2019survey}. In contrast to transductive transfer learning, inductive transfer learning involves transferring from one task to another where labeled data is available in the target domain. Inductive transfer learning can be approached in two main ways: learning multiple tasks simultaneously, known as \textit{multi-task learning}\footnote{\footnotesize{Throughout the thesis, \textit{multi-task learning} and \textit{multi-task fine-tuning} are used interchangeably, implying that the weights of the model are changed.}} \citep{zhang-etal-2023-survey}, or learning tasks in a sequence, referred to as \textit{sequential transfer learning}.

\subsection{Transductive Transfer Learning} \label{subsec:transductive}

This branch of transfer learning entails adaptation either to different domains or languages. However, since this thesis focuses exclusively on English, cross-lingual learning will not be discussed further. Instead, the focus in the context of transductive transfer learning will be put on \textit{domain adaptation}. Adapting PLMs to different domains is an important research direction. Even though large-scale pre-training covers a range of domains, including but not limited to Wikipedia, news, books, scientific, biomedical, and financial domains in the pre-training data mixture, there are still underrepresented domains that surface through suboptimal performance in the adaptation phase. Improvements on one domain exhibit limited transfer to other domains, emphasizing the need for a balanced pre-training data mixture consisting of various domains \citep{fu2024data}. Furthermore, empirical evidence suggests massive performance drops in zero- and low few-shot transfer from a high-resource source to a low(er)-resource target domain---a phenomenon commonly referred to as \textit{negative transfer} \citep{wang2019characterizing, ngo-trung-etal-2021-unsupervised, meftah-etal-2021-hidden, 9938381, dukic-etal-2024-leveraging}. Zero-shot transfer refers to the case when there is no labeled target data available, and the model has to generalize to a target using only source examples. On the contrary, a few-shot transfer implies having a limited number of annotated target data, alleviating the negative transfer when combined with source data.

More recently, domain adaptation for models based on PLMs has been conducted with general self-supervised language modeling on (unlabeled) domain-specific corpora \cite{gururangan-etal-2020-dont,hung-etal-2022-ds}. The authors of \cite{gururangan-etal-2020-dont} show that continual pre-training for MLM-based encoder on target domain or task data can bring significant gains when transferring to a target domain or task. These approaches are referred to as domain-adaptive pre-training (DAPT) and task-adaptive pre-training (TAPT).

\subsection{Inductive Transfer Learning}

\begin{figure*}
    \centering
    \includegraphics[width=1.0\linewidth]{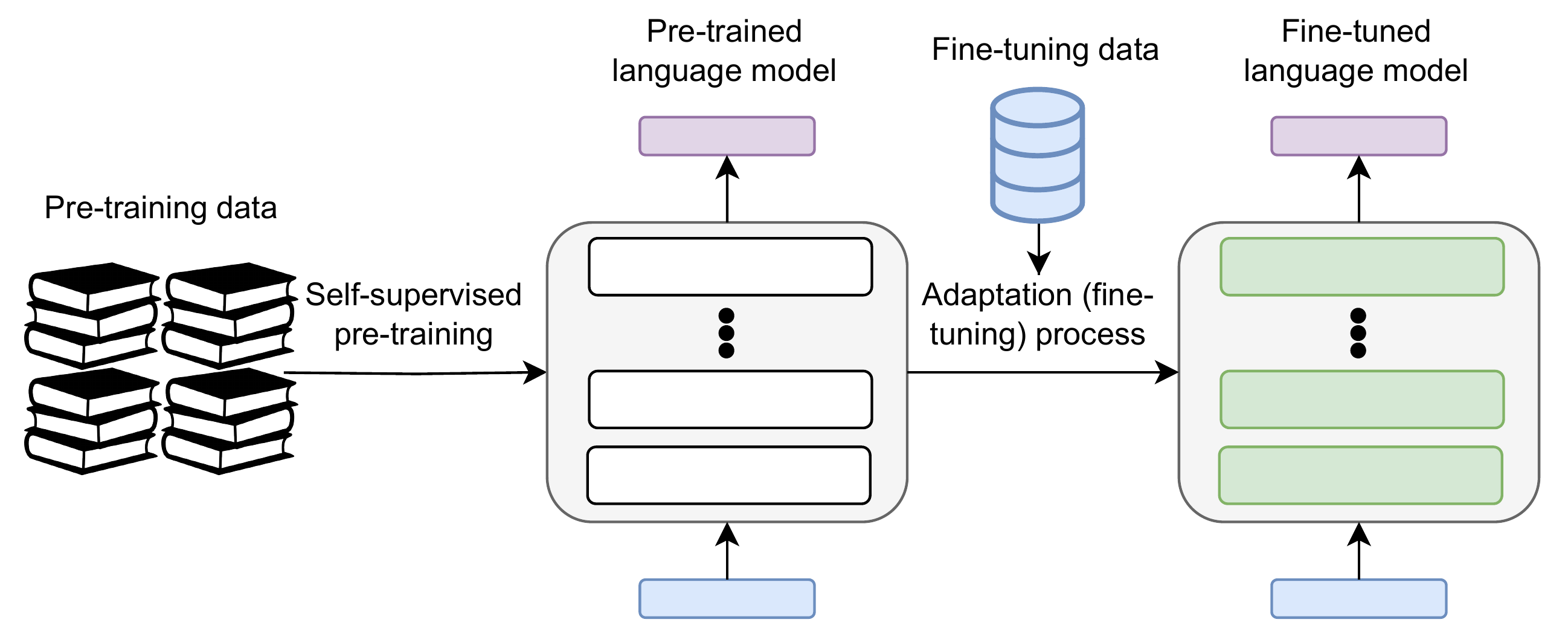}
    \caption{Sequential transfer learning (adapted from \citep{jm3})}
    \label{fig:seq_tl}
\end{figure*}

Inductive transfer learning paradigms differ in the order the tasks are learned---either simultaneously or sequentially. \textit{Multi-task learning} paradigm trains a model on multiple tasks at the same time, leveraging shared information across tasks that benefit from being learned together. Multi-task learning is typically implemented in Transformer-based models by adding task-specific heads (usually softmax classifiers) to a pre-trained model. The parameters of the model and task heads are commonly updated in one of two ways: (1) either by backpropagating over a shared loss function interchangeably between tasks as in \cite{baric-etal-2023-target} or (2) by having separate losses for each task and averaging them before applying backpropagation as in \citep{dukic-etal-2024-leveraging}. Alternating updates can be done on a mini-batch, level, epoch level, or even training set level. \textit{Sequential transfer learning} is the standard transfer paradigm in NLP. It starts with a neural language model pre-trained with self-supervision on unlabeled corpora (typically with MLM or CLM). Further, the model is adapted to a task or a dataset using labeled target data by updating the weights of the pre-trained model (cf.~Figure~\ref{fig:seq_tl}). If the adaptation process is successful, the model achieves better generalization on the target task than the starting pre-trained language model. This transfer approach led to the most significant advances on NLU tasks and benchmarks such as GLUE \citep{wang-etal-2018-glue} and is consequently the most researched transfer paradigm in modern NLP.

\subsection{Pre-trained Language Model Adaptation} \label{subsec:plm_adapt}

Since training PLMs with millions and billions of parameters is expensive, academic research directed its efforts to improve the adaptation step of transfer learning. According to the authors of \citep{ruder-etal-2019-transfer}, three standard methods to improve the adaptation step when transferring to a target task are: 
\begin{enumerate}
    \item Introducing an additional task-specific signal through multi-task learning, weak supervision, and/or ensembles;
    \item Architectural modifications;
    \item Different optimization schemes.
\end{enumerate}

Each adaptation method highly depends on the target task. Before performing transfer learning, it is essential to carefully consider the task's characteristics and the discrepancy between pre-training and the target task. Choosing the proper adaptation method aims to diminish this discrepancy, ultimately boosting the overall performance on the target task. Failing to recognize where exactly the discrepancy lies and ignoring the peculiarities of the target task leads to suboptimal performance. Adapting the model appropriately to the target task boils down to decisions on what additional signal to include, which part of the model's internals to change (if any), which weights to update, and how \cite{ruder2019neural}.

\subsubsection{Introducing More Signal} \label{subsubsec:intro_more_signal}

Adapting PLMs in a supervised manner to the target task necessitates having a labeled corpus, which can be costly to obtain. High-quality datasets for target tasks are limited, and many target tasks are low-resource. Domain-specific data are even more scarce, making adaptation even more challenging. When transferring to a low-resource, domain-specific task, we often encounter one of two extreme scenarios: either no data is available for the target domain (zero-shot) or only a small amount of labeled data is present (few-shot). In such cases, introducing more signal into the PLM through a related source task with more data (a high(er)-resource task) is an opportunity for improving the adaptation process. Options for introducing more signal surface in the form of \textit{sequential intermediate fine-tuning} and \textit{multi-task fine-tuning} on related tasks. \textit{Dataset slicing}, \textit{semi-supervised learning}, and \textit{ensembling} are other options for building additional signal into the PLM, but are not explored in this thesis. For more details, we refer the interested reader to the tutorial by Ruder et al. \citep{ruder-etal-2019-transfer} and to the works discussing these other options in more detail \citep{clark-etal-2018-semi, xie2020unsupervised}.

Sequential intermediate fine-tuning gets more signal into the PLM by first fine-tuning on a related task data followed by fine-tuning on a target task. There can also be a sequence of multiple intermediate supervised fine-tuning tasks before transferring to the target task. Furthermore, PLM can first be subjected to domain or task adaptation with a self-supervised language modeling objective before being fine-tuned on the available labeled target task examples as in \cite{gururangan-etal-2020-dont, repetal2024electras}, thus leveraging the unlabeled target data through self-supervision. 

Using related auxiliary tasks, multi-task fine-tuning trains the PLM with multiple training objectives simultaneously. Options for multi-task fine-tuning are many. If enough data is available, training can be performed by averaging target and auxiliary task(s) losses during a fixed set of epochs and then fine-tuning only on a target task for a few epochs towards the end of the training \citep{ruder-etal-2019-transfer}. However, when target data is scarce, one can resort to mixed batch fine-tuning, as in \citep{schmidt-etal-2022-dont}. A mixed batch consists predominantly of source auxiliary examples and a much lower fixed share of few-shot target examples. Intuitively, having fewer few-shot target examples should contribute to the update of model parameters with equal weight as the abundant source examples and ultimately prevent the model from overfitting on source data. Whether enough target task data is available or not, the losses can be averaged or kept separately for interchangeable updates of the parameters. Another option for multi-task fine-tuning is leveraging a set of unlabeled target task data through an auxiliary language modeling training objective, usually implemented by having a separate language modeling head (typically the same the model was pre-trained with, e.g., BERT with MLM head) alongside the target task-specific head on top of the PLM. The parameters are updated with a weighted average loss between the language modeling and the target task objectives.

\subsubsection{Modifying the Architecture of the Model} \label{subsubsec:modify_architecture}

Architectural modifications can be divided into two categories: (1) the backbone PLM is kept unchanged or (2) PLM's internals are modified \citep{ruder-etal-2019-transfer}. The former means we keep the PLM as is, remove the language modeling head used for pre-training, and attach additional layer(s) to model output, input, or both, whichever benefits the target task most. The latter pertains to modifications that would enable a more natural transfer to the target task since the target task is structurally different from the one used for pre-training. For example, autoregressive language models were pre-trained without bidirectionality, which aggravates the transfer to NLU tasks that benefit from bidirectionally \citep{dukic-snajder-2024-looking}. Therefore, the modifications in the PLM internals are driven by the requirements and the nature of the target task. These modifications can be in the form of removing layers from model internals, adding more parameters through additional layers \citep{houlsby2019parameter}, introducing more skip connections, or modifying the attention modules of the PLM \citep{ramachandran-etal-2017-unsupervised, dukic-snajder-2024-looking}.

\subsubsection{Choosing Optimization Scheme} \label{subsubsec:optim_scheme}

When choosing the proper optimization scheme, the design decisions pertain to (1) how and when to update the weights and (2) which model weights to update \citep{ruder-etal-2019-transfer}. The question of how and when to update the weights of the PLM for optimal target task performance boils down to (1) choosing the proper learning rate schedule and (2) modifying the loss function. 

An appropriate learning rate is crucial for balancing underfitting and overfitting in PLM adaptation. A learning rate that is too small may prevent the model from learning effectively, while one that is too large risks overwriting pre-trained knowledge. To address this, learning rate schedulers adjust the learning rate dynamically, ensuring optimal parameter updates throughout training. Some authors propose training the lower layers of the PLM with lower learning rates for better results \cite{howard-ruder-2018-universal}. Others propose scheduling the learning rate change using a cosine annealing learning rate schedule with restarts during the training process \citep{loshchilov2016sgdr}. Modern LLMs rely on some variant of cosine learning rate schedule with a warmup period \citep{liu2019variance} as in \citep{touvron2023llama, dubey2024llama}. Other scheduling strategies include starting with a higher learning rate and gradually reducing it throughout training by multiplying it with a fixed factor or leveraging a triangular learning rate schedule \citep{howard-ruder-2018-universal}, where the learning rate first increases linearly and then decays linearly toward the end of training.

PLMs can also improve their adaptation performance through modifications of the loss function. Most PLMs use the cross-entropy loss function for pre-training and adaptation phases. A typical modification of the cross-entropy loss function involves adding a regularization term to prevent overfitting and avoid catastrophic forgetting. This can be done, for example, by forcing the target model parameters to stay close to the pre-trained model parameters \citep{wiese-etal-2017-neural}. Other types of modifications are typically driven by the nuances of the target task. The loss is modified with weighting terms to amplify the effect of specific data points or auxiliary training objectives, which can be viewed as a more specific type of regularization \citep{tong-etal-2020-improving, sow2025dynamic}. Finally, some methods indirectly affect parameter updates through the loss function, although the loss function is not modified. For example, embeddings of the auxiliary task labels can be used to guide the parameter updates through the loss function as in \citep{dukic-etal-2024-leveraging}. Furthermore, some tokens can be included in loss calculation, and others can be excluded depending on the target task the PLM is adapted to. An interesting example is \citep{huerta-enochian-ko-2024-instruction}, which explores the impact of excluding tokens from loss calculation during SFT. The authors examine how adjusting the probability of masking tokens (which they refer to as token weighting) through the loss function affects the adaptation of CLM-based decoders. However, a more practical approach for CLM-based decoders is to compute the loss only on expected responses (i.e., completions), excluding preceding tokens in the prompt, such as in \citep{an2024response}. This adjustment ensures that the model is penalized only for incorrectly generated response tokens.

The question of which model weights to update is as crucial as choosing how and when to update the weights. Choosing a subset of weights to update from induces a trade-off between the training complexity and fine-tuning performance. The more weights we adapt during fine-tuning, the higher the computational complexity, but the model generalizes better. Fine-tuning the whole model, starting with the initialized pre-trained weights, is the most demanding approach. 

The least computationally demanding approach is freezing the pre-trained parameters of the model and using the PLM as a feature extractor. The embeddings from the last PLM layer are typically fed to a simple, fully connected linear classifier whose parameters are updated, or the embeddings are fed to an even simpler classifier such as logistic regression \cite{9260074}. Extensions of this approach are learning a linear combination of the embeddings of different layers taken from the frozen PLM \citep{peters-etal-2018-deep} or forwarding the embeddings from the frozen PLM to a more complex model with multiple hidden layers, embedding layer, etc. The feature extraction approach yields the best results if the source (usually the pre-training task) and target task are unalike \citep{peters-etal-2019-tune-new, ruder-etal-2019-transfer}. For example, the authors of \citep{peters-etal-2019-tune-new} show that using BERT as a feature extractor is more beneficial for classification tasks than semantic textual similarity tasks, as they uncover the latter's similarity to BERT's pre-training objective. However, as the PLMs grew, the community shifted from using them as feature extractors to adjusting the weights using parameter-efficient fine-tuning (PEFT) methods \citep{houlsby2019parameter}. 

PEFT methods provide a practical trade-off between computational efficiency and performance, often matching or even surpassing full fine-tuning \citep{lialin2023scaling} while updating only a small fraction of PLM parameters. These methods effectively reduce the computational demands of fine-tuning models with billions of parameters while significantly minimizing memory usage on disk. Saving modules trained with PEFT requires considerably less memory than PLM weights while encoding the knowledge about a domain, task, or dataset. This eliminates the need for full fine-tuning and reduces the chances of catastrophic forgetting \citep{MCCLOSKEY1989109}---the case when the model overwrites the previously learned information with new one due to the effects of fine-tuning.

\begin{figure*}
    \centering
    \includegraphics[width=1.0\linewidth]{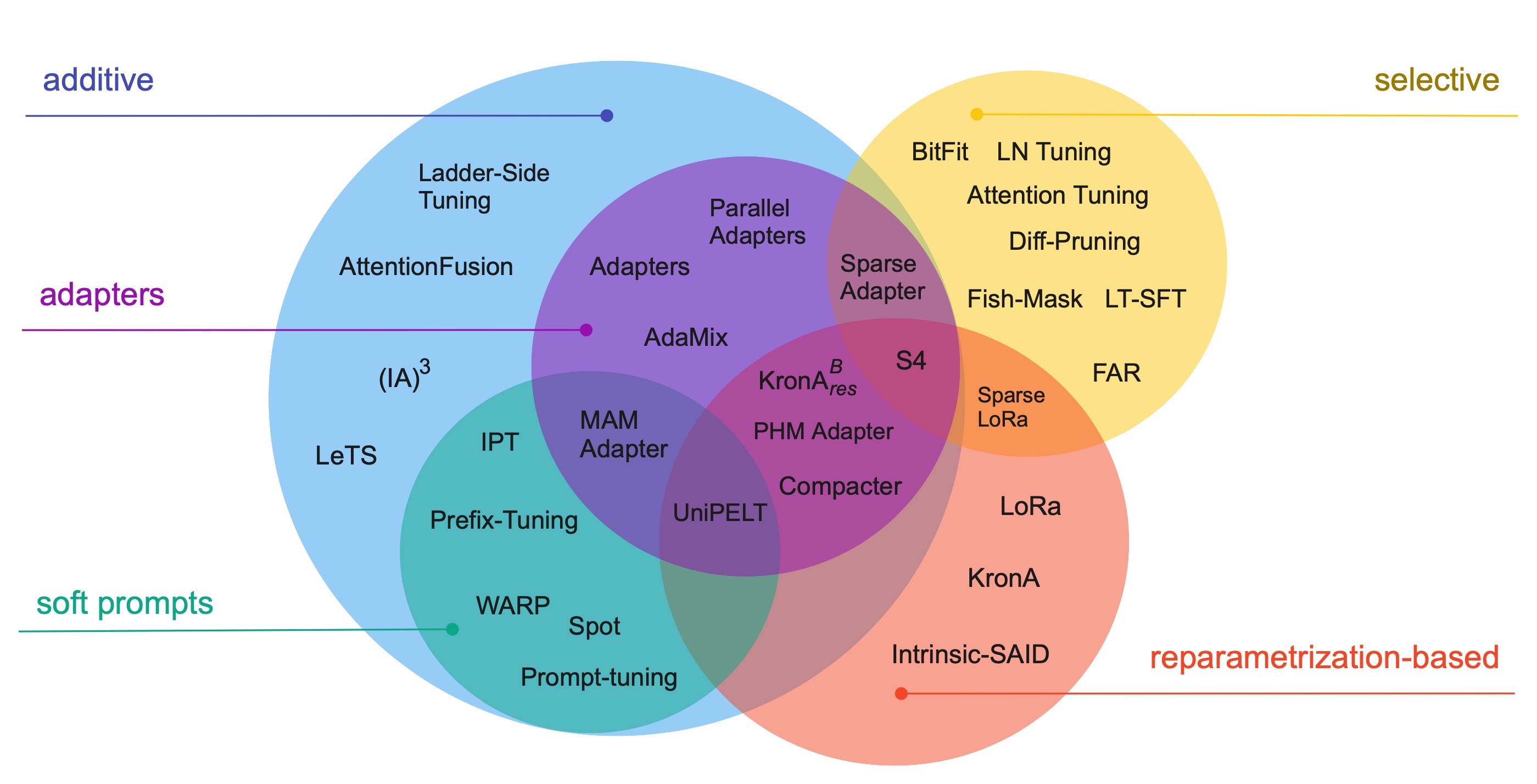}
    \caption{A taxonomy of PEFT methods \citep{lialin2023scaling}}
    \label{fig:peft_tax}
\end{figure*}

PEFT methods can be categorized into three main groups: (1) \textit{addition-based} methods, which introduce task-specific parameter sets between PLM layers \citep{houlsby2019parameter}, (2) \textit{selection-based} methods, which update only a subset of existing PLM parameters, e.g., only biases \citep{ben-zaken-etal-2022-bitfit}, and (3) \textit{reparametrization-based}, which disassemble the layers in PLM blocks into their low-rank representations such as low-rank adaptation (LoRA) \citep{hu2021lora}. Figure \ref{fig:peft_tax} gives the taxonomy of PEFT methods with examples for each of the three main groups. We focus in this thesis on the LoRA method and its upgrades.

\subsection{Low-rank Adaptation} \label{subsec:lora}

\begin{figure*}
    \centering
    \includegraphics[width=0.85\linewidth]{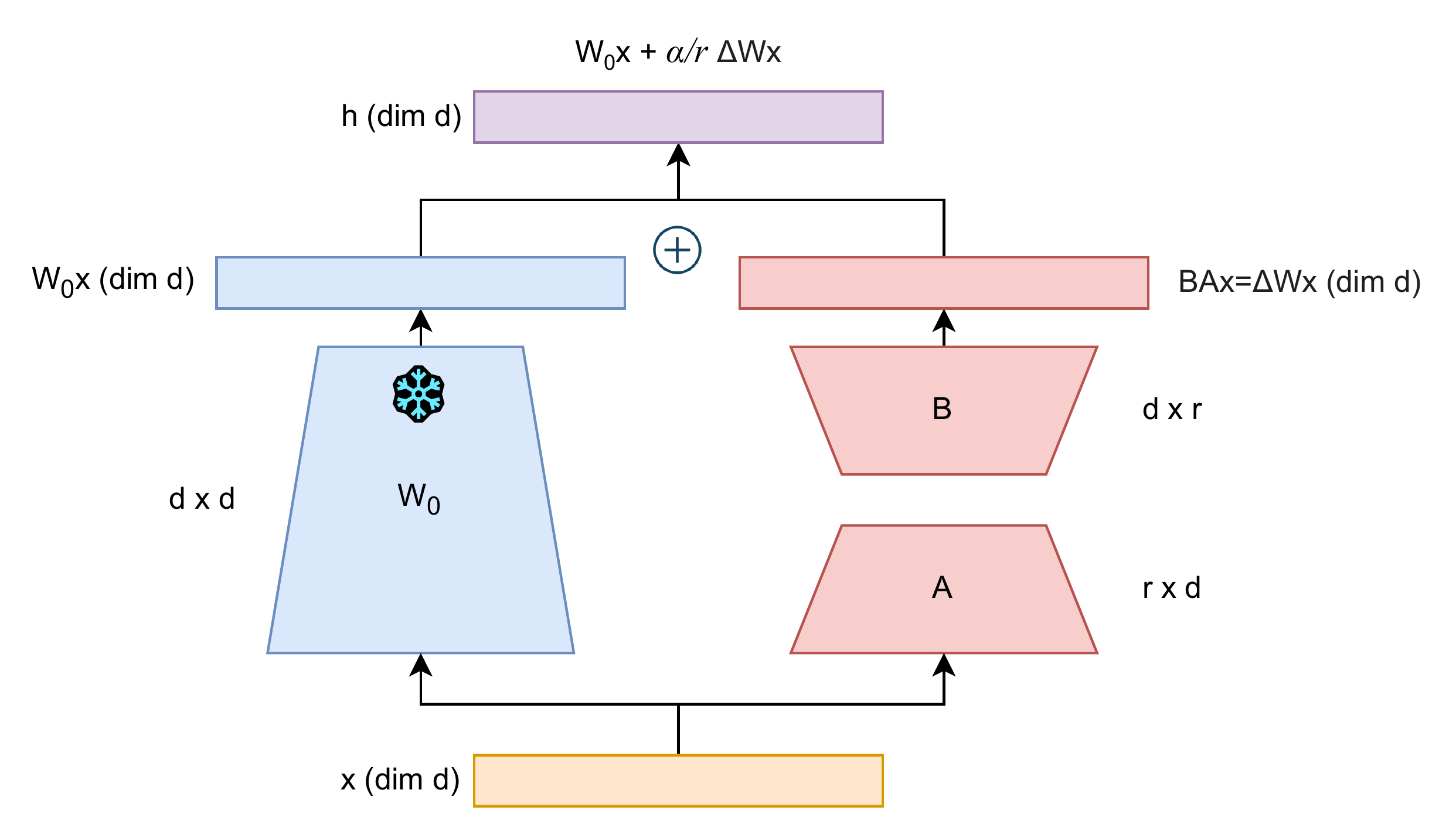}
    \caption[LoRA method visualized]{LoRA method visualized. Pre-trained weights $\mathbf{W_0}$ are frozen, while the low-rank decomposition $\mathbf{B}\mathbf{A}$ is updated during adaptation. For the Transformer architecture, $\mathbf{W_0}$ typically refers to either query, key, value, or output weight matrices, which are decomposed into two matrices with a lower rank. The input and output are vector representations of dimension $d$. The final output is obtained by summing the pre-trained weights with the scaled decomposed weights.}
    \label{fig:lora}
\end{figure*}

LoRA is arguably the most researched low-rank reparametrization method, which has made possible the adaptation of PLMs with billions of parameters. LoRA can be applied to any subset of weights in the neural network \citep{hu2021lora}. The core idea of applying LoRA to the Transformer models is decomposing the query, key, value, and output matrices. Figure \ref{fig:lora} depicts this decomposition. If we denote the pre-trained Transformer weight matrices as $\mathbf{W}_0^{q}, \mathbf{W}_0^{k}, \mathbf{W}_0^{v}, \mathbf{W}_0^{o}  \in \mathbb{R}^{d_{model} \times d_{model}}$, where $d_{model}$ is the input and output dimension size of the Transformer, the low-rank decompositions of these matrices are as follows:

\[
\mathbf{W}_0^{q} + \Delta \mathbf{W}^q = \mathbf{W}_0^{q} + \mathbf{B}^q\mathbf{A}^q, \quad \mathbf{B}^q \in \mathbb{R}^{d \times r}, \quad \mathbf{A}^q \in \mathbb{R}^{r \times d},
\]
\[
\mathbf{W}_0^{k} + \Delta \mathbf{W}^k =\mathbf{W}_0^{k} + \mathbf{B}^k\mathbf{A}^k, \quad \mathbf{B}^k \in \mathbb{R}^{d \times r}, \quad \mathbf{A}^k \in \mathbb{R}^{r \times d},
\]
\[
\mathbf{W}_0^{v} + \Delta \mathbf{W}^v = \mathbf{W}_0^{v} + \mathbf{B}^v\mathbf{A}^v, \quad \mathbf{B}^v \in \mathbb{R}^{d \times r}, \quad \mathbf{A}^v \in \mathbb{R}^{r \times d},
\]
\[
\mathbf{W}_0^{o} + \Delta \mathbf{W}^o = \mathbf{W}_0^{o} + \mathbf{B}^o\mathbf{A}^o, \quad \mathbf{B}^o \in \mathbb{R}^{d \times r}, \quad \mathbf{A}^o \in \mathbb{R}^{r \times d},
\]
\[
r \ll d.
\]

During training, $\mathbf{W}_0$ matrices are frozen, while the decomposed matrices $\mathbf{B}$ and $\mathbf{A}$ are updated. The key is to choose an appropriately low rank $r$ for the decomposition. Choosing a rank that is too small might not be enough for the model to adapt to the new task compared to the total number of the model parameters, while choosing too high a rank might cause the model to overfit to the newly trained task and potentially induce catastrophic forgetting \citep{MCCLOSKEY1989109}. The effect of the trained LoRA weights on the adapted model is also controlled with the scaling parameter $\alpha$ where the idea is to scale $\Delta W$ with $\frac{\alpha}{r}$ and is used for merging LoRA weights with the pre-trained weights $\mathbf{W}_0$. Higher $\alpha$ will accentuate the newly learned weights more. 

PEFT methods made LLM adaptation and experimentation possible on consumer-grade hardware by choosing only a tiny fraction of model parameters to update. However, even with most memory-efficient PEFT methods, the pre-trained model weights still occupy a significant portion of the GPU memory. To reduce the overhead of the PLM weights, the authors of \cite{dettmers2023qlora} introduced a quantized low-rank adaptation (QLoRA) method. QLoRA combines the quantization of pre-trained model weights with the LoRA method to minimize GPU memory usage, thus making the adaptation of larger LLMs possible. It achieves this through several key techniques: 4-bit NormalFloat quantization of the pre-trained weights, double quantization of the quantization constants, and paged optimizers to prevent memory spikes during training. NormalFloat was introduced as a novel information-theoretic optimal quantization data type suitable for normally distributed weights, which achieves better results than the 4-bit integer and 4-bit float data types \cite{dettmers2023qlora}. However, dealing with the NormalFloat data type introduces additional constants to keep in memory. Quantizing these constants using double quantization reduces the memory footprint even more. Finally, to manage the gradient checkpointing \citep{chen2016training} spikes that occur when processing batches with long sequence length, paged optimizers are introduced, which transfer optimizer states between CPU and GPU to prevent out-of-memory errors on the GPU. 

\section{Sequence Labeling} \label{sec:sl}

Effective self-supervised pre-training mechanisms created strong pre-trained neural language models. Combining PLMs with transfer learning paradigms has enabled solving a wide range of NLU tasks through supervised adaptation to labeled data from target tasks and domains. Many NLU tasks involve classification, where labels are assigned to text examples based on a predefined label set. These tasks can be broadly categorized into \textit{text classification} and \textit{sequence labeling}. 

\begin{figure*}
    \centering
    \begin{subfigure}{\textwidth}
        \centering
        \includegraphics[width=\linewidth]{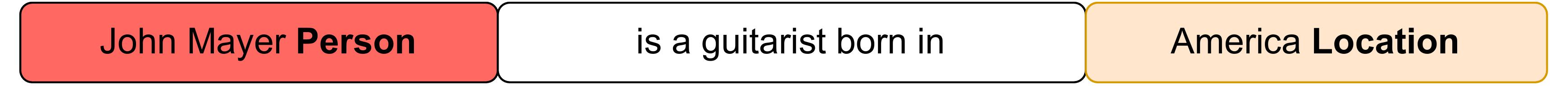}
        \caption{Named entity recognition}
        \label{fig:sl_ner}
    \end{subfigure}
    \vspace{0.01em}
    \\
    \begin{subfigure}{\textwidth}
        \centering
        \includegraphics[width=\linewidth]{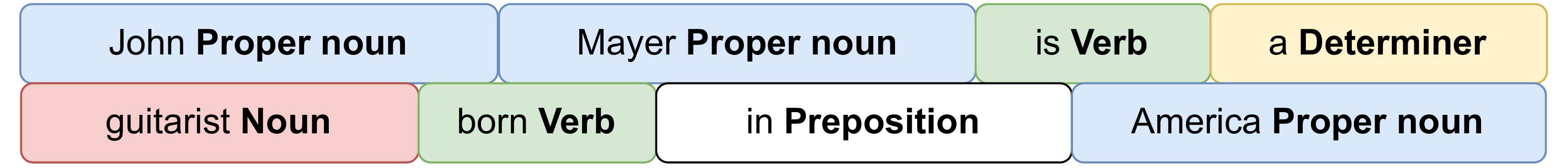}
        \caption{Part-of-speech tagging}
        \label{fig:sl_pos}
    \end{subfigure}
    \vspace{0.01em}
    \\
    \begin{subfigure}{\textwidth}
        \centering
        \includegraphics[width=\linewidth]{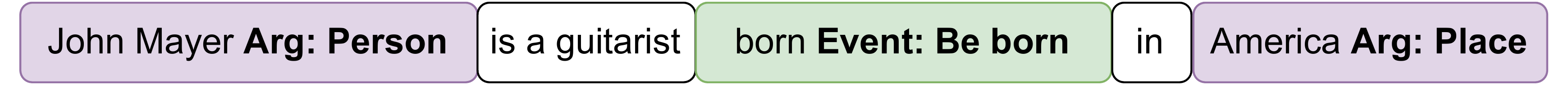}
        \caption{Event extraction}
        \label{fig:sl_ee}
    \end{subfigure}
    \caption[Examples of sequence labeling tasks]{Examples of three sequence labeling tasks---(a) named entity recognition, (b) part-of-speech-tagging, and (c) event extraction---on the sentence \textit{John Mayer is a guitarist born in America}. The labels for each task are shown in bold.} 
    \label{fig:sl_examples}
\end{figure*}

Within the IE pipeline \citep{Piskorski2013}, numerous tasks are naturally framed as sequence labeling, where each token in a text sequence is assigned a label. Prototypical sequence labeling tasks include NER, text chunking, and PoS tagging \citep{tjong-kim-sang-buchholz-2000-introduction, tjong-kim-sang-de-meulder-2003-introduction}. Other tasks, such as event extraction, aspect-based sentiment analysis, slot labeling, aspect-based argument mining, keyphrase extraction, and SRL, are often modeled as sequence labeling tasks. PoS tagging differs from other sequence labeling tasks in that every token is assigned a label---a part-of-speech tag. In contrast, most other sequence labeling tasks allow tokens or spans of tokens to remain unlabeled. For example, in NER, the goal is to identify and classify named entities such as persons, locations, and organizations. However, most token spans in a text do not correspond to any named entity. Figure \ref{fig:sl_examples} gives an example of three sequence labeling tasks on the same sentence. Between NER, PoS tagging, and event extraction, event extraction is the most complex task, as it involves extracting both events and their arguments.

Sequence labeling is a fundamental paradigm in NLP, and many systems rely on accurate outputs of sequence labeling models. For example, knowledge base construction and population typically depends on the quality outputs of sequence labeling models trained for NER, event extraction, and slot labeling. Named entities serve as a base for the named entity linking task, accurate event extraction drives alerting systems, and slot labeling is a foundation for task-oriented dialogue, which is crucial for implementing virtual assistants.  %
Specialized sequence labeling-based IE systems are needed whenever we want to identify and count information elements in the text. This is especially useful for computational social science applications \citep{fischer-etal-2024-concept, dukic2024takelab} and specialized monitoring and alerting systems such as news event monitoring \citep{acled, dukic2024closed}. 

\subsection{Modeling Sequence Labeling} \label{subsec:modeling_sl}

\begin{figure*}
    \centering
    \includegraphics[width=1.0\linewidth]{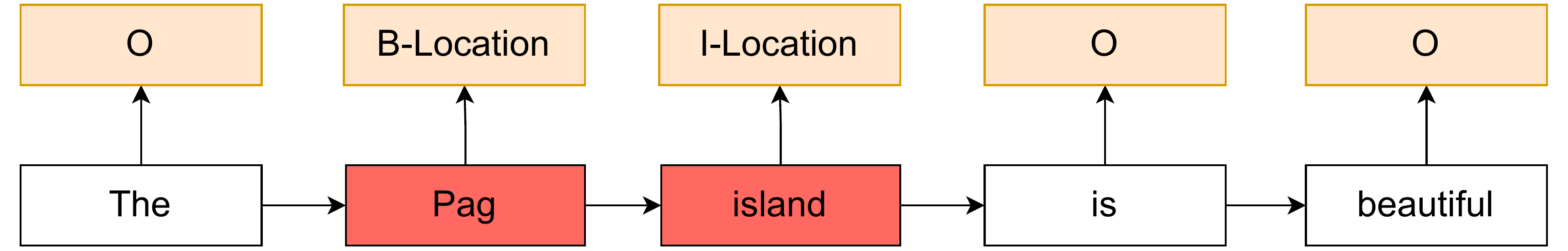}
    \caption[Sequence labeling cast as an IOB2 tagging task for NER]{Sequence labeling cast as an IOB2 tagging task for NER. The task is to detect and classify consecutive spans of tokens, which refer to the named entities in the text. In this example, \textit{Pag island} is a span that needs to be classified as a named entity of class \textit{location}. All other tokens in this example must be assigned the \textit{O} tag, meaning they do not belong to any predefined named entity class.}
    \label{fig:sl}
\end{figure*}

Sequence labeling datasets are typically accompanied by labels following a specific tagging scheme. The BIO (begin, inside, outside) tagging scheme is a standard way of annotating spans of tokens in the text. Since sequence labeling is done on word (token) spans, where each span often contains multiple words, we require a labeling scheme to encode spans composed of tokens as their elements. The labeling scheme enables boundaries to be set between annotated spans and tokens outside of these spans, ensuring consistency between data points and alleviating parsing model outputs since the models are trained to assign BIO tags instead of raw labels. A widely adopted variant of BIO tagging is the IOB2 tagging scheme,\footnote{\footnotesize{Another variant of BIO tagging is the IOB1 tagging scheme. Here, the \textit{B-} prefix is employed to separate two adjacent labeled spans of the same class.}} where the \textit{B-} prefix denotes the start of each annotated span, the \textit{I-} prefix indicates that the token is a part of the annotated span, and \textit{O} denotes that the token does not belong to any annotated span \citep{iob2}. Figure~\ref{fig:sl} shows sequence labeling cast as an IOB2 tagging task for NER. Each \textit{B-} and \textit{I-} prefix is accompanied by a class label. Considering a case of the NER with $N=4$ classes in a label set $L$, the total number of labels a sequence labeling model is trained on equals $|L|=2N + 1$.

Sequence labeling has a long tradition in the field of NLP. The first sequence labeling models were trained on manually designed features 
\citep{he2020survey}. These features were then fed into statistical models \citep{tjong-kim-sang-buchholz-2000-introduction, ahn-2006-stages} such as logistic regression, support vector machine (SVM), hidden Markov model (HMM), maximum entropy Markov model (MEMM), or conditional random fields (CRFs) \citep{lafferty2001conditional}. Then, just before the emergence of Transformer-based PLMs, the primary models were RNNs \citep{ELMAN1990179}, RNNs with the attention mechanism \citep{bahdanau2014neural}, and BiLSTMs \citep{akbik-etal-2018-contextual}. Pre-training mechanisms of Transformer-based models have enabled more effective knowledge transfer to sequence labeling tasks through the transfer learning process. 

Sequence labeling tasks are typically tackled with encoder-only models \citep{devlin-etal-2019-bert, fei-etal-2021-better-new}, pre-trained with a bidirectional language modeling pre-training objective (cf.~Subsection \ref{subsec:enc_dec}). Most NLU tasks benefit from the right-side context, which is omitted in decoder-only variants of the Transformer model. Consequently, sequence labeling tasks usually benefit from the right-side context, which encoders provide out-of-the-box. A typical bidirectional sequence labeling model uses a stack of pre-trained encoders as a backbone with a fully connected linear layer on top (colloquially referred to as a \textit{token classification head}), which projects each token from the input to the label space at the model output (cf.~Figure~\ref{fig:enc}). However, sequence labeling can also be done in an autoregressive way, by generating spans and their labels directly using natural language (cf.~Figure \ref{fig:dec}) \citep{wang2022instructionner, wang2023instructuie}. In this setup, each generated token comes from the PLM's vocabulary, and some form of decoding operation over the logits in the last hidden layer is performed to obtain the output. BIO tag generation for decoders is not enforced since they were trained to generate coherent natural language.

Transferring the pre-trained knowledge to sequence labeling tasks with pre-trained encoders is beneficial due to inherent bidirectionality of the model internals, but encoders suffer from pre-train--fine-tune discrepancy (cf.~Subsection \ref{subsec:enc_dec}), which aggravates the transfer. On the other hand, decoders are not bidirectional by design, meaning that using them as encoders should result in an even more pronounced negative effects of discrepancy. Luckily, they do not exhibit the pre-train--fine-tune discrepancy if used as text generators. This is because they can be fine-tuned generatively using the same CLM loss they leveraged in the pre-training phase. A witness to this phenomenon is the success of SFT methods, which rely on this pre-train--fine-tune alignment \citep{zhang2023instruction}.

Regardless of the underlying model, most sequence labeling tasks can be solved in a \textit{pipeline} or \textit{jointly}. Each sequence labeling task can be split into detection (identification) and classification steps. The detection step identifies spans (typically consecutive sequences of tokens) that carry semantic information, while the classification step classifies the identified spans into predefined classes. The sequence labeling is said to be solved in a pipeline when the detection step precedes the classification step. On the other hand, when both detection and classification are tackled simultaneously, the sequence labeling is said to be solved jointly. Today, most approaches tackle sequence labeling tasks jointly. However, with more complex IE systems, such as the ones built for the event extraction pipeline, it is meaningful to approach modeling the subtasks with sequence labeling models in a pipeline.

\subsection{Evaluating Sequence Labeling} \label{subsec:evaluating_sl}

Since PLMs are equipped with their own tokenizers, which split text into subwords, and sequence labeling datasets contain labels on a word level, training and evaluating sequence labeling models requires a method to align the reference labels and the labels predicted by the model for comparison. The possible approaches are many. However, what is typically done for training the model is either (1) duplicating the predicted label on all subwords of the tokenized word or (2) assigning the label only to the first subword of the tokenized word (i.e., the \textit{head word}), whilst ignoring other subwords of the tokenized word through modifications of the loss function. In both cases, the evaluation is done by aligning the number of predictions with the number of references. Alignment can be done, for example, by taking only the prediction on the \textit{head word} as a predicted label for each word. Evaluation is then performed on aligned predicted label spans with the reference spans. Section \ref{sec:evaluation} will provide details on specific evaluation metrics.\clearpage{}%
\clearpage{}%
\chapter{Tasks, Datasets, and Evaluation} \label{ch:tde}

The previous chapter introduced sequence labeling in NLP, providing examples of common sequence labeling tasks and outlining standard modeling approaches. Before delving into the improvements in the transfer learning process for sequence labeling tasks through interventions in the adaptation phase of pre-trained neural language models, we will first lay a foundation through a detailed discussion on sequence labeling tasks, the datasets used, as well as the evaluation metrics and procedures. These tasks, datasets, evaluation metrics, and procedures will be used for experimental assessment of the proposed improvements. Concretely, the following Section \ref{sec:tasks_datasets} gives an overview of sequence labeling tasks and corresponding datasets. Furthermore, we describe the evaluation procedure for sequence labeling models in Section \ref{sec:evaluation}.

\section{Tasks and Datasets} \label{sec:tasks_datasets}

Any task in NLP that can be cast in a way that a finite set of labels is assigned to consecutive spans of tokens, where each token is assigned only one label and overlap between spans is not allowed, is considered a \textit{standard sequence labeling} task. In other words, if a task can be solved with the BIO tagging scheme, it satisfies the aforementioned criteria. There are also non-standard sequence labeling tasks that allow for overlapping tokens (e.g., variants of the relation extraction task). In this thesis, we consider only tasks with consecutive, non-overlapping spans of tokens, where each span is assigned a single label. 

Training supervised sequence labeling models necessitates manual labeling of each token in the corpus. Such a process is costly, and the number of high-quality datasets for solving sequence labeling tasks is limited. Although there are many sequence labeling tasks in NLP, the high-quality ones with enough training data are scarce. The selection of sequence labeling tasks for experimental assessment described in subsequent chapters is therefore guided by two criteria: (1) widespread adoption in the research community and (2) adequate volume of annotated examples to ensure robust model training and evaluation. Transfer learning methods for sequence labeling proposed in this thesis will be tested on the following eight tasks: event extraction, relation extraction, NER, text chunking, aspect term extraction and polarity (ATE + ATP), slot labeling, aspect-based argument mining (ABAM), and semantic role labeling (SRL). These tasks are diverse and call for a broad range of language understanding capabilities. PoS tagging is excluded from this list as it is different in nature from other sequence labeling tasks, requiring that each token gets assigned a label. Other sequence labeling tasks operate on a span level and typically do not assign a label to every token. Furthermore, syntactic tasks are disqualifying in this context, as they can be solved fairly accurately using rules that conform to the natural language's grammar. Therefore, text chunking is not explored in much detail since it is a syntactic task. In addition, PoS tagging and text chunking were deemed highly saturated tasks already in 2018 \citep{akbik-etal-2018-contextual}, meaning that models trained to solve these tasks achieved near-perfect scores in terms of sequence labeling evaluation metrics. We mostly pick semantic NLP tasks and tasks from the IE pipeline, which require a much deeper understanding of natural language. We tackle all these tasks in a joint sequence labeling manner, detecting and classifying the spans simultaneously using a BIO tagging scheme (cf.~Subsection \ref{subsec:modeling_sl}). Next, we describe all the selected sequence labeling tasks in detail. We provide statistics for all datasets and corresponding sequence labeling tasks in Table \ref{tab:dataset_stats} and Table \ref{tab:dataset_stats2}. 

To characterize sequence labeling datasets, we propose a novel metric for sequence labeling tasks, which we call \emph{right-side dependency relations ratio} (RDRR). For each dataset in Table \ref{tab:dataset_stats2}, we calculate its RDRR score. RDRR is defined as the ratio of right-side dependency relationships to the total number of left-side and right-side relationships counted for all labeled spans in the training set. Essentially, this metric indicates the degree to which labeled spans depend on the context to their right. This is a useful dataset and task characterization metric for comparing encoder and decoder PLMs. Recall that encoder and decoder PLMs differ in causal masking, which is present with decoders and restrains them from utilizing the right-side context. Since most of the sequence labeling tasks by design depend on the left- and right-side context, it is beneficial to characterize how much they depend on the omitted right-side context with the RDRR metric. We use spaCy \citep{spacy} dependency parser to obtain the dependency relations. 

\subsection{Event Extraction} \label{subsec:ee}

The \textit{Automatic content extraction} (ACE) research program \citep{doddington-etal-2004-automatic} put on a vast amount of effort into defining the event terminology in computational linguistics. Based on that terminology, the guidelines were produced that alleviated the event data labeling. The ACE guidelines define an event as a specific occurrence involving participants, as something that happens and can frequently be described as a change of state~\cite{guidelines}.

When extracting events from text data in NLP, the fundamental task is the \textit{event extraction (EE)} task. Solving this task efficiently enables automatic extraction of events from established (newspapers) and less formal (social media posts) web sources on a large scale. The extracted events could then be used for various applications, including but not limited to event coreference resolution, event timeline extraction, crisis and disaster detection, topic detection and tracking, event linking, and event monitoring. With extracted event-related information from a corpus of recently published news, a system can be created that monitors events and reports to the user about novel events in the world in real-time. The extracted events can also be leveraged for many downstream tasks in NLP, including knowledge graph construction \citep{zhang-etal-2021-eventke-event}, information retrieval \citep{glavas-snajder-2013-event}, text summarization \citep{zhang-etal-2023-enhancing}, and aspect-based sentiment analysis \citep{tang-etal-2022-affective}.

The events can be extracted in either closed-domain or open-domain fashion. Closed-domain event extraction (CDEE) requires predefined event types and appropriate argument roles for each event type that need to be inferred from the data. CDEE systems are commonly bound to domain-specific schemes and fill predefined event-specific slots evoked by an event \textit{trigger}---a span of words that evokes a particular type of event. For example, Figure \ref{fig:sl_examples} shows the sentence \textit{John Mayer is a guitarist born in America}. In this sentence, the trigger for the event of type \textit{be born} is the word \textit{born}, indicating that the person came into existence. A typical domain-specific CDEE workflow starts with event trigger detection (ETD), which locates the trigger span in the text, and event trigger classification (ETC) \citep{xiang2019survey}, which assigns one of the predefined event types to the trigger span. With triggers identified, the next step is to detect and classify the corresponding arguments, e.g., participants, location, and time (cf.~Figure \ref{fig:sl_examples}), performed by \textit{event argument detection (EAD)} and \textit{event argument classification (EAC)} components of the pipeline. This thesis puts focus on the beginning of the CDEE pipeline, i.e., the ETD and ETC tasks. For ETD, we rely on the MAVEN dataset \citep{wang-etal-2020-maven}, ACE05 dataset \citep{doddington-etal-2004-automatic}, EDNYT \cite{maisonnave2022detecting}, and EVEXTRA dataset \citep{glavas_snajder_2015}. For ETC, we leverage the ACE05 dataset. We give the statistics for the ETD datasets in Table~\ref{tab:dataset_stats} and for the ETC dataset in Table~\ref{tab:dataset_stats2}.

\begin{table}
  \centering
  \adjustbox{width=0.9\columnwidth}{
  \begin{tabular}{@{}lrrrrrrrrr@{}}
    \toprule
    \multicolumn{1}{c}{\multirow{2}{*}{\textbf{Dataset}}} & \multicolumn{3}{c}{\textbf{Train}} & \multicolumn{3}{c}{\textbf{Validation}} & \multicolumn{3}{c}{\textbf{Test}} \\
    \cmidrule(lr){2-4} \cmidrule(lr){5-7} \cmidrule(lr){8-10}
    {} & \multicolumn{1}{c}{\#Sent} & \multicolumn{1}{c}{\#Tr} & \multicolumn{1}{c}{\#Re} & \multicolumn{1}{c}{\#Sent} & \multicolumn{1}{c}{\#Tr} & \multicolumn{1}{c}{\#Re} & \multicolumn{1}{c}{\#Sent} & \multicolumn{1}{c}{\#Tr} & \multicolumn{1}{c}{\#Re} \\
    \midrule
    MAVEN     & \num{25944}  & \num{24063} & \num{15590}  & \num{6487} & \num{6038} & \num{3940}  & \num{8042} & \num{7469} & \num{4805}    \\
    ACE05  & \num{14672} & \num{3256} & \num{7403} & 873 & 340 & 446 & 711 & 292 & 412     \\
    EDNYT     & \num{1842}  & \num{1500} & \num{1164} & 95 & 74 & 65 & 198 & 155 & 115     \\
    EVEXTRA   & \num{8534} & \num{7056} & \num{5461} & \num{1103} & 902 & 700 & \num{2482} & \num{2077} & \num{1590}  \\     
    \bottomrule
  \end{tabular}}
  \caption[Statistics for the four datasets for ETD and their splits]{Statistics for the four datasets for ETD and their splits. We show the number of sentences (\#Sent), the number of sentences with triggers (\#Tr), and the number of relations after post-processing of MinIE triple extractions (\#Re).}
  \label{tab:dataset_stats}
\end{table}

\begin{table}
  \centering
  \adjustbox{width=0.9\columnwidth}{
  \begin{tabular}{@{}crrrrrr@{}}
    \toprule
    \multicolumn{1}{c}{\multirow{1}{*}{\textbf{Dataset (Task)}}} & \multicolumn{1}{c}{\textbf{Train}} & \multicolumn{1}{r}{\textbf{Validation}} & \multicolumn{1}{r}{\textbf{Test}} & \multicolumn{1}{r}{\textbf{Total}} & \multicolumn{1}{r}{\textbf{\#Classes}} & \multicolumn{1}{r}{\textbf{RDRR}} \\
    \midrule
    OntoNotes v5.0 (SRL) & \num{21244} & \num{5385} & \num{6000} & \num{32629} & 27 & 0.50 \\  
    CoNLL03 (NER) & \num{14041}  & \num{3250} & \num{3453} & \num{20744} & 4 & 0.59 \\
    CoNLL03 (Chunking) & \num{14041}  & \num{3250} & \num{3453} & \num{20744} & 11 & 0.52 \\
    ACE05 (ETC) & \num{14672} & 873 & 711 & \num{16256} & 33 & 0.41 \\
    Rest14 (ATE+ATP) & \num{2737} & 304 & 800 & \num{3841} & 4 & 0.40 \\
    NLU++ (Slot Labeling) & \num{2152} & 309 & 619 & \num{3080} & 17 & 0.37 \\  
    AAC-MW (ABAM) & 670 & 95 & 193 & 958 & 12 & 0.47 \\  
    \bottomrule
  \end{tabular}}
  \caption[Statistics for the remaining datasets (tasks) and their splits]{Statistics for the remaining datasets (tasks) and their splits. We show the number of sentences per split, the total number of sentences, the total number of classes, and the right-side dependency relations ratio (RDRR). OntoNotes v5.0 was subsampled, while remaining datasets were used with the original number of examples.}
  \label{tab:dataset_stats2}
\end{table}

\subsubsection{MAVEN}

The MAssive eVENt detection dataset (MAVEN) \citep{wang-etal-2020-maven} from the English Wikipedia domain is the largest freely available dataset suitable for ETD. It covers more than $150$ events, ranging from violent events such as \textit{Attack}, \textit{Hostile encounter}, or \textit{Kidnapping} to everyday events such as \textit{Traveling} or \textit{Deciding}. MAVEN comes with tokenized sentences and a predefined train, validation, and test split. However, since no gold test set labels were published, we use the official validation set as a test set (only to measure the source model performance on it) and randomly sample $20\%$ of sentences from the training data as a new validation set. Here, we provide examples from the MAVEN dataset with annotated event triggers in bold:

\begin{exampleblock}{MAVEN dataset examples}
\begin{enumerate}
    \item The aircraft \textbf{broke} [TRIGGER] into two parts, but there was no fire.
    \item The naval \textbf{Battle} of Salis \textbf{took place} [TRIGGER] during the Polish–Swedish \textbf{War} [TRIGGER] (1600--1611) during the night of March 23–24, 1609.
\end{enumerate}
\end{exampleblock}

\subsubsection{ACE05}

The Automatic Content Extraction 2005 (ACE05) \citep{doddington-etal-2004-automatic} is a widely used event detection dataset consisting predominantly of articles from various news sources in multiple languages. We use only the English train, validation, and test split, obtained with the standard ACE pre-processing tool,\footnote{\footnotesize\url{https://github.com/nlpcl-lab/ace2005-preprocessing}} which we also use to obtain sentences and tokens, and create IOB2 tags. Although ACE is a sizable dataset, as noted by \cite{wang-etal-2020-maven}, many ACE sentences do not contain any triggers (cf.~Table~\ref{tab:dataset_stats}). This dataset has $33$ event classes. Here, we provide examples from the ACE05 dataset with annotated event triggers in bold and their classes in the brackets:

\begin{exampleblock}{ACE05 dataset examples}
\begin{enumerate}
    \item India's has been reeling under a heatwave since mid-May which has \textbf{killed} [DIE] 1,403 people.
    \item Orders went out today to \textbf{deploy} [TRANSPORT] 17,000 U.S. Army soldiers in the Persian Gulf region.
\end{enumerate}
\end{exampleblock}

\subsubsection{EDNYT}

The event detection New York Times dataset (EDNYT) \citep{maisonnave2022detecting} was compiled from the New York Times articles on financial crises, which makes the dataset more topically focused than the other datasets. The dataset was not tokenized, but it came with a train-test split, with the test set comprising $10\%$ of the data. We obtain a validation set by randomly sampling $5\%$ of the train data. We use spaCy \citep{spacy} to tokenize the sentences. We discarded $3\%$ of sentences with trigger spans that could not be aligned with spaCy tokenization. Here, we provide examples from the EDNYT dataset with annotated triggers in bold:

\begin{exampleblock}{EDNYT dataset examples}
\begin{enumerate}
    \item Mr. James \textbf{said} [TRIGGER] the owner also \textbf{wanted} [TRIGGER] to get rid of Mr. Cohen because he \textbf{disliked} [TRIGGER] the coverage Mr. Cohen had directed of the U.P.I.'s long financial crisis in 1984 and 1985.
    \item But P.R.I. officials now \textbf{say} [TRIGGER] that goal too has been \textbf{jeopardized} [TRIGGER] by the \textbf{worsening} [TRIGGER] of the economic situation.
\end{enumerate}
\end{exampleblock}

\subsubsection{EVEXTRA}

The event extraction dataset (EVEXTRA) \citep{glavas_snajder_2015} is an English newspaper corpus annotated with event triggers. It was created by labeling English articles collected with EMM NewsBrief, a service for collecting news stories.\footnote{\footnotesize\url{https://emm.newsbrief.eu/NewsBrief}} The dataset comes tokenized but with no predefined split. We randomly assign sentences to train, validation, and test sets in a 70/10/20 ratio, respectively, ensuring that sentences from the same article end up in the same set. Less than $1\%$ of sentences were dropped because aligning the trigger annotations with tokens was impossible. Here, we provide examples from the EVEXTRA dataset with annotated event triggers in bold:

\begin{exampleblock}{EVEXTRA dataset examples}
\begin{enumerate}
    \item The \textbf{Battle} [TRIGGER] of the 300 Champions was a \textbf{battle} [TRIGGER] \textbf{fought} [TRIGGER] in roughly 545 BC between Argos and Sparta. 
    \item Opposition parties and international observers \textbf{said} [TRIGGER] the \textbf{vote} [TRIGGER] was \textbf{marred} [TRIGGER] by \textbf{vote-rigging} [TRIGGER], including alleged ballot-box \textbf{stuffing} [TRIGGER] and false voter rolls.
\end{enumerate}
\end{exampleblock}

\subsection{Relation Extraction} \label{subsec:re}

Here were consider relation extraction as a subtask of Open Information Extraction (OIE). OIE systems are intended to facilitate various downstream tasks, including text summarization \cite{fan-etal-2019-using,ribeiro-etal-2022-factgraph}, question answering \cite{yan2018assertion, nagumothu-etal-2022-pie}, incomplete sentence reconstruction \citep{montella-etal-2020-denoising}, and event extraction \citep{chen-etal-2023-led}. Many event-related tasks, such as event schema induction \citep{balasubramanian-etal-2013-generating} and cross-domain event coreference \citep{pratapa-etal-2021-cross}, benefit from OIE. OIE systems \citep{banko2007open} automatically extract domain-agnostic subject-relation-object triples because they discover relations not predefined by any scheme \citep{biomedoie, sun-etal-2018-logician, gashteovski2019opiec}. In contrast, closed-domain information extraction systems rely on a predefined, fixed scheme. Therefore, OIE systems are not limited by domain-specific constraints and can be applied to a wider variety of texts and domains. This makes OIE systems more universal and broadly applicable than closed-domain information extraction systems.

Although most recent OIE systems are neural models trained in a supervised manner \citep{kolluru-etal-2020-openie6,kotnis-etal-2022-milie}, traditional OIE systems such as Stanford OIE \citep{angeli-etal-2015-leveraging} and MinIE \citep{gashteovski-etal-2017-minie} are rule-based and typically do not require domain-specific pre-processing of the input text \citep{lauscher2019minscie}. Here, we provide examples of MinIE subject-relation-object triple extraction applied to sentences from the MAVEN dataset (extractions in bold and classes in brackets):

\begin{exampleblock}{MinIE applied to MAVEN examples}
\begin{enumerate}
    \item The \textbf{aircraft} [SUBJECT] \textbf{broke into} [RELATION] \textbf{two parts} [OBJECT], but there was no fire.
    \item \textbf{Eight matches} [SUBJECT] \textbf{were contested at} [RELATION] the \textbf{event} [OBJECT].
\end{enumerate}
\end{exampleblock}

We formulate relation extraction in this context as a sequence labeling task where the models are trained for the assignment of OIE relation spans to an unlabeled corpus. In the sentence \textit{The aircraft broke into two parts, but there was no fire}, we want to label the span \textit{broke into} as a relation. Previously discussed ETD datasets are used as a source and are labeled using the rule-based OIE system MinIE. MinIE has proven useful for many downstream tasks by the BenchIE benchmark and evaluation framework \citep{gashteovski-etal-2022-benchie}.

Since MinIE extracts triples and introduces minor extraction errors, we use a set of heuristics to post-process the results and align them with the requirements of sequence labeling tasks. First, we remove implicit\footnote{\footnotesize{OIE systems often incorporate binding tokens (like the copula \textit{is}), which do not have to be present in the text.}} triple extractions and discard all non-consecutive subject, relation, or object extractions. Further, we remove non-triples, relations with more than five tokens (long OIE relations), and extractions not in the subject-relation-object order. Implicit relations cannot be used for sequence labeling since they introduce tokens not present in the text. Furthermore, it has been shown that long OIE relations are noisy \citep{broscheit-etal-2020-predict-new}. If the OIE system is presented with the sentence \textit{President Biden right now stands really worried about future economic growth.}~it might extract (1) the implicit triple (\textit{Biden}; \textit{is}; \textit{President}) and (2) the triple with long OIE relation (\textit{President Biden}; \textit{right now stands really worried about}; \textit{future economic growth}). Our heuristics would drop both extractions, and the implicit extraction would also be filtered out on account of not being in the order subject-relation-object in the input sentence. Moreover, we filter out all extractions that are incomplete triples, i.e., are missing either subject, relation, or object. If, after that, there are still multiple relation extractions for the same sentence, we try to merge the remaining relations. The merging process is designed to keep all the relations if the tokens are not shared between them. In the case of shared tokens, we keep only the relation extraction with the highest number of tokens that make up the relation. Finally, subject and object extractions and duplicates are dropped, only the relations are kept, and if our heuristics filter out all the relation extractions for the sentence, we do not discard it but consider it a sentence without relations and use it for training as an example with all \textit{O} token labels based on the IOB2 tagging scheme. We apply the OIE system, and the post-processing as described above, to each split of the source and target datasets. Table \ref{tab:dataset_stats} shows the final number of sentences containing relations in the post-processed datasets.

\subsection{Named Entity Recognition} 

Named entity recognition (NER) is the task of identifying and classifying named entities in the text \citep{9039685,nadeau2007survey}. It is the most popular sequence labeling task in the community and is often used to benchmark PLM performance in sequence labeling. Among sequence labeling datasets, those with the most extensive annotations are NER datasets, such as CoNLL03 \citep{tjong-kim-sang-de-meulder-2003-introduction} and OntoNotes \citep{pradhan-etal-2013-towards}. Additionally, datasets primarily designed for other tasks often include NER labels, such as ACE05 \citep{doddington-etal-2004-automatic}. 

Beyond its role as a standalone task, NER also serves as a basis for more complex NLP applications, including entity linking, coreference resolution, and knowledge base population. For our experiments, we choose CoNLL03 as the NER dataset. We use the version available in \textit{Hugging Face Datasets} \citep{lhoest-etal-2021-datasets}, which follows the IOB2 sequence tagging scheme and provides predefined train, validation, and test splits. The dataset has four classes of named entities: \textit{person}, \textit{organization}, \textit{location}, and \textit{miscellaneous}. Table \ref{tab:dataset_stats2} gives the statistics for the CoNLL03 dataset and NER task. Here, we provide examples from the CoNLL03 dataset with annotated named entity spans in bold and classes in brackets:

\begin{exampleblock}{CoNLL03 dataset examples}
\begin{enumerate}
    \item A \textbf{Florida} [LOCATION] restaurant paid 10,925 pounds (\$16,935) for the draft of ``\textbf{Ain't no telling}'' [MISCELLANEOUS], which \textbf{Hendrix} [PERSON] penned on a piece of \textbf{London} [LOCATION] hotel stationery in late 1966.
    \item \textbf{Germany} [LOCATION] imported 47,600 sheep from \textbf{Britain} [LOCATION] last year, nearly half of total imports.
\end{enumerate}
\end{exampleblock}

\subsection{Text Chunking} 

Text chunking is a syntactic NLP task where chunks of text are identified as constituents of the sentence and then linked to meta structures on a grammatical level. For example, chunking identifies noun phrases in the text. Text chunking is useful for preparing the stage for more sophisticated parsing methods \citep{tjong-kim-sang-buchholz-2000-introduction}, such as dependency parsing and constituency parsing. Text chunking is also known as shallow parsing since it does not perform the complete grammatical analysis of the sentence, unlike more sophisticated parsing methods, which often construct a detailed parse tree. This task was chosen to illustrate the adaptation performance of the PLMs on a syntactic sequence labeling task. For the experiments, we choose CoNLL03 as the chunking dataset, leveraging IOB2 annotations and $11$ classes of chunk tags, with predefined train, validation, and test splits. Table \ref{tab:dataset_stats2} gives the statistics for the CoNLL03 dataset and text chunking task. Here, we provide examples from the CoNLL03 dataset with annotated chunking spans in bold and classes in brackets:

\begin{exampleblock}{CoNLL03 dataset examples}
\begin{enumerate}
    \item \textbf{The European Commission} [NOUN PHRASE] \textbf{said} [VERB PHRASE] \textbf{on} [PREPOSITIONAL PHRASE] \textbf{Thursday} [NOUN PHRASE] \textbf{it} [NOUN PHRASE] \textbf{disagreed} [VERB PHRASE] \textbf{with} [PREPOSITIONAL PHRASE] \textbf{German advice} [NOUN PHRASE] \textbf{to} [PREPOSITIONAL PHRASE] \textbf{consumers} [NOUN PHRASE] \textbf{to shun} [VERB PHRASE] \textbf{British lamb} [NOUN PHRASE] \textbf{until} [SUBORDINATED CLAUSE] \textbf{scientists} [NOUN PHRASE] \textbf{determine} [VERB PHRASE] \textbf{whether} [SUBORDINATED CLAUSE] \textbf{mad cow disease} [NOUN PHRASE] \textbf{can be transmitted} [VERB PHRASE] \textbf{to} [PREPOSITIONAL PHRASE] \textbf{sheep} [NOUN PHRASE] . 
    \item \textbf{Tap sales} [NOUN PHRASE] \textbf{began} [VERB PHRASE] \textbf{on} [PREPOSITIONAL PHRASE] \textbf{Monday} [NOUN PHRASE] and \textbf{are being held} [VERB PHRASE] \textbf{daily} [ADVERB PHRASE] \textbf{from} [PREPOSITIONAL PHRASE] \textbf{07.00 GMT} [NOUN PHRASE] \textbf{to} [PREPOSITIONAL PHRASE] \textbf{15.00 GMT} [NOUN PHRASE] \textbf{until} [PREPOSITIONAL PHRASE] \textbf{further notice} [NOUN PHRASE] .
\end{enumerate}
\end{exampleblock}

\subsection{Aspect Term Extraction and Polarity}

Aspect-based sentiment analysis is the \textit{subjective} (evaluative language) task of automatically evaluating people's sentiment towards an aspect in the text, e.g., an entity or a particular aspect of an entity \citep{9996141}. Aspect-based sentiment analysis consists of the following subtasks: \textit{aspect term extraction (ATE)}, \textit{aspect term polarity (ATP)}, \textit{aspect category detection}, and \textit{aspect category polarity.} This task and its subtasks are useful for obtaining a fine-grained analysis of people's sentiments toward various aspects of the text. This can then be used to assess better a sentiment towards a particular entity or its aspect. The ability to automatically obtain this information is valuable in computational social science, where we can analyze how sentiment is expressed toward specific targets, e.g., politicians portrayed in news article titles \citep{baric-etal-2023-target}. Furthermore, aspect-based sentiment analysis is also valuable in other application areas, such as e-commerce, customer service, hospitality, and market research.

We focus on the first two subtasks of aspect-based sentiment analysis and merge them into one sequence labeling task (ATE+ATP). ATE is similar in a way to ETD (cf.~Subsection \ref{subsec:ee}) since its goal is to detect spans of aspect terms, while the connection can also be made between ATP and ETC where ATP aims to classify the aspect terms into classes of polarity (e.g., \textit{positive}, \textit{neutral}, \textit{negative}, and \textit{conflict}). We use the data from SemEval-2014 Task 4 and the restaurants domain (Rest14) \citep{pontiki-etal-2014-semeval}. We tokenize the dataset with spaCy and match given character spans of aspect terms with token spans to obtain IOB2 tags (we discard $13$ aspect terms that could not be matched in this way). Training and test split were predefined. Following prior work \citep{wang-etal-2021-automated}, we randomly sample $10\%$ of the sentences for the validation set. Table \ref{tab:dataset_stats2} gives the statistics for the Rest14 dataset. Here, we provide examples from the Rest14 dataset with annotated aspect terms in bold and their polarity in brackets:

\begin{exampleblock}{Rest14 dataset examples}
\begin{enumerate}
    \item It may be a bit \textbf{packed} [NEGATIVE] on weekends, but the \textbf{vibe} [POSITIVE] is good, and it is the best \textbf{French food} [POSITIVE] you will find in the area.
    \item \textbf{Food} [CONFLICT] is usually very good, though occasionally I wondered about freshness of \textbf{raw vegetables} [NEGATIVE] in side orders.
\end{enumerate}
\end{exampleblock}

\subsection{Aspect-based Argument Mining}

Aspect-based argument mining (ABAM) is the \textit{subjective} (evaluative language) task of automatically evaluating people's opinions towards an argument aspect in the text. It is a rather novel task as it was introduced in 2020 \citep{trautmann-2020-aspect}. The task boils down to automatic detection and classification of argument aspects in a text. Argument aspects refer to specific components of argumentative text that correspond to predefined categories of arguments \citep{ruckdeschel-wiedemann-2022-boundary}. Detected and classified arguments can then be leveraged for downstream applications such as argument ranking, argument summarization, and counter-argument extraction on the aspect level, enabling a deeper understanding of natural language arguments \citep{trautmann-2020-aspect}.

For this task, we leverage the Argument Aspect Corpus (AAC) with token-level annotations on four topics: \textit{minimum wage (AAC-MW)}, \textit{nuclear energy}, \textit{marijuana legalization}, and \textit{abortion}. For the experiments, the AAC-MW topic was picked, which contains $12$ classes and additional \textit{Other} class, which was dropped. Data splits were created using the code provided by the dataset authors.\footnote{\footnotesize{\url{https://github.com/Leibniz-HBI/argument-aspect-corpus-v1/blob/main/classification.py}}} The dataset was converted to IOB2 tags. Table \ref{tab:dataset_stats2} gives the statistics for the AAC-MW dataset. Here, we provide examples from the AAC-MW dataset with annotated argument aspect spans in bold and argument aspect types in brackets:

\begin{exampleblock}{AAC-MW dataset examples}
\begin{enumerate}
    \item Increasing the minimum wage would increase \textbf{worker productivity} [MOTIVATION/CHANCES] and \textbf{reduce employee turnover} [TURNOVER].
    \item If we raise the minimum wage to \$9, \textbf{is it going to make McDonald's cashiers super rich} [SOCIAL JUSTICE/INJUSTICE]?
\end{enumerate}
\end{exampleblock}

\subsection{Slot Labeling} 

Task-oriented dialogue systems typically rely on the NLU module, which extracts information from the user's prompts to the system. A critical component of this module is the \textit{slot labeling} part, which aims to detect and classify the slot values from user's prompts \citep{gupta-etal-2019-simple}. The classified slots can then be used for the improvement of many downstream applications powered by task-oriented dialogue systems in the entertainment and hospitality industry, such as customer support virtual assistants, travel and accommodation booking systems, and healthcare virtual assistants \citep{razumovskaia-etal-2022-natural}. 

The slot labeling task is typically framed as a sequence labeling task. For this task, we leverage the NLU++ dataset  \citep{casanueva-etal-2022-nlu} with the \textit{banking} and \textit{hotels} domains merged, counting a total of $17$ classes. We merge the domains to create a larger dataset for more reliable training and evaluation. We create IOB2 tags for this dataset by mapping slot value offsets in the sequences with spaCy tokenization (all except six out of \num{3080} sentences were perfectly aligned with the provided spans). The recommended evaluation approach by the authors was to use k-fold cross-validation. However, since this approach is resource-intensive in the context of LLM adaptation, we shuffle and randomly split the folds from two domains in a train/validation/test ratio of 70/10/20 to have fixed sets for performance estimation of the trained models. Table \ref{tab:dataset_stats2} gives the statistics for the NLU++ dataset. Here, we provide examples from the NLU++ dataset with annotated slot spans in bold and slot types in brackets:

\begin{exampleblock}{NLU++ dataset examples}
\begin{enumerate}
    \item From the \textbf{16th} [DATE FROM] to the \textbf{30th} [DATE TO], \textbf{a single room} [ROOMS] for \textbf{2} [ADULTS] adults and \textbf{1} [KIDS] child.
    \item It was a contactless payment of \textbf{53 dollars} [AMOUNT OF MONEY] for \textbf{Blues Kitchen} [COMPANY NAME] subscription.
\end{enumerate}
\end{exampleblock}

\subsection{Semantic Role Labeling}

Semantic role labeling (SRL) is a task of detecting argument spans conditioned on a predicate or a verb and classifying the arguments of a predicate into semantic roles \citep{jm3}. It is a standard sequence labeling task in NLP and resembles the closed-domain \textit{slot-filling} paradigm of event extraction but on a more general level. SRL is also known as \textit{shallow semantic parsing}. It goes beyond syntactic parsing by classifying the semantic roles of argument spans, but it does not reach the complexity of full semantic parsing. SRL outputs can be used to improve any task that relies on predicate-argument structures in the text data, such as event extraction subtasks, question answering, and text summarization.

The key difference between SRL and other introduced sequence labeling tasks is that SRL can assign different labels to the same sequence depending on the predicate in question. If there are no predicates, there are no semantic roles to be assigned. Moreover, SRL dictates that different argument spans for a predicate invoke non-overlapping frames, making it a natural and standard sequence labeling task \citep{mehta-etal-2024-promptly}. 

For this task, we use the OntoNotes v5.0 dataset with the English v12 subset \citep{pradhan-etal-2013-towards}. Compared to the other sequence labeling datasets presented, this one stands out as the largest in terms of data. It also provides predefined train, validation, and test splits. However, due to the large size of the training set, we sampled instances from the training and validation sets to facilitate experiments with multiple models. Examples containing semantic roles that appeared less than \num{1000} times in the original 200k-sentence train corpus were removed, reducing the total number of labels from $67$ to $27$. For the training and validation sets, sentences with less than three tokens or more than $50$ tokens were removed. Sampling was then performed from two subsets: sentences with a single predicate and those with multiple predicates. Duplicates based on tokens and BIO tags were dropped. Finally, sentences without predicates were also discarded. The test set, on the other hand, was primarily filtered by excluding the examples with labels filtered from training and validation. Also, sentences without predicates and duplicate sentences were dropped. After pre-processing, we randomly sample \num{6000} test set sentences out of \num{26355} with a fixed seed for efficiency reasons. Table \ref{tab:dataset_stats2} gives the statistics for the OntoNotes v5.0 dataset. Here, we provide examples from the OntoNotes v5.0 dataset with annotated argument spans of the given verbs (1) \textit{establishing} and (2) \textit{is} in bold and their semantic roles in brackets:

\begin{exampleblock}{OntoNotes v5.0 dataset examples}
\begin{enumerate}
    \item Like \textbf{other foreign banks} [ARG0] \textbf{establishing} [VERB] \textbf{a presence} [ARG1] \textbf{here} [ARGM-LOC], the family describes its move as a calculated decision to set up a financial services outlet in Europe's largest economy ahead of the integration of European Community markets after 1992.
    \item \textbf{Development and Reform Commission} [ARG1] \textbf{is} [VERB] \textbf{the organization in charge of prices} [ARG2] .
\end{enumerate}
\end{exampleblock}

\section{Evaluation of Sequence Labeling Models} \label{sec:evaluation}

\subsection{Evaluation Metrics for Sequence Labeling}

Sequence labeling models are typically evaluated using standard precision (P), recall (R), and F1 score metrics. These are classification metrics calculated using three types of predictions: true positives, false positives, and false negatives. Precision is calculated as the ratio of true positives in all instances predicted as positive (true positives and false positives). Recall is calculated as the ratio of true positives in all instances that were supposed to be identified as positive (true positives that were identified and the false negatives that should have been identified but were not). The F1 score is the harmonic mean of precision and recall, resulting in an equal contribution of both metrics to the final F1 score. 

The naive evaluation of sequence labeling would involve computing contingency matrices over BIO-tagged tokens. However, this would fail to capture the inherent connections between consecutive sequences of tokens---spans. Instead, we focus on spans as atomic evaluation elements, thereby properly respecting the sequential nature of sequence labeling tasks. The classification metrics are calculated over assigned BIO tags by extracting the predicted spans and their classes (cf.~Subsection \ref{subsec:modeling_sl}). However, since in most sequence labeling tasks, most tokens have the \textit{O} tags assigned, these \textit{outside} labels are not included in the metrics if they are correctly labeled, but only if they are assigned to tokens that constitute the reference span, i.e., in the case of incorrect predictions.~\textit{O} tags are assigned to individual tokens and are another reason to avoid token-level evaluation. Since models trained for sequence labeling typically detect and classify consecutive spans of tokens in the text, token-level evaluation would amplify the effect of correct \textit{O} predictions. In that case, the model that predicts most tokens as \textit{O} tokens would be considered a strong one. 

Each sequence labeling task consists of a span detection (SD) and a span classification (SC) subtask. As introduced in Section \ref{sec:sl}, sequence labeling tasks are solved either in a pipeline, where SD precedes SC, or jointly by detecting and classifying spans at the same time. Furthermore, both SD and SC have their \textit{strict} and \textit{partial} evaluation setups. 

The sequence labeling models for SD assign \textit{B}, \textit{I}, and \textit{O} tags to tokens, which are then converted to predicted spans and their left and right boundaries. Table \ref{tab:sequence labeling_eval_notation} details the notation used for various types of spans and their boundaries. For example, let $PS$ and $RS$ be the predicted and reference span, respectively, denoting predicted and expected sequences of BIO tags. Furthermore, let $PS_L, RS_L$, and $PS_R, RS_R$ be indices of left and right boundaries of predicted and reference spans, respectively. Then, simplified \textit{strict} evaluation of sequence labeling models trained for span detection (SD) is defined as follows:

\begin{equation}
P^{\mathit{SD}}=\frac{\sum_i 1\left\{\mathit{PS^i}=\mathit{RS^i} \wedge \mathit{PS}_{L}^i=\mathit{RS}_{L}^i \wedge \mathit{PS}_{R}^i=\mathit{RS}_{R}^i\right\}}{N_{\mathit{PS}}}
\end{equation}

\begin{equation}
R^{\mathit{SD}}=\frac{\sum_i 1\left\{\mathit{PS}^i=\mathit{RS}^i \wedge \mathit{PS}_{L}^i=\mathit{RS}_{L}^i \wedge \mathit{PS}_{R}^i=\mathit{RS}_{R}^i\right\}}{N_{\mathit{RS}}}
\end{equation}

\begin{equation}
{F_1}^{\mathit{SD}}=\frac{2 P^{\mathit{SD}} R^{\mathit{SD}}}{P^{\mathit{SD}}+R^{\mathit{SD}}}
\end{equation}

\noindent where the predicted span (PS) must exactly match the reference span (RS) in the exact place in the sequence, comparing by the left and right boundaries of the predicted and reference spans. 

The sequence labeling models trained to classify spans indirectly tackle span detection. These models are trained on BIO tags representing classes with suffixes (e.g., \textit{B-person}, \textit{I-person}, \textit{O}). The evaluation methods for strict span classification (SC) must take into account both the correctness of the predicted span and its assigned class, ignoring the correctly predicted \textit{O} tags. Simplified \textit{strict} evaluation of sequence labeling models trained for SC is defined as follows:

\begin{equation}
P^{\mathit{SC}}=\frac{\sum_i 1\left\{\mathit{PS}^i=\mathit{RS}^i \wedge \mathit{PS}_{c}^i=\mathit{RS}_{c}^i \wedge \mathit{PS}_{L}^i=\mathit{RS}_{L}^i \wedge \mathit{PS}_{R}^i=\mathit{RS}_{R}^i\right\}}{N_{\mathit{PS}}}
\end{equation}

\begin{equation}
R^{\mathit{SC}}=\frac{\sum_i 1\left\{\mathit{PS}^i=\mathit{RS}^i \wedge \mathit{PS}_{c}^i=\mathit{RS}_{c}^i \wedge \mathit{PS}_{L}^i=\mathit{RS}_{L}^i \wedge \mathit{PS}_{R}^i=\mathit{RS}_{R}^i\right\}}{N_{RS}}
\end{equation}

\begin{equation}
{F_1}^{\mathit{SC}}=\frac{2 P^{\mathit{SC}} R^{\mathit{SC}}}{P^{\mathit{SC}}+R^{\mathit{SC}}}
\end{equation}

\noindent where the predicted span must exactly match the reference span in the exact place in the sequence, comparing by the left and right boundaries of the predicted and reference spans, and the predicted span class must exactly match the reference span class (cf.~Table \ref{tab:sequence labeling_eval_notation} for detailed notation descriptions). 

Strict SD and SC metrics count partially correct predictions as invalid. However, partial evaluations are possible, which perform more fine-grained evaluations by counting the correct predictions token by token. These evaluations are suitable when we want to allow for some leniency, i.e., a mismatch between reference and predicted spans. We do not use partial evaluations in this thesis as no task from Section \ref{sec:tasks_datasets} necessitates it. Also, evaluating with partial evaluations can result in an overestimation of model capabilities. Thus, we stick to strict matching in this thesis for evaluation. However, partial evaluations are still useful and can be used in cases when we do not require perfectly detected span boundaries.

\begin{table}
    \centering
    \begin{adjustbox}{width=0.5\columnwidth}
    \begin{tabular}{cc}
    \toprule
    \textbf{Notation} & \textbf{Description} \\
    \midrule
    $PS$ & The predicted span \\
    
    $PS_L$ & The left boundary of the predicted span $PS$ \\
    
    $PS_R$ & The right boundary of  the predicted span $PS$ \\
    
    $RS$ & The reference span \\
    
    $RS_L$ & The left boundary of the reference span $RS$ \\
    
    $RS_R$ & The right boundary of the reference span $RS$ \\
    
    $N_{PS}$ & The number of predicted spans \\
    
    $N_{RS}$ & The number of reference spans \\
    
    $\mathit{PS}_{c}$ & The predicted span class \\
    
    $\mathit{RS}_{c}$ & The reference span class \\
    \bottomrule
    \end{tabular}
    \end{adjustbox}
    \caption{Notation used in the evaluation formulas}
    \label{tab:sequence labeling_eval_notation}
\end{table}

Similar to standard multiclass evaluation metrics in machine learning, sequence labeling metrics have their micro and macro variants. The micro metrics are calculated globally, counting the total number of true positives, false positives, and false negatives, irrespective of the class. On the other hand, the macro metrics are calculated for each class, and the total numbers of true positives, false positives, and false negatives are averaged class-wise. Then, the final score is produced by averaging the results of all the classes. Since sequence labeling tasks often come with many classes, micro metrics are typically better at reflecting the actual performance of sequence labeling models, accounting for class imbalance in the data. 

The results on sequence labeling tasks presented in the following chapters are evaluated with a micro F1 score on IOB2 tag predictions with strict matching using \textit{seqeval} library \citep{seqeval}, where the predicted span must exactly match the reference (expected) span. \textit{Seqeval} with micro F1 score as evaluation metric is standardly used in sequence labeling evaluations in the NLP community. Since modern PLMs rely on proprietary tokenizers, the predictions are evaluated only on the \textit{head word} from the tokenized input sequence to obtain the same number of predictions as there are reference IOB2 tags, which are on a word level (cf.~Subsection \ref{subsec:evaluating_sl}). 

\subsubsection{Evaluation Example} \label{subsubsec:evaluation_example}

Here, we provide a short example of micro and macro precision, recall, and F1 scores calculation for the SC task with strict matching using the IOB2 tagging scheme. We start with the following sentence, true (reference), and predicted labels:
\[
\begin{split} 
    & \mathit{sent} = [\mathrm{Paul}, \mathrm{McCartney}, \mathrm{performed}, \mathrm{on}, \mathrm{the}, \mathrm{rooftop}, \mathrm{in}, \mathrm{United}, \mathrm{Kingdom}, \mathrm{with}, \mathrm{The}, \mathrm{Beatles}], \\
    & \mathit{y_{true}}=[\texttt{B-PER}, \texttt{I-PER}, \texttt{O}, \texttt{O}, \texttt{O}, \texttt{O}, \texttt{O}, \texttt{B-LOC}, \texttt{I-LOC}, \texttt{O}, \texttt{B-ORG}, \texttt{I-ORG}], \\
    &\mathit{y_{pred}}=[\texttt{B-PER}, \texttt{I-PER}, \texttt{O}, \texttt{O}, \texttt{O}, \texttt{O}, \texttt{O}, \texttt{B-LOC}, \texttt{B-ORG}, \texttt{O}, \texttt{B-ORG}, \texttt{I-LOC}].
\end{split}
\]

\paragraph{Micro Scores.} To calculate the micro precision, recall, and F1 scores, we need to count the number of exact matches where the reference span matches the predicted span and the predicted span class is the same as the reference span class. We perform counting altogether, not distinguishing between different classes. For the given example, we identify only one case where the reference span matches the predicted span ($\texttt{B-PER}, \texttt{I-PER}$) and we count a total of $N_\mathit{PS}=4$ predicted spans, and a total of $N_\mathit{RS}=3$ reference spans. Now, we calculate the micro $P^{\mathit{SC}}_{\mathit{micro}}$, $R^{\mathit{SC}}_{\mathit{micro}}$, and $F_{1,\mathit{micro}}^{\mathit{SC}}$ metrics: 

\[
P^{\mathit{SC}}_{\mathit{micro}} = \frac{1}{4} = 0.25, \quad
R^{\mathit{SC}}_{\mathit{micro}} = \frac{1}{3} \approx 0.33, \quad
F_{1,\mathit{micro}}^{\mathit{SC}} = \frac{2 \cdot 0.25 \cdot 0.33}{0.25 + 0.33} \approx 0.29.
\]

\paragraph{Macro Scores.} To calculate the macro precision, recall, and F1 scores, we need to evaluate each class independently. This approach differs from calculating micro scores, where all classes are considered together. Using the previously provided example, we compute the scores for each class (\textit{PER}, \textit{LOC}, \textit{ORG}) separately. We proceed by counting the number of exact matches where the reference span matches the predicted span for each class. For the \textit{PER} class, we count one such span out of $N_\mathit{PS}=1$ predicted spans, and a total of $N_\mathit{RS}=1$ reference spans. We calculate the scores for the \textit{PER} class as follows:

\[
P^{\mathit{SC}}_{\mathit{PER}} = \frac{1}{1} = 1.00, \quad
R^{\mathit{SC}}_{\mathit{PER}} = \frac{1}{1} = 1.00, \quad
F_{1,\mathit{PER}}^{\mathit{SC}} = \frac{2 \cdot 1 \cdot 1}{1 + 1} = 1.00.
\]

\noindent Next, for the \textit{LOC} class, we count zero cases where the reference span matches the predicted span out of $N_\mathit{PS}=1$ predicted spans and $N_\mathit{RS}=1$ reference spans. We calculate the scores for the \textit{LOC} class as follows:

\[
P^{\mathit{SC}}_{\mathit{LOC}} = \frac{0}{1} = 0.00, \quad
R^{\mathit{SC}}_{\mathit{LOC}} = \frac{0}{1} = 0.00, \quad
F_{1,\mathit{LOC}}^{\mathit{SC}} = 0.00.
\]

\noindent We set $F_1=0.00$ score as the harmonic means goes to zero if either precision or recall is zero. Finally, for the \textit{ORG} class, we count zero cases where the reference span matches the predicted span out of $N_\mathit{PS}=2$ predicted spans and $N_\mathit{RS}=1$ reference spans. We calculate the scores for the \textit{ORG} class as follows:

\[
P^{\mathit{SC}}_{\mathit{ORG}} = \frac{0}{2} = 0.00, \quad
R^{\mathit{SC}}_{\mathit{ORG}} = \frac{0}{1} = 0.00, \quad
F_{1,\mathit{ORG}}^{\mathit{SC}} = 0.00.
\]

\noindent Now that we have calculated precision, recall, and F1 scores for each class, we can calculate the overall macro values for these metrics simply by averaging the calculation per class as follows:  

\[
P_{\mathit{macro}}^{\mathit{SC}} = \frac{P_{\mathit{PER}}^{\mathit{SC}} + P_{\mathit{LOC}}^{\mathit{SC}} + P_{\mathit{ORG}}^{\mathit{SC}}}{3} = \frac{1}{3} \approx 0.33,
\]
\[
R_{\mathit{macro}}^{\mathit{SC}} = \frac{R_{\mathit{PER}}^{\mathit{SC}} + R_{\mathit{LOC}}^{\mathit{SC}} + R_{\mathit{ORG}}^{\mathit{SC}}}{3} = \frac{1}{3} \approx 0.33,
\]
\[
F_{1,\mathit{macro}}^{\mathit{SC}} = \frac{F_{1,\mathit{PER}}^{\mathit{SC}} + F_{1,\mathit{LOC}}^{\mathit{SC}} + F_{1,\mathit{ORG}}^{\mathit{SC}}}{3} = \frac{1}{3} \approx 0.33.
\]

\subsection{Evaluation of Decoders} \label{subsubsec:evaluation_decoders}

In the context of PLMs for sequence labeling, we tackle the tasks with either encoders or decoders. Encoders solve the task directly, assigning BIO tags to tokens in the input sequence. In contrast, decoders do not assign tags to tokens but rather generate tokens from their vocabulary since they are trained to generate natural language. Therefore, a different approach is needed for predicting and mapping generated natural language to BIO tags for a fair evaluation across various PLMs (cf.~Subsection \ref{subsec:modeling_sl}). In this thesis, we first generate spans and their classes using a simple generation scheme applied to each trained decoder model and then map the generated tokens to BIO tags. We describe the scheme we use for the sequence labeling dataset that has four classes using the following regular expression:

\begin{center}
\begin{BVerbatim}
NA|([^:;]+:(class_1|class_2|class_3|class_4);)*
[^:;]+:(class_1|class_2|class_3|class_4)
\end{BVerbatim}    
\end{center}

\noindent where we divide spans and their classes with colons, multiple extractions with semicolons, and allow \textit{NA} generation as a signal that no spans are present in the query example. Following this, we form instructions similar to the ones used for NER by \cite{wang2022instructionner}, although we require a more strict output response from decoder-only LLMs, in line with the output format used by \cite{wang2023instructuie}. We give some examples from reformatted sequence labeling datasets in Tables~\ref{tab:it_examples} and \ref{tab:icl_sift_examples}.

To ensure a fair evaluation consistent with models fine-tuned directly for sequence labeling, we heuristically map response spans of decoders to IOB2 tags. We employ greedy span-based matching of predicted spans and their classes with input tokens, similar to \cite{wang2022instructionner}. We treat all cases in which no predictions are made, all cases where predicted spans do not align with input tokens, or an exception arises during matching due to output generation stochasticity, as if the \textit{O} tag was predicted for every input token. Further, we consider only the first line of the generated model answer as the prediction since some models tend to overestimate the number of required generated tokens to complete the task. After we finish the parsing of generated model outputs, evaluation is conducted in the same manner as for encoders, using the micro F1 score on IOB2 tag predictions with strict matching.\clearpage{}%
\clearpage{}%
\chapter{External Signal Through Multi-task Model} \label{ch:method1}

This thesis builds upon prior work on domain adaptation, multi-task learning, and sequential transfer learning (cf.~Section \ref{sec:tl}) to improve transfer learning for sequence labeling tasks starting from pre-trained neural language models with the encoder-only and decoder-only Transformer model architecture (cf.~Subsections \ref{subsec:seq_proc} and \ref{subsec:plm_adapt}). In the remainder of this chapter, we demonstrate how transfer learning for sequence labeling tasks can be improved through a model for reducing the effects of negative transfer in low(er)-resource domain transfer scenarios and encoder-only models for the ETD task.

The introduction of an additional, task-specific signal into the model is the first of the three pillars of PLM adaptation. More specifically, we achieve this by building on the encoder PLMs (cf.~Subsection \ref{subsec:enc_dec}) and methods for introducing more signal into the PLM (cf.~Subsection \ref{subsubsec:intro_more_signal}). In line with this and as one of the contributions of this thesis, we provide a multi-task model for integrating an additional signal to improve the performance of ETD tasks in low(er)-resource domain transfer scenarios. With this model, we aim to address RQ1: \textit{How to improve the domain transfer with little to no data for sequence labeling?} (cf.~Chapter \ref{ch:intro}). The emphasis in building the sequence labeling models is put on the nuances of the ETD task, a pivotal part of the IE pipeline in NLP. ETD is a necessary prerequisite for solving more complex tasks in the EE pipeline---ETC, EAD, and EAC. In the following sections, we present the motivation, models, experiments, and results for introducing the additional signal into PLMs.\footnote{\footnotesize{This chapter is adapted from: David Dukić, Kiril Gashteovski, Goran Glavaš, and Jan Šnajder. 2024. Leveraging Open Information Extraction for More Robust Domain Transfer of Event Trigger Detection. In Findings of the Association for Computational Linguistics: EACL 2024, pages 1197–1213, St. Julian's, Malta. Association for Computational Linguistics.}}

\section{Motivation}

ETD is much more complex than canonical sequence labeling tasks such as NER, PoS tagging, and chunking due to the complexity of event semantics, frequent implicitness of event triggers, and their high contextual dependency. While the notion of an event trigger is intuitive and universal (i.e., events and their triggers exist in all text domains), NLP research has struggled to provide a clear-cut operational definition of an event, giving rise to diverse annotation schemes, e.g., \cite{doddington-etal-2004-automatic,pustejovsky2005specification,shaw2009lode,cybulska2014guidelines,song-etal-2015-light}. The differences between annotation schemes, alongside the usual distribution shifts between text domains, make domain transfer of ETD very challenging. For this reason, PLMs adapted to one dataset and one domain for the ETD task will not perform well in another domain, leading to the effect of negative transfer. Since annotating domain-specific data is costly and time-consuming, one of the contributions of this thesis is to reduce negative domain transfer in the ETD task in scenarios with unlabeled domain-specific data (zero-shot) and scenarios with very few available labeled data in the target domain (few-shot). 

One way to facilitate domain transfer of ETD may be by means of a proxy task that exhibits a smaller distributional shift across domains and could thus mediate representational alignment between triggers of different domains. In principle, all tasks that extract structures that relate to event semantics, such as syntactic or predicate-argument structures, make good candidates for such a mediator \citep{liu-etal-2016-leveraging}. Recent work by \cite{deng-etal-2022-title2event} demonstrated that trigger and argument detection can be effectively aligned with subject-relation-object triples as mediators (in Chinese), where subjects and objects are mapped to arguments and relations to triggers. In other words, both events and subject-relation-object triples represent predicate-argument structures, pointing to tasks that extract the latter as potentially good mediators for domain transfer of ETD. 

\begin{figure}
    \centering
    \includegraphics[width=0.6\linewidth]{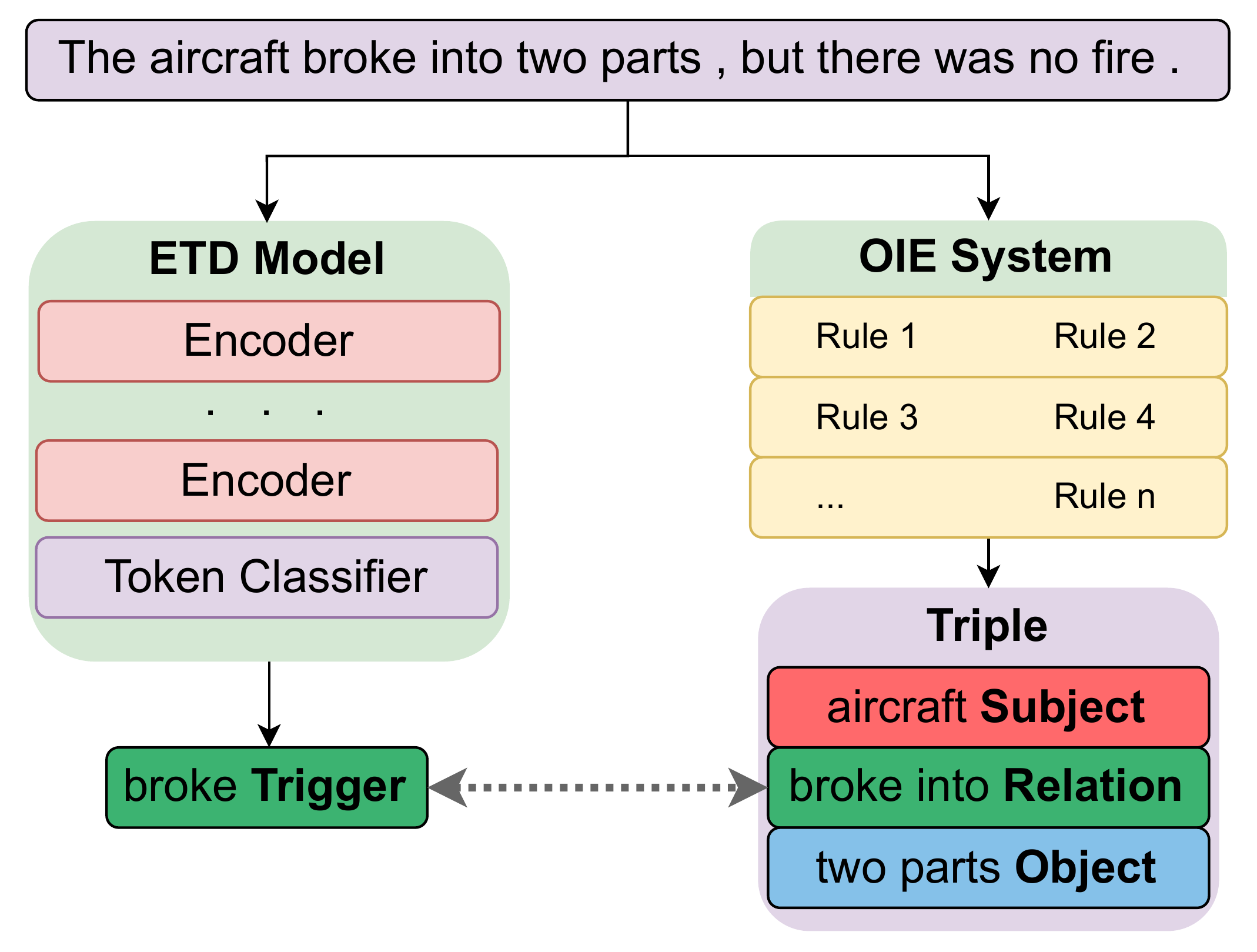}
    \caption[An example of ETD and subject-relation-object extraction with an OIE system]{An example of ETD and subject-relation-object extraction with an OIE system. The detected trigger and the extracted OIE relation often overlap to a significant degree, which can be leveraged for creating more robust trigger detection models across domains.}
    \label{fig:triple_extraction}
\end{figure}

OIE systems automatically extract subject-relation-object triples in a domain-independent manner. Figure~\ref{fig:triple_extraction} illustrates the overlap between the trigger \textit{broke} detected by the ETD model and an OIE relation \textit{broke into}, extracted by the OIE system. Following the observation of correspondence between event triggers and OIE relations,  we couple the domain-specific trigger annotations with the relation extractions obtained with a domain-agnostic rule-based OIE system through different multi-task model architectures and zero- and few-shot transfer regimes. The intuition is that, by coupling trigger annotations with OIE relations, we effectively couple event triggers between domains with OIE relations as mediators, ultimately achieving better transfer performance. Here, relation extraction is used as a proxy task, providing an additional signal that is universal enough to facilitate the transfer of the ETD task from the source to the target domain.

\section{Model} \label{sec:external_signal_model}

We leverage the IOB2 tagging scheme (cf.~Subsection \ref{subsec:modeling_sl}) with our models to assign \textit{B-Trigger}, \textit{I-Trigger}, or \textit{O} labels to sequences of tokens. Analogously, we model the relation extraction task---for which we use post-processed OIE relation extractions as ground-truth labels---also as a sequence labeling task with its own tags (\textit{B-Relation}, \textit{I-Relation}, and \textit{O}). 

We tackle domain transfer for ETD with two different model architectures (based on an encoder PLM) that couple OIE relations with ETD annotations, which we refer to as \textit{implicit} and \textit{explicit} OIE-ETD multi-task models. We next describe both variants in detail. The two multi-task setups use gold trigger labels from both the source and target (in few-shot scenarios) domains. Additionally, we use silver post-processed relation extractions obtained via the rule-based MinIE OIE system (cf.~Subsection \ref{subsec:re}) to couple triggers across domains through relations. For the sake of completeness, we also describe the \textit{vanilla} baseline model for ETD.

\subsection{Vanilla Model}

\begin{figure*}
    \centering
    \begin{subfigure}{0.47\textwidth}
        \centering
        \includegraphics[width=1.0\linewidth]{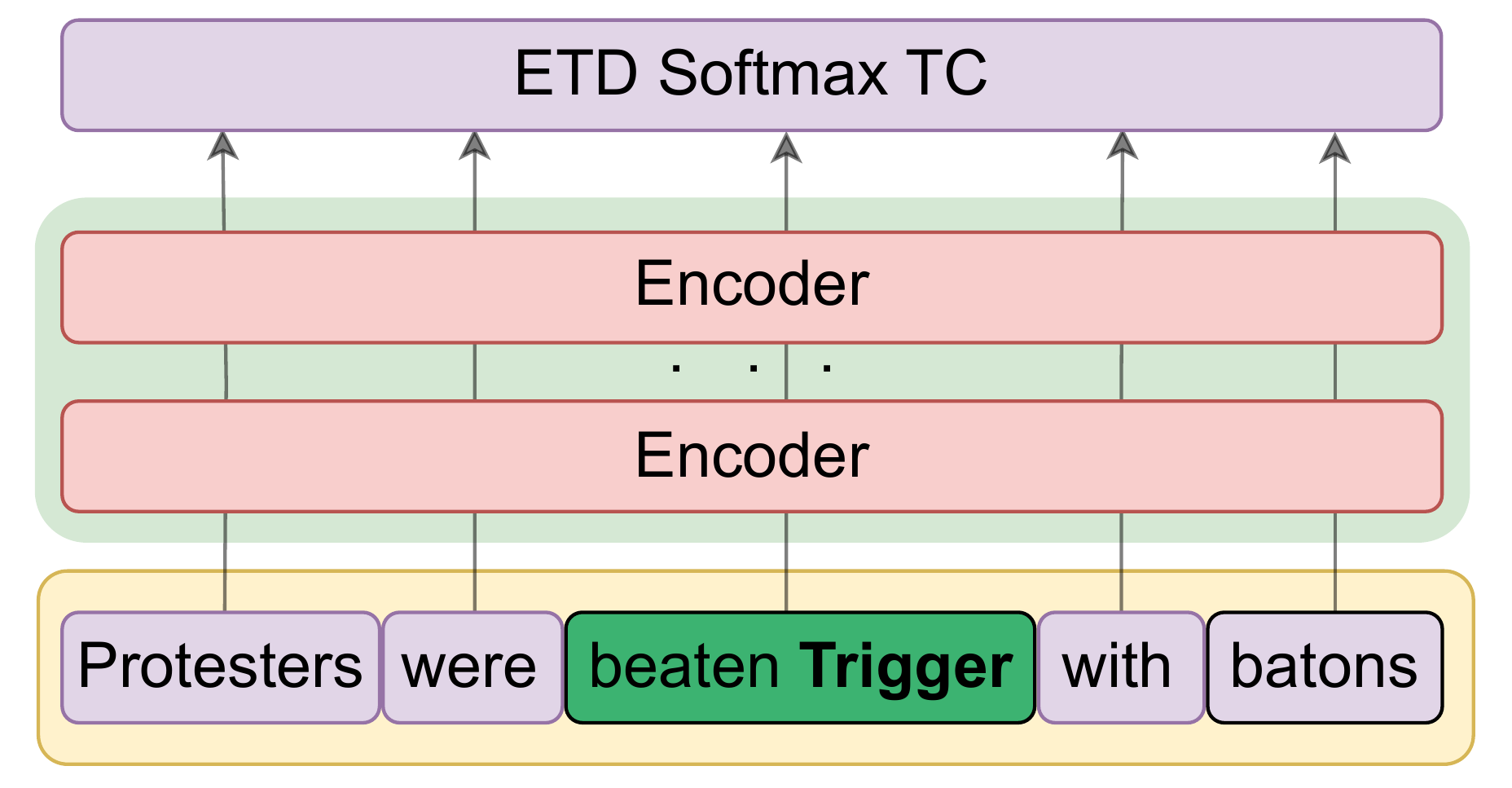} 
        \caption{}
        \label{fig:vanilla_model}
    \end{subfigure} \\
    \vspace{0.5cm}
    \begin{subfigure}{0.47\textwidth}
        \centering
        \includegraphics[width=1.0\linewidth]{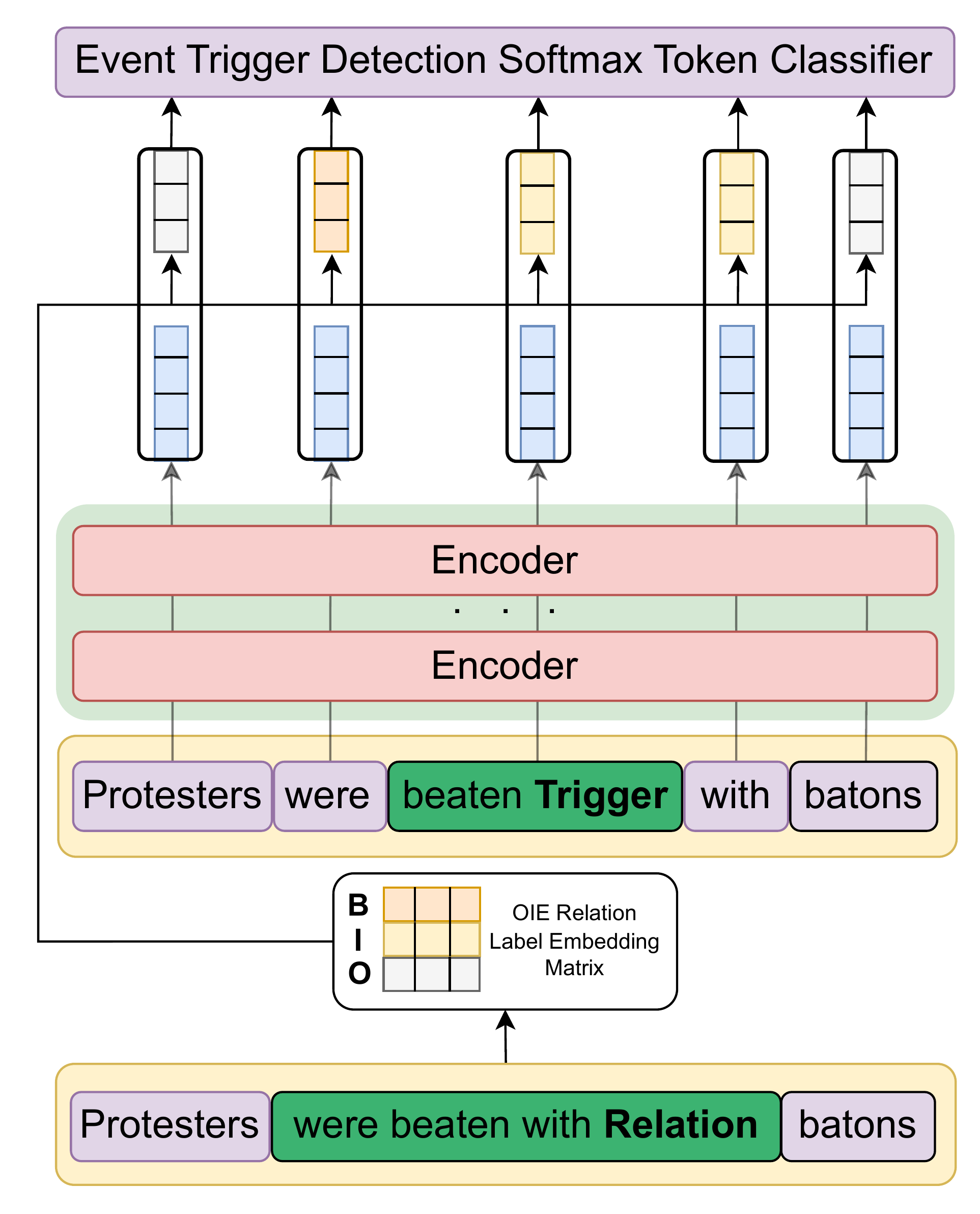}
        \caption{}
        \label{fig:implicit_model}
    \end{subfigure}%
    \hfill
    \begin{subfigure}{0.47\textwidth}
        \centering
        \includegraphics[width=1.0\linewidth]{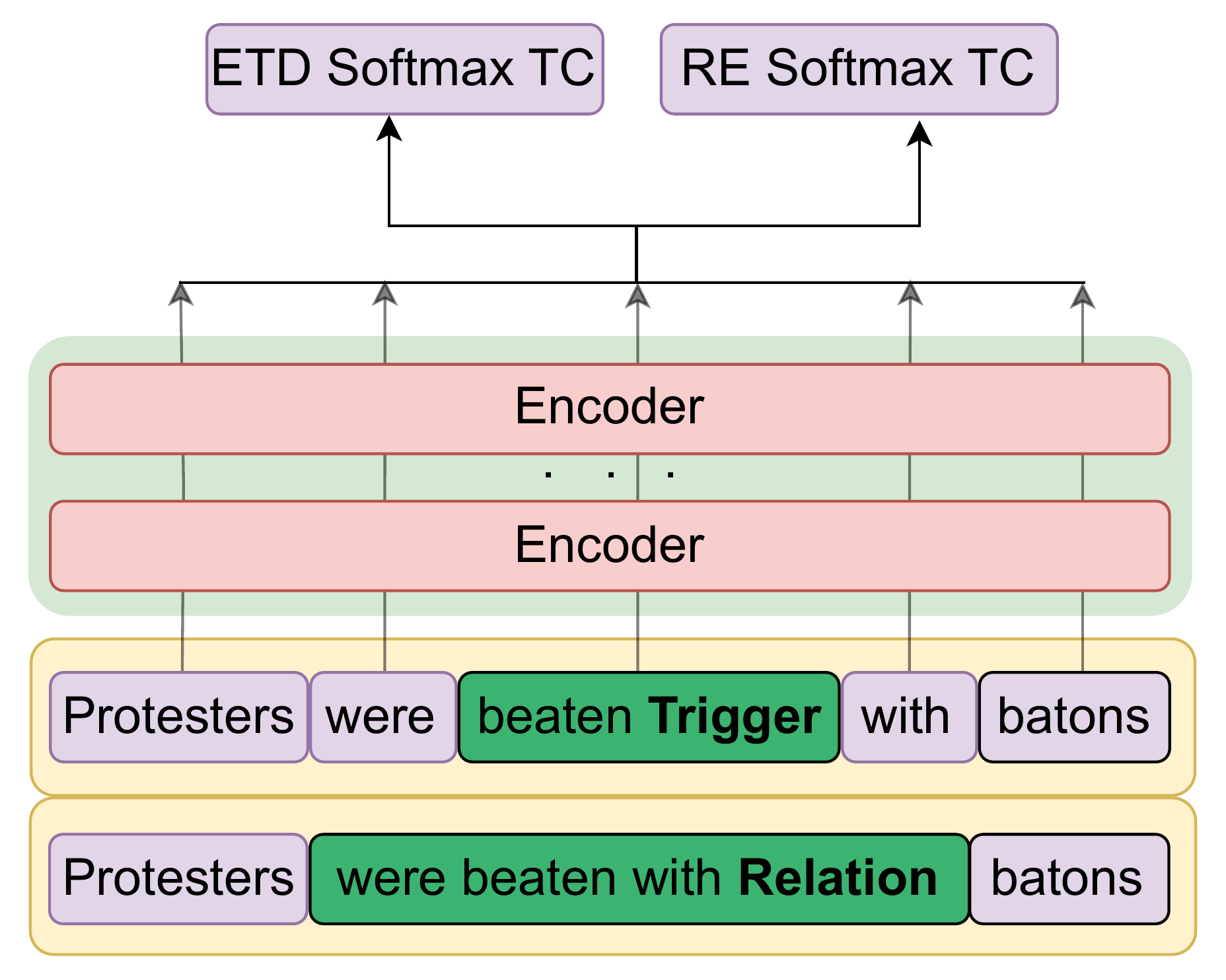}
        \vspace{1.1cm}
        \caption{}
        \label{fig:explicit_model}
    \end{subfigure}
    \caption[Comparison of the (a) \textit{vanilla}, (b) \textit{implicit}, and (c) \textit{explicit} models during training]{Comparison of the three model variants for the ETD task during training: (a) \textit{vanilla}, (b) \textit{implicit}, and (c) \textit{explicit}. (a) \textit{Vanilla} model during training. There is only one softmax token classifier used for assigning the trigger IOB2 tags. (b) \textit{Implicit} model during training. The input sentence is fed twice: once with trigger IOB2 tags through PLM encoders and once with OIE relation IOB2 tags by indexing the corresponding label embedding matrix. At the \textit{implicit} output, PLM's last hidden state embeddings are concatenated with OIE relation label embeddings per token and passed through the ETD softmax token classifier. (c) \textit{Explicit} model during training. The input sentence is fed once, but the loss is calculated separately for two sets of IOB2 tags---trigger tags and OIE relation labels---and passed through either the ETD or relation extraction (RE) softmax token classifier.}
    \label{fig:multi_task}
\end{figure*}

The \textit{vanilla} model is simply a Transformer-based encoder PLM with a standard softmax token classifier. The representation of each token $\mathbf{x}_\mathrm{PLM} \in \mathbb{R}^h$, from PLM's last layer, is forwarded to the ETD softmax classifier $\mathit{softmax}(\mathbf{W}_\mathrm{etd}^\mathrm{T}\mathbf{x}_\mathrm{PLM} + \mathbf{b}_\mathrm{etd})$, which predicts the IOB2 event trigger label for the token, with $\mathbf{W}_\mathrm{etd} \in \mathbb{R}^{h\times3}$ and $\mathbf{b}_\mathrm{etd} \in \mathbb{R}^3$ as trainable parameters of the softmax token classifier. All parameters are tuned by minimizing the (multi-class) cross-entropy loss. These include (1) all of the PLM's parameters and (2) the classifier's parameters $\mathbf{W}_\mathrm{etd}$ and $\mathbf{b}_\mathrm{etd}$. Figure \ref{fig:vanilla_model} shows the architecture of the \textit{vanilla} model.

\subsection{Implicit Multi-task Model} \label{subsec:implicit}

In the \textit{implicit} model, we train and use embeddings for token labels of OIE relations: one randomly initialized vector for each of the three IOB2 tags. The model concatenates the embedding $\mathbf{x}_\mathrm{OIE} \in \mathbb{R}^d$ of the OIE relation label of each token embedding to the contextualized token embedding of the token $\mathbf{x}_\mathrm{PLM} \in \mathbb{R}^h$ (the output of the last PLM layer), where $d$ is the dimension of the trainable OIE relation label embeddings (hyperparameter of the model), and $h$ is the PLM's hidden size. The final token representation is as a concatenation of the embedding from the last PLM layer and OIE relation label embedding: $\mathbf{x} = [\mathbf{x}_\mathrm{PLM}; \mathbf{x}_\mathrm{OIE}]$, which is fed to the standard softmax token classifier. The softmax token classifier predicts the IOB2 event trigger label for the token, $\mathit{softmax}(\mathbf{W}_\mathrm{cl}^\mathrm{T}\mathbf{x} + \mathbf{b}_\mathrm{cl})$, with $\mathbf{W}_\mathrm{cl} \in \mathbb{R}^{(d+h)\times 3}$ and $\mathbf{b}_\mathrm{cl} \in \mathbb{R}^{3}$ as trainable parameters of the classifier. As is common in multi-class classification, we tune all parameters by minimizing the (multi-class) cross-entropy loss. The \textit{implicit} model is illustrated in Figure~\ref{fig:implicit_model}. 

We train the model on ETD in the source domain, optimizing (1) all of the PLM's parameters, (2) classifier's parameters $\mathbf{W}_\mathrm{cl}$ and $\mathbf{b}_\mathrm{cl}$, and (3) embedding matrix $\mathbf{X}_\mathrm{OIE} \in \mathbb{R}^{3\times d}$ containing the trainable embeddings of the OIE labels. At inference time in the target domain, we run the OIE system on test sentences to obtain the OIE relation labels for tokens and then perform inference using the \textit{implicit} PLM for ETD and embeddings of OIE labels obtained in training. We hypothesize that the \textit{implicit} model is incentivized to establish---within the OIE label embeddings trained via ETD---contextualized associations between the two tasks. Intuitively, this should improve the recall of ETD in the target domain as long as the OIE---which is rule-based and thus more domain agnostic---is resilient to distribution shifts between domains. Similar approaches based on training label embeddings exist \citep{nguyen-grishman-2015-event, liu-etal-2017-exploiting, ji-etal-2019-exploiting}. However, they typically concatenate the label and token embeddings at the encoder's input and rely on encoders shallower than common Transformer-based PLMs.

\subsection{Explicit Multi-task Model}

The \textit{explicit} model exploits relation labels only during training. The model's training utilizes two standard softmax token classifiers and a shared PLM encoder. The representation of each token $\mathbf{x}_\mathrm{PLM} \in \mathbb{R}^h$, from PLM's last layer, is forwarded to the (1) ETD softmax classifier $\mathit{softmax}(\mathbf{W}_\mathrm{etd}^\mathrm{T}\mathbf{x}_\mathrm{PLM} + \mathbf{b}_\mathrm{etd})$, which predicts the IOB2 event trigger label for the token and (2) relation extraction softmax classifier $\mathit{softmax}(\mathbf{W}_\mathrm{re}^\mathrm{T}\mathbf{x}_\mathrm{PLM} + \mathbf{b}_\mathrm{re})$, which predicts the IOB2 relation label for the token, with $\mathbf{W}_\mathrm{etd},\mathbf{W}_\mathrm{re} \in \mathbb{R}^{h\times3}$ and $\mathbf{b}_\mathrm{etd},\mathbf{b}_\mathrm{re} \in \mathbb{R}^3$ as trainable parameters of two classifiers. Figure \ref{fig:explicit_model} shows the architecture of the \textit{explicit} model.

Based on the predictions, the (multi-class) cross-entropy loss is calculated for each classifier separately on a mini-batch basis. The average of calculated ETD and relation extraction losses is used to update PLM's and classifiers' parameters during training. This is where knowledge from both tasks interacts, allowing for the integration of additional signal from the relation extraction examples into PLM's and ETD classifier's parameters. At inference time, we do not use OIE relation labels in any way. The intuition is that if the notion of triggers is universal across domains and the OIE relations are indeed domain-independent, it should be sufficient only to leverage the in-domain trigger-relation connection during training to improve the transfer learning performance. Considering that the ETD and relation extraction tasks have the same number of corresponding labels, we tried to share the softmax classifier between ETD and relation extraction, but that led to worse overall performance. 

\section{Experiments} \label{sec:external_signal_exp}

This section describes the experimental setup used to explore incorporating an additional signal into PLMs. The proposed multi-task models, i.e., the \textit{implicit} and \textit{explicit} multi-task models from the previous section, are employed in these investigations. To improve the domain transfer of ETD with more signal, we use the outputs of the OIE extractor (cf.~Subsection \ref{subsec:re}) and combine them with available annotated source and target ETD examples through a multi-task model architecture and various transfer regimes. We also examine whether the transfer succeeds when we use the other rule-based OIE relation extractor. Our experiments investigate the transfer from a high-resource source domain to a low-resource target domain, which is the common transfer direction (cf.~Section \ref{sec:tl}). As a dataset from a high-resource source domain, we use MAVEN, a dataset of Wikipedia articles. In the low-resource target domain, we use datasets from the news domain---ACE05, EDNYT, and the EVEXTRA (cf.~Subsection  \ref{subsec:ee}). Essentially, we investigate the transfer from Wikipedia to the news domain. 

\subsection{Transfer Training Regimes}

For facilitating few-shot domain transfer of ETD, we employ \textit{joint} and \textit{sequential} transfer training regimes in combination with multi-task models. These training regimes are the extensions and combinations of inductive transfer learning paradigms (cf.~Section \ref{sec:tl}). In addition to using OIE relations with multi-task models to couple triggers with relations, we take inspiration from recent findings in language transfer \citep{meftah-etal-2021-hidden, schmidt-etal-2022-dont} and experiment with three transfer training regimes: \textit{joint training}, \textit{joint transfer}, and \textit{sequential transfer}. For the sake of completeness, we also consider \textit{in-domain training}, which reduces to fine-tuning each model on few-shot target domain examples. Finally, we experiment with the transductive transfer learning paradigm of domain adaptation through MLM, in the vein of \cite{gururangan-etal-2020-dont}, as an additional auxiliary objective next to open relation extraction. We achieve this by incorporating token-level MLM as an auxiliary training objective, utilizing an additional MLM head across all model variants. The head's parameters are updated during training and not used during inference.

\paragraph{Joint Training.}

The \textit{joint training} regime relies on mixed batches, adopted from the work on language transfer \citep{schmidt-etal-2022-dont}. A mixed batch consists predominantly of source trigger examples combined with a much lower fixed share of few-shot target trigger examples. Intuitively, having fewer few-shot examples should contribute to the update of model parameters with equal weight as the abundant source examples and ultimately prevent the model from overfitting on source data. We create mixed mini-batches consisting of $B \!=\! n \!+\! m$ examples, where $n$ are source examples, $m$ are randomly sampled few-shot target examples, and $n\!\gg\!m$. If more than $m$ few-shot examples are available, $m$ are consistently sampled from the few-shot pool. We fix $B\!=\!32$ with $n\!=\!27,m\!=\!5$ in our experiments. Fine-tuning is performed for a fixed number of epochs based on mixed mini-batch loss, calculated as the average of the source loss and $m$-shot target loss. In our experiments, \textit{joint training} amounts to mixed batch fine-tuning from either single- or multi-task PLMs. Single-task PLMs leverage ETD, while multi-task PLMs leverage ETD and relation extraction.

\paragraph{Joint Transfer.}

Similar to \textit{joint training}, the \textit{joint transfer} regime also uses mixed batches. However, instead of fine-tuning from PLM, we first train each PLM on source training data and then fine-tune with mixed batches in the same manner as in \textit{joint training}. \textit{Joint transfer} applied to multi-task models utilizes source OIE relations twice and target relations once during mixed batch fine-tuning.

\paragraph{Sequential Transfer.}

Analogous to \textit{joint transfer}, in the \textit{sequential transfer} regime, we fine-tune for a fixed number of epochs from the PLM trained on the source domain training data. However, unlike in \textit{joint transfer}, fine-tuning is done only with target few-shot examples.

\subsection{Training Details and Hyperparameters}

In this subsection, we provide details on the training process, the models employed, and optimization caveats for adapting ETD models in combination with transfer regimes. Additionally, we give an overview of the hyperparameter optimization for the \textit{implicit} model (cf.~Subsection \ref{subsec:implicit}).

\subsubsection{Models and Optimization} 

We use the RoBERTa-base \citep{liu2019roberta} PLM for token classification with $125$ million parameters, implemented in \textit{Hugging Face Transformers} \citep{wolf-etal-2020-transformers}. We evaluate ETD by micro F1 score on IOB2 tag predictions using strict matching. Since RoBERTa-base works on input split into subwords, the ETD cross-entropy loss is adjusted to take into account only the first token of each tokenized word from the input sequence. The models are trained with cross-entropy loss and Adam optimizer \citep{kingma2014adam} with the learning rate of $1\mathrm{e}{-5}$ for $10$ epochs. Our preliminary experiments found that incorporating a learning rate scheduler is beneficial. We use a multiplicative learning rate scheduler with a multiplying factor of $0.99$, which multiplies the learning rate in each epoch, lowering it throughout training.

\subsubsection{Transfer Training Details} 

When training on the source domain, we use the source validation set to select the best model based on the ETD micro F1 score. Specifically, we choose the model from the epoch that yields the highest ETD validation performance.\footnote{\footnotesize We also experimented with selecting the model based on the MLM perplexity on the target validation set, but that led to worse performance than optimizing for ETD F1 on the source validation set. The two options present a trade-off between adequately learning ETD and adjusting to the target domain, which comes at the expense of ETD performance.} 
Fine-tuning in \textit{joint/sequential transfer} regimes starts from the best model selected on the source validation set. In \textit{joint transfer} with the \textit{implicit} model, we perform mixed batch fine-tuning by averaging the source ETD and target few-shot ETD losses. Similarly, we average the source ETD and relation extraction losses with the target few-shot ETD and relation extraction losses in the \textit{joint transfer} with the \textit{explicit} model. Throughout experiments, we use a batch size of $B\!=\!32$. For each mini-batch, padding is applied to match the length of the longest example in the batch. Also, we employ gradient clipping of model parameters to a maximum of $1.0$ before each mini-batch update. We do transfer experiments with $0, 5, 10, 50, 100, 250$, and $500$ shots. 

\subsubsection{Masked Language Modeling Setup} 

For MLM and \textit{in-domain training}, we update the models' parameters in an alternating fashion within each epoch: first, based on the target training data MLM loss, and then based on the target few-shot loss. The MLM \textit{sequential transfer} is similar as without MLM. The difference is in the starting model, which is obtained by first training in the previously described alternate fashion, but with updates based on MLM loss on target training data and ETD loss on source training data. For all MLM experiments, we use a token-level masking probability of $15\%$, and the masking procedure is inherited from \cite{devlin-etal-2019-bert}. Specifically, out of $15\%$ of randomly chosen tokens, we mask $80\%$ tokens, replace $10\%$ tokens with random tokens from the vocabulary, and leave the remaining $10\%$ of the tokens unchanged.

\subsubsection{Hyperparameter Optimization} 

When training on the source domain, the \textit{implicit} model is additionally optimized on the source validation set (based on the ETD micro F1 score) with a simple grid search over the dimension of the trainable OIE-label embeddings $d$ and the learning rate for it. We try dimensions of $10, 50, 100$, and $300$ and learning rates of $1\mathrm{e}{-4}$, $5\mathrm{e}{-5}$, and $1\mathrm{e}{-5}$. When performing target few-shot fine-tuning in \textit{joint transfer} and \textit{sequential transfer}, we fix the dimension to the one that produced the highest source validation set ETD micro F1 score. In the \textit{joint training} and \textit{in-domain training} experiments, we arbitrarily fix the embedding size of the \textit{implicit} model to $300$ and $10$ across all the experiments, respectively.

\section{Results} \label{sec:external_signal_results}
 
We report the results of our experiments for three model architectures (cf.~Figure\ref{fig:multi_task}): \textit{vanilla}, \textit{implicit}, and \textit{explicit}. \textit{Vanilla} is trained in the same fashion as our proposed \textit{implicit} and \textit{explicit} variants, but without incorporating in any way the OIE relation information. For all experiments, we average results over three seeds and report micro F1 ETD scores on the held-out target test sets. For few-shot experiments, we additionally perform averaging on five different randomly sampled subsets from the target data training set. Moreover, we take precautions to ensure that samples from each draw are consistent across experiments and exclusively contain examples with triggers.

\subsection{Main Results} 

Table~\ref{tab:main_results} shows the main results of our experiments, with MinIE as a relation extractor for the multi-task models. Zero-shot domain transfer of ETD from MAVEN as the source to news datasets as targets exhibits noticeable negative transfer. The drops are massive compared to the performance of the models trained on all ACE05, EDNYT, or EVEXTRA training data. Even in this worst-case zero-shot setup, multi-task \textit{implicit} and \textit{explicit} models bring gains compared to \textit{vanilla} ones. Some interesting trends emerge when the number of shots increases. On average, relations help achieve higher target domain ETD performance for a low-to-moderate number of shots. However, when the number of shots reaches $500$ (or even $250$ in some cases) target examples, the effects of relations become negligible, except for the EVEXTRA dataset, where the gains from relations are consistent regardless of the number of shots or training regime. When considering all training regimes, the \textit{implicit} model outperforms the \textit{explicit} model. 

Contrary to the findings from studies on language transfer where the use of \textit{joint} transfer training regime was the best option \citep{schmidt-etal-2022-dont}, \textit{joint} transfer training regimes in our experiments were almost consistently worse compared to \textit{sequential transfer} and \textit{in-domain training}.\footnote{\footnotesize{Except for $5$ and $10$ shots.}} These findings are of practical interest since \textit{joint} is worse performance-wise and takes far more resources and time to train. With $500$ shots, \textit{sequential transfer} and \textit{in-domain training} come close to the full in-domain training performance for each news dataset. For a low number of shots ($5$ and $10$), doing \textit{in-domain training} is useless, and in this case, \textit{sequential transfer} is a better option. However, a higher number of shots in combination with \textit{in-domain training} can lead to a better performance than \textit{sequential transfer}.

\begin{table*}
\adjustbox{width=\linewidth}{
\small{\begin{tabular}{l|lccccccccc}
\toprule
\multicolumn{2}{c}{\multirow{2}{*}{\textbf{Training Regime}}} & \multicolumn{3}{c}{\textbf{ACE05 (0.706)}} & \multicolumn{3}{c}{\textbf{EDNYT (0.702)}} & \multicolumn{3}{c}{\textbf{EVEXTRA (0.893)}} \\
\cmidrule(lr){3-5} \cmidrule(lr){6-8} \cmidrule(lr){9-11}
\multicolumn{1}{l}{} & & {} \textbf{Vanilla} & \textbf{Implicit} & \textbf{Explicit} & \textbf{Vanilla} & \textbf{Implicit} & \textbf{Explicit} & \textbf{Vanilla} & \textbf{Implicit} & \textbf{Explicit} \\
\midrule
\multicolumn{1}{l}{} &   0-Shot          & 0.234           & 0.237          & \textbf{0.240} & 0.392          & 0.399          & \textbf{0.408} & 0.650           & 0.650          & \textbf{0.653} \\
\midrule
\multirow{6}{*}{\rotatebox[origin=c]{90}{\shortstack{joint \\ training}}} & 5-Shot        & 0.246           & 0.250          & \textbf{0.256} & 0.451         & 0.455          & \textbf{0.457} & 0.643           & 0.643          & \textbf{0.654} \\
{}  &   10-Shot    & 0.251           & 0.253          & \textbf{0.262} & 0.482          & \textbf{0.484} & \textbf{0.484} & 0.645           & 0.645          & \textbf{0.658} \\
{}  &   50-Shot       & 0.265           & 0.268          & \textbf{0.283} & 0.566          & \textbf{0.575} & 0.567          & 0.679           & 0.681          & \textbf{0.687} \\
{}  &   100-Shot      & 0.286           & 0.286          & \textbf{0.310} & 0.597          & \textbf{0.602} & 0.596          & 0.715           & 0.721          & \textbf{0.725} \\
{}  &   250-Shot      & 0.332           & 0.330          & \textbf{0.357} & 0.628          & \textbf{0.629} & \textbf{0.629} & 0.766           & \textbf{0.767} & 0.765          \\
{}  &   500-shot      & 0.382           & 0.378          & \textbf{0.398} & \textbf{0.649} & \textbf{0.649} & 0.646          & 0.793           & \textbf{0.798} & 0.792          \\
\midrule
\multirow{6}{*}{\rotatebox[origin=c]{90}{\shortstack{joint \\ transfer}}} &   5-Shot   & 0.248           & 0.248          & \textbf{0.254} & 0.433          & 0.436          & \textbf{0.440} & 0.631           & 0.633          & \textbf{0.636} \\
{}  &   10-Shot  & 0.251           & 0.250          & \textbf{0.256} & 0.448          & \textbf{0.451} & 0.450          & 0.632           & 0.634          & \textbf{0.638} \\
{}  &   50-Shot  & 0.262           & 0.265          & \textbf{0.267} & 0.524          & \textbf{0.536} & 0.507          & 0.650           & \textbf{0.656} & 0.648          \\
{} &    100-Shot    & 0.283           & 0.283          & \textbf{0.284} & 0.569          & \textbf{0.573} & 0.551          & 0.676           & \textbf{0.684} & 0.667          \\
{}  &   250-Shot & \textbf{0.328}           & \textbf{0.328} & 0.318          & 0.608          & \textbf{0.611} & 0.592          & 0.727           & \textbf{0.735} & 0.705          \\
{}  &   500-Shot & \textbf{0.388}  & 0.381          & 0.369          & 0.637          & \textbf{0.641} & 0.621          & 0.770           & \textbf{0.777} & 0.744          \\
\midrule
\multirow{6}{*}{\rotatebox[origin=c]{90}{\shortstack{sequential \\ transfer}}}   &   5-Shot     & \textbf{0.294}  & \textbf{0.294} & 0.276          & 0.458          & \textbf{0.466} & 0.448          & 0.659           & \textbf{0.661} & 0.653          \\
{}  &   10-Shot    & 0.372           & \textbf{0.374} & 0.330          & 0.512          & \textbf{0.521} & 0.490          & 0.688           & \textbf{0.693} & 0.680          \\
{}  &   50-Shot    & \textbf{0.511}  & 0.506          & 0.463          & 0.581          & \textbf{0.592} & 0.568          & 0.750           & \textbf{0.764} & 0.741          \\
{}  &   100-Shot   & 0.538           & \textbf{0.548} & 0.501          & 0.605          & \textbf{0.616} & 0.584          & 0.786           & \textbf{0.795} & 0.773          \\
{}  &   250-Shot   & \textbf{0.587}  & 0.577          & 0.556          & 0.631          & \textbf{0.644} & 0.607          & 0.824           & \textbf{0.835} & 0.813          \\
{}  &   500-Shot   & \textbf{0.610}  & 0.609          & 0.586          & \textbf{0.653} & 0.652          & 0.640          & 0.852           & \textbf{0.857} & 0.836          \\
\midrule
\midrule
\multirow{6}{*}{\rotatebox[origin=c]{90}{\shortstack{in-domain \\ training}}}  &   5-Shot              & 0.000           & 0.000          & 0.000          & 0.000          & 0.000          & 0.000          & 0.000           & 0.000          & 0.000          \\
{}  &   10-Shot              & 0.000           & 0.000          & 0.000          & 0.000          & 0.000          & 0.000          & 0.000           & 0.000          & 0.000          \\
{}  &   50-Shot              & 0.464           & \textbf{0.466} & 0.417          & \textbf{0.607} & 0.601          & 0.597          & 0.768           & \textbf{0.774} & 0.757          \\
{}  &   100-Shot             & 0.510           & \textbf{0.529} & 0.511          & 0.626          & \textbf{0.632} & 0.611          & 0.807           & \textbf{0.812} & 0.801          \\
{}  &   250-shot             & \textbf{0.570}  & 0.569          & 0.550          & 0.649          & \textbf{0.654} & 0.642          & 0.845           & \textbf{0.847} & 0.835          \\
{}  &    500-Shot            & 0.598           & \textbf{0.600} & 0.584          & 0.660          & 0.658          & \textbf{0.666} & 0.858           & \textbf{0.862} & 0.854          \\
\bottomrule
\end{tabular}}}
\caption[ETD domain transfer micro F1 scores when transferring from MAVEN as a source to ACE05, EDNYT, and EVEXTRA as targets]{ETD domain transfer micro F1 scores when transferring from MAVEN as a source to ACE05, EDNYT, and EVEXTRA as targets (zero-shot, three few-shot transfer training regimes, and in-domain, with six varying numbers of shots). The numbers in parentheses next to the target dataset are the in-domain performance test set scores when using all target training data. \textit{Joint/in-domain training}---target fine-tuning from PLM. \textit{Joint/sequential transfer}---target fine-tuning from PLM trained for ETD on MAVEN source training data. The best results by dataset and model per training regime are in bold. \textit{Implicit} and \textit{explicit} models leverage MinIE relation labels, unlike the \textit{vanilla} model. All scores were obtained by averaging over three seeds, and all few-shot experiments were additionally averaged across five different few-shot samples.}
\label{tab:main_results}
\end{table*}
 
\subsection{Auxiliary MLM Objective Results} \label{subsec:auxiliary_mlm}

Building on recent findings from work on PLM domain adaptation \citep{gururangan-etal-2020-dont}, we investigate whether MLM can further boost ETD transfer from Wikipedia to the news domain (cf.~Subsection \ref{subsec:transductive}). Since \textit{joint} regimes were consistently worse in main results, we examine the MLM effect only for \textit{in-domain training} and \textit{sequential transfer}. Table~\ref{tab:mlm_results} gives the results. \textit{Sequential transfer} proved to be more efficient than \textit{in-domain training}. On average, MLM with relations embodied into \textit{implicit} model in \textit{sequential transfer} regime outperforms the best results without MLM. An exception is the EVEXTRA dataset, where using OIE relations in conjunction with MLM and \textit{sequential transfer} does not lead to performance improvements compared to using only MLM.

\begin{table*}
\centering
\begin{subtable}{0.8\textwidth}
\adjustbox{width=\linewidth}{
\small{\begin{tabular}{l|lccccccccc}
\toprule
\multicolumn{2}{c}{\multirow{2}{*}{\textbf{Training Regime}}} & \multicolumn{3}{c}{\textbf{ACE05 (0.706)}} & \multicolumn{3}{c}{\textbf{EDNYT (0.702)}} & \multicolumn{3}{c}{\textbf{EVEXTRA (0.893)}} \\
\cmidrule(lr){3-5} \cmidrule(lr){6-8} \cmidrule(lr){9-11}
\multicolumn{1}{l}{} & & {} \textbf{Vanilla} & \textbf{Implicit} & \textbf{Explicit} & \textbf{Vanilla} & \textbf{Implicit} & \textbf{Explicit} & \textbf{Vanilla} & \textbf{Implicit} & \textbf{Explicit} \\
\midrule
\multicolumn{1}{l}{} &   0-Shot          & 0.234           & 0.237          & \textbf{0.240} & 0.392          & 0.399          & \textbf{0.408} & 0.650           & 0.650          & \textbf{0.653} \\
\midrule
\multirow{6}{*}{\rotatebox[origin=c]{90}{\shortstack{sequential \\ transfer}}}   &   5-Shot     & \textbf{0.294}  & \textbf{0.294} & 0.276          & 0.458          & \textbf{0.466} & 0.448          & 0.659           & \textbf{0.661} & 0.653          \\
{}  &   10-Shot    & 0.372           & \textbf{0.374} & 0.330          & 0.512          & \textbf{0.521} & 0.490          & 0.688           & \textbf{0.693} & 0.680          \\
{}  &   50-Shot    & \textbf{0.511}  & 0.506          & 0.463          & 0.581          & \textbf{0.592} & 0.568          & 0.750           & \textbf{0.764} & 0.741          \\
{}  &   100-Shot   & 0.538           & \textbf{0.548} & 0.501          & 0.605          & \textbf{0.616} & 0.584          & 0.786           & \textbf{0.795} & 0.773          \\
{}  &   250-Shot   & \textbf{0.587}  & 0.577          & 0.556          & 0.631          & \textbf{0.644} & 0.607          & 0.824           & \textbf{0.835} & 0.813          \\
{}  &   500-Shot   & \textbf{0.610}  & 0.609          & 0.586          & \textbf{0.653} & 0.652          & 0.640          & 0.852           & \textbf{0.857} & 0.836          \\
\midrule
\midrule
\multirow{6}{*}{\rotatebox[origin=c]{90}{\shortstack{in-domain \\ training}}}  &   5-Shot              & 0.000           & 0.000          & 0.000          & 0.000          & 0.000          & 0.000          & 0.000           & 0.000          & 0.000          \\
{}  &   10-Shot              & 0.000           & 0.000          & 0.000          & 0.000          & 0.000          & 0.000          & 0.000           & 0.000          & 0.000          \\
{}  &   50-Shot              & 0.464           & \textbf{0.466} & 0.417          & \textbf{0.607} & 0.601          & 0.597          & 0.768           & \textbf{0.774} & 0.757          \\
{}  &   100-Shot             & 0.510           & \textbf{0.529} & 0.511          & 0.626          & \textbf{0.632} & 0.611          & 0.807           & \textbf{0.812} & 0.801          \\
{}  &   250-shot             & \textbf{0.570}  & 0.569          & 0.550          & 0.649          & \textbf{0.654} & 0.642          & 0.845           & \textbf{0.847} & 0.835          \\
{}  &    500-Shot            & 0.598           & \textbf{0.600} & 0.584          & 0.660          & 0.658          & \textbf{0.666} & 0.858           & \textbf{0.862} & 0.854          \\
\bottomrule
\end{tabular}}}
\subcaption{Without MLM}
\end{subtable}

\vspace{0.15cm}

\begin{subtable}{0.8\linewidth}
\adjustbox{width=\linewidth}{
\small{\begin{tabular}{l|lccccccccc}
\toprule
\multicolumn{2}{c}{\multirow{2}{*}{\textbf{Training Regime}}} & \multicolumn{3}{c}{\textbf{ACE05 (0.706)}} & \multicolumn{3}{c}{\textbf{EDNYT (0.702)}} & \multicolumn{3}{c}{\textbf{EVEXTRA (0.893)}} \\
\cmidrule(lr){3-5} \cmidrule(lr){6-8} \cmidrule(lr){9-11}
\multicolumn{1}{l}{} & & {} \textbf{Vanilla} & \textbf{Implicit} & \textbf{Explicit} & \textbf{Vanilla} & \textbf{Implicit} & \textbf{Explicit} & \textbf{Vanilla} & \textbf{Implicit} & \textbf{Explicit} \\
\midrule
\multicolumn{1}{l}{} &   0-Shot          & 0.226 &              0.233 &              \textbf{0.241} & 0.396 &              \textbf{0.405} &              0.389 & 0.658 &              0.657 &              \textbf{0.659} \\
\midrule
\multirow{6}{*}{\rotatebox[origin=c]{90}{\shortstack{sequential \\ transfer}}}   &   5-Shot     & \textbf{0.311} &              0.309 &              0.303          & 0.469 &              \textbf{0.480} &              0.468          & 0.680 &              \textbf{0.681} &              0.666          \\
{}  &   10-Shot    & 0.390 &              \textbf{0.395} &              0.359          & \textbf{0.532} &              0.531 &              0.509          & \textbf{0.707} &              0.702 &              0.697          \\
{}  &   50-Shot    & \textbf{0.525} &              0.520 &              0.495          & 0.595 &              \textbf{0.600} &              0.577          & 0.774 &              \textbf{0.775} &              0.760          \\
{}  &   100-Shot   & 0.549 &              \textbf{0.561} &              0.519          & 0.612 &              \textbf{0.615} &              0.599          & \textbf{0.809} &              \textbf{0.809} &              0.791          \\
{}  &   250-Shot   & 0.587 &              \textbf{0.591} &              0.574          & 0.640 &              \textbf{0.645} &              0.627           & 0.843 &              \textbf{0.845} &              0.828          \\
{}  &   500-Shot   & \textbf{0.614} &              \textbf{0.614} &              0.604          & \textbf{0.661} &              \textbf{0.661} &              0.645          & \textbf{0.862} &              0.861 &              0.848          \\
\midrule
\midrule
\multirow{6}{*}{\rotatebox[origin=c]{90}{\shortstack{in-domain \\ training}}}  &   5-Shot              & 0.010 &              0.018 &              \textbf{0.034}          & 0.007 &              0.012 &              \textbf{0.037}          & 0.019 &              0.046 &              \textbf{0.085}          \\
{}  &   10-Shot              & \textbf{0.002} &              \textbf{0.002} &              0.000          & 0.002 &              0.000 &              \textbf{0.003}          & 0.001 &              0.002 &              \textbf{0.007}          \\
{}  &   50-Shot              & 0.366 &              \textbf{0.383} &              0.288          & 0.548 &              \textbf{0.557} &              0.552          & 0.685 &              \textbf{0.695} &              0.649          \\
{}  &   100-Shot             & \textbf{0.545} &              0.543 &              0.526          & 0.633 &              \textbf{0.638} &              0.623          & \textbf{0.796} &              0.794 &              0.790          \\
{}  &   250-shot             & 0.579 &              \textbf{0.584} &              0.564          & \textbf{0.661} &              \textbf{0.661} &              0.650          & 0.841 &              \textbf{0.844} &              0.835          \\
{}  &    500-Shot            & \textbf{0.612} &              0.607 &              0.596          & 0.670 &              \textbf{0.674} &              0.671 & \textbf{0.861} &              \textbf{0.861} &              0.852          \\
\bottomrule
\end{tabular}}}
\subcaption{With MLM}
\end{subtable}

\caption[ETD domain transfer micro F1 scores when transferring from MAVEN as a source to ACE05, EDNYT, and EVEXTRA as targets---with and without MLM comparison]{ETD domain transfer micro F1 scores when transferring from MAVEN as a source to ACE05, EDNYT, and EVEXTRA as targets (zero-shot, \emph{sequential transfer}, and \emph{in-domain training}, with six varying numbers of shots). Table (a) shows results without an auxiliary MLM objective, while Table (b) depicts results with an auxiliary MLM training objective on target domain training data. The numbers in parentheses next to the target dataset are the in-domain performance scores when using all target training data. \emph{In-domain training} results---target fine-tuning starting from PLM. \emph{Sequential transfer} results---target fine-tuning starting from PLM trained for ETD on MAVEN source training data. \emph{Implicit} and \emph{explicit} models leverage MinIE relation labels, unlike the \emph{vanilla} model. All scores were obtained by averaging over three seeds, and all few-shot experiments were additionally averaged across five different few-shot samples.}
\label{tab:mlm_results}
\end{table*}
 
\subsection{The Choice of the OIE System}

Finally, to examine if our results are specific to the OIE system, we replace MinIE with Stanford OIE (cf.~Subsection \ref{subsec:re} for details on the systems). We post-process the relations in the same manner as for MinIE. The experiments are conducted without MLM and for \textit{sequential transfer} and \textit{in-domain training} regimes. Table~\ref{tab:stanford_mini} shows the results. The difference between using MinIE and Stanford OIE is negligible for \textit{implicit} model but exists for \textit{explicit} model. Since \textit{explicit} outperformed \textit{implicit} in only five out of $156$ cases from Table~\ref{tab:stanford_mini}, we conclude that the gains from leveraging OIE relations in multi-task models are not due to the higher quality of MinIE extractions and persist for Stanford OIE. One can achieve similar, if not almost identical, gains using either extractor.

\begin{table*}
\centering
\adjustbox{width=\linewidth}{\small{\begin{tabular}{l|lcccccccccccc}
\toprule
\multicolumn{2}{c}{\multirow{4}{*}{\textbf{Training Regime}}} & \multicolumn{4}{c}{\textbf{ACE05 (0.706)}} & \multicolumn{4}{c}{\textbf{EDNYT (0.702)}} & \multicolumn{4}{c}{\textbf{EVEXTRA (0.893)}} \\
\cmidrule(lr){3-6} \cmidrule(lr){7-10} \cmidrule(lr){11-14}
\multicolumn{1}{l}{} & {} & \multicolumn{2}{c}{\textbf{MinIE}} & \multicolumn{2}{c}{\textbf{Stanford OIE}} & \multicolumn{2}{c}{\textbf{MinIE}} & \multicolumn{2}{c}{\textbf{Stanford OIE}} & \multicolumn{2}{c}{\textbf{MinIE}} & \multicolumn{2}{c}{\textbf{Stanford OIE}} \\
\cmidrule(lr){3-4} \cmidrule(lr){5-6} \cmidrule(lr){7-8} \cmidrule(lr){9-10} \cmidrule(lr){11-12} \cmidrule(lr){13-14}
\multicolumn{1}{l}{} & {} & \textbf{Implicit} & \textbf{Explicit} & \textbf{Implicit} & \textbf{Explicit} & \textbf{Implicit} & \textbf{Explicit} & \textbf{Implicit} & \textbf{Explicit} & \textbf{Implicit} & \textbf{Explicit} & \textbf{Implicit} & \textbf{Explicit} \\
\midrule
\multicolumn{1}{l}{} & 0-Shot   & 0.237    &0.240    & 0.237    & \textbf{0.242}    & 0.399    & \textbf{0.408}    & 0.401    & 0.406    &0.650    & 0.653    &0.650    & \textbf{0.657}     \\
\midrule
\multirow{6}{*}{\rotatebox[origin=c]{90}{\shortstack{sequential \\ transfer}}}   & 5-Shot   & 0.294    & 0.276    & \textbf{0.296}    & 0.283    & 0.466    & 0.448    & \textbf{0.468}    & 0.464    & \textbf{0.661}    & 0.653    & \textbf{0.661}    & 0.658     \\
{} & 10-Shot  & 0.374    & 0.330    & \textbf{0.375}    &0.350    & \textbf{0.521}    & 0.490    &0.520    & 0.512    & \textbf{0.693}    &0.680    & \textbf{0.693}    & 0.688     \\
{} & 50-Shot  & \textbf{0.506}    & 0.463    & \textbf{0.506}    & 0.476    & \textbf{0.592}    & 0.568    & 0.591    &0.570    & \textbf{0.764}    & 0.741    & 0.763    & 0.747     \\
{} & 100-Shot & \textbf{0.548}    & 0.501    & \textbf{0.548}    & 0.525    & \textbf{0.616}    & 0.584    & 0.615    & 0.587    & 0.795    & 0.773    & \textbf{0.796}    & 0.775     \\
{} & 250-Shot & \textbf{0.577}    & 0.556    & \textbf{0.577}    & 0.568    & 0.644    & 0.607    & \textbf{0.647}    & 0.602    & \textbf{0.835}    & 0.813    & 0.834    & 0.818     \\
{} & 500-Shot & \textbf{0.609}    & 0.586    & 0.602    & 0.584    & 0.652    &0.640    & \textbf{0.653}    & 0.627    & \textbf{0.857}    & 0.836    & 0.856    & 0.845 \\
\midrule
\midrule
\multirow{6}{*}{\rotatebox[origin=c]{90}{\shortstack{in-domain \\ training}}} & 5-Shot   &0.000       &0.000       &0.000       &0.000       &0.000       &0.000       &0.000       &0.000       &0.000       &0.000       &0.000       &0.000        \\
{} & 10-Shot  &0.000       &0.000       &0.000       &0.000       &0.000       &0.000       &0.000       &0.000       &0.000       &0.000       &0.000       &0.000        \\
{} & 50-Shot  & 0.466    & 0.417    & \textbf{0.467}    & 0.446    & 0.601    & 0.597    & 0.601    & \textbf{0.605}    & 0.774    & 0.757    & \textbf{0.775}    & 0.765     \\
{} & 100-Shot & \textbf{0.529}    & 0.511    & \textbf{0.529}    & 0.515    & 0.632    & 0.611    & \textbf{0.633}    & 0.615    & 0.812    & 0.801    & \textbf{0.814}    & 0.805     \\
{} & 250-Shot & \textbf{0.569}    &0.550    & \textbf{0.569}    & 0.557    & \textbf{0.654}    & 0.642    & 0.652    & 0.638    & \textbf{0.847}    & 0.835    & 0.846    & 0.840     \\
{} & 500-Shot & \textbf{0.600}      & 0.584    & 0.598    & 0.585    & 0.658    & \textbf{0.666}    & 0.657    & 0.662    & \textbf{0.862}    & 0.854    & 0.861    & 0.852     \\
\bottomrule
\end{tabular}}}
\caption[ETD domain transfer micro F1 scores when transferring from MAVEN as a source to ACE05, EDNYT, and EVEXTRA as targets---MinIE and Stanford OIE systems comparison]{ETD domain transfer micro F1 scores when transferring from MAVEN as a source to ACE05, EDNYT, and EVEXTRA as targets w.r.t. MinIE and Stanford OIE systems (zero-shot, \textit{sequential transfer}, and \textit{in-domain training}, with six varying numbers of shots). The numbers in parentheses next to the target dataset are the in-domain performance test set scores when using all target training data. \textit{Sequential transfer}---target fine-tuning from PLM trained for ETD on MAVEN source training data. \textit{In-domain training}---target fine-tuning from PLM. The best results by dataset, \textit{implicit} or \textit{explicit} relation-leveraging models, per training regime and OIE system, are in bold. All scores were obtained by averaging over three seeds, and all few-shot experiments were additionally averaged across five different few-shot samples.}
\label{tab:stanford_mini}
\end{table*} 
\subsection{Summary}

We demonstrated that multi-task fine-tuning of a PLM for OIE relation extraction and ETD and transfer regimes adopted from the body of work on language transfer \citep{lauscher-etal-2020-zero,schmidt-etal-2022-dont} reduce the trigger distribution shift between domains and consequently improve ETD performance in the low-resource target domain. The improvements of the transfer learning persist in zero- and few-shot setups. This addresses the RQ1 (cf.~Chapter \ref{ch:intro}), which focuses on enhancing transfer performance with minimal to no domain-specific target data. Addressing the RQ1 in the proposed manner paves the way for further advancements, particularly in the context of EE subtasks. The key to reducing the trigger distribution shift lies in choosing the right combination of a multi-task model and a transfer training regime. 

The best improvements were achieved with \textit{implicit} multi-task model and \textit{sequential transfer} regime. We also demonstrated that more substantial gains can be reached when combining OIE relations with MLM as an auxiliary task. This is especially evident for the models pre-trained with the ETD task on the source domain and with the MLM training objective on the target domain in the \textit{implicit} multi-task model. Replacing MinIE with Stanford OIE revealed that gains on the target domain for the ETD task persist when using the other OIE extractor. 

To summarize, introducing an external signal using a combination of inductive and transductive transfer learning methods with multi-task model architectures mitigates the negative transfer for the sequence labeling ETD task, consequently improving the transfer of the ETD task between domains and laying a foundation for universally more effective EE.

\clearpage{}%
\clearpage{}%
\chapter{Layer-wise Causal Mask Removal} \label{ch:method2}

The second pillar of the PLM adaptation pertains to the modifications of the model's architecture to improve the transfer to the target task (cf.~Subsection \ref{subsubsec:modify_architecture}). The architecture modifications depend on the PLM internals and the target task specificities. Encoders have shown remarkable performance in transfer to the NLU tasks and, more specifically, sequence labeling tasks \citep{devlin-etal-2019-bert, fei-etal-2021-better-new} due to the bidirectional information flow enabled through all the layers during both pre-training and adaptation \citep{artetxe-etal-2022-role}. However, today, the models of choice for many NLU tasks have become decoder-only LLMs, scaled to billions of parameters. Despite their good performance on NLU tasks, there is still considerable room for improvement, even for the largest decoder models, such as ChatGPT, which fall far behind SOTA results on fundamental NLP tasks. This holds particularly for IE tasks \citep{han2023information}, such as NER, aspect-based sentiment analysis, and EE. Tackling these tasks by prompting LLMs proved quite difficult \citep{wang2023gpt}. As the community stopped scaling up encoders, LLMs became the field's de facto standard for pre-training and adaptation to target tasks. 

In line with this, as one of the contributions of this thesis, we present a method for improved decoder-only LLM adaptation to sequence labeling tasks. The proposed method experiments with the decoder's causal mask (cf.~Subsection \ref{subsec:enc_dec}) in various LLM layers to improve its performance on sequence labeling tasks, addressing RQ2: \textit{How to efficiently bridge the pre-train--fine-tune gap in decoder-only LLMs for enhanced sequence labeling performance?} (cf.~Chapter \ref{ch:intro}) in a way that narrows the pre-train--fine-tune discrepancy effectively (cf.~Subsection \ref{subsec:enc_dec}) while simultaneously ensuring parameter-efficiency (cf.~Subsections \ref{subsubsec:optim_scheme} and \ref{subsec:lora}). In the following sections, we provide a detailed motivation for investigating the causal mask, present the method, and give the experiments and results.\footnote{\footnotesize{This chapter is adapted from: David Dukić and Jan Šnajder. 2024. Looking Right is Sometimes Right: Investigating the Capabilities of Decoder-only LLMs for Sequence Labeling. In Findings of the Association for Computational Linguistics: ACL 2024, pages 14168–14181, Bangkok, Thailand. Association for Computational Linguistics.
}}

\section{Motivation}

Although many SOTA LLMs are accessible only through paywalls, the community has responded by training and publicly releasing open-weight LLMs with multiple billions of parameters, such as Llama2 \citep{touvron2023llama}, Llama3 \citep{dubey2024llama}, Mistral \citep{jiang2023mistral}, and Gemma \citep{team2024gemma}. PEFT techniques, such as QLoRA \citep{dettmers2023qlora} (cf.~Subsection \ref{subsec:lora}), facilitate experimenting with adaptation of LLMs to task-specific data. These strategies allow leveraging open-weight LLMs' hidden states for task-specific classifiers---de facto using decoders as encoders. Using the hidden state of the decoder's last token serves as a reliable feature for classifying the input sequence. Here, the idea is that the representation of the last token in the sequence has \textit{seen} the whole context. 

However, using the distributed representation of each token at the model output for classification does not suffice for sequence labeling. Pre-training decoders with CLM teaches the model to generate coherent text. It achieves this with the causal mask (CM), which limits bidirectional information flow, preventing the model from attending to tokens in the sequence positioned to the right of the current token. Restricting the model \textit{to look right} is detrimental if one wants to use the decoder to produce fully contextualized token-level embeddings. What is worse, for many sequence labeling tasks, the token's label depends on the succeeding tokens, and omitting this context often results in subpar performance. For the given reasons, many released open-weight pre-trained decoder-only models implement sequence classification heads on top of decoder blocks but not sequence labeling (i.e., token classification) heads \citep{wolf-etal-2020-transformers}. This shows that decoders are not meant to be used as token classifiers. Fortunately, an effective method exists to adapt them for improved sequence labeling performance.

In line with previous findings and to address RQ2, one of the contributions of this thesis is a layer-wise CM removal method applied during fine-tuning and inference for improved sequence labeling performance of decoder-only LLMs. This method is surprisingly helpful in mitigating the mismatch between PLM's architecture and the nuances of target sequence labeling tasks. Concurrently with our findings, researchers have experimented with similar architectural modifications in decoder-only LLMs, removing the CM entirely during fine-tuning and inference \citep{li2023label,behnamghader2024llm2vec,lee2024nv,HUANG2025112907}. The findings showed improvements over a limited number of sequence labeling tasks. However, these works did not examine in depth why this phenomenon occurs, whether it occurs with layer-wise CM removal, and whether it persists across a range of sequence labeling tasks. We build on this gap and propose a method for removing the CM from groups of layers of open-weight decoder-only LLMs, which we refer to as \textit{layer group unmasking}.

\section{Method} \label{sec:cm_removal_method}

The CM is a crucial component of CLM-based decoders, as it prevents the model from attending to future tokens and facilitates autoregressive text generation. This constraint is enforced by defining the CM as a triangular matrix and adding this matrix to the dot product of the query and key attention matrices. The resulting sum is passed through the softmax function in the scaled dot-product attention mechanism, as introduced in \cite{vaswani2017attention}. Formally:
\begin{align*}
&\mathbf{CM} = \begin{pmatrix}
    0 & -\infty & -\infty & \dots  & -\infty \\
    0 & 0 & -\infty & \dots  & -\infty \\
    \vdots & \vdots & \vdots & \ddots & \vdots \\
    0 & 0 & 0 & \dots  & 0
\end{pmatrix},\\\\
&\mathrm{Attn}(\mathbf{Q},\mathbf{K},\mathbf{V})=\mathit{softmax}\left(\frac{\mathbf{Q}\mathbf{K}^T + \mathbf{CM}}{\sqrt{d_k}}\right)\mathbf{V}.
\end{align*}

\noindent Here, $\mathbf{Q}$, $\mathbf{K}$, and $\mathbf{V}$ are the query, key, and value attention matrices, respectively, and $d_k$ is the dimension of queries and keys. Effectively, applying softmax for tokens with value $-\infty$ in the CM results in attention scores being 0.

\begin{figure}
    \centering
    \includegraphics[width=\linewidth]{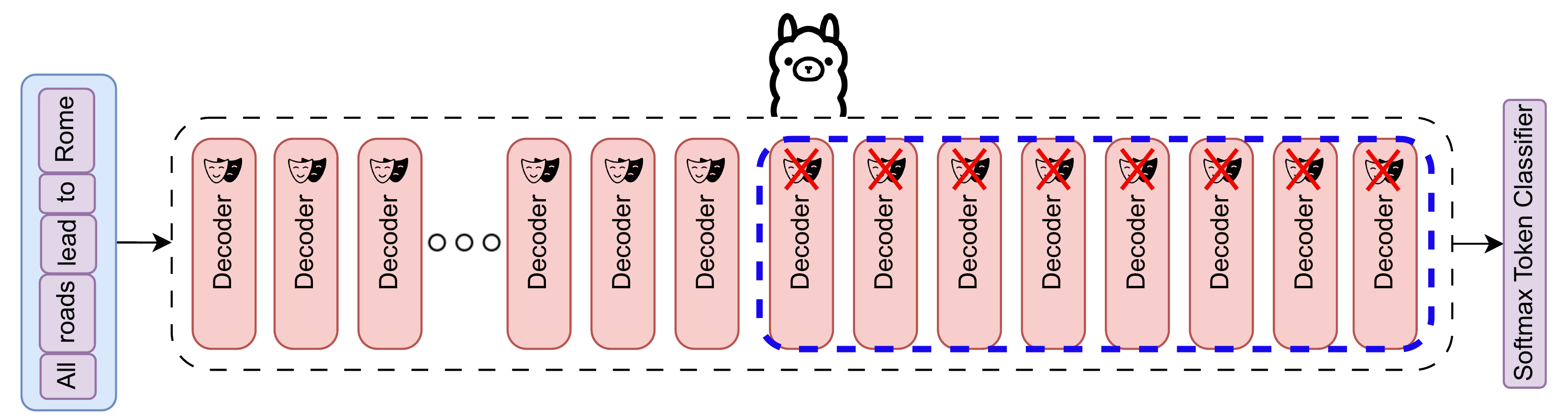}
    \caption[Layer-wise causal mask removal from decoder block groups in a decoder-only LLM]{Layer-wise causal mask removal from decoder block groups in a decoder-only LLM. Here, the causal mask is removed from the top eight decoder blocks of the Llama2-7B model to enable bidirectionality during fine-tuning, which proves beneficial for many sequence labeling tasks.}
    \label{fig:llama_unmasking}
\end{figure}

As reported in \cite{li2023label}, removing CM across all decoder layers of Llama2 during fine-tuning increases the model's performance on the NER sequence labeling task by a large margin. This is surprising because, although the CM mask was in place during pre-training and the model was restricted from utilizing the right-side context, it learned to attend to future tokens by backpropagating on the training data over a few epochs with CM removed. Building on this, we select a subset of decoder blocks for which we replace all $-\infty$ entries in the CM with zeros, effectively removing the CM. Since we group layers into groups of $b$ blocks, we refer to these decisions as \textit{layer group unmasking} configurations. We try out combinations of bidirectional information flow across decoder layers. Figure~\ref{fig:llama_unmasking} shows the removal of the CM from the top eight decoder blocks in the Llama2-7B model, where these eight blocks form a group.

Considering that we experiment with open-weight LLMs made up of $n=32$ decoder blocks, and each block can either remove or keep its CM, this gives a search space of $2^n=2^{32}$ possibilities. To reduce the search space to a manageable size, we group the $n$ decoder blocks into $m=4$ layer groups with $b=8$ consecutive decoder blocks per layer group and then jointly mask or unmask all blocks in each layer group. This leaves us with $2^m=16$ unmasking configurations per LLM and task. For ease of reference, we encode our unmasking configurations with binary four-digit codes ranging from $0000$ (all four layer groups masked) to $1111$ (all four layer groups unmasked), where each $0$ and $1$ denote a layer group that is masked or unmasked, respectively. Unmasking configurations are interpreted left to right (the first digit pertains to the layer group closest to the model's input).

\section{Experiments} \label{sec:cm_exp}

This section describes the experimental setup for the layer-wise CM removal inspections. We hypothesize that removing the CM from all decoder layers may not benefit all sequence labeling tasks. To confirm this, we experiment with layer-wise CM removal across open-weight LLM blocks. We extend our analysis to a series of sequence labeling tasks, namely NER, aspect-based sentiment analysis, text chunking, ETC, and demonstrate that layer-wise right-side context utilization is highly task-dependent. We compare against strong encoder-only sequence taggers and instruction-tuned LLMs. Finally, we compare with encoder-only models, assessing the relative strengths of the decoder and encoder architectures regarding parameter scale, training data, and the particular sequence labeling task (cf.~Subsection \ref{subsec:enc_dec} for a refresher on model architectures). In a controlled experiment, we pre-train small encoder and decoder models from scratch with an identical number of parameters on the same data. Throughout pre-training, we monitor and analyze the evolution of their performance on sequence labeling tasks and explore the effects of CM removal on a smaller scale. 

\subsection{Training}

In this subsection, we present the encoder and decoder models utilized for sequence labeling experiments, describe the optimization procedures, and detail the experimental setup for instruction tuning (IT) (refer to Section \ref{subsec:enc_dec}). Here, we merely utilize instruction-tuned models as an experimental choice to evaluate and compare their performance against encoder and decoder models utilized as encoders for sequence labeling. Furthermore, we outline the experimental setup for training small language models, highlighting key implementation details and configurations.

\subsubsection{Models}

We choose open-weight LLMs for unmasking experiments and IT: Llama2-7B \citep{touvron2023llama} and Mistral-7B \citep{jiang2023mistral}, both with 7B parameters. We use the base versions of the models, with their \textit{Hugging Face Hub} identifiers \texttt{meta-llama/Llama-2-7b-hf} and \texttt{mistralai/Mistral-7B-v0.1}, respectively. For encoders, we use RoBERTa-base and RoBERTa-large PLMs with $125$ and $355$ million parameters, respectively \citep{liu2019roberta}. 

We employ a classic softmax token classification head on top of the PLMs for sequence labeling experiments. Formally, assuming an IOB2 label set of cardinality $|L|$: the representation of each token $\mathbf{x}_\mathrm{PLM} \in \mathbb{R}^h$, from PLM's last layer, is forwarded to the sequence labeling softmax classifier $\mathit{softmax}(\mathbf{W}_\mathrm{sl}^\mathrm{T}\mathbf{x}_\mathrm{PLM} + \mathbf{b}_\mathrm{sl})$, which predicts the IOB2 label for the token, with $\mathbf{W}_\mathrm{sl} \in \mathbb{R}^{h\times|L|}$ and $\mathbf{b}_\mathrm{sl} \in \mathbb{R}^{|L|}$ as trainable parameters of the softmax token classifier. Here, the PLM is either an encoder or decoder model with CM removed in a subset of layers. PLM implementations and weights are taken from \textit{Hugging Face Transformers} \cite{wolf-etal-2020-transformers}. If the sequence labeling head is not implemented, we add our implementation, mimicking the one for RoBERTa models.

\subsubsection{Optimization} 

We use QLoRA \citep{dettmers2023qlora} for fine-tuning all models to ensure a fair comparison between the smallest and the largest models and enable fine-tuning under constrained computing resources. QLoRA is applied to query and value attention matrices inside each encoder or decoder block with a fixed rank of $r=64$, a scaling parameter of $\alpha=16$, and a dropout probability of $p=0.1$. 
This approach optimizes only the decomposed query and value matrices, along with the sequence labeling softmax classifier parameters, using cross-entropy loss, which results in a significant reduction in trainable parameters per model. The models are trained in bfloat16 precision, with loaded pre-trained weights in 4-bit NormalFloat data type, and we leverage double quantization. Using this setup, we were able to fit all models into 40GB of GPU memory of Ampere A100. The models are trained with a consistent batch size of $16$ per experiment. To use this batch size across models, we pre-process all datasets to a maximum tokenized sequence length of $128$. This cutoff is optimal as there are less than $10$ sentences for each dataset and split that end up truncated independent of the tokenizer used. We pad the sequences to the longest example in the batch and randomly sample examples for training depending on the seed.

We train the models with a paged 8-bit AdamW \citep{loshchilov2017decoupled} optimizer to handle the memory spikes \cite{dettmers2023qlora}. The parameters of AdamW are fixed to $\beta_1=0.9, \beta_2=0.95, \epsilon=1\mathrm{e}{-5}, \lambda=0.1$. For the learning rate scheduler, we choose the cosine annealing scheduler \citep{loshchilov2016sgdr}. We apply gradient clipping set to $1.0$, gradient accumulation with four steps, and gradient checkpointing. We use a consistent learning rate of $2\mathrm{e}{-4}$ across all experiments and fine-tune models over a fixed number of five epochs. All results are averages of five runs with different seeds, and we always pick the last model for each seed.

\subsubsection{Instruction Tuning} 

Each dataset was pre-processed to enable LLM training using IT on prompts. We create instruction prompts by parsing IOB2 tags from sequence labeling datasets to produce desired outputs. We show training and evaluation examples in Table~\ref{tab:it_examples}. Each training prompt template consists of an \textit{instruction} part (including the label set under \textit{options}), a query \textit{sentence}, and an expected \textit{response}. Evaluation prompts follow a similar template. The difference lies in the absence of the expected \textit{response}, as we require the model to generate this output at test time. We use Llama2-7B and Mistral-7B with CLM heads for IT. Training is done with QLoRA based on the vanilla cross-entropy CLM loss. Loss is computed over all the tokens in the prompt.

All hyperparameters for IT experiments are inherited from the previous subsection except for the batch size. The batch size depends on the maximum sequence length for packing the dataset examples. Short examples are packed together in the same input to increase training efficiency. This length is set to $512$ for the Rest14 dataset and \num{1024} for other datasets since Rest14 has shorter instruction examples on average and has the smallest training set among all the datasets used for experiments in this chapter (cf.~Table~\ref{tab:dataset_stats2}). We observe that ATE+ATP models require twice as many epochs ($10$ epochs) to learn to extract aspect terms and their polarity, and we hypothesize that this is due to the low number of training examples compared to other datasets (\textasciitilde3k vs.~\textasciitilde14k). Since RoBERTa requires tokenization with added prefix space, we enforce the same for open-weight LLMs' tokenizers. The cross-entropy loss is adjusted to consider only the first token of each tokenized word from the input sequence. We report all results as averages of five runs with different seeds.

\begin{table*}
    \centering
    \begin{adjustbox}{width=1.0\linewidth}
    \begin{tabular}{p{2cm}  L{5cm}  L{5cm}}
    \toprule
         \multicolumn{1}{c}{Task (dataset)} & \multicolumn{1}{c}{Example for training} & \multicolumn{1}{c}{Example for evaluation} \\
         \midrule
         \multicolumn{1}{c}{\multirow{3}{*}{NER (CoNLL03)}} & \tiny
         \#\#\# Instruction: 

please extract named entities and their type from the input sentence, all entity types are in options 

\#\#\# Options:

person, location, organization, miscellaneous

\#\#\# Sentence:

" What we have to be extremely careful of is how other countries are going to take Germany 's lead , " Welsh National Farmers ' Union ( NFU ) chairman John Lloyd Jones said on BBC radio .

\#\#\# Response:

Germany:location;Welsh National Farmers ' Union:organization;NFU:organization;John Lloyd Jones:person;BBC radio:organization & \tiny
         \#\#\# Instruction: 

please extract named entities and their type from the input sentence, all entity types are in options 

\#\#\# Options:

person, location, organization, miscellaneous

\#\#\# Sentence:

" What we have to be extremely careful of is how other countries are going to take Germany 's lead , " Welsh National Farmers ' Union ( NFU ) chairman John Lloyd Jones said on BBC radio .

\#\#\# Response:  \\\midrule

\multicolumn{1}{c}{\multirow{3}{*}{\shortstack{ATE+ATP \\ (Rest14)}}} & \tiny \#\#\# Instruction:

please extract aspect terms and their polarity from the input sentence, all polarity types are in options 

\#\#\# Options:

positive, negative, neutral, conflict

\#\#\# Sentence:

The lobster sandwich is \$ 24 and although it was good it was not nearly enough to warrant that price .

\#\#\# Response:

lobster sandwich:conflict;price:negative & \tiny \#\#\# Instruction:

please extract aspect terms and their polarity from the input sentence, all polarity types are in options 

\#\#\# Options:

positive, negative, neutral, conflict

\#\#\# Sentence:

The lobster sandwich is \$ 24 and although it was good it was not nearly enough to warrant that price .

\#\#\# Response: \\\midrule
\multicolumn{1}{c}{\multirow{3}{*}{\shortstack{ Event Trigger Classification \\ (ACE05)}}} & \tiny \#\#\# Instruction: 

please extract events and their types from the input sentence, all event types are in options 

\#\#\# Options:

merge organization, start organization, declare bankruptcy, end organization, grant pardon, extradite, execute, impose fine, conduct trial hearing, issue sentence, file appeal, convict, file lawsuit, release on parole, arrest and send to jail, charge and indict, acquit, participate in protest or demonstration, attack, contact via written or telephone communication, meet, start position, elect, end position, nominate, transfer ownership, transfer money, marry, divorce, be born, die, sustain injury, transport

\#\#\# Sentence:

In his previous letter home , Apache pilot Joe Bruhl did n't tell his family the full details about his first combat mission

\#\#\# Response:

tell:contact via written or telephone communication;combat:attack & \tiny \#\#\# Instruction: 

please extract events and their types from the input sentence, all event types are in options 

\#\#\# Options:

merge organization, start organization, declare bankruptcy, end organization, grant pardon, extradite, execute, impose fine, conduct trial hearing, issue sentence, file appeal, convict, file lawsuit, release on parole, arrest and send to jail, charge and indict, acquit, participate in protest or demonstration, attack, contact via written or telephone communication, meet, start position, elect, end position, nominate, transfer ownership, transfer money, marry, divorce, be born, die, sustain injury, transport

\#\#\# Sentence:

In his previous letter home , Apache pilot Joe Bruhl did n't tell his family the full details about his first combat mission

\#\#\# Response: \\\midrule
\multicolumn{1}{c}{\multirow{3}{*}{\shortstack{Text Chunking \\ (CoNLL03)}}} & \tiny
\#\#\# Instruction: 

please extract chunks and their type from the input sentence, all chunk types are in options 

\#\#\# Options:

noun phrase, verb phrase, prepositional phrase, adverb phrase, subordinated clause, adjective phrase, particles, conjunction phrase, interjection, list marker, unlike coordinated phrase

\#\#\# Sentence:

Rare Hendrix song draft sells for almost \$ 17,000 .

\#\#\# Response:

Rare Hendrix song draft:noun phrase;sells:verb phrase;for:prepositional phrase;almost \$ 17,000:noun phrase & \tiny \#\#\# Instruction: 

please extract chunks and their type from the input sentence, all chunk types are in options 

\#\#\# Options:

noun phrase, verb phrase, prepositional phrase, adverb phrase, subordinated clause, adjective phrase, particles, conjunction phrase, interjection, list marker, unlike coordinated phrase

\#\#\# Sentence:

Rare Hendrix song draft sells for almost \$ 17,000 .

\#\#\# Response: \\
    \bottomrule
    \end{tabular}
    \end{adjustbox}
    \caption{Examples for instruction tuning from four sequence labeling tasks: NER, ATE+ATP, ETC, and text chunking}
    \label{tab:it_examples}
\end{table*} 
\subsubsection{Training Small Language Models} 

For pre-training an MLM-based encoder and a CLM-based decoder, we 
randomly initialize small LMs with four encoder or decoder blocks following RoBERTa-base architecture with a language modeling head on top and a newly initialized embedding matrix with the size of RoBERTa-base's vocabulary. We inherit all other RoBERTa-base hyperparameters and produce a small LM with 68M parameters. We use the RoBERTa-base tokenizer with added prefix space to pre-process the BookCorpus dataset \citep{zhu2015aligning}, which consists of 74M sentences, and use this dataset to pre-train small LMs. Tokenized BookCorpus sentences are grouped to form chunks of size $512$. 

We train the small encoder and decoder with MLM and CLM, respectively, with AdamW \citep{loshchilov2017decoupled} optimizer, a cosine annealing learning rate scheduler, eight gradient accumulation steps, and bfloat16 precision with a batch size of $64$ and a learning rate of $2\mathrm{e}{-4}$. MLM probability is set at $0.15$. AdamW parameters are fixed to $\beta_1=0.9, \beta_2=0.95, \epsilon=1\mathrm{e}{-5}, \lambda=0.1$ and we apply gradient clipping to $1.0$. We save the model weights immediately after random initialization and then at five equally spaced intervals throughout each epoch ($51$ checkpoints). The training continues for a fixed number of $10$ epochs (roughly 200K steps). We pre-train both models with the same seed. After pre-training, we load the weights of each checkpoint, replace the LM head with the sequence labeling head for the appropriate task, and fine-tune all parameters for five epochs on a task-specific training set with a batch size of $16$. All results are averages of five runs with different seeds, and we always pick the last model for each seed.

\subsection{Evaluation}

Recall that our evaluation is conducted on IOB2 tags using strict matching and is conducted with the micro F1 score (cf.~Subsection \ref{subsubsec:evaluation_example} for more details on how this metric is calculated). Since IT models do not explicitly generate BIO tags, we adopt a greedy span-based matching approach to align predicted spans and their associated classes with the input tokens. Detailed evaluation of decoder models employed as text generators can be found in Subsection \ref{subsubsec:evaluation_decoders}.

For IT experiments, we generate the outputs for evaluation with default generation settings for the Llama2-7B model. We generate tokens using a temperature of $0.6$ and top-p sampling with a threshold of $0.9$. To speed up the generation, we decrease the total maximum length of the input instruction prompt combined with newly generated tokens to \num{1024}. The same generation configuration is used for Mistral-7B.

\section{Results} \label{sec:cm_removal_results}

We report the results for decoders combined with the \textit{layer group unmasking} method, strong encoder baselines, and instruction-tuned decoders using vanilla CLM. Additionally, we compare with SOTA models and report the results of the language model pre-training and adaptation to sequence labeling tasks on a smaller scale.

\subsection{Layer Group Unmasking Results}

\begin{figure*}
\begin{center}
\hspace*{\fill}%
\begin{subfigure}{0.25\textwidth}
    \centering
    \includegraphics[width=1.0\linewidth]{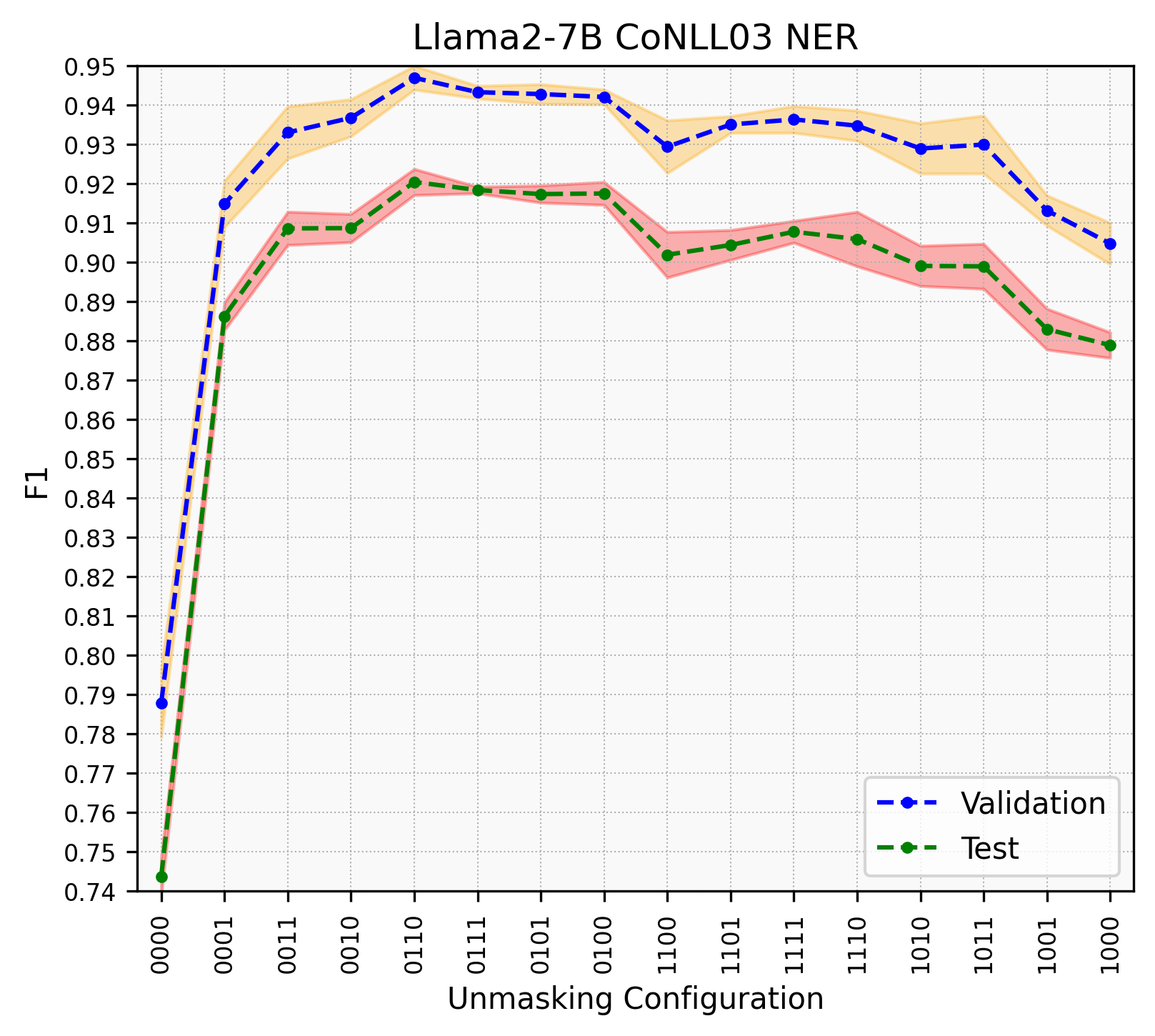} \\
    \includegraphics[width=1.0\linewidth]{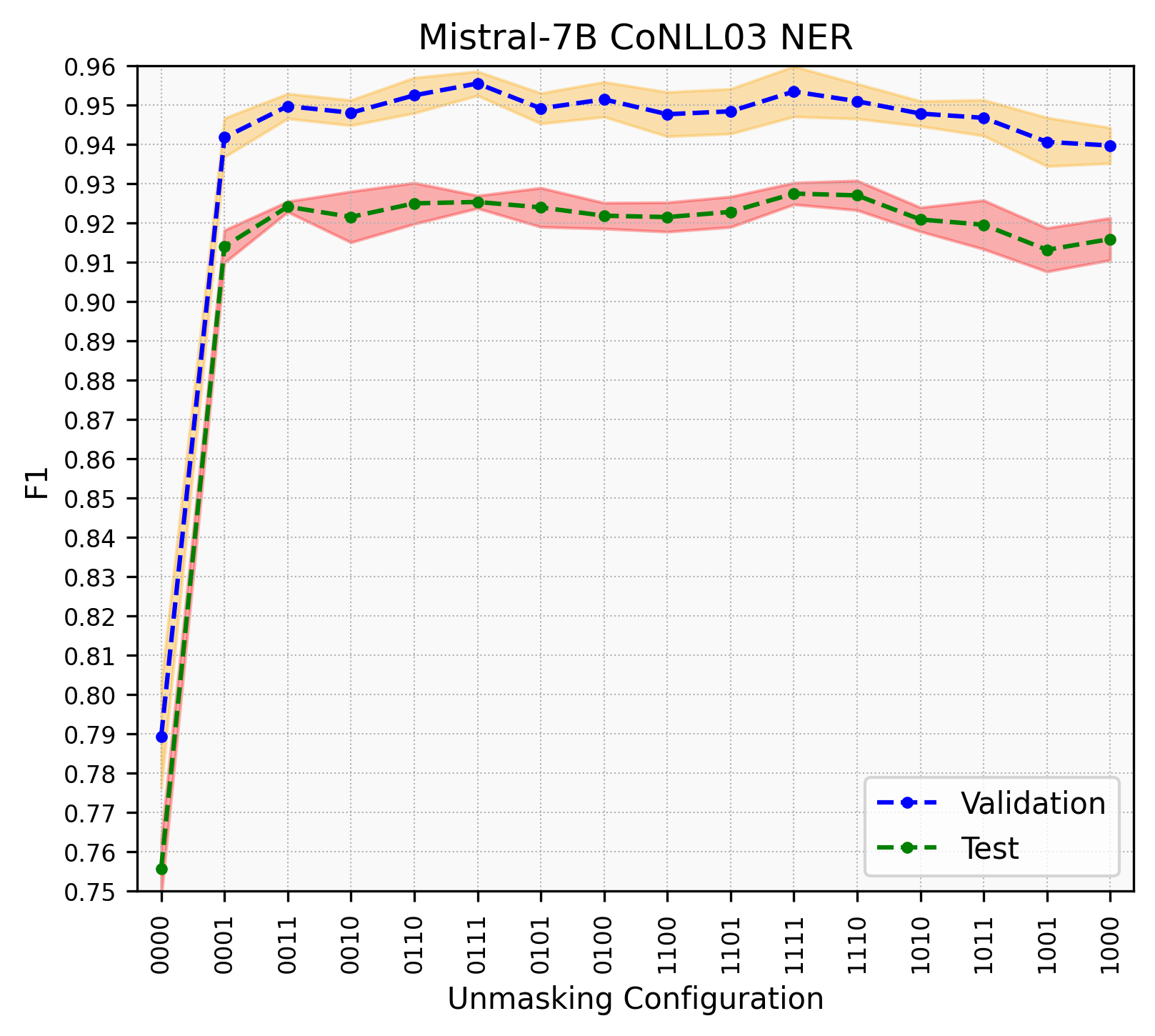}
    \caption{NER}
\end{subfigure}\hfill%
\begin{subfigure}{0.25\textwidth}
    \centering
    \includegraphics[width=1.0\linewidth]{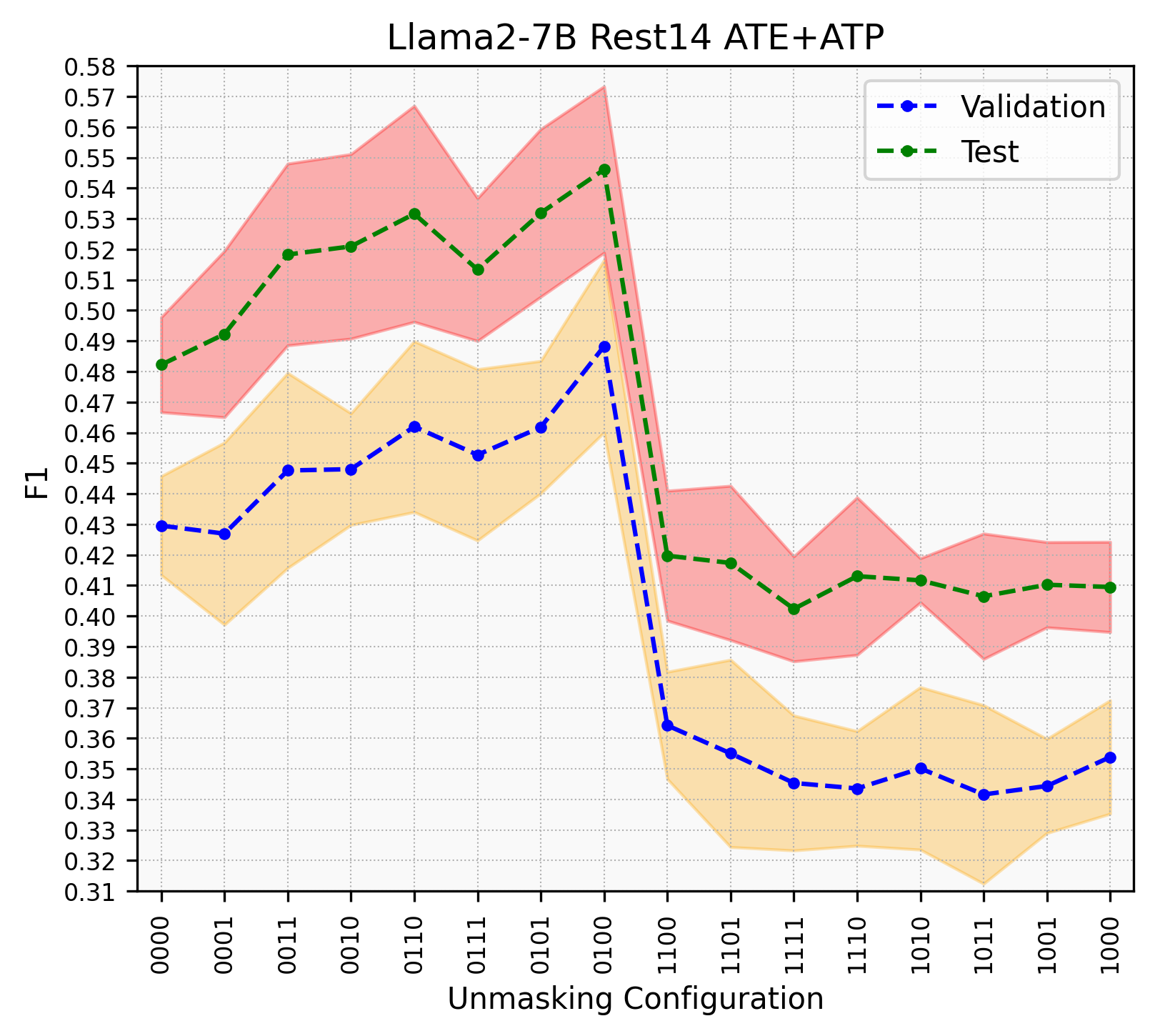} \\
    \includegraphics[width=1.0\linewidth]{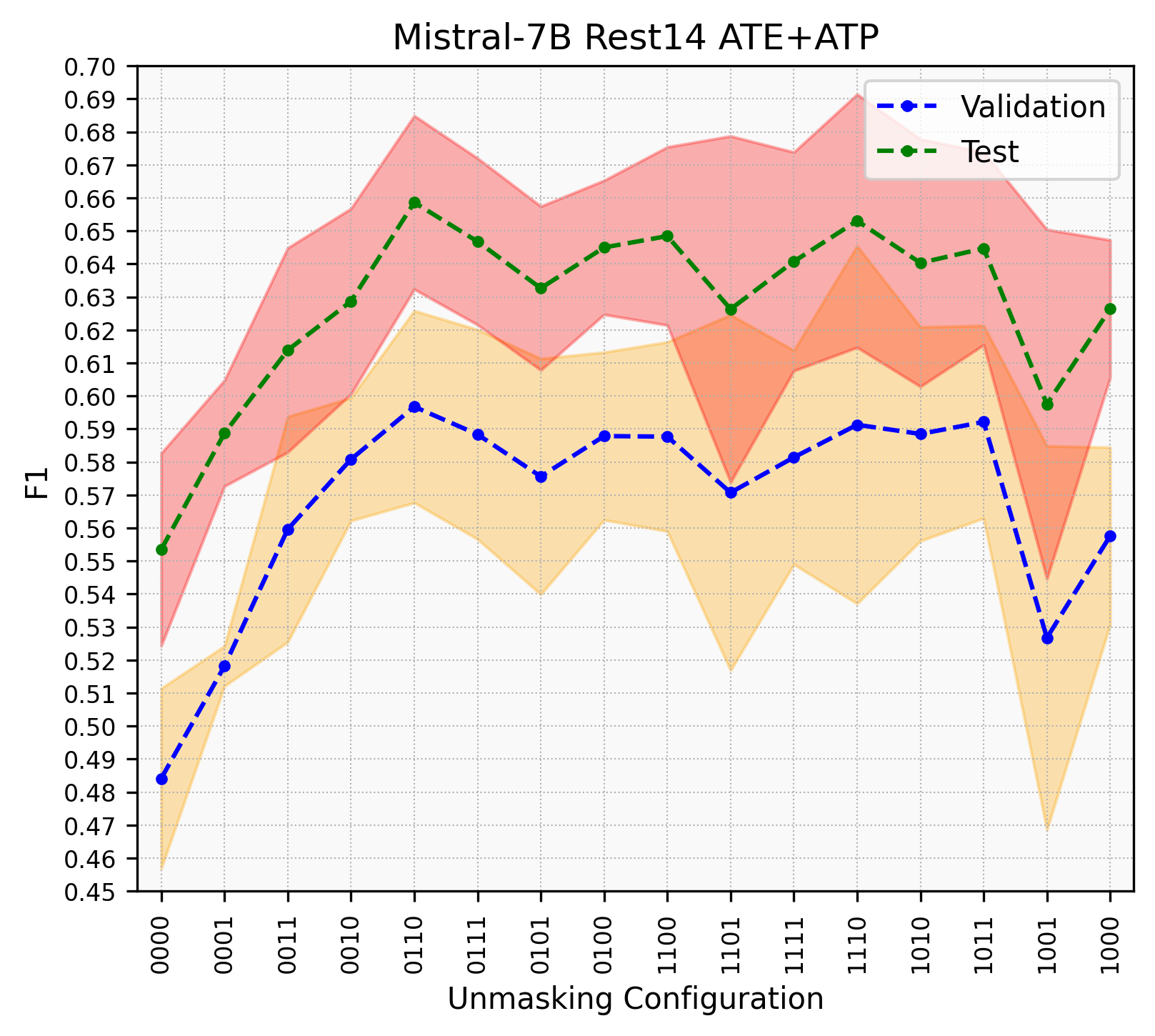}
    \caption{ATE+ATP}
\end{subfigure}\hfill%
\begin{subfigure}{0.25\textwidth}
    \centering
    \includegraphics[width=1.0\linewidth]{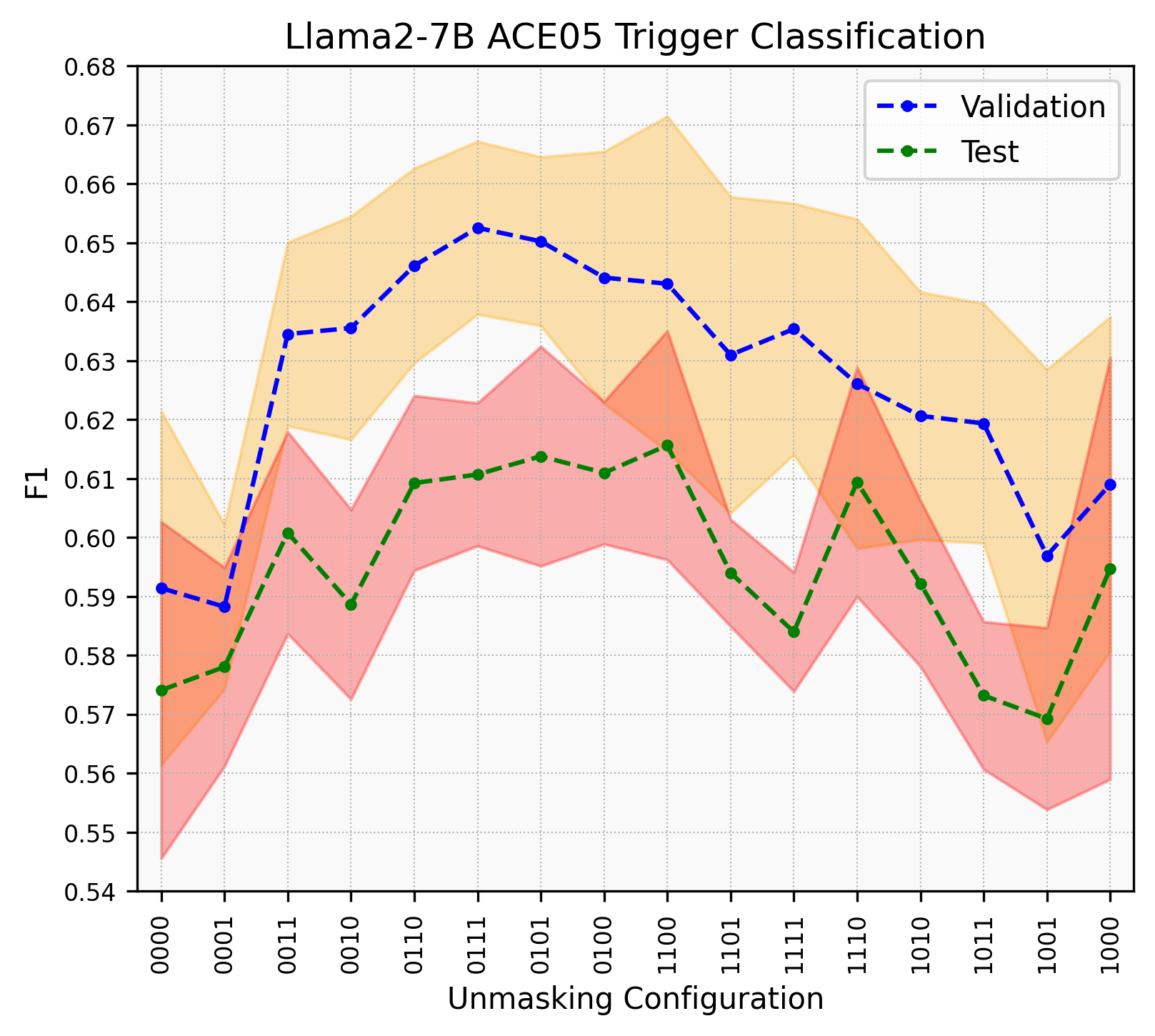} \\
    \includegraphics[width=1.0\linewidth]{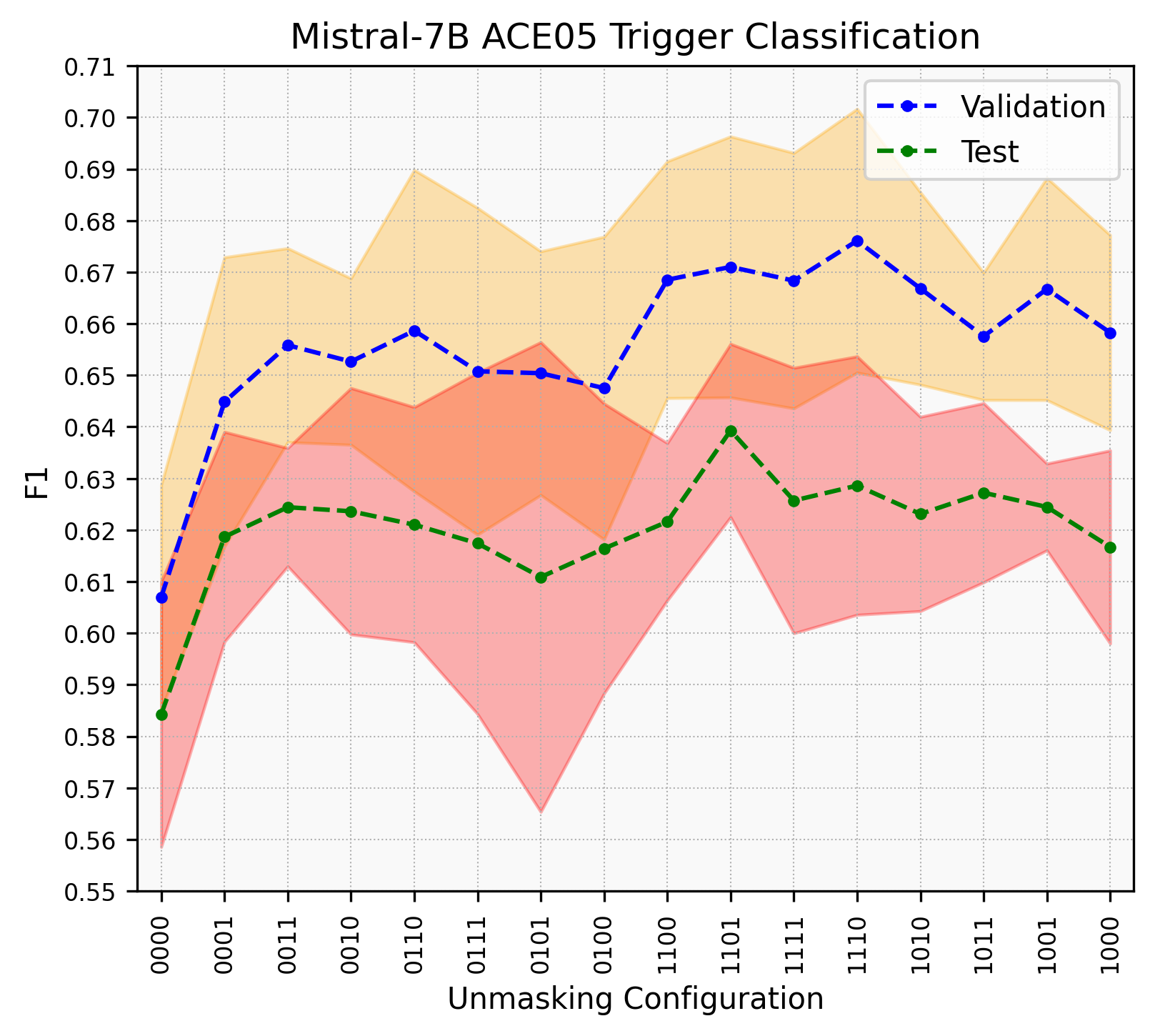}
    \caption{Trigger Classification}
\end{subfigure}\hfill%
\begin{subfigure}{0.25\textwidth}
    \centering
    \includegraphics[width=1.0\linewidth]{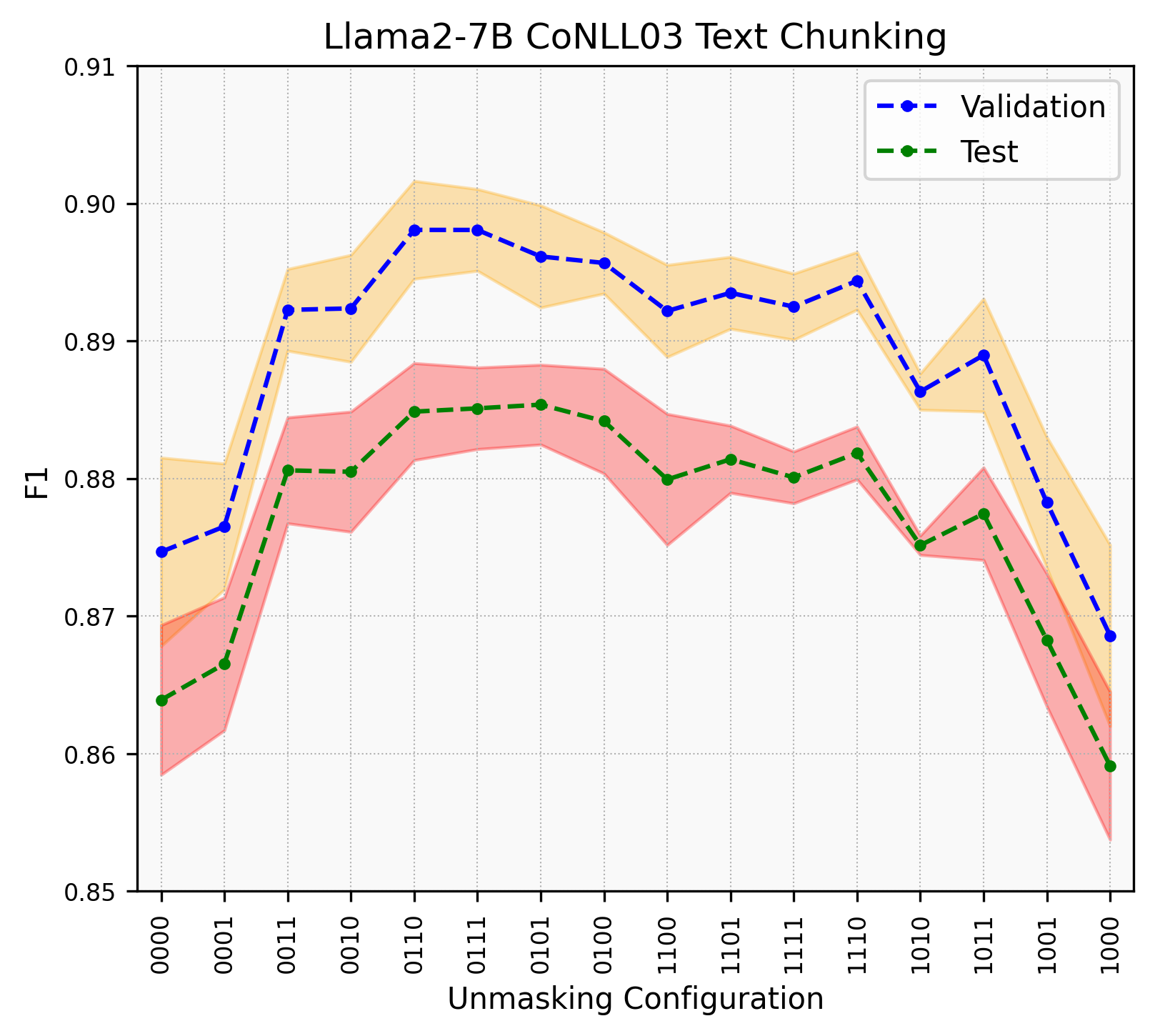} \\
    \includegraphics[width=1.0\linewidth]{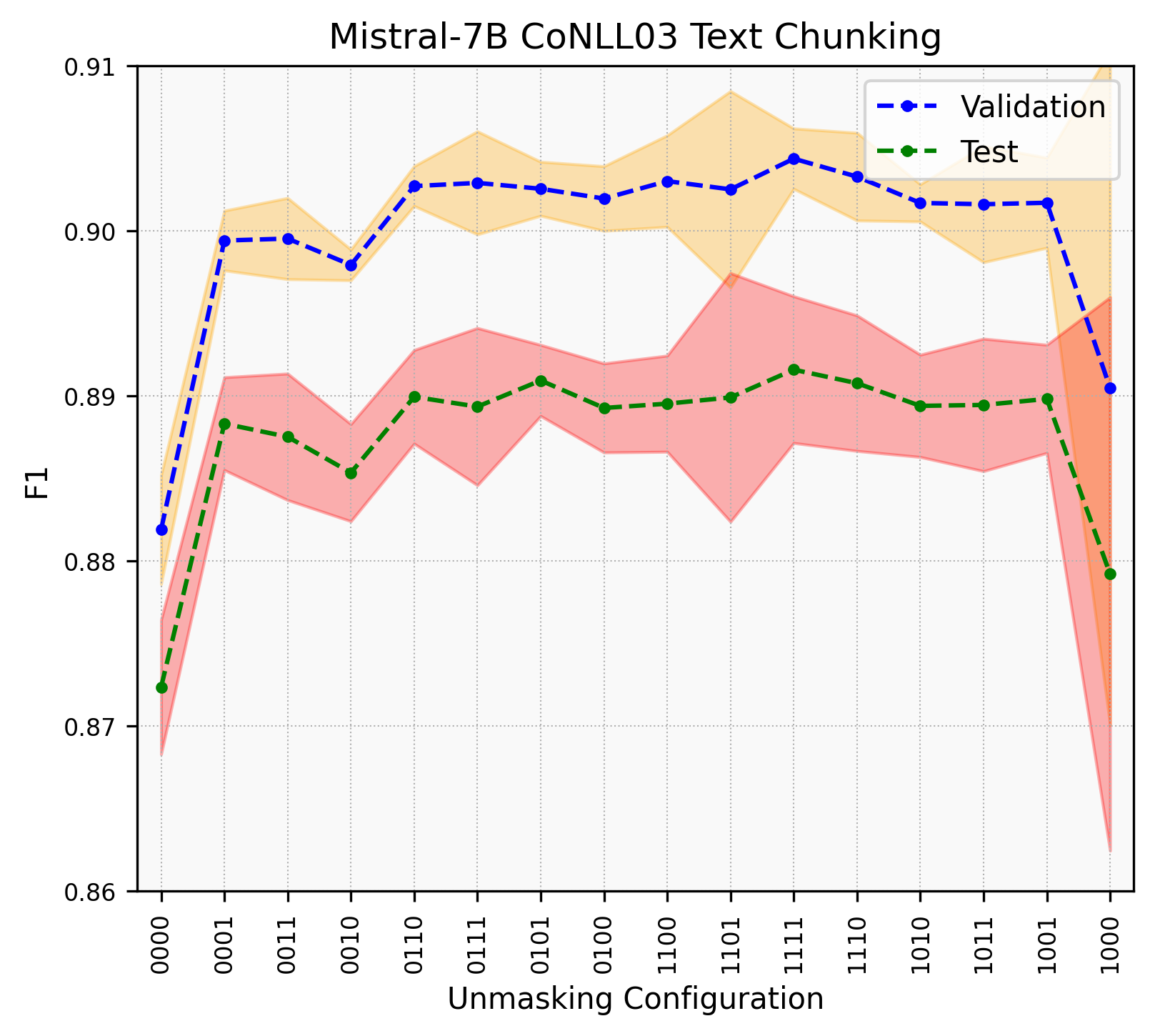}
    \caption{Text Chunking}
\end{subfigure}
\hspace*{\fill}%
\caption[Micro F1 sequence labeling scores with decoder-based LLMs and different unmasking configurations]{Micro F1 sequence labeling scores with decoder-based LLMs and different unmasking configurations (sorted by Gray code starting with all decoder layers masked---configuration 0000). Upper row plots show Llama2-7B model results, while lower row plots show Mistral-7B model results on validation and test sets of four sequence labeling tasks (left to right, one dataset per task). All results are averages of five runs. The shaded area corresponds to standard deviation.}
\label{fig:unmasking}
\end{center}
\end{figure*}

Figure~\ref{fig:unmasking} shows the validation and test evaluation sets performance of Llama2-7B and Mistral-7B for different unmasking configurations and four sequence labeling tasks. We observe a number of interesting phenomena: 

\begin{enumerate}
    \item In most cases, removing the CM in as little as one layer group significantly boosts the F1 score compared to keeping the CM in all layers (configuration $0000$);
    \item It is rarely the case that removing CM from all layer groups (configuration $1111$), compared to removing it only from some layer groups, yields the highest score on evaluation sets;
    \item Depending on the task, the boosts from the CM removal can vary (highest for NER, lowest for chunking) and deviate from minimal (NER) to substantial (ETC);
    \item Mistral-7B shows superior performance over Llama2-7B model across tasks and best configurations;
    \item The shapes of validation and test curves show high overlap across configurations for a fixed model and dataset.
\end{enumerate}

Unmasking gains are the highest for NER, and the results are overall the most consistent for CoNLL03 data (NER and chunking tasks). This aligns perfectly with the values of the RDRR score (cf.~Section \ref{sec:tasks_datasets} and Table \ref{tab:dataset_stats2}), suggesting that labeled NER spans depend more on the right context than on the left. We generally observe that unmasking layers closer to the model's output yields higher gains than unmasking layers closer to the model's input, except for ATE+ATP and ETC tasks. Such a regularity would align with the finding from the literature that higher layers in neural language models are more localized and task-specific \citep{wu-etal-2020-similarity}. To verify whether such a regularity generally holds, we conduct a one-sided two-sample t-test for the difference of F1 score means on validation and test sets per task. We compare the means of each unmasking configuration, which contains at least one unmasked layer group, with all other possible combinations where at least one additional layer group is unmasked closer to the model's output. For example, we compare the F1 score of unmasking configuration $0100$ to configurations $\{0100, 0101, 0110, 0111\}$. We repeat the same procedure for the following configurations: $0100$, $0110$, $1000$, $1010$, $1100$, and $1110$. The differences are significant for NER and chunking evaluation sets ($p<.01$) but not significant for ATE+ATP and ETC evaluation sets ($p>.01$). Applying CM removal to individual layers, rather than layer groups, could yield even larger F1 score boosts. The performance of Llama2-7B on the ATE+ATP task displays an unusual pattern where better performance is achieved when CM is preserved in all layer groups, as opposed to its complete removal. Here, allowing the model to leverage right-side context in all layers hurts the performance, which aligns with its low RDRR value. This trend is not present for any other task and model combination. Micro F1 score deviations across evaluation sets are the largest on ATE+ATP and ETC tasks, possibly due to the overall low number of examples per evaluation set for Rest14 and ACE05 datasets (less than a thousand).

\subsection{Comparison with Baselines}

Table~\ref{tab:comparison} shows the performance scores of the best unmasking configurations and the strong encoder and IT baselines. We compare against RoBERTa PLMs fine-tuned for the sequence labeling task and instruction-tuned Llama2-7B and Mistral-7B LLMs. The standard $0000$ and $1111$ configurations are compared with the best configurations per model and task. We also report the Pearson correlation coefficient $\rho$ between all validation and test unmasking configurations. For all sequence labeling tasks and the Llama2-7B model, we observe a consistent improvement of the best configurations over configuration $1111$. Similar holds for Mistral-7B, except for chunking, where $1111$ yields the highest F1 scores. Conforming to the findings from \cite{scaria-etal-2024-instructabsa}, IT is most beneficial for the ATE+ATP task, outperforming any unmasking configuration on evaluation sets. RoBERTa-large surpasses all other models on the ACE05 validation set but fails to do so on the test set. RoBERTa baselines achieve high scores on sequence labeling tasks, except for ATE+ATP. Training with QLoRA combined with \textit{layer group unmasking} creates high-performing and compact models, requiring a small number of additional parameters trained for each sequence labeling task. Correlations between configurations on evaluation sets are high for each task, while models trained on ACE05 exhibit the lowest $\rho$. A high overall $\rho$ indicates that the optimal unmasking configuration can be determined using the validation set.

\begin{table*}
\centering
\begin{adjustbox}{width=1.0\linewidth}
\small{\begin{tabular}{l|cccccccccc}
\toprule
\multicolumn{3}{c}{\multirow{1}{*}{\textbf{Model}}} & \multicolumn{2}{c}{\textbf{CoNLL03 NER}} & \multicolumn{2}{c}{\textbf{Rest14 ATE+ATP}} & \multicolumn{2}{c}{\textbf{ACE05 Trigger Clf.}} 
& \multicolumn{2}{c}{\textbf{CoNLL03 Chunking}}  \\
\midrule
\multicolumn{3}{c}{\multirow{2}{*}{Llama2-7B-SL}}     &  Valid F1 & Test F1 &  Valid F1 & Test F1 &  Valid F1 & Test F1 &  Valid F1 & Test F1  \\
\cmidrule(lr){4-4} \cmidrule(lr){5-5} \cmidrule(lr){6-6} \cmidrule(lr){7-7} \cmidrule(lr){8-8} \cmidrule(lr){9-9} \cmidrule(lr){10-10} \cmidrule(lr){11-11} \multicolumn{3}{c}{} & \multicolumn{2}{c}{$\rho=0.999$} & \multicolumn{2}{c}{$\rho=0.994$} & \multicolumn{2}{c}{$\rho=0.806$} & \multicolumn{2}{c}{$\rho=0.998$} \\
\cmidrule(lr){1-3} \cmidrule(lr){4-5} \cmidrule(lr){6-7} \cmidrule(lr){8-9} \cmidrule(lr){10-11}
\multirow{7}{*}{\rotatebox[origin=c]{90}{\shortstack{Unmask Config.}}} 
& & \multicolumn{1}{l}{0000} & 0.788 & 0.744 & 0.430 & 0.482 & 0.591 & 0.574 & 0.875 & 0.864 \\
& & \multicolumn{1}{l}{1111} & 0.936 & 0.908 & 0.345 & 0.402 & 0.635 & 0.584 & 0.892 & 0.880 \\
\cmidrule(lr){2-3}
& \multirow{5}{*}{\rotatebox[origin=c]{90}{\shortstack{Best}}} & \multicolumn{1}{l}{0100} & -- & -- & 0.488 & 0.546 & -- & -- & -- & -- \\
& & \multicolumn{1}{l}{0101} & -- & -- & -- & -- & -- & -- & -- & 0.885 \\
& & \multicolumn{1}{l}{0110} & 0.947 & 0.920 & -- & -- & -- & -- & 0.898 & -- \\
& & \multicolumn{1}{l}{0111} & -- & -- & -- & -- & 0.653 & -- & -- & -- \\
& & \multicolumn{1}{l}{1100} & -- & -- & -- & -- & -- & 0.616 & -- & -- \\
\cmidrule(lr){2-3}
\multicolumn{3}{c}{\multirow{2}{*}{Mistral-7B-SL}}         &  Valid F1 & Test F1 &  Valid F1 & Test F1 &  Valid F1 & Test F1 &  Valid F1 & Test F1  \\
\cmidrule(lr){4-4} \cmidrule(lr){5-5} \cmidrule(lr){6-6} \cmidrule(lr){7-7} \cmidrule(lr){8-8} \cmidrule(lr){9-9} \cmidrule(lr){10-10} \cmidrule(lr){11-11} \multicolumn{3}{c}{} & \multicolumn{2}{c}{$\rho=0.999$} & \multicolumn{2}{c}{$\rho=0.983$} & \multicolumn{2}{c}{$\rho=0.899$} & \multicolumn{2}{c}{$\rho=0.994$} \\
\cmidrule(lr){1-3} \cmidrule(lr){4-5} \cmidrule(lr){6-7} \cmidrule(lr){8-9} \cmidrule(lr){10-11}
\multirow{6}{*}{\rotatebox[origin=c]{90}{\shortstack{Unmask Config.}}} 
& & \multicolumn{1}{l}{0000} & 0.789 & 0.756 & 0.484 & 0.553 & 0.607 & 0.584 & 0.882 & 0.872 \\
& & \multicolumn{1}{l}{1111} & 0.953 & \textbf{0.927} & 0.581 & 0.641 & 0.668 & 0.626 & \textbf{0.904} & \textbf{0.892} \\
\cmidrule(lr){2-3}
& \multirow{4}{*}{\rotatebox[origin=c]{90}{\shortstack{Best}}} & \multicolumn{1}{l}{0110} & -- & -- & 0.597 & 0.659 & -- & -- & -- & -- \\
& & \multicolumn{1}{l}{0111} & \textbf{0.956} & -- & -- & -- & -- & -- & -- & -- \\
& & \multicolumn{1}{l}{1101} & -- & -- & -- & -- & -- & \textbf{0.639} & -- & -- \\
& & \multicolumn{1}{l}{1110} & -- & \textbf{0.927} & -- & -- & 0.676 & -- & -- & -- \\
\cmidrule(lr){2-3}
\midrule
\multicolumn{3}{c}{RoBERTa-base-SL}  & 0.897 & 0.883 & 0.313 & 0.369 & 0.609 & 0.508 & 0.889 & 0.877     \\
\multicolumn{3}{c}{RoBERTa-large-SL} & 0.924 & 0.900 & 0.403 & 0.474 & \textbf{0.698} & 0.628 & 0.891 & 0.877    \\
\multicolumn{3}{c}{Llama2-7B-IT}     & 0.778 & 0.771 & 0.523 & 0.608 & 0.375 & 0.347 & 0.833 & 0.818   \\
\multicolumn{3}{c}{Mistral-7B-IT}    & 0.897 & 0.887 & \textbf{0.646} & \textbf{0.733} & 0.477 & 0.461 & 0.873 & 0.860  \\
\bottomrule
\end{tabular}}
\end{adjustbox}
\caption[Validation and test micro F1 sequence labeling scores for Llama2-7B and Mistral-7B models with various unmasking configurations]{Validation and test micro F1 sequence labeling scores for Llama2-7B and Mistral-7B models with various unmasking configurations ($0000$, $1111$, and other configurations which surpass the F1 score of $1111$ over sequence labeling datasets---denoted as \textit{Best}) are in the upper table part. The results for sequence labeling encoders and IT baselines are in the lower table part. The best results by dataset and evaluation set are in bold. For Llama2-7B and Mistral-7B, we report the Pearson correlation coefficient between validation and test unmasking configurations ($\rho$). All results are averages over five runs.}
\label{tab:comparison}
\end{table*} 
\subsection{Comparison with SOTA Models}

Our results are competitive with SOTA. However, a fair comparison is challenging due to differences in training and evaluation. For example, the SOTA F1 score for CoNLL03 NER, reported by \cite{wang-etal-2021-automated}, is $0.946$. This result, however, was obtained using a model trained on the merged training and validation sets. Strong results on CoNLL03 NER were reported by \cite{liu-etal-2022-autoregressive}, reaching an F1 score of $0.941$ without task-specific feature engineering, relying solely on a conditional language model with explicit modeling of the target structure. Further, the authors whose work we build upon, \cite{li2023label}, report SOTA results on CoNLL03 NER. The score they report is $0.932$, which is competitive, but not SOTA. Upon code inspection, we found that they truncate sequences longer than $64$ tokens to fit the data into GPU memory. In contrast, we managed to keep the sequence length at $128$ tokens. 

These decisions have a significant impact on overall performance. Although the reported results in related work are SOTA or close to SOTA, these studies are representative of a common problem in the field, namely the fact that the important evaluation details are not always appropriately communicated. The authors of \cite{scaria-etal-2024-instructabsa} achieve SOTA F1 score of $0.928$ on Rest14 dataset and ATE task with IT. Furthermore, the authors of \cite{yang-li-2024-modeling} report SOTA macro F1 of $0.870$ on ATP task with DeBERTa model \cite{he2020deberta}. Finally, the SOTA F1 score of $0.698$ on ACE05 for ETC is achieved by \cite{wang-etal-2022-deepstruct}. Key choices---such as micro vs.~macro F1 score, token- vs.~span-based evaluation, and evaluating the model prediction on the first token of tokenized words from the input sequence vs.~on all tokens---are often not explicitly stated or justified in papers that report SOTA results.

\subsection{Investigating the Effect of Scale}

We observed gains upon CM removal on a 7B parameters scale. Our experiments prompt the question of whether similar findings would apply to models of different parameter scales. More specifically, whether CM removal from a small CLM-based decoder would exhibit the same trend, and also whether MLM-based pre-training is more beneficial for success on sequence labeling tasks when the number of parameters, pre-training data and steps, and all other hyperparameters between decoders and encoders are equal. To investigate this, we consider two models of comparable size---a small MLM-based encoder and a CLM-based decoder. We pre-train the encoder and decoder and fine-tune three variants: an encoder, a decoder with CM during fine-tuning, and a decoder without CM. We report averages over five fine-tuning runs, with the last model from each run evaluated on the validation set. 

\begin{figure}
\begin{center}
\includegraphics[width=0.5\columnwidth]{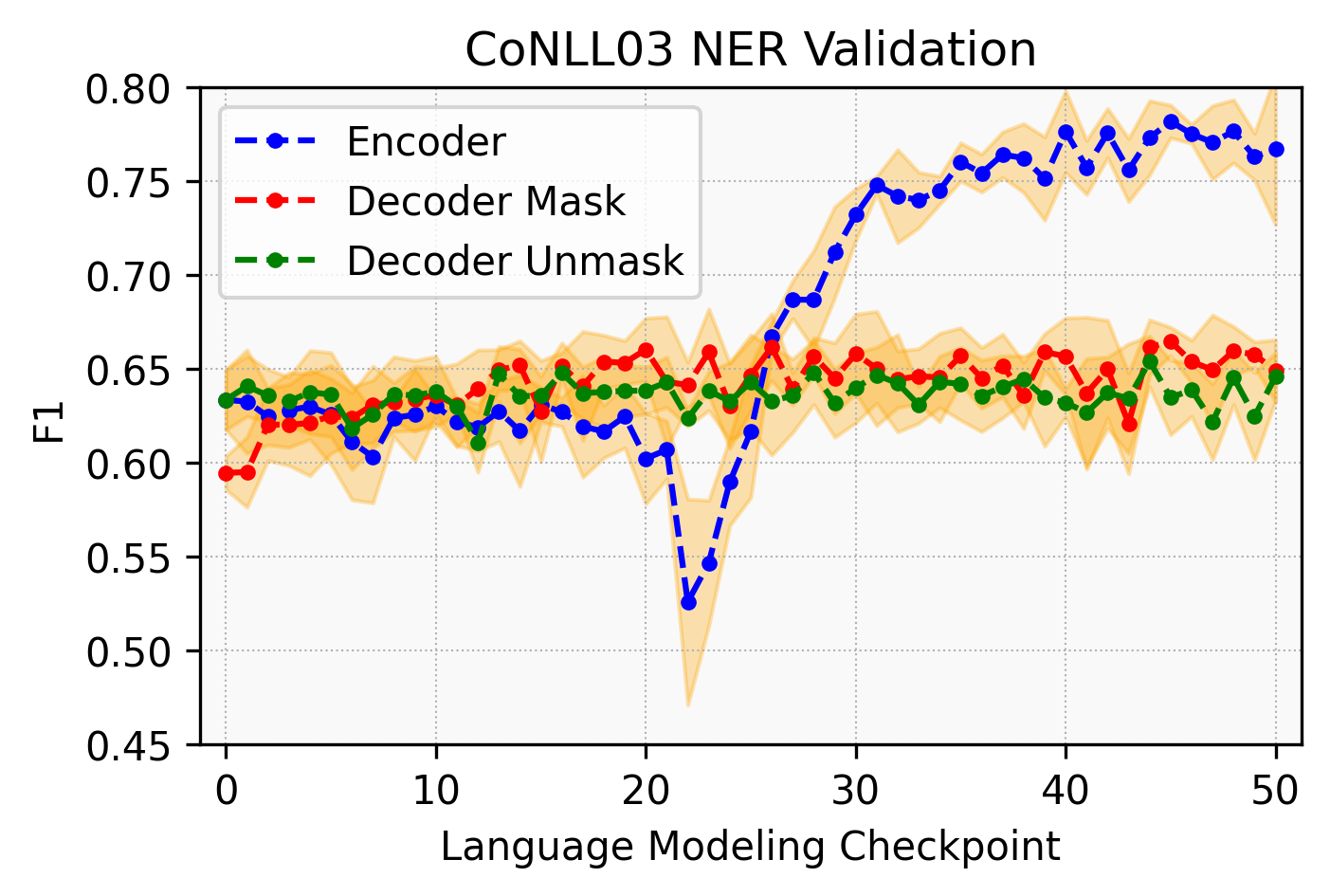}%
\includegraphics[width=0.5\columnwidth]{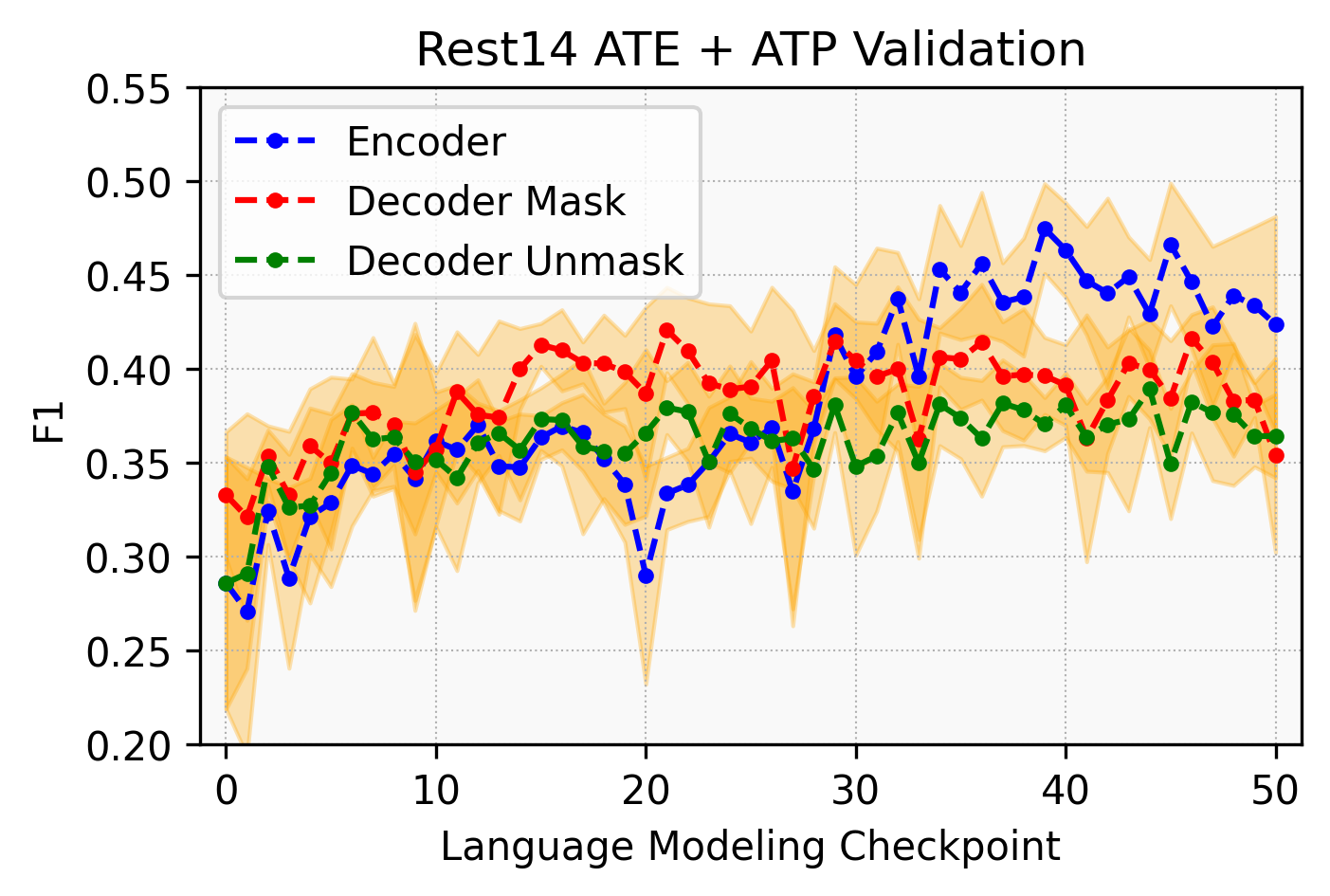}\\
\includegraphics[width=0.5\columnwidth]{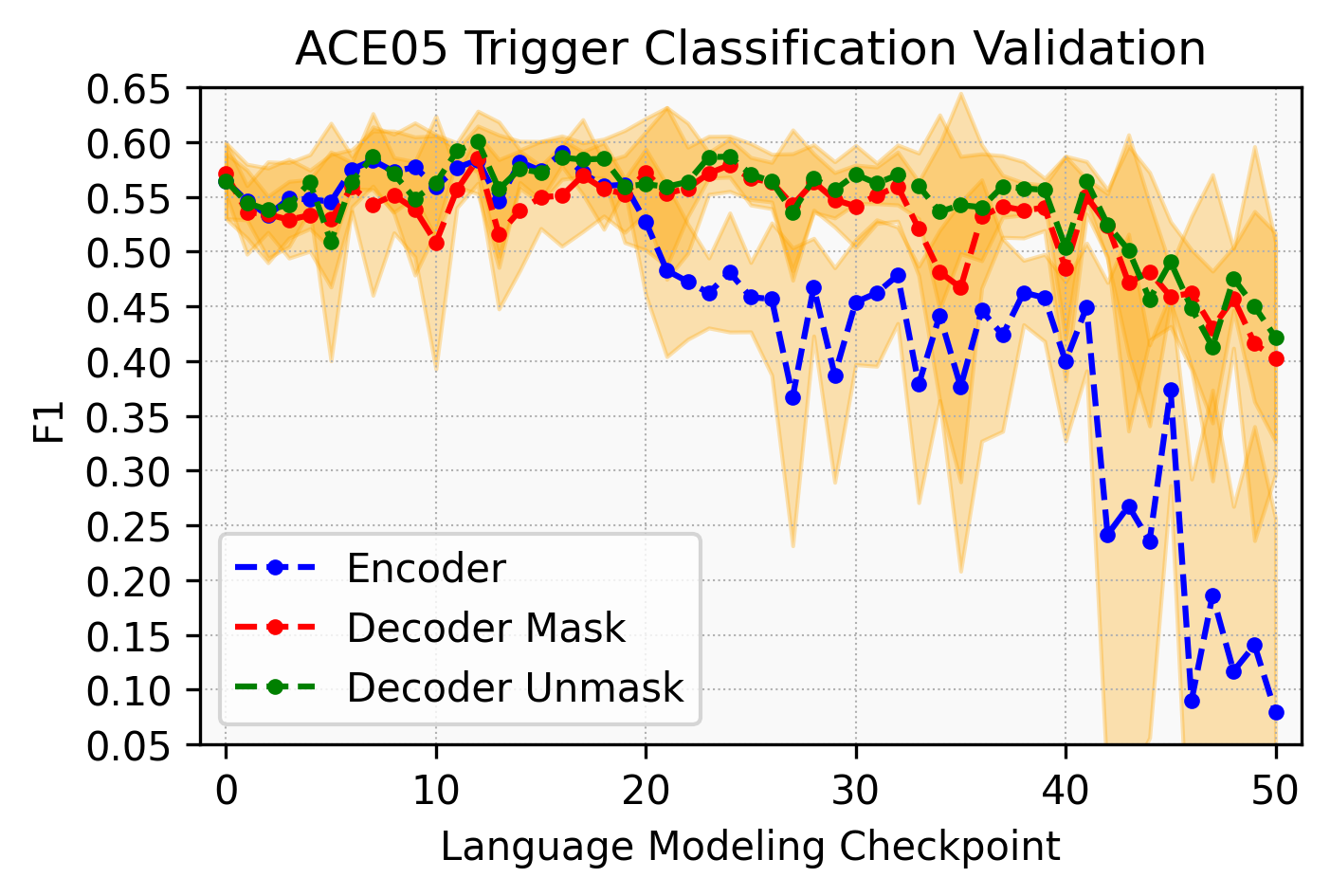}%
\includegraphics[width=0.5\columnwidth]{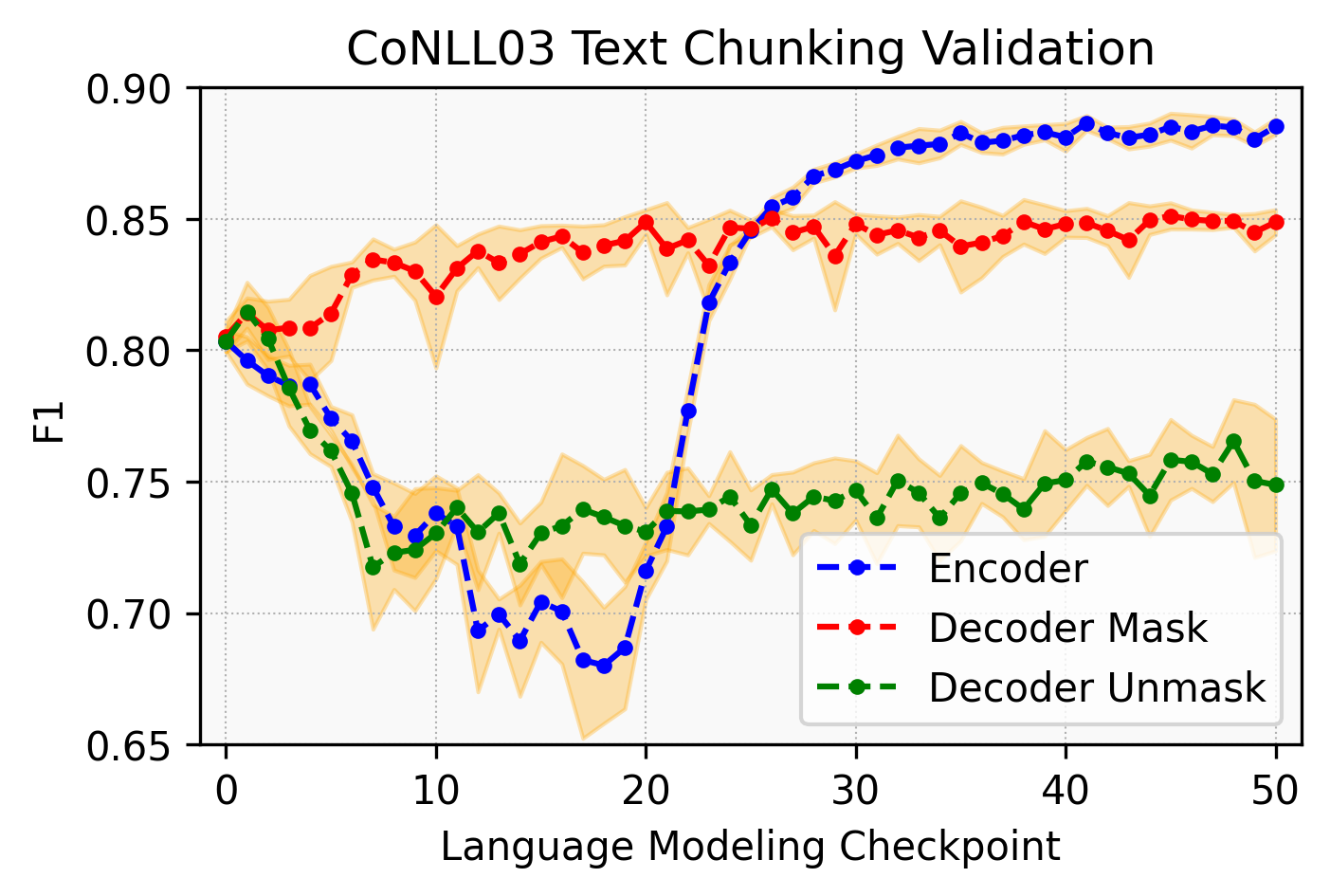}
\caption[Validation set micro F1 sequence labeling scores of small pre-trained encoders and decoders]{Validation set micro F1 sequence labeling scores of small pre-trained MLM-based encoder and CLM-based decoder models after fine-tuning on sequence labeling tasks training data starting from a particular language modeling checkpoint. Decoder Unmask: model pre-trained with CLM and a CM but fine-tuned without the CM. All results are averages over five runs. The shaded area represents the standard deviation.}
\label{fig:pretraining_roberta}
\end{center}
\end{figure}

The results in Figure~\ref{fig:pretraining_roberta} reveal that removing the CM on the 68M parameters scale produces no gains. On average, Decoder Unmask performs worse than Decoder Mask. Moreover, encoders struggle to keep up with decoders until around the 20th checkpoint (fourth pre-training epoch), when they start prevailing on all sequence labeling tasks except ETC. MLM training for over four epochs drastically hurts performance on ACE05. The score RoBERTa-base achieves on ACE05 (cf.~Table~\ref{tab:comparison}) is close to the best small encoder. The gains from continued MLM training on ACE05 become less important than the overall number of model parameters. This can be explained by the fact that RoBERTa-base and RoBERTa-large were both trained for 500k steps \citep{liu2019roberta}, and RoBERTa-large is drastically better on ACE05.

\subsection{Summary}

A series of experiments showed that LLMs can yield competitive performance with SOTA on sequence labeling tasks when the CM is removed from specific decoder blocks. The \textit{layer group unmasking} method offers a simple yet highly effective approach for transforming decoders pre-trained with CLM into encoders. This method requires only a few thousand annotated examples to adapt the decoder to the sequence labeling task. 

Addressing RQ2, our experiments demonstrated that selectively removing the CM from a subset of decoder layers effectively narrows the pre-train--fine-tune discrepancy, resulting in significant performance improvements in downstream sequence labeling tasks. Given the LLMs' scale, with billions of parameters, the \textit{layer group unmasking} method required the use of PEFT methods to ensure computational feasibility. The proposed unmasking method encodes transfer learning knowledge into the PEFT module while keeping the PLM weights unchanged. This approach preserves the integrity of the original model while enabling efficient adaptation by introducing and fine-tuning new parameters that are only a fraction of the size (often less than $1\%$) of PLM's total parameter number. In that way, the proposed CM removal method also successfully demonstrates that targeted architectural modifications combined with PEFT can indeed transform decoders into encoders that deliver substantial performance gains on downstream sequence labeling tasks.

Selecting the optimal unmasking configuration is critical for maximizing generalization performance, particularly in achieving significant improvements over baseline models without CM removal. The RDRR metric serves as an oracle for predicting the degree of right-side dependency in a task. Tasks with annotated data that result in low RDRR values show fewer benefits from removing CM across many decoder blocks. Comparisons with encoder-based models stress the importance of architecture and scale in sequence labeling tasks. Notably, a substantial performance gap exists between small-scale encoder and decoder models, with decoder models at small scales deriving little to no benefit from CM removal.

\clearpage{}%
\clearpage{}%
\chapter{Supervised In-context Fine-tuning} \label{ch:method3}

Decoder-only LLMs are powerful but not trivial to use for sequence labeling. As we have seen in Chapter \ref{ch:method2}, using decoders as encoders by fiddling with the causal mask across decoder layers can yield significant gains. However, using decoders as text generators to solve sequence labeling tasks would be beneficial, as this does not trigger the undesirable consequences of pre-train--fine-tune discrepancy (cf.~Subsection \ref{subsec:enc_dec}). To this end, we explore how autoregressive LLMs can be adapted effectively for sequence labeling through targeted modifications and extensions of (standard) supervised fine-tuning (SFT) methods under a unified framework. We build the framework by relying on the third pillar of model adaptation---choosing the appropriate optimization scheme (cf.~Subsection \ref{subsubsec:optim_scheme}) and aiming at addressing the RQ3: \textit{How to improve sequence labeling with decoder-only LLMs within a unified framework?} (cf.~Chapter \ref{ch:intro}). In line with this, we investigate how to leverage best multiple demonstrations in the context for improved ICL capabilities and also compare with the method from the previous chapter. The following sections present motivation, models, experiments, and results for a unified framework for adapting autoregressive LLMs using modifications and extensions of SFT methods with many demonstrations in the context.

\section{Motivation}

ICL is a powerful technique for employing LLMs, providing LLMs with task-specific demonstrations in the prompt. This surprising phenomenon has emerged with increasing the scale of pre-training decoder-only LLMs to billions of parameters \citep{brown2020language,wies2023learnability}. With ICL, the model can perform tasks unseen during pre-training and infer (learn) over the provided description of a task (\textit{instruction}) and demonstrations (examples and responses). The core advantage of ICL over parameter-updating methods lies in its reliance on inference alone, eliminating the need for fine-tuning to solve novel tasks. This way, ICL achieves lower computational costs relative to performance gains, which has driven its widespread adoption in NLP.

ICL can be categorized by the number of demonstrations provided in the context. With earlier models, ICL was limited to \textit{few-shot ICL} \cite{mosbach2023few},\footnote{\footnotesize{For simplicity, we consider the case of one demonstration in the context as part of few-shot ICL.}} as their smaller context could only accommodate a \textit{few} demonstrations. More recent models, however, with context lengths in the tens or hundreds of thousands of tokens, now enable many-shot ICL \cite{agarwal2024many}. When there are no labeled instances in the context and the model receives no explicit demonstrations and relies entirely on natural language instructions to interpret the task, \textit{zero-shot inference} is used as a term in the literature. Few-shot learning utilizes multiple demonstrations in the input to guide the model's inference. LLMs typically improve performance when presented with more demonstrations, especially for datasets with large label spaces \citep{bertsch-etal-2025-context}. When the model can generalize to unseen instructions, this is referred to as \textit{instruction following} \citep{hewitt2024instruction}. We can also provide the model with examples but no expected responses in an unsupervised ICL scenario \citep{min-etal-2022-rethinking,lampinen2024broader}.

ICL has been compared to fine-tuning in the literature numerous times \citep{duan-etal-2024-exploring, mosbach-etal-2023-shot, yin-etal-2024-deeper}. As opposed to fine-tuning, ICL does not change the model's parameters. When comparing ICL in decoders with fine-tuning, we typically compare SFT with ICL methods. Both SFT and ICL rely on labeled task-specific examples. ICL leverages labeled examples, called demonstrations, by incorporating them into the prompt without altering the model’s parameters in any way. SFT, in contrast, uses labeled examples as actual training instances based on which a loss function is computed and the model's parameters are updated. While SFT can leverage a virtually unlimited number of labeled instances, ICL is constrained by the LLM's context size, limiting the number of demonstrations.

In both ICL and SFT, one can start the context with an instruction, that is, a natural language description of the task. This, intuitively, makes more sense if the underlying model is an instruction-tuned LLM, that is, a model that has been fine-tuned on instruction-response pairs spanning many diverse tasks \cite{mishra-etal-2022-cross}, although instruction following may also emerge without instruction-tuning \cite{hewitt2024instruction}. While ICL is more flexible (i.e., no updates to the underlying LLM) and computationally cheaper than SFT, it typically cannot reach the performance of task-specific SFT: fine-tuning beats long-context ICL once presented with enough training examples \citep{bertsch-etal-2025-context}. These observations suggest that an effective sequence labeling strategy may integrate ICL and SFT by leveraging their complementary strengths. Therefore, we explore the potential of \textit{generative sequence labeling with LLMs} in this chapter and introduce response adaptation strategies for supervised in-context fine-tuning (SIFT)---a framework that unifies SFT and ICL for sequence labeling.  

\begin{figure*}
    \centering
    \includegraphics[width=1.0\linewidth]{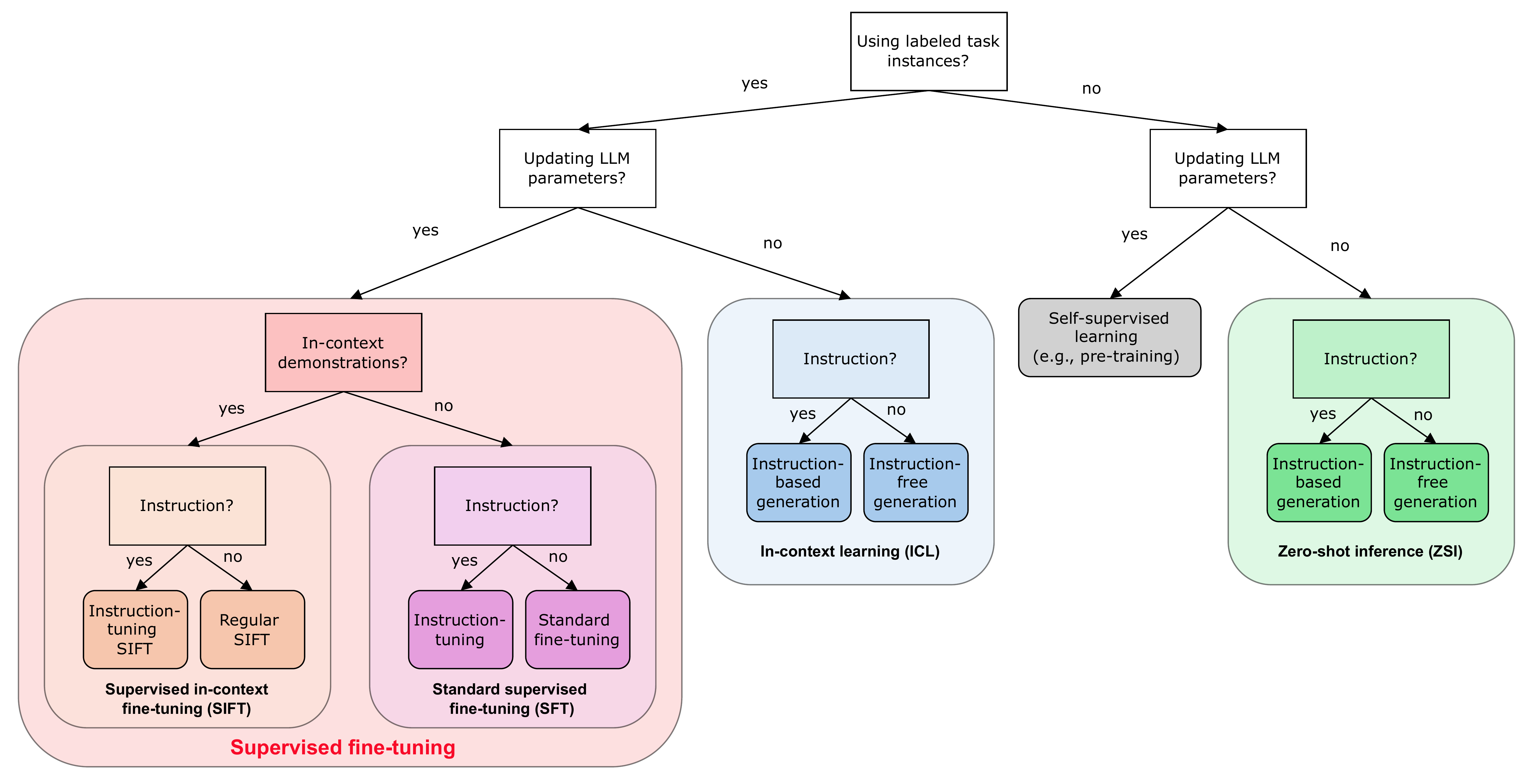}
    \caption[Supervised in-context fine-tuning in relation to (standard) supervised fine-tuning and in-context learning]{Supervised in-context fine-tuning (SIFT; in orange box) as a task-specific learning paradigm for LLMs, in relation to (standard) supervised fine-tuning (SFT; in purple box) and in-context learning (ICL; in blue box). For completion, zero-shot inference (no labeled instances) is shown in the green box.}
    \label{fig:sft_icl_tax}
\end{figure*}

As the fields of ICL and SFT evolve rapidly, the community faces challenges in establishing precise nomenclature and systematically categorizing available options for the two paradigms. These paradigms vary based on (1) the inclusion of task instructions in prompts, (2) the number of demonstrations used, and (3) whether parameter updates are performed or omitted entirely. We fill this gap by introducing the taxonomy for SFT and ICL in Figure \ref{fig:sft_icl_tax}, pointing out key differences, introducing a more precise overview of the interplay between the two paradigms, and ultimately situating SIFT within the broader landscape of LLM learning approaches.

SIFT---shown in the orange box in the decision tree in Figure~\ref{fig:sft_icl_tax}---can be seen as a hybrid between SFT and ICL: we update the model's parameters (as in SFT) based on the labeled examples in the context, but the context, besides the last, \textit{target} instance, additionally contains in-context demonstrations (as in ICL). Like both standard SFT and ICL, SIFT may or may not include the task instruction at the beginning of the prompt. 

The field has explored SIFT as a \textit{learning to learn in context} paradigm, explicitly improving ICL \citep{zhuang2024vector}. The paradigm was introduced first under pseudonyms \textit{meta ICL} \citep{min-etal-2022-metaicl} and \textit{in-context tuning} \citep{chen-etal-2022-meta}. Few-shot learning was also explored and compared with ICL on slot labeling and NER tasks \citep{chen-etal-2023-learning,li2023task,razumovskaia2024analyzing}. However, these works employ SIFT in combination with vanilla causal language modeling (CLM). Some works propose fine-tuning the model on the demonstration and not on the instruction in the prompt, but do not conduct few-shot learning \citep{an2024response,hewitt2024instruction}. Based on these insights, we build on the SIFT for sequence labeling tasks and advance the field with a framework for response-oriented adaptation strategies tailored for decoders. We dub the adaptation strategies inside the proposed framework as \textit{response-oriented} since we introduce modifications into the CLM loss function that amplify the response tokens' contribution during the adaptation phase of transfer learning. Finally, we also experiment with the effect of the instruction on fine-tuning and performance in different adaptation scenarios, training the models without the instruction and performing the inference with variations of the instruction included.

What remains to be defined for a concrete SIFT training run is the actual learning objective. Since we focus on decoder-based LLMs, the objective has to be generative, i.e., token prediction. However, the question of which tokens in the context to predict, including the (optional) instruction, demonstrations, and the last target instance, remains unanswered. We investigate three different SIFT training strategies for sequence labeling tasks described in detail in the next section.           

\section{Framework} \label{sec:framework}

\begin{figure*}
    \centering
    \includegraphics[width=1.0\linewidth]{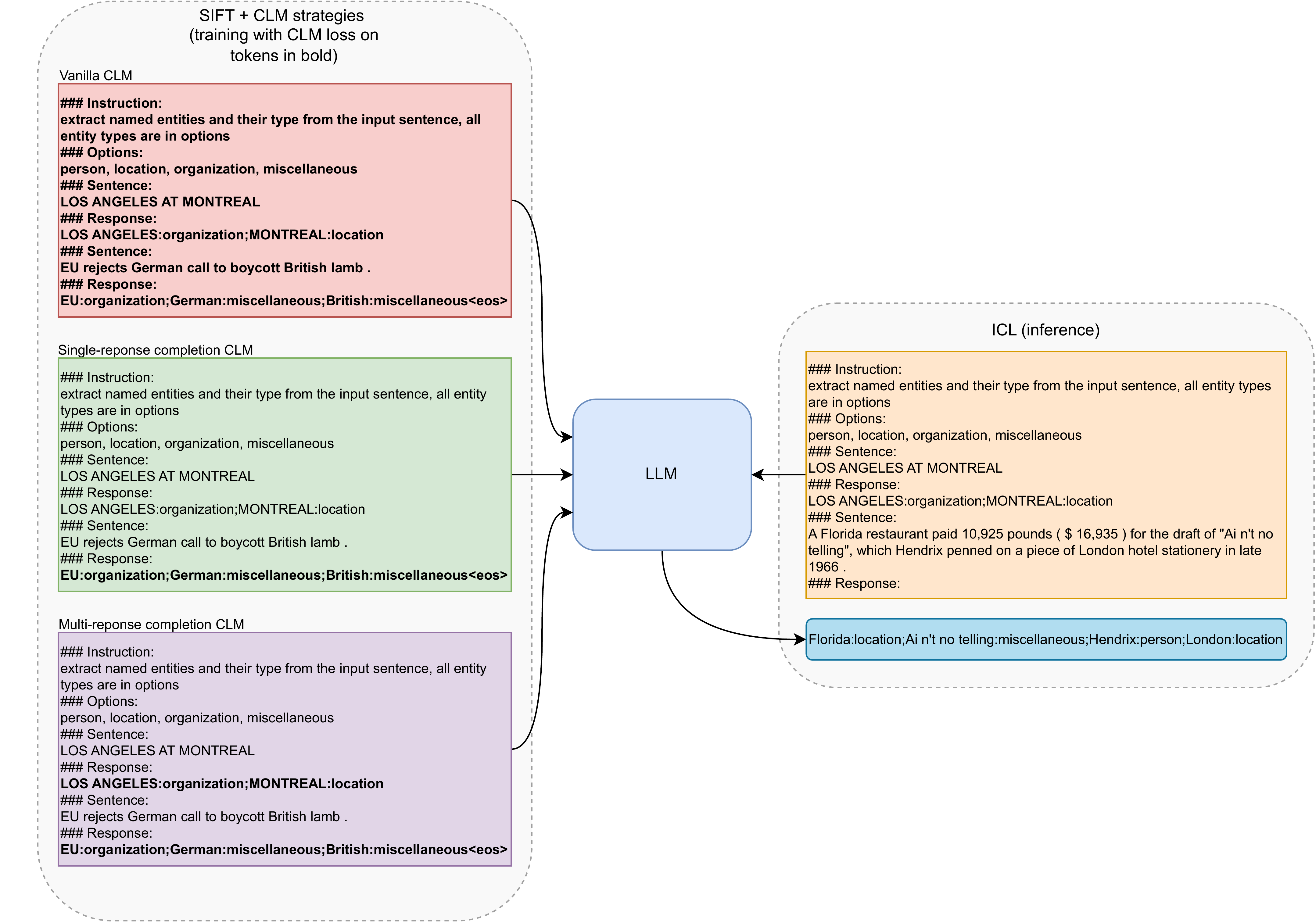}
    \caption[Supervised in-context fine-tuning combined with CLM strategies for training vs.~ICL for inference]{Supervised in-context fine-tuning combined with vanilla, single-response completion, and multi-response completion CLM strategies for training (left) vs.~ICL for inference (right). For illustrative purposes, we show only one example in the context and simplify the instruction. The examples are taken from CoNLL03. The LLM is penalized using CLM loss only for wrong predictions on tokens in bold.}
    \label{fig:clm_strategies}
\end{figure*}

We propose a SIFT framework for sequence labeling tasks, investigating different fine-tuning objectives and comparing SIFT empirically to standard SFT and ICL. Figure \ref{fig:clm_strategies} illustrates our proposed framework during training (SIFT) and inference (ICL). The SIFT framework gives rise to three sensible strategies for generative fine-tuning with in-context demonstrations: (1) \textit{vanilla}, where standard CLM is applied to the entire prompt (i.e., the loss is computed on all tokens), (2) \textit{single-response completion (SRC)}, where the model predicts only the response tokens of the \textit{target} instance (i.e., the last response, excluding demonstrations), and (3) \textit{multi-response completion (MRC)}, where the model predicts the response tokens for both the demonstrations and the final, \textit{target} instance. At inference time, we carry out ICL with constrained token generation using the model adapted with SIFT. 

A SIFT prompt consists of three parts: (1) the instruction, (2) the demonstrations, i.e., in-context examples with gold responses, and (3) the query: the final example. At training time, the query example is also coupled with the gold response, and as such it does not really differ from the in-context demonstrations (it is essentially the last demonstration); at inference time, in a standard ICL setup, we have the demonstrations and the query example and the model generates its response; in-context demonstrations still consist of the example and gold response. Note that removing in-context demonstrations (i.e., part (2)) reduces SIFT to standard SFT. We provide both training and inference templates for both SIFT (with in-context demonstrations) and standard SFT (no in-context demonstrations) in \Cref{tab:token_div}. Additionally, we also evaluate both SIFT and standard SFT without the instruction in the prompt. 

\begin{table*}
    \centering
    \begin{adjustbox}{width=1.0\linewidth}
    \begin{tabular}{p{2.5cm}  p{4.9cm}  p{4.9cm}}
    \toprule
         \multicolumn{1}{c}{Number of demonstrations} & \multicolumn{1}{c}{Prompt template for training} & \multicolumn{1}{c}{Prompt template for evaluation} \\
         \midrule
         \multicolumn{1}{c}{\multirow{3}{*}{Standard SFT}} & \scriptsize
         <instruction\_tokens\_start>
         
         \#\#\# Instruction: 

\{instruction\}

\#\#\# Options:

\{available\_classes\_for\_task\}

<instruction\_tokens\_end>

<query\_tokens\_start>

\#\#\# Sentence:

\{query\_example\}

\#\#\# Response:

\{query\_response\_completion\}

<query\_tokens\_end> & \scriptsize
         <instruction\_tokens\_start>
         
         \#\#\# Instruction: 

\{instruction\}

\#\#\# Options:

\{available\_classes\_for\_task\}

<instruction\_tokens\_end>

<query\_tokens\_start>

\#\#\# Sentence:

\{query\_example\}

\#\#\# Response:

<query\_tokens\_end>  \\\midrule

\multicolumn{1}{c}{\multirow{3}{*}{\shortstack{SIFT}}} & \scriptsize <instruction\_tokens\_start>
         
         \#\#\# Instruction: 

\{instruction\}

\#\#\# Options:

\{available\_classes\_for\_task\}

<instruction\_tokens\_end>

<context\_tokens\_start>

<demonstration\_\#1\_tokens\_start>

\#\#\# Sentence:

\{example\_\#\_1\}

\#\#\# Response:

\{response\_completion\_\#\_1\}

<demonstration\_\#1\_tokens\_end>

\dots

<demonstration\_\#n\_tokens\_start>

\#\#\# Sentence:

\{example\_\#\_n\}

\#\#\# Response:

\{response\_completion\_\#\_n\}

<demonstration\_\#n\_tokens\_end>

<context\_tokens\_end>

<query\_tokens\_start>

\#\#\# Sentence:

\{query\_example\}

\#\#\# Response:

\{query\_response\_completion\}

<query\_tokens\_end> & \scriptsize 
<instruction\_tokens\_start>
         
         \#\#\# Instruction: 

\{instruction\}

\#\#\# Options:

\{available\_classes\_for\_task\}

<instruction\_tokens\_end>

<context\_tokens\_start>

<demonstration\_\#1\_tokens\_start>

\#\#\# Sentence:

\{example\_\#\_1\}

\#\#\# Response:

\{response\_completion\_\#\_1\}

<demonstration\_\#1\_tokens\_end>

\dots

<demonstration\_\#n\_tokens\_start>

\#\#\# Sentence:

\{example\_\#\_n\}

\#\#\# Response:

\{response\_completion\_\#\_n\}

<demonstration\_\#n\_tokens\_end>

<context\_tokens\_end>

<query\_tokens\_start>

\#\#\# Sentence:

\{query\_example\}

\#\#\# Response:

<query\_tokens\_end> \\
    \bottomrule
    \end{tabular}
    \end{adjustbox}
    \caption{Supervised fine-tuning prompt templates for training and evaluation concerning the number of demonstrations in the prompt}
    \label{tab:token_div}
\end{table*} 
Since we are in the realm of generative decoder-only LLMs, our SIFT objectives are generative, i.e., based on token prediction: (1) \textit{vanilla} SIFT trains on all tokens, i.e., amounts to standard CLM on the entire prompt; (2) \textit{single-response completion} (SRC) predicts only the tokens of the response to the query (i.e., the last response)---we will denote this set of query response tokens with $\mathit{QR}$, and (3) \textit{multi-response completion} predicts all tokens in responses of all demonstrations as well as the query response---we will denote the set of all tokens from responses of all in-context demonstrations with $\mathit{DR}$. SIFT models require in-context demonstrations for both training and inference. Unless specified otherwise, the number of demonstrations at inference matches the number used for training.

\subsection{Vanilla SIFT} 

This is the standard CLM objective, as used in LLM pre-training. Formally, we define the vanilla training loss over every token of the training input sequence $\mathbf{t}=(t_1, \dots, t_N)$:
\[
L_V(\mathbf{t};\mathbf{\theta}) = - \sum_{i=1}^{N}\log P(t_i | t_{<i};\mathbf{\theta})
\]
\noindent where $\mathbf{\theta}$ are trainable model parameters and $t_i$ is the $i$-th token in the training sequence out of $N$ tokens, conditioned on preceding context $t_{<i}$.

\subsection{Single-response Completion}

This strategy adapts the decoder by masking out all tokens except those in $\mathit{QR}$ from the loss computation, steering the model to generate valid responses to an example query. Formally, for each training input sequence $\mathbf{t}$, we define the SRC loss over a subset of training tokens:
\[
L_{SRC}(\mathbf{t};\mathbf{\theta}) = - \sum_{i=1}^{N} \delta_i \log P(t_i | t_{<i};\mathbf{\theta}),
\]

\[
\delta_i =
\begin{cases} 
1 & \text{if } t_i \in \mathit{QR} \\
0 & \text{otherwise}
\end{cases}
\]

\noindent where $\mathbf{\theta}$ are trainable model parameters, $t_i$ is is the $i$-th token in the training sequence out of $N$ tokens, conditioned on preceding token context $t_{<i}$, and $\delta_i$ is the indicator function, which keeps the loss only for the $\mathit{QR}$ tokens.

\subsection{Multi-response Completion}

Here we extend SRC to all responses in the context, i.e., we do not compute the CLM loss only for the query response $\mathit{QR}$ but also for the responses of all in-context demonstrations, i.e., all tokens in $\mathit{DR}$. This is meant to force the model to generate correct answers for multiple instances simultaneously, while at the same time using for each generated response all other instances as context. Formally, for each training input sequence $\mathbf{t}$, we define MRC loss as follows:
\[
L_{MRC}(\mathbf{t};\mathbf{\theta}) = - \sum_{i=1}^{N} \delta_i \log P(t_i | t_{<i};\mathbf{\theta}),
\]

\[
\delta_i =
\begin{cases} 
1 & \text{if } t_i \in \mathit{QR} \cup \mathit{DR} \\
0 & \text{otherwise}
\end{cases}
\]
\noindent where $\mathbf{\theta}$ are trainable model parameters, $t_i$ is is the $i$-th token in the training sequence out of $N$ tokens, conditioned on preceding token context $t_{<i}$, and $\delta_i$ is the indicator function, which holds true only for tokens in $\mathit{QR}$ and tokens in $\mathit{DR}$.    

\section{Experiments}

In this section, we describe the experimental setup for ICL and our proposed response-oriented SIFT framework. Further, we describe our model optimization and text generation setups. We compare SIFT with standard SFT and ICL. Experiments underline the effect of the increasing number of demonstrations during fine-tuning and inference on overall PLM adaptation performance to sequence labeling tasks. Moreover, we compare the \textit{base} PLMs with their instructed and dialogue-optimized variants (\textit{instruct}). Finally, we compare SIFT results with the unmasking method from the previous chapter, where the CM is removed from all the layers (cf.~Section \ref{sec:cm_removal_method}). For all experiments, we rely on SOTA open-weight decoder-only LLMs. Furthermore, we compare the CLM strategies and highlight the differences in performance across various response-oriented adaptation strategies, providing recommendations for achieving improved adaptation of autoregressive language models on sequence labeling tasks. Finally, to observe how much decoders depend on the instruction, we experiment with removing the instruction during fine-tuning, and including variations of it during inference. We experiment with four sequence labeling tasks and four sequence labeling datasets. We train the models for NER on CoNLL03, ABAM on AAC-MW, slot labeling on NLU++, and SRL on OntoNotes v5.0. Training is performed on the training sets, and evaluation is done on the validation and test sets. For information on data splits and statistics, see Section \ref{sec:tasks_datasets} and Table \ref{tab:dataset_stats2}.

\subsection{Training}

This subsection details the models and optimization procedures employed for SFT and SIFT. Furthermore, we reflect on the nuances of handling special tokens and ensuring consistency between training setups for different PLMs. Finally, we discuss the formats of our training examples for zero- and few-shot training setups.

\subsubsection{Models}

We leverage five open-weight LLMs for our experiments with either seven or eight billion parameters. We choose Gemma-7B \citep{team2024gemma}, 
Llama2-7B \citep{touvron2023llama}, Llama3-8B and Llama3.1-8B \citep{dubey2024llama}, and Mistral-7B \citep{jiang2023mistral}. Each model is tested as a \textit{base} variant, which was pre-trained using vanilla CLM, and an \textit{instruct} variant. \textit{Instruct} variants of the PLMs were additionally optimized with either reinforcement learning with human feedback (RLHF) \citep{NIPS2017_d5e2c0ad} or direct preference optimization (DPO) \citep{NEURIPS2023_a85b405e}. These techniques equip the models with the ability to effectively solve tasks, follow instructions, and generate human-preferred text. Table~\ref{tab:sift_models} gives the exact models with their \textit{Hugging Face Hub} identifiers that we leverage for our experiments. Models were picked based on the popularity on the \textit{Hugging Face Hub}. PLMs are used with their pre-trained CLM head for experiments inside the framework. 

For experiments with the CM removal, we use the token classification head on top of decoders as in Section \ref{sec:cm_exp}, with the CM removed from all decoder blocks. We use the standard softmax token classifier for all tasks except SRL. For the case of the SRL task, we guide the model's prediction based on the \textit{head word} of the verb using a modification of the model architecture, similar to our proposed \textit{implicit} model from Chapter \ref{ch:method1}. More specifically, the model concatenates the embedding $\mathbf{x}_\mathrm{verb} \in \mathbb{R}^d$ from its embedding matrix with each contextualized token embedding $\mathbf{x}_\mathrm{PLM} \in \mathbb{R}^d$ (the output of the last PLM layer), where $d$ is the PLM's hidden size. The final token representation is a concatenation of the embedding from the last PLM layer and verb embedding: $\mathbf{x} = [\mathbf{x}_\mathrm{PLM}; \mathbf{x}_\mathrm{verb}]$, which is fed to the standard softmax token classifier.

\begin{table}
\centering
\begin{adjustbox}{width=0.8\linewidth}
\begin{tabular}{lll}
\toprule
PLM & Base Variant Identifier & Instruct Variant Identifier \\ \hline
Gemma-7B & \texttt{google/gemma-7b} & \texttt{google/gemma-1.1-7b-it} \\ \hline
Llama2-7B & \texttt{meta-llama/Llama-2-7b-hf} & \texttt{meta-llama/Llama-2-7b-chat-hf} \\ \hline
Llama3-8B & \texttt{meta-llama/Meta-Llama-3-8B} & \texttt{meta-llama/Meta-Llama-3-8B-Instruct} \\ \hline
Llama3.1-8B & \texttt{meta-llama/Llama-3.1-8B} & \texttt{meta-llama/Llama-3.1-8B-Instruct} \\ \hline
Mistral-7B & \texttt{mistralai/Mistral-7B-v0.1} & \texttt{mistralai/Mistral-7B-Instruct-v0.2} \\
\bottomrule
\end{tabular}
\end{adjustbox}
\caption{PLMs and their \textit{Hugging Face Hub} identifiers for \textit{base} and \textit{instruct} model variants}
\label{tab:sift_models}
\end{table}

\subsubsection{Optimization}

We consider the three PLM fine-tuning strategies: (1) vanilla CLM, (2) SRC CLM, and (3) MRC CLM. QLoRA is applied to all fine-tuning experiments to enable PLM adaptation on consumer-grade hardware. We leverage QLoRA on query and value attention matrices inside each decoder block with a fixed rank of $r=16$,\footnote{\footnotesize{We also experimented with a higher rank of $r=64$ and got similar results on the validation set, but with substantially more trained parameters per model, which slowed down the overall fine-tuning process.}} a scaling parameter of $\alpha=16$, and a dropout probability of $p=0.1$. Hence, only decomposed query and value matrices are optimized with cross-entropy loss, which is modified in the case of response-oriented CLM strategies.

The models are trained in bfloat16 precision, with loaded pre-trained weights in 4-bit NormalFloat data type, and we use double quantization. Using this setup, we fit all models into 40GB of GPU memory of Ampere A100 and train with a consistent batch size of $8$ per experiment. For vanilla CLM, we exploit example packing, where we pack short examples in the same input sequence to maximize efficiency during training. We set the maximum input sequence length to \num{1024} for all experiments. To experiment with SRC CLM, we use \textit{DataCollatorForCompletionOnlyLM} implementation from the TRL library \citep{vonwerra2022trl}. This library enables SFT training and IT, implementing techniques such as example packing, loss calculation only on specified completion tokens, RLHF, and DPO. With \textit{DataCollatorForCompletionOnlyLM}, we mask all the tokens from the loss function except the QR tokens. For MRC CLM, we implement our own data collator, which masks all the tokens from the loss function except the DR and the QR tokens (cf.~Table~\ref{tab:token_div}). These tokens can be easily identified since we provide the collator with a prompt template for training. However, we do not leverage example packing due to the discontinuity of the tokens incurred with token masking through the loss function. 

We employ gradient accumulation to address out-of-memory issues while maintaining a fixed batch size. If all examples fit into GPU RAM, we use a batch size of $8$ with $4$ gradient accumulation steps. Otherwise, we reduce the batch size to $4$ and increase the gradient accumulation steps to $8$. This approach ensures consistent training performance without exceeding memory limits. We train the models with a paged 8-bit AdamW \citep{loshchilov2017decoupled} optimizer to handle the memory spikes \cite{dettmers2023qlora}. The parameters of AdamW are fixed to $\beta_1=0.9, \beta_2=0.95, \epsilon=1\mathrm{e}{-5}, \lambda=0.1$. We choose the cosine annealing scheduler \citep{loshchilov2016sgdr} for the learning rate scheduler. We apply gradient clipping set to $1.0$ and use gradient checkpointing. We use a consistent learning rate of $2\mathrm{e}{-4}$ across all experiments and fine-tune models over a fixed number of five epochs. All results are averages of four runs with different seeds, and we always pick the last model for each seed. For CM removal experiments, we inherit the optimization hyperparameters and the training procedure. We set the maximum sequence length to $256$ in these experiments.

\subsubsection{Handling Special Tokens} 

We pad the sequences using left-side padding since the model should be prevented from learning to generate text starting with the padding token, which occurs with right-side padding. Furthermore, we define a padding token for PLMs that do not define it explicitly. Padding token is set to either some of the unique reserved tokens or to \texttt{<unk>} token in the case of LLama2-7B and Mistral-7B models. We avoid using the end-of-sequence (\texttt{<eos>}) token as a padding token because the loss is masked out for padding tokens. If the \texttt{<eos>} tokens were used for padding, the model would never learn when to stop generating text. Instead, we explicitly teach the model to recognize when to stop generation by preserving the loss for the \texttt{<eos>} token and including it at the end of each sequence during training (cf.~Table \ref{tab:icl_sift_examples}).

\subsubsection{Forming Training Examples} 

The training examples are constructed to adhere to the specified format with instruction, demonstrations (for SIFT experiments), and query parts. We conduct training with $0$, $1$, $5$, and $10$ demonstrations ($n$ shots) in the context part. We refer to the scenario of $0$ demonstrations in the context as SFT. When the context includes one or more examples, we employ the SIFT setup. These zero- and few-shot fine-tuning setups are combined with three CLM strategies. Since we always have at least the query example in the training prompt template, we effectively apply CLM to $n+1$ examples from the training set. The demonstrations are sampled from the training set, and we fix the sampling to be dependent on the seed and the query example to ensure that the whole training setup is shared between PLMs. Table~\ref{tab:icl_sift_examples} shows the training examples for the case of $1$-shot SIFT. Analogous to the $1$-shot setup, we prepare the training examples for SIFT experiments with more than one demonstration in the context. Similar holds for SFT experiments, where we have zero demonstrations in the context. The expected responses are formatted to adhere to the regular expression introduced in Subsection \ref{subsubsec:evaluation_decoders}. This way, we train the models to learn to generate the spans and the classes simultaneously, saving on the total number of tokens required for training and generation.

Inspired by previous work \citep{razumovskaia-etal-2024-sqatin}, we also experimented with the prompts adhering to the question-answering style in our preliminary experiments. In this design, the model is prompted to answer with a span for each span class in the question or provide an \textit{NA} response if there are no spans for the class in the question. However, this design led to worse results on the validation set for most CLM strategies and PLMs, introducing long context problems where the prompt lengths grew significantly in size with the increase in the number of classes and length of the context examples. These findings align with previous work, where it has been shown that LLMs struggle to utilize the long contexts \citep{10.1162/tacl_a_00638}.

\begin{table*}
    \centering
    \begin{adjustbox}{width=0.75\linewidth}
    \begin{tabular}{p{2cm}  L{5cm}  L{5cm}}
    \toprule
         \multicolumn{1}{c}{Task (dataset)} & \multicolumn{1}{c}{Example for training} & \multicolumn{1}{c}{Example for evaluation} \\
         \midrule
         \multicolumn{1}{l}{\multirow{3}{*}{NER (CoNLL03)}} & \tiny \#\#\# Instruction: 

extract named entities and their type from the input sentence, all entity types are in options
if there are no named entities in the sentence the output should just be 'NA'
if there are multiple extractions from the sentence, the extraction format should be entity\_1\_span:entity\_1\_class;entity\_2\_span:entity\_2\_class;...

\#\#\# Options:

person, location, organization, miscellaneous

\#\#\# Sentence:

LOS ANGELES AT MONTREAL

\#\#\# Response:

LOS ANGELES:organization;MONTREAL:location

\#\#\# Sentence:

EU rejects German call to boycott British lamb .

\#\#\# Response:

EU:organization;German:miscellaneous;British: miscellaneous<eos> & \tiny \#\#\# Instruction: 

extract named entities and their type from the input sentence, all entity types are in options
if there are no named entities in the sentence the output should just be 'NA'
if there are multiple extractions from the sentence, the extraction format should be entity\_1\_span:entity\_1\_class;entity\_2\_span:entity\_2\_class;...

\#\#\# Options:

person, location, organization, miscellaneous

\#\#\# Sentence:

LOS ANGELES AT MONTREAL

\#\#\# Response:

LOS ANGELES:organization;MONTREAL:location

\#\#\# Sentence:

EU rejects German call to boycott British lamb .

\#\#\# Response: \\\midrule
\multicolumn{1}{c}{\multirow{3}{*}{\shortstack{ABAM  \\ (AAC-MW)}}} & \tiny \#\#\# Instruction: 

extract argument aspects and their type from the input sentence, all aspect types are in options
if there are no argument aspects in the sentence the output should just be 'NA'
if there are multiple extractions from the sentence, the extraction format should be aspect\_1\_span:aspect\_1\_class;aspect\_2\_span:aspect\_2\_class;... 

\#\#\# Options:

capital\_vs\_labor, social\_justice/injustice, economic\_impact, prices, low\_skilled, turnover, government, youth\_and\_secondary\_wage\_earners, competition/business\_challenges, motivation/chances, welfare, un/employment\_rate

\#\#\# Sentence:

Reduced Expense for Social Programs : Employees surviving at minimum wage are also often the same people who must rely on additional support of government run social programs to support themselves and their families on such a small amount of income .

\#\#\# Response:

Reduced Expense for Social Programs:welfare;government run social programs:welfare

\#\#\# Sentence:

As the cost of living has jumped by leaps and bounds minimum wage has barely made an impact .

\#\#\# Response:

the cost of living has jumped:social\_justice/injustice<eos> & \tiny \#\#\# Instruction: 

extract argument aspects and their type from the input sentence, all aspect types are in options
if there are no argument aspects in the sentence the output should just be 'NA'
if there are multiple extractions from the sentence, the extraction format should be aspect\_1\_span:aspect\_1\_class;aspect\_2\_span:aspect\_2\_class;... 

\#\#\# Options:

capital\_vs\_labor, social\_justice/injustice, economic\_impact, prices, low\_skilled, turnover, government, youth\_and\_secondary\_wage\_earners, competition/business\_challenges, motivation/chances, welfare, un/employment\_rate

\#\#\# Sentence:

Reduced Expense for Social Programs : Employees surviving at minimum wage are also often the same people who must rely on additional support of government run social programs to support themselves and their families on such a small amount of income .

\#\#\# Response:

Reduced Expense for Social Programs:welfare;government run social programs:welfare

\#\#\# Sentence:

As the cost of living has jumped by leaps and bounds minimum wage has barely made an impact .

\#\#\# Response: \\\midrule
\multicolumn{1}{c}{\multirow{3}{*}{\shortstack{Slot labeling \\ (NLU++)}}} & \tiny \#\#\# Instruction: 

extract slots and their type from the input sentence, all slot label types are in options
if there are no slots in the sentence the output should just be 'NA'
if there are multiple extractions from the sentence, the extraction format should be slot\_1\_span:slot\_1\_class;slot\_2\_span:slot\_2\_class;... 

\#\#\# Options:

time\_from, person\_name, shopping\_category, date\_from, date, number, adults, rooms, amount\_of\_money, kids, people, date\_to, date\_period, time, company\_name, time\_period, time\_to

\#\#\# Sentence:

book a skincare session Saturday at quarter past 5 afternoon

\#\#\# Response:

Saturday:date;quarter past 5 afternoon:time

\#\#\# Sentence:

send 4900 euros to domineque curl after half past 17 today

\#\#\# Response:

4900 euros:amount\_of\_money;domineque curl:person\_name;half past 17:time\_from;today:date<eos> & \tiny \#\#\# Instruction: 

extract slots and their type from the input sentence, all slot label types are in options
if there are no slots in the sentence the output should just be 'NA'
if there are multiple extractions from the sentence, the extraction format should be slot\_1\_span:slot\_1\_class;slot\_2\_span:slot\_2\_class;... 

\#\#\# Options:

time\_from, person\_name, shopping\_category, date\_from, date, number, adults, rooms, amount\_of\_money, kids, people, date\_to, date\_period, time, company\_name, time\_period, time\_to

\#\#\# Sentence:

book a skincare session Saturday at quarter past 5 afternoon

\#\#\# Response:

Saturday:date;quarter past 5 afternoon:time

\#\#\# Sentence:

send 4900 euros to domineque curl after half past 17 today

\#\#\# Response: \\\midrule
\multicolumn{1}{c}{\multirow{3}{*}{\shortstack{SRL \\ (OntoNotes v5.0)}}} & \tiny \#\#\# Instruction: 

extract arguments of the given verb and their semantic roles from the input sentence, all semantic roles are in options
if there are multiple extractions from the sentence, the extraction format should be argument\_1\_span:argument\_1\_role;argument\_2\_span: argument\_2\_role;... 

\#\#\# Options:

ARG0, ARG1, ARG2, ARG3, ARG4, ARGM-ADJ, ARGM-ADV, ARGM-CAU, ARGM-COM, ARGM-DIR, ARGM-DIS, ARGM-EXT, ARGM-GOL, ARGM-LOC, ARGM-MNR, ARGM-MOD, ARGM-NEG, ARGM-PNC, ARGM-PRD, ARGM-PRP, ARGM-TMP, C-ARG0, C-ARG1, C-ARG2, R-ARG0, R-ARG1, V

\#\#\# Sentence:

The wrong things the sinful self does are clear :

\#\#\# Verb:

does

\#\#\# Response:

The wrong things:ARG1;the sinful self:ARG0;does:V

\#\#\# Sentence:

But using foreign - funded banks to scare people has absolutely no meaning with regard to solving the problem .

\#\#\# Verb:

scare

\#\#\# Response:

scare:V;people:ARG1<eos> & \tiny \#\#\# Instruction: 

extract arguments of the given verb and their semantic roles from the input sentence, all semantic roles are in options
if there are multiple extractions from the sentence, the extraction format should be argument\_1\_span:argument\_1\_role;argument\_2\_span: argument\_2\_role;... 

\#\#\# Options:

ARG0, ARG1, ARG2, ARG3, ARG4, ARGM-ADJ, ARGM-ADV, ARGM-CAU, ARGM-COM, ARGM-DIR, ARGM-DIS, ARGM-EXT, ARGM-GOL, ARGM-LOC, ARGM-MNR, ARGM-MOD, ARGM-NEG, ARGM-PNC, ARGM-PRD, ARGM-PRP, ARGM-TMP, C-ARG0, C-ARG1, C-ARG2, R-ARG0, R-ARG1, V

\#\#\# Sentence:

The wrong things the sinful self does are clear :

\#\#\# Verb:

does

\#\#\# Response:

The wrong things:ARG1;the sinful self:ARG0;does:V

\#\#\# Sentence:

But using foreign - funded banks to scare people has absolutely no meaning with regard to solving the problem .

\#\#\# Verb:

scare

\#\#\# Response: \\
    \bottomrule
    \end{tabular}
    \end{adjustbox}
    \caption{Examples from four sequence labeling datasets for $1$-shot ICL and SIFT experiments}
    \label{tab:icl_sift_examples}
\end{table*} 
\subsection{Evaluation}

This subsection reflects on our evaluation setup for ICL, SFT, and SIFT. The evaluation setup is shared between ICL over raw PLMs and those adapted with SFT and SIFT to sequence labeling tasks. We point out differences in training and evaluation examples. Furthermore, we introduce constrained generation to ensure the models follow the expected generation scheme. This is enforced to alleviate the parsing of the generated model output since our evaluation is done on IOB2 tags with strict matching and evaluated using the micro F1 score. SFT and SIFT models do not provide explicit BIO tags, so we use greedy span-based matching of predicted spans and their classes with input tokens (cf.~Subsection \ref{subsubsec:evaluation_decoders}).

\subsubsection{Forming Evaluation Examples} 

Examples of prompts for $1$-shot evaluation are shown in Table~\ref{tab:icl_sift_examples}. Analogous to the $1$-shot setup, we prepare the training examples for evaluation with more than one demonstration in the context. We match the format with the training prompt templates. We omit the \texttt{<eos>} token from the evaluation examples since the model trained to stop with the \texttt{<eos>} token will not continue generating the response. The training and evaluation prompts differ only in the last part, where the evaluation examples do not provide the QR tokens. The model is prompted to complete the output, and the generated output is parsed to obtain IOB2 tags. Matching the number of demonstrations in the context for training and evaluation gave the best performance on the validation set, so we stuck to this type of evaluation. For example, if the model was fine-tuned with five demonstrations in the context, we evaluate the model with the five demonstrations in the context. The alternative was to always prompt the model in the zero-shot style. However, our preliminary experiments have shown that this approach introduces a mismatch between pre-training and fine-tuning, which in turn aggravates the transfer in few-shot scenarios.

We sample the demonstrations from the training set as context for evaluations on the examples from the validation and test sets. Importantly, we share the demonstrations in the context between PLMs trained under $n$-shot setups and CLM strategies to ensure a fair comparison. Contexts depend on the seed the model was trained on, meaning that we randomly sample new demonstrations for each example, but have four sets of demonstrations in total (due to four seeds) for each example in the validation and test set. These are shared between ICL, SFT, and SIFT experiments for the fairness of evaluation.

\subsubsection{Constrained Generation and Evaluation} 

We leverage the \textit{outlines} library \citep{willard2023efficient} to ensure the models follow the generation scheme specified with a regular expression. This library relies on a finite-state machine formulation of the provided regular expression to guide generation for decoder models by operating on the model logits. We use the \textit{vLLM} library \citep{kwon2023efficient} to speed up generation. Since \textit{vLLM} does not support integration with PEFT methods, we merge the weights of the QLoRA-trained module with the PLM and perform generation. The merging is performed by summing up the pre-trained weights and scaled LoRA weights (cf.~Figure \ref{fig:lora}). The merged model is saved on the disk, loaded in memory, and moved to the GPU, and the inference is executed. After performing inference, the saved model is removed from the disk, and we keep only the saved LoRA weights, which leaves a negligible memory footprint.

Generation setup is shared between all models. We generate tokens using a temperature of $0.1$ and top-p sampling with a threshold of $0.9$. We require the model to generate up to $200$ tokens since there is no query response in any evaluation set longer than $200$ tokens. As described in Subsection \ref{subsubsec:evaluation_decoders}, we heuristically map response spans of decoders to IOB2 tags. We employ greedy span-based matching of predicted spans and their classes with input tokens. Finally, during parsing, we consider only the first line of the generated response and discard the rest since we notice on the validation set that models trained in few-shot setups tend to overgenerate even when we keep the loss on the \texttt{<eos>} token during training. However, they complete the task in the first line of the generated response, so we allow this bias in the parsing of model outputs.

\subsection{Setup for the Instruction Ablation}

To measure the effect of instruction on the performance of standard SFT and SIFT models, we first train the model for the task without the instruction and then evaluate using ICL with variations of the task instruction, matching the number of demonstrations used for training. We conduct an evaluation using ICL and match the number of demonstrations that were used for training. We evaluate on the validation sets for four sequence labeling datasets. The proposed variations are:
\begin{enumerate}
    \item \textit{Vanilla}, where we include the usual task-specific instruction (same as the ones in Table \ref{tab:icl_sift_examples}) into the evaluation prompt;
    \item \textit{Permuted}, where we randomly permute the order of the tokens of the \textit{vanilla} instruction (with a fixed permutation seed) to observe if the model relies on the lexical presence of key words rather than the compositional structure of the instruction;
    \item \textit{Nonsense}, where we include a text snippet entirely unrelated to the task we trained the model for, to test whether the model depends on the instruction’s meaning to complete the task.
\end{enumerate}

Here, we demonstrate the exact instructions that we used for each proposed variation, on the example of the NER task:

\begin{exampleblock}{Instruction variation examples for NER task}
\begin{enumerate}
    \item Vanilla \\
\#\#\# Instruction: \\ extract named entities and their type from the input sentence, all entity types are in options
if there are no named entities in the sentence the output should just be ``NA''\\
if there are multiple extractions from the sentence, the extraction format should be entity\_1\_span:entity\_1\_class;entity\_2\_span:entity\_2\_class;... \\
\#\#\# Options:
person, location, organization, miscellaneous
    \item Permuted \\
\#\#\# Instruction: \\
the the no entity their Options: output sentence, be if if entites types and the sentence, all sentence extractions extract be are are organization, format ``NA'' just named in should person, from there entity\_1\_span:entity\_1\_class;entity\_2\_span:entity\_2\_class;...\#\#\# are miscellaneous location, entities should the type multiple from input in options there named the extraction
    \item Nonsense \\
\#\#\# Instruction: 
``The Funniest Joke in the World'' (also ``Joke Warfare'' and ``Killer Joke'') is a Monty Python comedy sketch revolving around a joke that is so funny that anyone who reads or hears it promptly dies from laughter. Ernest Scribbler (Michael Palin), a British ``manufacturer of jokes,'' writes the joke on a piece of paper only to die laughing. His mother (Eric Idle) also immediately dies laughing after reading it, as do the first constables on the scene. Eventually the joke is contained, weaponized, and deployed against Germany during World War II.\footnotemark
\end{enumerate}
\end{exampleblock}
\footnotetext{\footnotesize{Taken from \url{https://en.wikipedia.org/wiki/The_Funniest_Joke_in_the_World}}}

\section{Results} \label{sec:sift_results}

Comparisons between ICL, SFT, CM removal, and SIFT are given in the remainder of this section. We provide the results for five open-weight decoder-only LLMs with \textit{base} and \textit{instruct} variants. We outline the differences in performance on four sequence labeling tasks and comment on the effects of including more demonstrations in the prompt during training and inference. Additionally, we compare the impact of having a response-oriented adaptation strategy as opposed to treating all tokens during the adaptation phase of transfer learning equally. Likewise, comparisons are given between the three CLM strategies for SFT and SIFT alongside the analysis of the effect of the instruction on overall performance. Finally, we conclude the chapter with recommendations for sequence labeling tasks based on our proposed SIFT framework in Section \ref{sec:sift_summary}.

\subsection{In-context Learning Results}

\begin{figure*}
\begin{center}
\includegraphics[width=\textwidth]{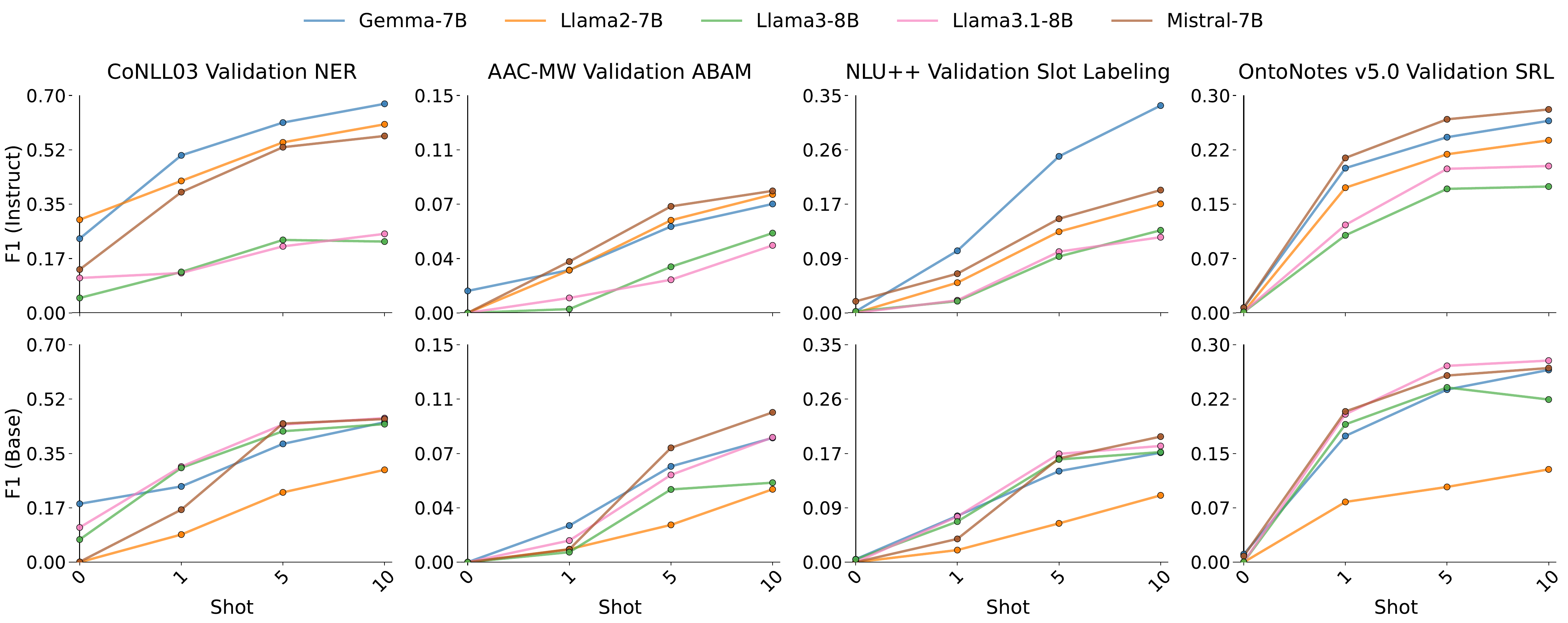}
\caption[Micro F1 sequence labeling scores on the validation set using decoder-only PLMs on ICL]{Micro F1 scores using five \textit{base} and \textit{instruct} variants of decoders on ICL and for a varying number of shots. The x-axis shows the number of shots on an ordinal scale. The results are given for the validation set on four tasks (left to right), with top row plots corresponding to \textit{instruct} variants and bottom row plots corresponding to \textit{base} variants. All results are averages of four runs. The context for the higher number of shots was kept fixed for a fixed seed.}
\label{fig:icl_sl}
\end{center}
\end{figure*}

Figure \ref{fig:icl_sl} shows the validation set performance for ICL experiments. We note an increasing F1 trend as we raise the number of shots. This trend is more pronounced for \textit{instruct} variants of PLMs than \textit{base} variants, as \textit{instruct} models reach higher overall ICL performance per task as we increase the number of shots. Gemma-7B-Instruct dominates the performance on the NER and slot labeling tasks, while Mistral-7B reaches the best results for ICL on ABAM, and Mistral-7B-Instruct reaches the best results with ICL on the SRL task. The highest F1 score was reached with Gemma-7B-Instruct for the NER task, leveraging $10$ demonstrations in the context. 

The most challenging task to complete was ABAM, as we recorded the lowest performance out of all four sequence labeling tasks. Generally, we observe fairly low performance for all tasks except NER, demonstrating that, in standard ICL, LLMs struggle with sequence labeling task completion across many classes, as the model cannot observe all the classes in the prompts for this low number of demonstrations in the context. In our experiments, CoNLL03 has only four classes, while AAC-MW, NLU++, and OntoNotes v5.0 have $12$, $17$, and $27$ classes, respectively. On the other hand, increasing the number of shots for sequence labeling tasks incurs too long a context, which is detrimental to ICL performance \citep{10.1162/tacl_a_00638} and causes potential out-of-memory issues. 

Notably, we also observed this phenomenon for ICL on question-answering style sequence labeling tasks, where the model has to utilize a much longer context, as the context length depends on the number of classes. However, we do not report these results, as the performance was significantly worse than our proposed setup.

\subsection{Main Results for SIFT and Standard SFT}

\begin{figure*}
\begin{center}
\begin{subfigure}{1.0\textwidth}
    \centering
    \includegraphics[width=\textwidth]{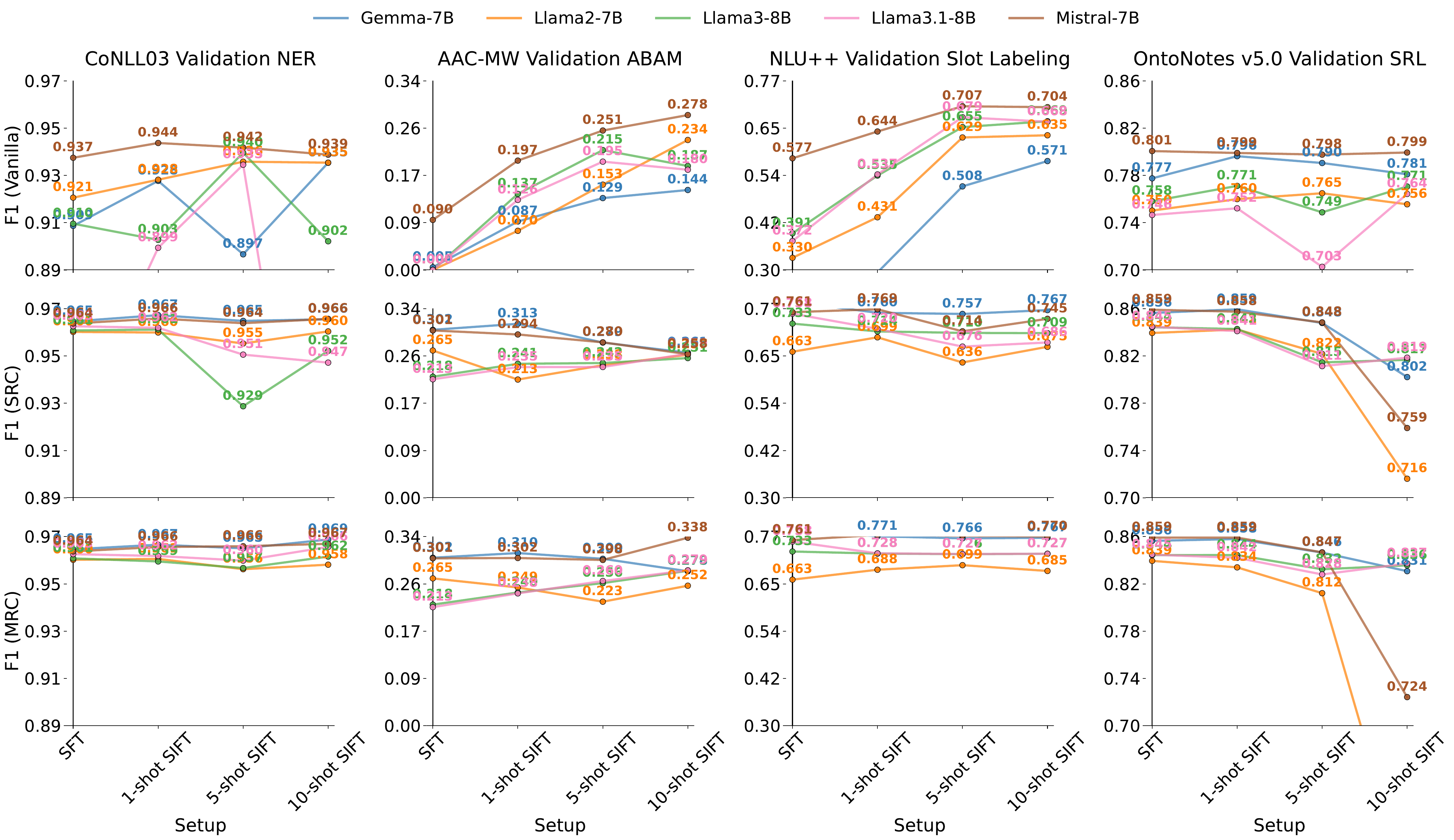}
    \caption{\textit{Base}}
\end{subfigure}
\begin{subfigure}{1.0\textwidth}
    \centering
    \includegraphics[width=\textwidth]{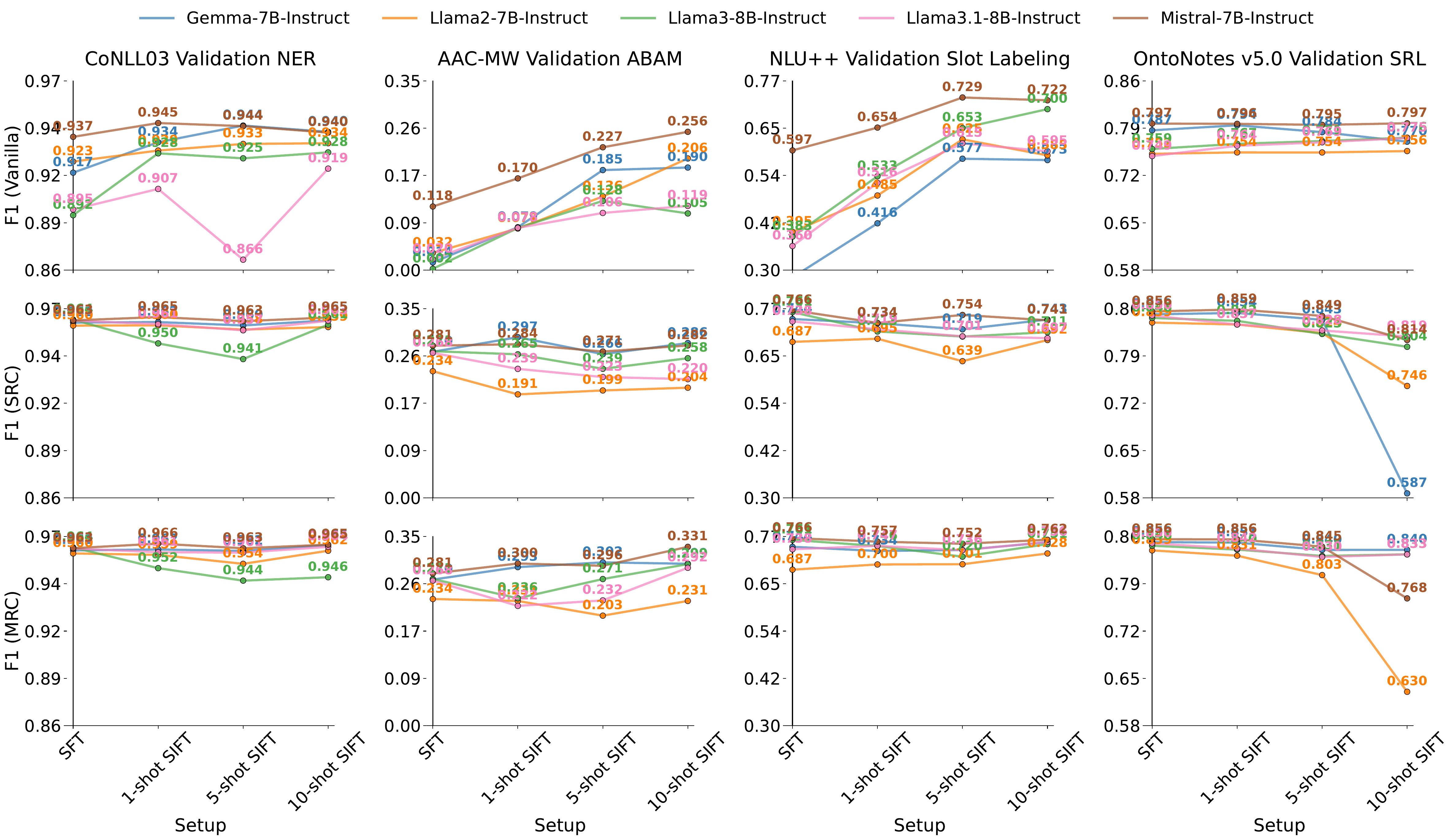}
    \caption{\textit{Instruct}}
\end{subfigure}
\caption[Micro F1 sequence labeling scores on validation sets using decoder-only PLMs on SFT and SIFT]{Micro F1 scores for five \textit{base} (a) and \textit{instruct} (b) variants of decoders on standard SFT and SIFT for a varying number of shots. The results are given for the validation set on four tasks (left to right) and for three CLM strategies (top to bottom). The models are evaluated with the same number of shots in the context that they used for fine-tuning. All results are averages of four runs. The context for the higher number of shots was kept fixed for a fixed seed.}
\label{fig:it_sift_validation_sl}
\end{center}
\end{figure*}

\begin{figure*}
\begin{center}
\begin{subfigure}{1.0\textwidth}
    \centering
    \includegraphics[width=\textwidth]{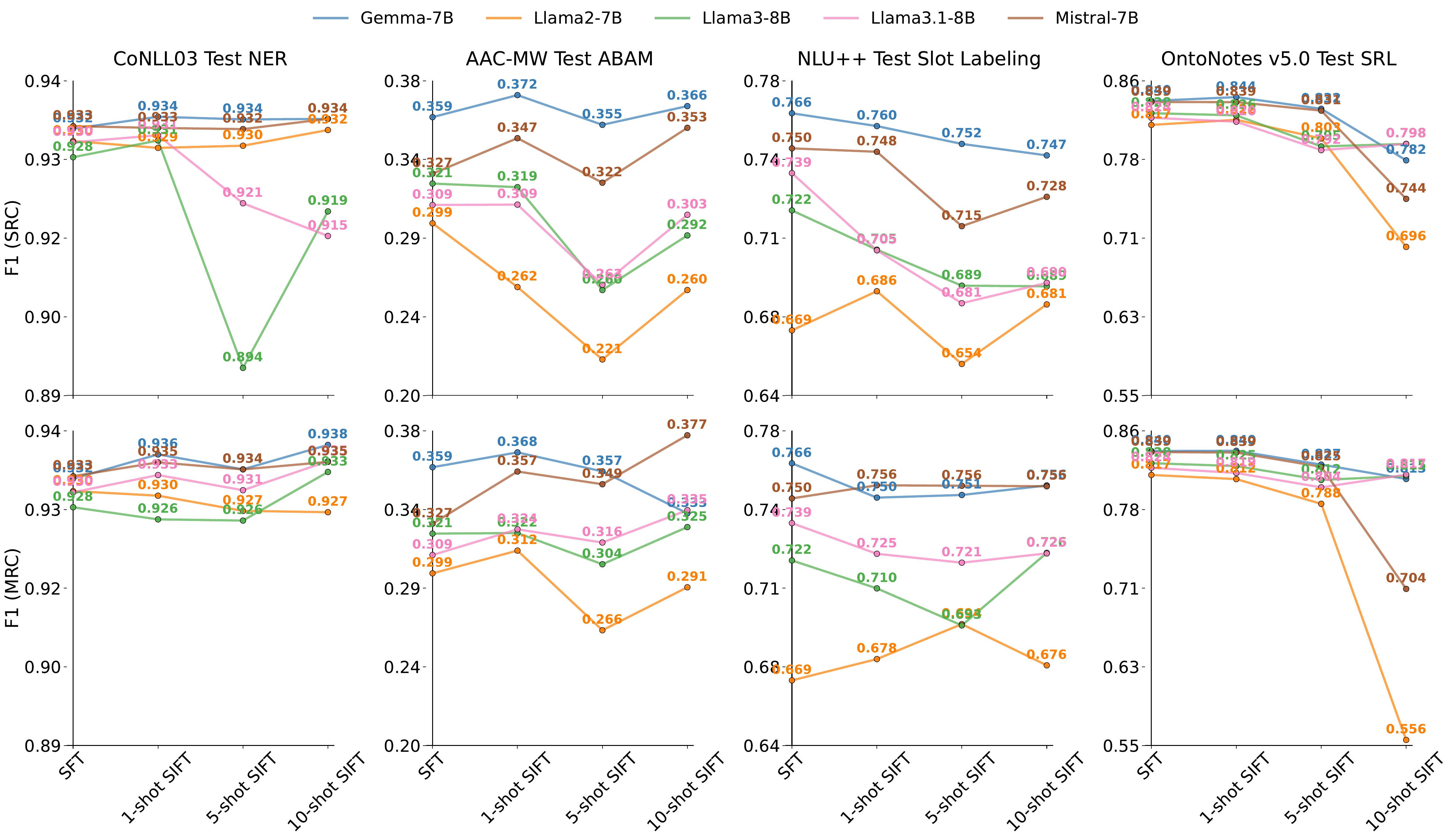}
    \caption{\textit{Base}}
\end{subfigure}
\begin{subfigure}{1.0\textwidth}
    \centering
    \includegraphics[width=\textwidth]{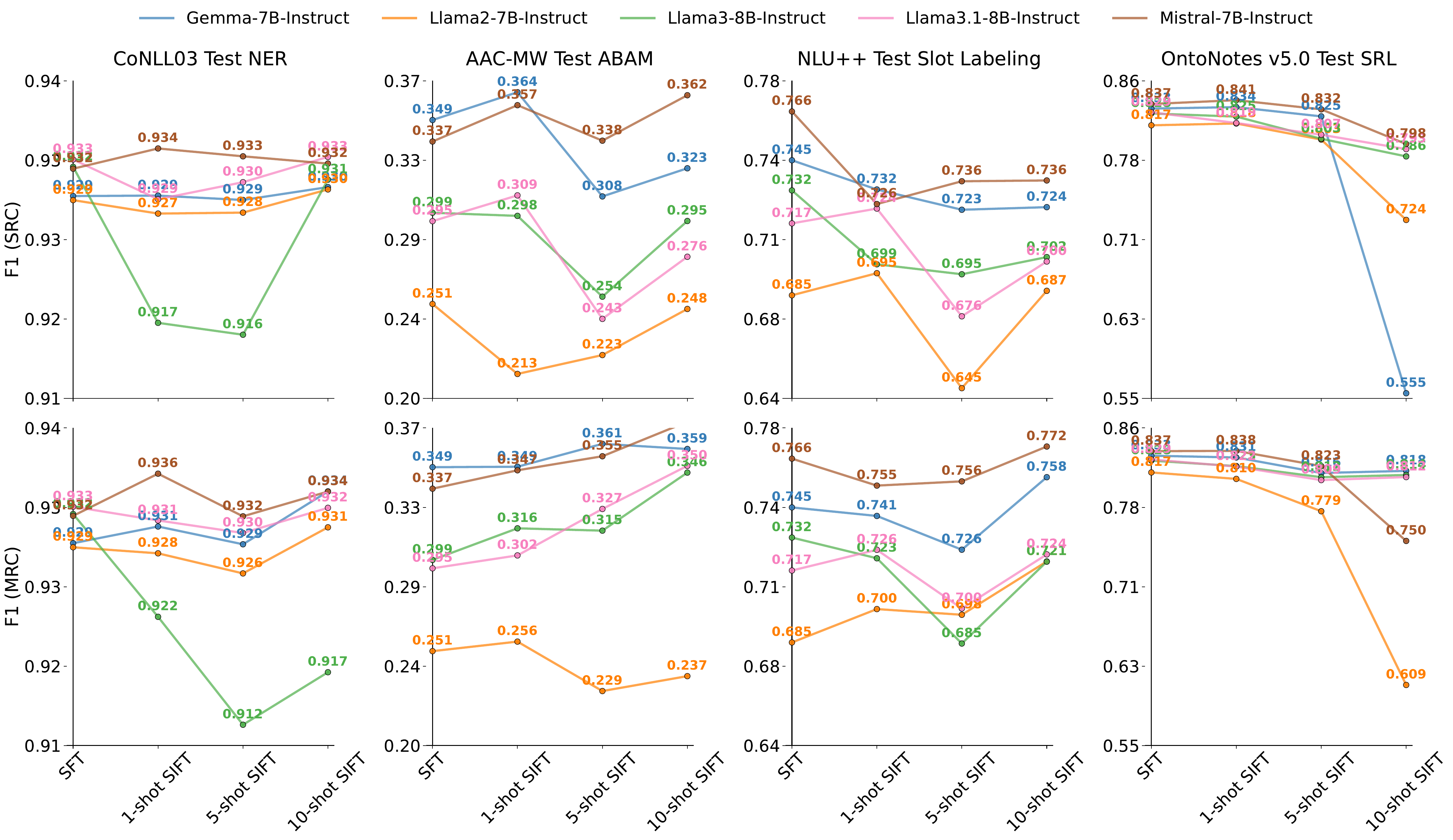}
    \caption{\textit{Instruct}}
\end{subfigure}
\caption[Micro F1 sequence labeling scores on test sets using decoder-only PLMs on SFT and SIFT]{Micro F1 scores for five \textit{base} (a) and \textit{instruct} (b) variants of decoders on standard SFT and SIFT for a varying number of shots. The results are given for the test set on four tasks (left to right) and for three CLM strategies (top to bottom). The models are evaluated with the same number of shots in the context that they used for fine-tuning. All results are averages of four runs. The context for the higher number of shots was kept fixed for a fixed seed.}
\label{fig:it_sift_test_sl}
\end{center}
\end{figure*}

Figures \ref{fig:it_sift_validation_sl} and \ref{fig:it_sift_test_sl} give all the results for SFT experiments. More detailed results are given in Tables \ref{tab:sift_instruct_validation_results} and \ref{tab:sift_base_validation_results} for the validation set and Tables \ref{tab:sift_instruct_test_results} and \ref{tab:sift_base_test_results} for the test set,  where we report results for response-oriented CLM strategies. Inside fine-tuning experiments, we include SFT and SIFT experiments. We note that training is done with three CLM strategies, and we provide the results for these strategies under varying numbers of demonstrations in the context during fine-tuning and inference. We split the results into validation and test set results. For the validation set results, we also include CM removal results for comparison with SFT and SIFT in Tables \ref{tab:sift_instruct_validation_results} and \ref{tab:sift_base_validation_results}.  

\begin{table*}
\centering
\adjustbox{width=0.85\linewidth}{
\small{\begin{tabular}{l|lcccccccc}
\toprule
\multicolumn{2}{l}{\multirow{2}{*}{}} & \multicolumn{2}{c}{\textbf{CoNLL03}} & \multicolumn{2}{c}{\textbf{AAC-MW}} & \multicolumn{2}{c}{\textbf{NLU++}} & \multicolumn{2}{c}{\textbf{OntoNotes v5.0}} \\
\cmidrule(lr){3-4} \cmidrule(lr){5-6} \cmidrule(lr){7-8} \cmidrule(lr){9-10}
\multicolumn{1}{l}{Model} & \multicolumn{1}{l}{Setup} & \textbf{SRC} & \textbf{MRC} & \textbf{SRC} & \textbf{MRC} & \textbf{SRC} & \textbf{MRC} & \textbf{SRC} & \textbf{MRC} \\
\midrule
\multirow{4}{*}{\rotatebox[origin=c]{90}{\shortstack{Gemma-7B \\ Instruct}}}   &   0-Shot/SFT  & $96.2_{0.3}$ & $96.2_{0.3}$ & $27.0_{1.1}$ & $27.0_{1.1}$ & $\mathbf{74.4}_{0.9}$ & $74.4_{0.9}$ & $85.2_{0.3}$ & $\mathbf{85.2}_{0.3}$ \\
{} & 1-Shot/SIFT  & $96.2_{0.2}$ & $96.2_{0.2}$ & $\mathbf{29.7}_{3.1}$ & $29.3_{1.5}$ & $73.4_{2.2}$ & $73.4_{1.3}$ & $\underline{\mathbf{85.4}}_{0.1}$ & $85.1_{0.2}$ \\
{} & 5-Shot/SIFT  & $96.0_{0.1}$ & $96.2_{0.2}$ & $26.6_{2.9}$ & $\underline{\mathbf{30.2}}_{1.1}$ & $71.9_{1.1}$ & $73.6_{1.7}$ & $84.3_{0.2}$ & $84.0_{0.3}$ \\
{} & 10-Shot/SIFT  & $\mathbf{96.3}_{0.2}$ & $\underline{\mathbf{96.5}}_{0.2}$ & $28.6_{4.7}$ & $29.9_{6.1}$ & $74.3_{3.2}$ & $\underline{\mathbf{75.6}}_{0.8}$ & $58.7_{7.5}$ & $84.0_{0.3}$ \\
\midrule

\multirow{4}{*}{\rotatebox[origin=c]{90}{\shortstack{Llama2-7B \\ Instruct}}}   &   0-Shot/SFT  & $\mathbf{96.0}_{0.2}$ & $96.0_{0.2}$ & $\underline{\mathbf{23.4}}_{3.8}$ & $\underline{\mathbf{23.4}}_{3.8}$ & $68.7_{3.3}$ & $68.7_{3.3}$ & $\underline{\mathbf{83.9}}_{0.2}$ & $\underline{\mathbf{83.9}}_{0.2}$ \\
{} & 1-Shot/SIFT  & $96.0_{0.3}$ & $95.9_{0.2}$ & $19.1_{2.7}$ & $23.1_{3.7}$ & $\mathbf{69.5}_{0.4}$ & $70.0_{3.3}$ & $83.7_{0.3}$ & $83.1_{0.1}$ \\
{} & 5-Shot/SIFT  & $95.8_{0.1}$ & $95.4_{0.2}$ & $19.9_{1.5}$ & $20.3_{0.5}$ & $63.9_{2.1}$ & $70.1_{1.4}$ & $82.4_{0.5}$ & $80.3_{0.6}$ \\
{} & 10-Shot/SIFT  & $95.9_{0.2}$ & $\underline{\mathbf{96.2}}_{0.2}$ & $20.4_{1.2}$ & $23.1_{4.4}$ & $69.2_{3.1}$ & $\underline{\mathbf{72.8}}_{1.3}$ & $74.6_{0.2}$ & $63.0_{3.6}$ \\
\midrule

\multirow{4}{*}{\rotatebox[origin=c]{90}{\shortstack{Llama3-8B \\ Instruct}}}   &   0-Shot/SFT  & $\underline{\mathbf{96.4}}_{0.2}$ & $\underline{\mathbf{96.4}}_{0.2}$ & $\mathbf{27.1}_{2.1}$ & $27.1_{2.1}$ & $\underline{\mathbf{76.1}}_{2.2}$ & $\underline{\mathbf{76.1}}_{2.2}$ & $\underline{\mathbf{84.6}}_{0.1}$ & $\underline{\mathbf{84.6}}_{0.1}$ \\
{} & 1-Shot/SIFT  & $95.0_{0.7}$ & $95.2_{0.2}$ & $26.5_{2.1}$ & $23.6_{2.4}$ & $71.3_{2.8}$ & $74.6_{1.6}$ & $84.2_{0.5}$ & $84.0_{0.4}$ \\
{} & 5-Shot/SIFT  & $94.1_{1.3}$ & $94.4_{0.4}$ & $23.9_{2.5}$ & $27.1_{3.2}$ & $70.1_{3.2}$ & $72.0_{2.3}$ & $82.3_{0.6}$ & $83.1_{0.1}$ \\
{} & 10-Shot/SIFT  & $96.1_{0.0}$ & $94.6_{0.4}$ & $25.8_{1.6}$ & $\underline{\mathbf{29.9}}_{1.6}$ & $71.1_{2.2}$ & $75.1_{1.6}$ & $80.4_{0.6}$ & $83.3_{0.1}$ \\
\midrule

\multirow{4}{*}{\rotatebox[origin=c]{90}{\shortstack{Llama3.1-8B \\ Instruct}}}   &   0-Shot/SFT  & $96.3_{0.1}$ & $96.3_{0.1}$ & $\mathbf{26.8}_{1.0}$ & $26.8_{1.0}$ & $\mathbf{73.8}_{4.2}$ & $73.8_{4.2}$ & $\underline{\mathbf{85.0}}_{0.1}$ & $\underline{\mathbf{85.0}}_{0.1}$ \\
{} & 1-Shot/SIFT  & $96.1_{0.4}$ & $96.1_{0.3}$ & $23.9_{1.3}$ & $22.2_{1.4}$ & $71.9_{2.7}$ & $74.7_{1.7}$ & $83.7_{0.3}$ & $84.2_{0.1}$ \\
{} & 5-Shot/SIFT  & $95.7_{0.4}$ & $96.1_{0.1}$ & $22.3_{3.8}$ & $23.2_{2.5}$ & $70.1_{1.7}$ & $73.6_{1.4}$ & $82.8_{0.5}$ & $83.0_{0.4}$ \\
{} & 10-Shot/SIFT  & $\mathbf{96.3}_{0.1}$ & $\underline{\mathbf{96.4}}_{0.1}$ & $22.0_{0.9}$ & $\underline{\mathbf{29.2}}_{3.2}$ & $69.7_{1.5}$ & $\underline{\mathbf{75.6}}_{1.7}$ & $81.9_{0.5}$ & $83.3_{0.2}$ \\
\midrule

\multirow{4}{*}{\rotatebox[origin=c]{90}{\shortstack{Mistral-7B \\ Instruct}}}   &   0-Shot/SFT  & $96.3_{0.1}$ & $96.3_{0.1}$ & $28.1_{2.3}$ & $28.1_{2.3}$ & $\underline{\mathbf{76.6}}_{1.4}$ & $\underline{\mathbf{76.6}}_{1.4}$ & $85.6_{0.2}$ & $85.6_{0.2}$ \\
{} & 1-Shot/SIFT  & $96.5_{0.1}$ & $\underline{\mathbf{96.6}}_{0.2}$ & $\mathbf{28.4}_{2.3}$ & $30.0_{2.7}$ & $73.4_{3.6}$ & $75.7_{2.5}$ & $\underline{\mathbf{85.9}}_{0.2}$ & $\mathbf{85.6}_{0.1}$ \\
{} & 5-Shot/SIFT  & $96.3_{0.2}$ & $96.3_{0.1}$ & $27.1_{2.5}$ & $29.6_{3.9}$ & $75.4_{1.2}$ & $75.2_{1.5}$ & $84.9_{0.2}$ & $84.5_{0.2}$ \\
{} & 10-Shot/SIFT  & $\mathbf{96.5}_{0.1}$ & $96.5_{0.1}$ & $28.2_{1.8}$ & $\underline{\mathbf{33.1}}_{2.2}$ & $74.1_{1.5}$ & $76.2_{0.2}$ & $81.4_{1.0}$ & $76.8_{1.2}$ \\
\midrule
\multicolumn{2}{l}{Gemma-7B (No CM)} & \multicolumn{2}{c}{$95.2_{0.4}$} & \multicolumn{2}{c}{$5.2_{3.0}$} & \multicolumn{2}{c}{$68.5_{3.1}$} & \multicolumn{2}{c}{$39.9_{2.6}$} \\
\multicolumn{2}{l}{Llama2-7B (No CM)} & \multicolumn{2}{c}{$94.4_{1.9}$} & \multicolumn{2}{c}{$0.0_{0.0}$} & \multicolumn{2}{c}{$51.6_{1.4}$} & \multicolumn{2}{c}{$42.1_{2.8}$} \\
\multicolumn{2}{l}{Llama3-8B (No CM)} & \multicolumn{2}{c}{$95.9_{0.2}$} & \multicolumn{2}{c}{$1.5_{0.5}$} & \multicolumn{2}{c}{$66.9_{4.0}$} & \multicolumn{2}{c}{$38.0_{3.1}$} \\
\multicolumn{2}{l}{Llama3.1-8B (No CM)} & \multicolumn{2}{c}{$95.9_{0.2}$} & \multicolumn{2}{c}{$7.2_{1.6}$} & \multicolumn{2}{c}{$70.6_{2.0}$} & \multicolumn{2}{c}{$39.7_{3.1}$} \\
\multicolumn{2}{l}{Mistral-7B (No CM)} & \multicolumn{2}{c}{$96.3_{0.2}$} & \multicolumn{2}{c}{$9.7_{4.5}$} & \multicolumn{2}{c}{$73.3_{1.4}$} & \multicolumn{2}{c}{$38.1_{2.3}$} \\\bottomrule
\end{tabular}}}
\caption[Micro F1 sequence labeling scores on the validation set using \textit{instruct} variants of decoder-only PLMs under SFT, SIFT, and CM removal]{Micro F1 sequence labeling scores on the validation set using five \textit{instruct} variants of decoder-only PLMs on SFT and SIFT and for a varying number of shots. We report scores for SRC and MRC CLM strategies. We also include experiments for the CM removal (the results are given for the same \textit{instruct} PLMs). Zero shots are the results of SFT experiments, while one, five, and ten shots are SIFT experiment results. The models are evaluated with the same number of in-context demonstrations as during fine-tuning. The results are given for four sequence labeling datasets. All results are averages of four runs. The context for the higher number of shots was kept fixed for a fixed seed. Standard deviations are shown in subscript. Best F1 scores per model, dataset, and CLM strategy are shown in bold. Additionally, the overall best F1 scores between the SRC and MRC CLM strategies per model and dataset are underlined. For convenience, we treat zero-shot MRC as SFT and report the corresponding SRC results.} \label{tab:sift_instruct_validation_results}
\end{table*}

\begin{table*}
\centering
\adjustbox{width=0.85\linewidth}{
\small{\begin{tabular}{l|lcccccccc}
\toprule
\multicolumn{2}{l}{\multirow{2}{*}{}} & \multicolumn{2}{c}{\textbf{CoNLL03}} & \multicolumn{2}{c}{\textbf{AAC-MW}} & \multicolumn{2}{c}{\textbf{NLU++}} & \multicolumn{2}{c}{\textbf{OntoNotes v5.0}} \\
\cmidrule(lr){3-4} \cmidrule(lr){5-6} \cmidrule(lr){7-8} \cmidrule(lr){9-10}
\multicolumn{1}{l}{Model} & \multicolumn{1}{l}{Setup} & \textbf{SRC} & \textbf{MRC} & \textbf{SRC} & \textbf{MRC} & \textbf{SRC} & \textbf{MRC} & \textbf{SRC} & \textbf{MRC} \\
\midrule
\multirow{4}{*}{\rotatebox[origin=c]{90}{\shortstack{Gemma-7B}}}   &   0-Shot/SFT  & $96.5_{0.3}$ & $96.5_{0.3}$ & $30.2_{0.9}$ & $30.2_{0.9}$ & $\underline{\mathbf{78.8}}_{1.4}$ & $\underline{\mathbf{78.8}}_{1.4}$ & $85.6_{0.4}$ & $85.6_{0.4}$ \\
{} & 1-Shot/SIFT  & $\mathbf{96.7}_{0.2}$ & $96.7_{0.1}$ & $\underline{\mathbf{31.3}}_{1.3}$ & $\mathbf{31.0}_{3.6}$ & $76.0_{1.8}$ & $77.1_{1.8}$ & $\underline{\mathbf{85.9}}_{0.1}$ & $\mathbf{85.8}_{0.1}$ \\
{} & 5-Shot/SIFT  & $96.5_{0.2}$ & $96.5_{0.1}$ & $27.9_{1.2}$ & $30.0_{3.2}$ & $75.7_{2.5}$ & $76.6_{2.4}$ & $84.8_{0.3}$ & $84.6_{0.5}$ \\
{} & 10-Shot/SIFT  & $96.6_{0.2}$ & $\underline{\mathbf{96.9}}_{0.1}$ & $26.1_{0.9}$ & $27.8_{2.0}$ & $76.7_{2.5}$ & $76.7_{1.2}$ & $80.2_{8.7}$ & $83.1_{0.7}$ \\
\midrule

\multirow{4}{*}{\rotatebox[origin=c]{90}{\shortstack{Llama2-7B}}}   &   0-Shot/SFT  & $96.0_{0.2}$ & $96.0_{0.2}$ & $\underline{\mathbf{26.5}}_{1.8}$ & $\underline{\mathbf{26.5}}_{1.8}$ & $66.3_{2.9}$ & $66.3_{2.9}$ & $83.9_{0.5}$ & $\mathbf{83.9}_{0.5}$ \\
{} & 1-Shot/SIFT  & $\mathbf{96.0}_{0.0}$ & $\underline{\mathbf{96.1}}_{0.1}$ & $21.3_{3.4}$ & $24.9_{3.9}$ & $\underline{\mathbf{69.9}}_{1.6}$ & $68.8_{2.8}$ & $\underline{\mathbf{84.2}}_{0.3}$ & $83.4_{0.8}$ \\
{} & 5-Shot/SIFT  & $95.5_{0.3}$ & $95.6_{0.4}$ & $23.8_{2.4}$ & $22.3_{1.4}$ & $63.6_{2.3}$ & $\mathbf{69.9}_{5.3}$ & $82.2_{0.3}$ & $81.2_{0.9}$ \\
{} & 10-Shot/SIFT  & $96.0_{0.1}$ & $95.8_{0.2}$ & $25.7_{0.4}$ & $25.2_{3.7}$ & $67.5_{2.0}$ & $68.5_{3.0}$ & $71.6_{2.1}$ & $56.6_{7.6}$ \\
\midrule

\multirow{4}{*}{\rotatebox[origin=c]{90}{\shortstack{Llama3-8B}}}   &   0-Shot/SFT  & $96.1_{0.3}$ & $96.1_{0.3}$ & $21.8_{1.5}$ & $21.8_{1.5}$ & $\underline{\mathbf{73.3}}_{3.2}$ & $\underline{\mathbf{73.3}}_{3.2}$ & $\underline{\mathbf{84.4}}_{0.2}$ & $84.4_{0.2}$ \\
{} & 1-Shot/SIFT  & $\mathbf{96.1}_{0.2}$ & $95.9_{0.2}$ & $24.1_{2.1}$ & $24.0_{1.4}$ & $71.4_{2.3}$ & $72.8_{1.4}$ & $84.3_{0.3}$ & $\underline{\mathbf{84.4}}_{0.2}$ \\
{} & 5-Shot/SIFT  & $92.9_{2.9}$ & $95.7_{0.2}$ & $24.2_{2.3}$ & $25.6_{1.4}$ & $71.0_{2.4}$ & $72.6_{1.9}$ & $81.5_{0.6}$ & $83.2_{0.4}$ \\
{} & 10-Shot/SIFT  & $95.2_{0.4}$ & $\underline{\mathbf{96.2}}_{0.2}$ & $\mathbf{25.1}_{0.8}$ & $\underline{\mathbf{27.9}}_{2.2}$ & $70.9_{2.5}$ & $72.7_{1.3}$ & $81.7_{2.4}$ & $83.6_{0.5}$ \\
\midrule

\multirow{4}{*}{\rotatebox[origin=c]{90}{\shortstack{Llama3.1-8B }}}   &   0-Shot/SFT  & $\mathbf{96.3}_{0.2}$ & $96.3_{0.2}$ & $21.3_{2.2}$ & $21.3_{2.2}$ & $\underline{\mathbf{75.8}}_{1.8}$ & $\underline{\mathbf{75.8}}_{1.8}$ & $\underline{\mathbf{84.5}}_{0.3}$ & $\underline{\mathbf{84.5}}_{0.3}$ \\
{} & 1-Shot/SIFT  & $96.2_{0.2}$ & $96.2_{0.2}$ & $23.5_{3.9}$ & $23.8_{2.0}$ & $72.0_{0.6}$ & $72.8_{2.0}$ & $84.1_{0.6}$ & $84.2_{0.3}$ \\
{} & 5-Shot/SIFT  & $95.1_{0.6}$ & $96.0_{0.2}$ & $23.5_{4.0}$ & $26.0_{1.9}$ & $67.6_{2.7}$ & $72.7_{0.9}$ & $81.1_{1.1}$ & $82.8_{0.3}$ \\
{} & 10-Shot/SIFT  & $94.7_{0.5}$ & $\underline{\mathbf{96.6}}_{0.1}$ & $\mathbf{25.9}_{1.2}$ & $\underline{\mathbf{27.9}}_{3.4}$ & $68.6_{1.8}$ & $72.7_{1.3}$ & $81.9_{0.6}$ & $83.7_{0.3}$ \\
\midrule

\multirow{4}{*}{\rotatebox[origin=c]{90}{\shortstack{Mistral-7B}}}   &   0-Shot/SFT  & $96.4_{0.2}$ & $96.4_{0.2}$ & $\mathbf{30.1}_{2.4}$ & $30.1_{2.4}$ & $76.1_{2.0}$ & $76.1_{2.0}$ & $\mathbf{85.9}_{0.3}$ & $85.9_{0.3}$ \\
{} & 1-Shot/SIFT  & $96.6_{0.2}$ & $96.6_{0.2}$ & $29.4_{3.5}$ & $30.2_{3.3}$ & $\mathbf{76.9}_{2.5}$ & $\underline{\mathbf{77.4}}_{1.7}$ & $85.8_{0.2}$ & $\underline{\mathbf{85.9}}_{0.1}$ \\
{} & 5-Shot/SIFT  & $96.4_{0.2}$ & $96.6_{0.1}$ & $28.0_{3.0}$ & $29.8_{3.8}$ & $71.4_{2.2}$ & $77.2_{1.0}$ & $84.8_{0.3}$ & $84.7_{0.2}$ \\
{} & 10-Shot/SIFT  & $\mathbf{96.6}_{0.2}$ & $\underline{\mathbf{96.7}}_{0.2}$ & $25.8_{1.3}$ & $\underline{\mathbf{33.8}}_{4.0}$ & $74.5_{1.6}$ & $77.0_{1.4}$ & $75.9_{0.7}$ & $72.4_{6.4}$ \\
\midrule
\multicolumn{2}{l}{Gemma-7B (No CM)} & \multicolumn{2}{c}{$95.2_{0.6}$} & \multicolumn{2}{c}{$0.0_{0.0}$} & \multicolumn{2}{c}{$54.7_{1.2}$} & \multicolumn{2}{c}{$40.1_{3.1}$} \\
\multicolumn{2}{l}{Llama2-7B (No CM)} & \multicolumn{2}{c}{$95.3_{0.4}$} & \multicolumn{2}{c}{$0.0_{0.0}$} & \multicolumn{2}{c}{$51.5_{0.5}$} & \multicolumn{2}{c}{$40.9_{2.8}$} \\
\multicolumn{2}{l}{Llama3-8B (No CM)} & \multicolumn{2}{c}{$95.9_{0.2}$} & \multicolumn{2}{c}{$5.7_{0.4}$} & \multicolumn{2}{c}{$69.6_{3.0}$} & \multicolumn{2}{c}{$38.4_{3.2}$} \\
\multicolumn{2}{l}{Llama3.1-8B (No CM)} & \multicolumn{2}{c}{$95.9_{0.3}$} & \multicolumn{2}{c}{$7.2_{2.1}$} & \multicolumn{2}{c}{$70.2_{1.9}$} & \multicolumn{2}{c}{$38.9_{3.0}$} \\
\multicolumn{2}{l}{Mistral-7B (No CM)} & \multicolumn{2}{c}{$96.4_{0.2}$} & \multicolumn{2}{c}{$9.9_{4.0}$} & \multicolumn{2}{c}{$73.5_{1.8}$} & \multicolumn{2}{c}{$37.7_{3.1}$} \\ 
\bottomrule
\end{tabular}}}
\caption[Micro F1 sequence labeling scores on the validation set using \textit{base} variants of decoder-only PLMs under SFT, SIFT, and CM removal]{Micro F1 sequence labeling scores on the validation set using five \textit{base} variants of decoder-only PLMs on SFT and SIFT and for a varying number of shots. We report scores for SRC and MRC CLM strategies. We also include experiments for the CM removal. Zero shots are the results of SFT experiments, while one, five, and ten shots are SIFT experiment results. The models are evaluated with the same number of in-context demonstrations as during fine-tuning. The results are given for four sequence labeling datasets. All results are averages of four runs. The context for the higher number of shots was kept fixed for a fixed seed. Standard deviations are shown in subscript. Best F1 scores per model, dataset, and CLM strategy are shown in bold. Additionally, the overall best F1 scores between the SRC and MRC CLM strategies per model and dataset are underlined. For convenience, we treat zero-shot MRC as SFT and report the corresponding SRC results.} \label{tab:sift_base_validation_results}
\end{table*}

\begin{table*}
\centering
\adjustbox{width=0.85\linewidth}{
\small{\begin{tabular}{l|lcccccccc}
\toprule
\multicolumn{2}{l}{\multirow{2}{*}{}} & \multicolumn{2}{c}{\textbf{CoNLL03}} & \multicolumn{2}{c}{\textbf{AAC-MW}} & \multicolumn{2}{c}{\textbf{NLU++}} & \multicolumn{2}{c}{\textbf{OntoNotes v5.0}} \\
\cmidrule(lr){3-4} \cmidrule(lr){5-6} \cmidrule(lr){7-8} \cmidrule(lr){9-10}
\multicolumn{1}{l}{Model} & \multicolumn{1}{l}{Setup} & \textbf{SRC} & \textbf{MRC} & \textbf{SRC} & \textbf{MRC} & \textbf{SRC} & \textbf{MRC} & \textbf{SRC} & \textbf{MRC} \\
\midrule
\multirow{4}{*}{\rotatebox[origin=c]{90}{\shortstack{Gemma-7B \\ Instruct}}}   &   0-Shot/SFT  & $92.9_{0.2}$ & $92.9_{0.2}$ & $34.9_{2.1}$ & $34.9_{2.1}$ & $\mathbf{74.5}_{2.2}$ & $74.5_{2.2}$ & $83.3_{0.2}$ & $\mathbf{83.3}_{0.2}$ \\
{} & 1-Shot/SIFT  & $92.9_{0.2}$ & $93.1_{0.2}$ & $\underline{\mathbf{36.4}}_{1.4}$ & $34.9_{2.4}$ & $73.2_{1.6}$ & $74.1_{1.4}$ & $\underline{\mathbf{83.4}}_{0.1}$ & $83.1_{0.2}$ \\
{} & 5-Shot/SIFT  & $92.9_{0.3}$ & $92.9_{0.3}$ & $30.8_{1.5}$ & $\mathbf{36.1}_{2.2}$ & $72.3_{2.7}$ & $72.6_{1.2}$ & $82.5_{0.1}$ & $81.6_{0.4}$ \\
{} & 10-Shot/SIFT  & $\mathbf{93.0}_{0.3}$ & $\underline{\mathbf{93.4}}_{0.2}$ & $32.3_{1.7}$ & $35.9_{2.7}$ & $72.4_{2.0}$ & $\underline{\mathbf{75.8}}_{1.0}$ & $55.5_{7.7}$ & $81.8_{0.1}$ \\
\midrule

\multirow{4}{*}{\rotatebox[origin=c]{90}{\shortstack{Llama2-7B \\ Instruct}}}   &   0-Shot/SFT  & $92.9_{0.2}$ & $92.9_{0.2}$ & $\mathbf{25.1}_{1.9}$ & $25.1_{1.9}$ & $68.5_{2.2}$ & $68.5_{2.2}$ & $81.7_{0.2}$ & $\mathbf{81.7}_{0.2}$ \\
{} & 1-Shot/SIFT  & $92.7_{0.5}$ & $92.8_{0.2}$ & $21.3_{2.7}$ & $\underline{\mathbf{25.6}}_{1.6}$ & $\mathbf{69.5}_{0.5}$ & $70.0_{3.0}$ & $\underline{\mathbf{81.8}}_{0.1}$ & $81.0_{0.3}$ \\
{} & 5-Shot/SIFT  & $92.8_{0.2}$ & $92.6_{0.3}$ & $22.3_{1.0}$ & $22.9_{2.8}$ & $64.5_{2.3}$ & $69.8_{1.6}$ & $80.3_{0.3}$ & $77.9_{0.9}$ \\
{} & 10-Shot/SIFT  & $\mathbf{93.0}_{0.3}$ & $\underline{\mathbf{93.1}}_{0.3}$ & $24.8_{1.0}$ & $23.7_{3.7}$ & $68.7_{3.4}$ & $\underline{\mathbf{72.1}}_{1.7}$ & $72.4_{0.5}$ & $60.9_{3.5}$ \\
\midrule

\multirow{4}{*}{\rotatebox[origin=c]{90}{\shortstack{Llama3-8B \\ Instruct}}}   &   0-Shot/SFT  & $\underline{\mathbf{93.2}}_{0.3}$ & $\underline{\mathbf{93.2}}_{0.3}$ & $\mathbf{29.9}_{1.9}$ & $29.9_{1.9}$ & $\underline{\mathbf{73.2}}_{1.2}$ & $\underline{\mathbf{73.2}}_{1.2}$ & $\underline{\mathbf{82.8}}_{0.4}$ & $\underline{\mathbf{82.8}}_{0.4}$ \\
{} & 1-Shot/SIFT  & $91.7_{0.5}$ & $92.2_{0.2}$ & $29.8_{1.9}$ & $31.6_{2.7}$ & $69.9_{1.3}$ & $72.3_{0.3}$ & $82.5_{0.6}$ & $82.3_{0.3}$ \\
{} & 5-Shot/SIFT  & $91.6_{1.0}$ & $91.2_{0.5}$ & $25.4_{1.2}$ & $31.5_{1.2}$ & $69.5_{1.9}$ & $68.5_{2.7}$ & $80.3_{0.5}$ & $81.2_{0.1}$ \\
{} & 10-Shot/SIFT  & $93.1_{0.4}$ & $91.7_{0.5}$ & $29.5_{3.1}$ & $\underline{\mathbf{34.6}}_{1.3}$ & $70.2_{2.6}$ & $72.1_{0.5}$ & $78.6_{0.5}$ & $81.4_{0.2}$ \\
\midrule

\multirow{4}{*}{\rotatebox[origin=c]{90}{\shortstack{Llama3.1-8B \\ Instruct}}}   &   0-Shot/SFT  & $\underline{\mathbf{93.3}}_{0.1}$ & $\underline{\mathbf{93.3}}_{0.1}$ & $29.5_{2.0}$ & $29.5_{2.0}$ & $71.7_{2.3}$ & $71.7_{2.3}$ & $\underline{\mathbf{82.9}}_{0.2}$ & $\underline{\mathbf{82.9}}_{0.2}$ \\
{} & 1-Shot/SIFT  & $92.9_{0.4}$ & $93.1_{0.3}$ & $\mathbf{30.9}_{2.3}$ & $30.2_{1.9}$ & $\mathbf{72.4}_{0.8}$ & $\underline{\mathbf{72.6}}_{2.3}$ & $81.9_{0.4}$ & $82.2_{0.1}$ \\
{} & 5-Shot/SIFT  & $93.0_{0.4}$ & $93.0_{0.3}$ & $24.3_{1.9}$ & $32.7_{0.9}$ & $67.6_{0.8}$ & $70.0_{2.4}$ & $80.7_{0.5}$ & $80.9_{0.3}$ \\
{} & 10-Shot/SIFT  & $93.3_{0.2}$ & $93.2_{0.1}$ & $27.6_{2.2}$ & $\underline{\mathbf{35.0}}_{0.4}$ & $70.0_{2.5}$ & $72.4_{1.0}$ & $79.3_{0.5}$ & $81.2_{0.1}$ \\
\midrule

\multirow{4}{*}{\rotatebox[origin=c]{90}{\shortstack{Mistral-7B \\ Instruct}}}   &   0-Shot/SFT  & $93.2_{0.1}$ & $93.2_{0.1}$ & $33.7_{1.5}$ & $33.7_{1.5}$ & $\mathbf{76.6}_{1.5}$ & $76.6_{1.5}$ & $83.7_{0.2}$ & $83.7_{0.2}$ \\
{} & 1-Shot/SIFT  & $\mathbf{93.4}_{0.4}$ & $\underline{\mathbf{93.6}}_{0.4}$ & $35.7_{1.8}$ & $34.7_{2.4}$ & $72.6_{1.7}$ & $75.5_{0.8}$ & $\underline{\mathbf{84.1}}_{0.1}$ & $\mathbf{83.8}_{0.1}$ \\
{} & 5-Shot/SIFT  & $93.3_{0.3}$ & $93.2_{0.2}$ & $33.8_{2.4}$ & $35.5_{1.6}$ & $73.6_{1.3}$ & $75.6_{1.5}$ & $83.2_{0.2}$ & $82.3_{0.2}$ \\
{} & 10-Shot/SIFT  & $93.2_{0.1}$ & $93.4_{0.2}$ & $\mathbf{36.2}_{2.2}$ & $\underline{\mathbf{37.3}}_{0.2}$ & $73.6_{1.0}$ & $\underline{\mathbf{77.2}}_{0.7}$ & $79.8_{0.9}$ & $75.0_{1.5}$ \\
\bottomrule
\end{tabular}}}
\caption[Micro F1 sequence labeling scores on the test set using \textit{instruct} variants of decoder-only PLMs on SFT and SIFT]{Micro F1 sequence labeling scores on the test set using five \textit{instruct} variants of decoder-only PLMs on SFT and SIFT and for a varying number of shots. We report scores for SRC and MRC CLM strategies. Zero shots are the results of SFT experiments, while one, five, and ten shots are SIFT experiment results. The models are evaluated with the same number of in-context demonstrations as during fine-tuning. The results are given for four sequence labeling datasets. All results are averages of four runs. The context for the higher number of shots was kept fixed for a fixed seed. Standard deviations are shown in subscript. Best F1 scores per model, dataset, and CLM strategy are shown in bold. Additionally, the overall best F1 scores between the SRC and MRC CLM strategies per model and dataset are underlined. For convenience, we treat zero-shot MRC as SFT and report the corresponding SRC results.} \label{tab:sift_instruct_test_results}
\end{table*}

\begin{table*}
\centering
\adjustbox{width=0.85\linewidth}{
\small{\begin{tabular}{l|lcccccccc}
\toprule
\multicolumn{2}{l}{\multirow{2}{*}{}} & \multicolumn{2}{c}{\textbf{CoNLL03}} & \multicolumn{2}{c}{\textbf{AAC-MW}} & \multicolumn{2}{c}{\textbf{NLU++}} & \multicolumn{2}{c}{\textbf{OntoNotes v5.0}} \\
\cmidrule(lr){3-4} \cmidrule(lr){5-6} \cmidrule(lr){7-8} \cmidrule(lr){9-10}
\multicolumn{1}{l}{Model} & \multicolumn{1}{l}{Setup} & \textbf{SRC} & \textbf{MRC} & \textbf{SRC} & \textbf{MRC} & \textbf{SRC} & \textbf{MRC} & \textbf{SRC} & \textbf{MRC} \\
\midrule
\multirow{4}{*}{\rotatebox[origin=c]{90}{\shortstack{Gemma-7B}}}   &   0-Shot/SFT  & $93.2_{0.3}$ & $93.2_{0.3}$ & $35.9_{1.4}$ & $35.9_{1.4}$ & $\underline{\mathbf{76.6}}_{0.7}$ & $\underline{\mathbf{76.6}}_{0.7}$ & $84.0_{0.1}$ & $\mathbf{84.0}_{0.1}$ \\
{} & 1-Shot/SIFT  & $93.4_{0.3}$ & $93.6_{0.2}$ & $\underline{\mathbf{37.2}}_{1.7}$ & $\mathbf{36.8}_{1.4}$ & $76.0_{1.0}$ & $75.0_{0.6}$ & $\underline{\mathbf{84.4}}_{0.1}$ & $84.0_{0.2}$ \\
{} & 5-Shot/SIFT  & $93.4_{0.2}$ & $93.4_{0.1}$ & $35.5_{0.5}$ & $35.7_{1.6}$ & $75.2_{0.8}$ & $75.1_{0.7}$ & $83.2_{0.2}$ & $82.7_{0.3}$ \\
{} & 10-Shot/SIFT  & $\mathbf{93.4}_{0.2}$ & $\underline{\mathbf{93.8}}_{0.1}$ & $36.6_{1.7}$ & $33.3_{1.0}$ & $74.7_{0.7}$ & $75.6_{1.7}$ & $78.2_{9.0}$ & $81.3_{0.7}$ \\
\midrule

\multirow{4}{*}{\rotatebox[origin=c]{90}{\shortstack{Llama2-7B}}}   &   0-Shot/SFT  & $93.0_{0.1}$ & $\mathbf{93.0}_{0.1}$ & $\mathbf{29.9}_{1.2}$ & $29.9_{1.2}$ & $66.9_{3.0}$ & $66.9_{3.0}$ & $81.7_{0.3}$ & $\mathbf{81.7}_{0.3}$ \\
{} & 1-Shot/SIFT  & $92.9_{0.2}$ & $93.0_{0.2}$ & $26.2_{1.3}$ & $\underline{\mathbf{31.2}}_{3.2}$ & $\mathbf{68.6}_{2.1}$ & $67.8_{1.3}$ & $\underline{\mathbf{82.2}}_{0.3}$ & $81.2_{0.7}$ \\
{} & 5-Shot/SIFT  & $93.0_{0.4}$ & $92.7_{0.3}$ & $22.1_{1.7}$ & $26.6_{0.4}$ & $65.4_{2.1}$ & $\underline{\mathbf{69.4}}_{3.3}$ & $80.3_{0.3}$ & $78.8_{1.0}$ \\
{} & 10-Shot/SIFT  & $\underline{\mathbf{93.2}}_{0.2}$ & $92.7_{0.2}$ & $26.0_{1.6}$ & $29.1_{4.4}$ & $68.1_{0.9}$ & $67.6_{1.3}$ & $69.6_{1.9}$ & $55.6_{7.5}$ \\
\midrule

\multirow{4}{*}{\rotatebox[origin=c]{90}{\shortstack{Llama3-8B}}}   &   0-Shot/SFT  & $92.8_{0.1}$ & $92.8_{0.1}$ & $\mathbf{32.1}_{0.5}$ & $32.1_{0.5}$ & $\mathbf{72.2}_{0.7}$ & $72.2_{0.7}$ & $\underline{\mathbf{82.8}}_{0.2}$ & $\underline{\mathbf{82.8}}_{0.2}$ \\
{} & 1-Shot/SIFT  & $\mathbf{93.1}_{0.3}$ & $92.6_{0.1}$ & $31.9_{1.1}$ & $32.2_{2.3}$ & $70.5_{1.0}$ & $71.0_{1.0}$ & $82.6_{0.3}$ & $82.5_{0.3}$ \\
{} & 5-Shot/SIFT  & $89.4_{3.1}$ & $92.6_{0.2}$ & $26.0_{0.5}$ & $30.4_{1.0}$ & $68.9_{1.8}$ & $69.3_{0.9}$ & $79.5_{0.6}$ & $81.2_{0.2}$ \\
{} & 10-Shot/SIFT  & $91.9_{0.8}$ & $\underline{\mathbf{93.3}}_{0.1}$ & $29.2_{2.7}$ & $\underline{\mathbf{32.5}}_{1.8}$ & $68.9_{1.6}$ & $\underline{\mathbf{72.6}}_{1.5}$ & $79.8_{2.0}$ & $81.5_{0.5}$ \\
\midrule

\multirow{4}{*}{\rotatebox[origin=c]{90}{\shortstack{Llama3.1-8B }}}   &   0-Shot/SFT  & $93.0_{0.2}$ & $93.0_{0.2}$ & $\mathbf{30.9}_{1.1}$ & $30.9_{1.1}$ & $\underline{\mathbf{73.9}}_{1.0}$ & $\underline{\mathbf{73.9}}_{1.0}$ & $\underline{\mathbf{82.4}}_{0.2}$ & $\underline{\mathbf{82.4}}_{0.2}$ \\
{} & 1-Shot/SIFT  & $\mathbf{93.1}_{0.2}$ & $93.3_{0.2}$ & $30.9_{2.1}$ & $32.4_{1.8}$ & $70.5_{0.5}$ & $72.5_{1.1}$ & $82.0_{0.5}$ & $81.9_{0.3}$ \\
{} & 5-Shot/SIFT  & $92.1_{0.3}$ & $93.1_{0.2}$ & $26.3_{2.3}$ & $31.6_{2.3}$ & $68.1_{1.3}$ & $72.1_{0.4}$ & $79.2_{0.8}$ & $80.4_{0.2}$ \\
{} & 10-Shot/SIFT  & $91.5_{0.7}$ & $\underline{\mathbf{93.5}}_{0.2}$ & $30.3_{1.7}$ & $\underline{\mathbf{33.5}}_{2.5}$ & $69.0_{1.1}$ & $72.5_{0.7}$ & $79.8_{0.5}$ & $81.7_{0.2}$ \\
\midrule

\multirow{4}{*}{\rotatebox[origin=c]{90}{\shortstack{Mistral-7B}}}   &   0-Shot/SFT  & $93.3_{0.2}$ & $93.3_{0.2}$ & $32.7_{2.5}$ & $32.7_{2.5}$ & $\mathbf{75.0}_{1.8}$ & $75.0_{1.8}$ & $\mathbf{83.9}_{0.2}$ & $83.9_{0.2}$ \\
{} & 1-Shot/SIFT  & $93.3_{0.2}$ & $93.5_{0.3}$ & $34.7_{3.1}$ & $35.7_{2.3}$ & $74.8_{1.4}$ & $\underline{\mathbf{75.6}}_{0.1}$ & $83.9_{0.3}$ & $\underline{\mathbf{83.9}}_{0.1}$ \\
{} & 5-Shot/SIFT  & $93.2_{0.5}$ & $93.4_{0.2}$ & $32.2_{2.0}$ & $34.9_{2.0}$ & $71.5_{4.0}$ & $75.6_{0.5}$ & $83.1_{0.2}$ & $82.5_{0.3}$ \\
{} & 10-Shot/SIFT  & $\mathbf{93.4}_{0.1}$ & $\underline{\mathbf{93.5}}_{0.2}$ & $\mathbf{35.3}_{1.3}$ & $\underline{\mathbf{37.7}}_{1.1}$ & $72.8_{0.8}$ & $75.5_{0.5}$ & $74.4_{0.8}$ & $70.4_{6.2}$ \\
\bottomrule
\end{tabular}}}
\caption[Micro F1 sequence labeling scores on the test set using \textit{base} variants of decoder-only PLMs on SFT and SIFT]{Micro F1 sequence labeling scores on the test set using five \textit{base} variants of decoder-only PLMs on SFT and SIFT and for a varying number of shots. We report scores for SRC and MRC CLM paradigms. Zero shots are the results of SFT experiments, while one, five, and ten shots are SIFT experiment results. The models are evaluated with the same number of in-context demonstrations as during fine-tuning. The results are given for four sequence labeling datasets. All results are averages of four runs. The context for the higher number of shots was kept fixed for a fixed seed. Standard deviations are shown in subscript. Best F1 scores per model, dataset, and CLM strategy are shown in bold. Additionally, the overall best F1 scores between the SRC and MRC CLM strategies per model and dataset are underlined. For convenience, we treat zero-shot MRC as SFT and report the corresponding SRC results.} \label{tab:sift_base_test_results}
\end{table*}
 
\subsubsection{Validation Set Results}

For the validation set, we run all the evaluation experiments for the four sequence labeling tasks of NER, ABAM, slot labeling, and SRL (cf.~Figure \ref{fig:it_sift_validation_sl}). We also include the results for the three CLM strategies and \textit{instruct} and \textit{base} PLMs. Results reveal that vanilla CLM performs worse than SRC and MRC CLM. This is consistent across all datasets. Nevertheless, increasing the number of demonstrations for vanilla CLM brings significant gains. On average, adding more demonstrations at least does not degrade performance. These gains are most apparent for the AAC-MW and NLU++ datasets.

Consistent with vanilla CLM, we find that SRC and MRC CLM also benefit from an increased number of demonstrations during fine-tuning and inference, except for the SRL task. Providing the model with more than one demonstration for the SRL task typically results in a decline in performance. All other tasks measure at least some benefits from more demonstrations in the context. NER shows fewer gains than ABAM and slot labeling. ABAM appears to be the most challenging task for the models to learn, presumably due to the small overall number of training examples. 

On average, the MRC strategy shows improvements more often than SRC with the increase in the number of shots---aggregating the loss over multiple responses thus generally seems beneficial. MRC stabilizes the training and helps the model learn to leverage multiple responses in the context better than SRC. For both strategies, \textit{base} model variants score higher (winning in $96$ out of $160$ experiments) than the respective \textit{instruct} variants. Interestingly, Llama models perform worse than Gemma and Mistral in all experiments. Finally, when comparing standard SFT with SIFT, we observe that combinations of either (1) SRC + standard SFT or (2) MRC + SIFT yield the highest overall F1 score per model.

In comparison with the CM removal from all layers, SIFT strategies outperform the models with CM removed from all layers. To be more specific, CM removal performs on a par with best-performing SIFT models on NER, but falls well behind SIFT on other datasets, suggesting that perhaps a more nuanced, task-specific unmasking configurations are needed. The discrepancy between pre-training and fine-tuning is harder to narrow using CM removal when there is no training data in abundance. This is especially true for AAC-MW and NLU++ datasets and is reflected in the obtained results. Since the results were worse on the validation set for CM removal compared to SIFT, we omitted the results for CM removal on the test set. Results from Section \ref{sec:cm_removal_results} showed that the best setup was not the one where the CM was removed from all layers. However, finding the best configuration is computationally expensive.

\subsubsection{Test Set Results}

For the test set, we run all the evaluation experiments for the four sequence labeling tasks of NER, ABAM, slot labeling, and SRL (cf.~Figure \ref{fig:it_sift_test_sl}). However, we do not compare response-oriented CLM strategies with the vanilla CLM since vanilla delivered worse performance for sequence labeling tasks on the validation set than the SRC and MRC CLM strategies. Here, we also include the results for the \textit{instruct} and \textit{base} PLMs (cf.~Tables \ref{tab:sift_instruct_test_results} and \ref{tab:sift_base_test_results}). The trends observed on the validation set hold for the test set, as indicated by the similar shapes of the curves. Looking at the CLM-ing strategies/objectives, we find that, on average, MRC utilizes (more) shots better than SRC.

Furthermore, similar to validation set results, we observe notable improvements across all datasets except for SRL when increasing the number of demonstrations to $10$. However, even though SRL does not benefit from multiple demonstrations in the context often, the best results on SRL are obtained with one demonstration in the context during training and evaluation. The SRL dataset has the most classes of all the datasets ($27$), and this high number of classes and options results in longer contexts, which negatively impact performance. Notably, $1$-shot SIFT can achieve strong results (cf.~ABAM and SRL tasks), demonstrating the power of even minimal in-context supervision. Again, we observe lower performance for Llama models than Gemma and Mistral, indicating that Gemma and Mistral models and their pre-training procedures are better suited for transfer to sequence labeling tasks. \textit{Base} models show a better utilization of a higher number of demonstrations than \textit{instruct} models (winning in $94$ out of $160$ experiments).

\subsection{Comparison of ICL and SIFT}

The performance gap between ICL and SIFT is substantial, ranging from around $0.1$ (ABAM) to more than $0.6$ (NER, slot labeling, and SRL) F1 points in favor of fine-tuning, which is more than a double increase in the performance. \textit{Instruct} variants dominate ICL, while \textit{base} variants perform best in SIFT. Expectedly, ICL benefits more from a larger number of demonstrations than SIFT. With a large enough size of the fine-tuning dataset, the models saturate and do not benefit further from in-context demonstrations: This is consistent with prior work \citep{bertsch-etal-2025-context}. While more shots improve SIFT results, the benefits diminish with larger contexts (in the number of tokens), suggesting limits in the model's ability to effectively process extensive inputs \citep{10.1162/tacl_a_00638}.

\subsection{Instruction Ablation Results}

\begin{figure*}
\begin{center}
\includegraphics[width=\textwidth]{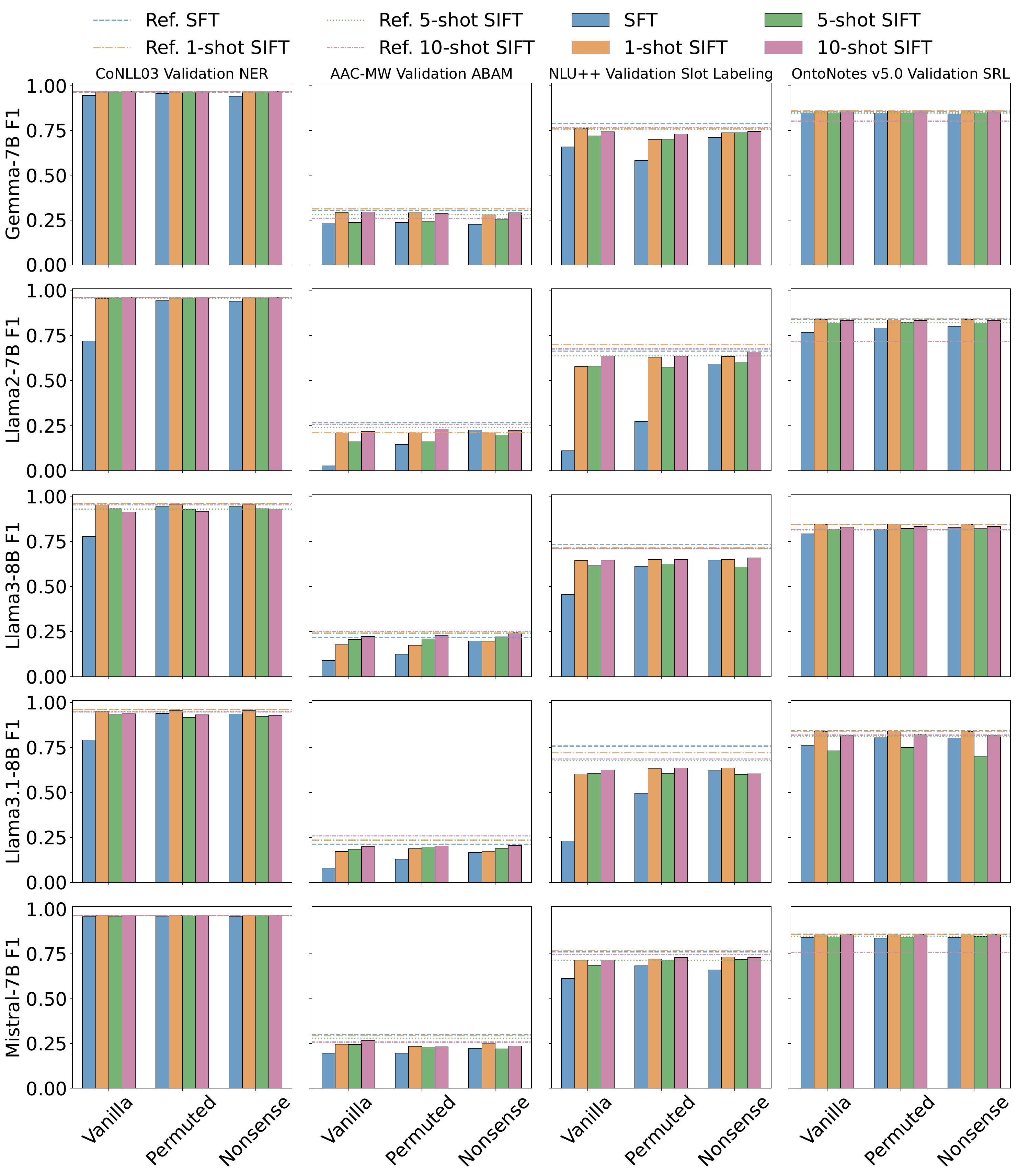}
\caption[Instruction ablation experiment results on the validation set for base decoder-only PLMs trained with SFT and SIFT (SRC and without instruction)]{Instruction variants applied on the validation set at inference time for five \textit{base} variants of decoder-only PLMs trained with SFT and SIFT using the SRC strategy and without instruction. The reference (Ref.) lines show the results for the same models trained with instructions. All results are averages of four runs.}
\label{fig:sift_src_base_noinstr}
\end{center}
\end{figure*}

\begin{figure*}
\begin{center}
\includegraphics[width=\textwidth]{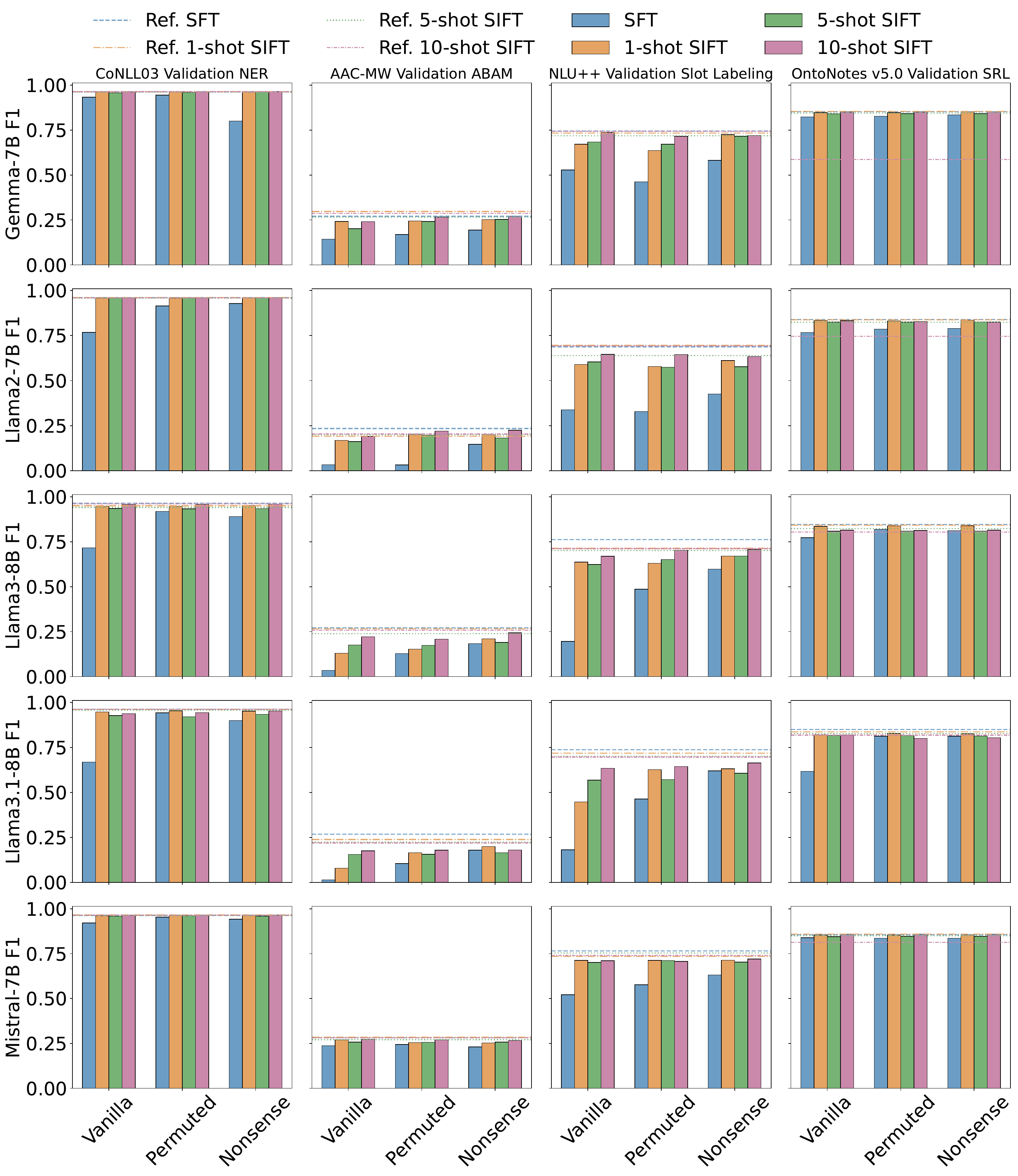}
\caption[Instruction ablation experiment results on the validation set for instruct decoder-only PLMs trained with SFT and SIFT (SRC and without instruction)]{Instruction variants applied on the validation set at inference time for five \textit{instruct} variants of decoder-only PLMs trained with SFT and SIFT using the SRC strategy and without instruction. The reference (Ref.) lines show the results for the same models trained with instructions. All results are averages of four runs.}
\label{fig:sift_src_instruct_noinstr}
\end{center}
\end{figure*}

\begin{figure*}
\begin{center}
\includegraphics[width=\textwidth]{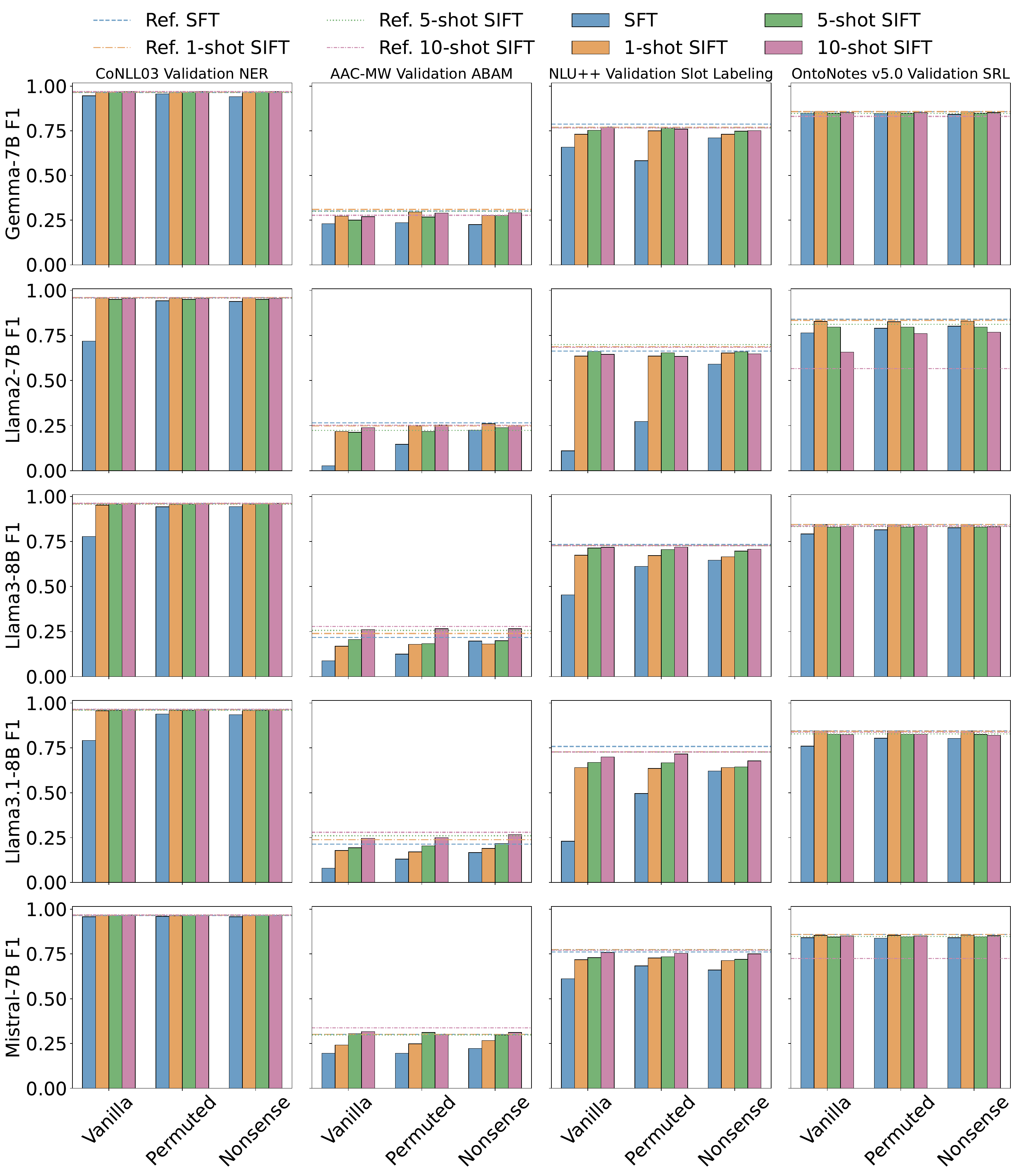}
\caption[Instruction ablation experiment results on the validation set for base decoder-only PLMs trained with SFT and SIFT (MRC and without instruction)]{Instruction variants applied on the validation set at inference time for five \textit{base} variants of decoder-only PLMs trained with SFT and SIFT using the MRC strategy and without instruction. The reference (Ref.) lines show the results for the same models trained with instructions. All results are averages of four runs.}
\label{fig:sift_mrc_base_noinstr}
\end{center}
\end{figure*}

\begin{figure*}
\begin{center}
\includegraphics[width=\textwidth]{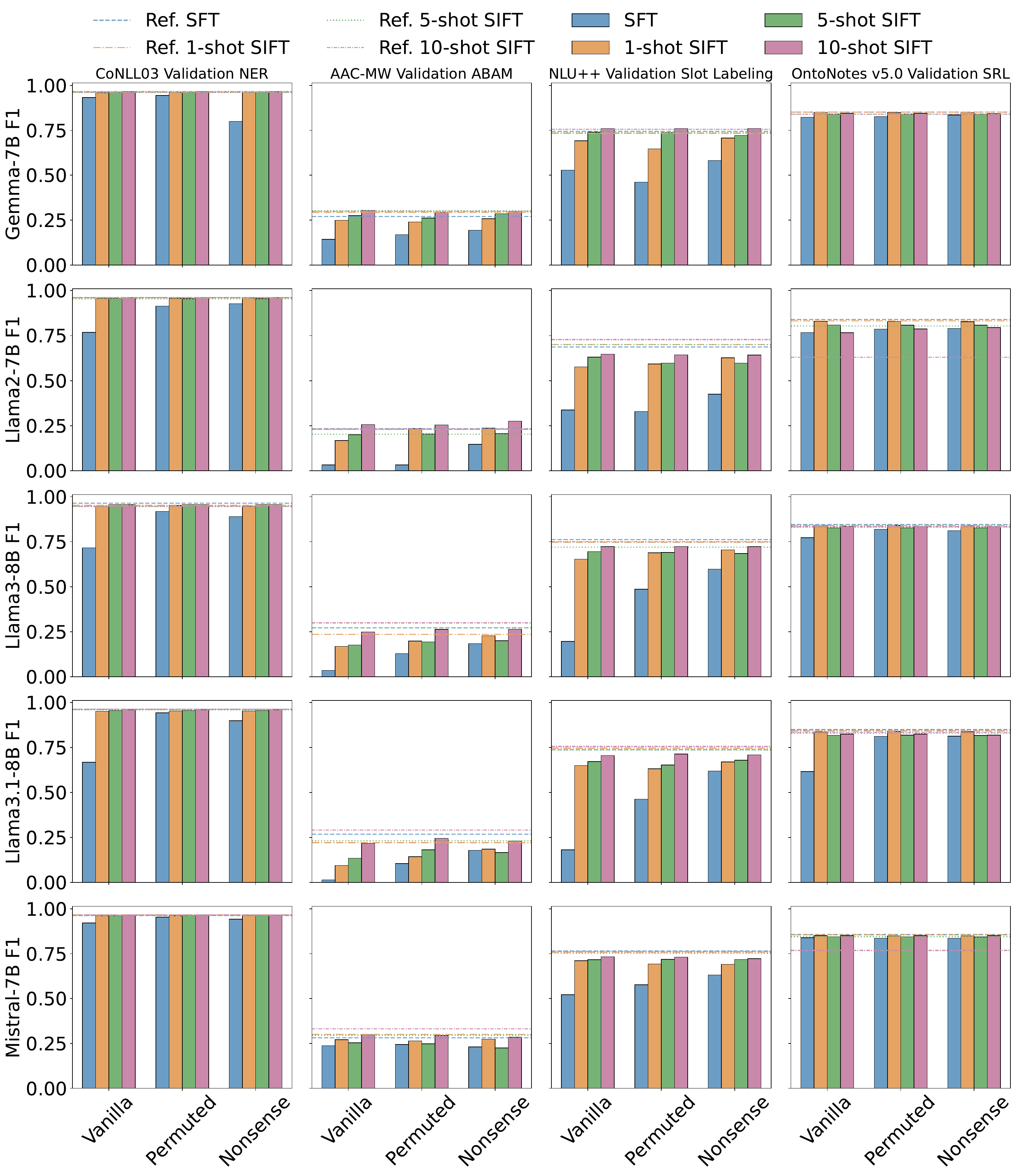}
\caption[Instruction ablation experiment results on the validation set for instruct decoder-only PLMs trained with SFT and SIFT (MRC and without instruction)]{Instruction variants applied on the validation set at inference time for five \textit{instruct} variants of decoder-only PLMs trained with SFT and SIFT using the MRC strategy and without instruction. The reference (Ref.) lines show the results for the same models trained with instructions. All results are averages of four runs.}
\label{fig:sift_mrc_instruct_noinstr}
\end{center}
\end{figure*}

Figures \ref{fig:sift_src_base_noinstr}, \ref{fig:sift_src_instruct_noinstr}, \ref{fig:sift_mrc_base_noinstr}, and \ref{fig:sift_mrc_instruct_noinstr} present the results of our experiments where the instruction was removed during fine-tuning with SFT and SIFT but reinstated during inference on the validation set. We report results with SRC and MRC strategies and for \textit{base} and \textit{instruct} decoder variants. We observe that including the instruction has a significantly negative impact on SFT performance, but to a lesser extent in SIFT. The SIFT performance is expectedly worse compared to having the instruction included in both training and inference, but the gap narrows with more in-context demonstrations. This suggests that at inference time, a SIFT model (trained without instruction) benefits more from informative in-context demonstrations than from the instruction. Interestingly, for SRL, SIFT models with more shots (see $10$-shot) trained without the instruction even perform better than their counterparts trained with the instruction: we believe that this is due to the drastic context length increase brought about by the instruction---due to the SRL's large number of classes ($27$)---which reduces the model's ability to effectively \textit{consume} the entire context including the provided in-context demonstrations.

\section{Summary} \label{sec:sift_summary}

The SIFT framework offered many interesting insights into solving sequence labeling tasks using the generative abilities of decoder-only LLMs. From a practical standpoint, we recommend using either SFT with SRC CLM or SIFT with MRC CLM during training for optimal results. We recommend matching the number of demonstrations used for fine-tuning during inference. These demonstrations can be the same as those used during training, as we merely use them to reduce the discrepancy between fine-tuning and inference while simultaneously making the model recall the examples seen during training through the ICL mechanism. ICL combined with constrained decoding during inference is key to getting the best out of decoders fine-tuned with in-context examples.  

The results showed that the proposed approaches from the SIFT framework yield better results than the CM removal when the mask is removed from all layers of the model. However, fine-tuning with in-context examples and inference in a generative manner comes with a higher computational cost. Finally, we recommend using task-specific instructions during training and inference, even though experiments with variations of the instruction showed that the instruction in the prompt is not pivotal for achieving top performance. 

For SFT and SIFT, we relied on PEFT, encoding the knowledge about the sequence labeling task into a PEFT module. By training models with CLM objectives, we do not observe the effects of the pre-train--fine-tune discrepancy. Adapting the model with a loss on specific tokens in the training prompts and PEFT under a unified framework addresses RQ3, demonstrating that decoder-only LLMs are optimally utilized for sequence labeling as text generators by leveraging in-context examples through fine-tuning. Our results indicate that this specific approach yields optimal performance for the sequence labeling tasks considered.

ICL, on its own, can serve as a baseline for sequence labeling tasks and datasets with few classes. However, getting the best generalization performance is always obtained with fine-tuning in combination with ICL. We recommend using \textit{base} PLMs for best results, as they exhibit greater SIFT consistency than \textit{instruct} PLMs, reducing variability across runs and configurations. Additionally, we recommend MRC CLM as a go-to response-oriented adaptation strategy. MRC CLM stabilizes performance drops when using a higher number of shots more frequently than SRC CLM, ensuring the model focuses on relevant response information during the adaptation phase for sequence labeling.

\clearpage{}%
\clearpage{}%
\chapter{Discussion}
\label{ch:discussion}

The improvements in the transfer learning process for sequence labeling tasks, as presented in this thesis, are achieved in the second phase of transfer learning: the adaptation phase. Improvements included (1) introducing an additional signal for domain-specific knowledge transfer for encoders (Chapter \ref{ch:method1}), (2) architectural modifications for decoders (Chapter \ref{ch:method2}), and (3) a framework for adapting autoregressive LLMs through changes in the optimization scheme (Chapter \ref{ch:method3}). In the following sections, we outline the findings with respect to the research questions from Chapter ~\ref{ch:intro}, as well as the implications and limitations of our findings and contributions.

\section{Findings}

In Chapter \ref{ch:intro}, we set out to answer three research questions, aiming to improve the performance of PLMs on various and diverse sequence labeling tasks through effective and efficient adaptation. This section provides a summary of our findings for each of these questions.

\paragraph{RQ1: How to improve the transfer with little (few-shot) to none (zero-shot) domain-specific data for sequence labeling?} In Chapter \ref{ch:method1} we showcased how the transfer for the ETD task from a high-resource source domain to the low(er)-resource target domain can be improved using an auxiliary, domain-independent OIE signal. This holds for both zero and few-shot cases. However, we also demonstrated that the effects of improved transfer diminish when we acquire a substantial number of target domain examples for fine-tuning. Generally, the models that leverage the signal in the \textit{implicit} (utilizing implicitly the OIE relation signal) way with \textit{sequential} (training first on the source data followed by training on the target domain data) transfer training achieve the best transfer performance. The transfer can be further improved with the domain adaptation methods and does not depend on the specific OIE system. Improving the performance for the sequence labeling task of ETD is a strong case of utilizing both inductive and transductive transfer learning methods to achieve optimal improvement in subsequent tasks within the EE pipeline, and, by extension, other tasks in the IE pipeline that rely on EE outputs.

\paragraph{RQ2: How to narrow the pre-train--fine-tune discrepancy effectively and efficiently during the adaptation phase of decoder-only LLMs for improved downstream sequence labeling performance?} To answer this research question in the context of sequence labeling tasks, we have focused on decoder-only LLMs---de facto standard for tackling many tasks in contemporary NLP. We identified their limitation, i.e., the discrepancy between pre-training and fine-tuning that arises when we try to use them as encoders for sequence labeling. To overcome this limitation and narrow the discrepancy, we introduced a novel \textit{layer group unmasking} method in Chapter \ref{ch:method2}, which enables the decoder models to simply and effectively leverage the right-side context, and which turned out to be a missing piece of the puzzle for improved performance on various sequence labeling tasks. By integrating this unmasking technique with the most advanced PEFT methods, which leverage reparameterization of PLM weights and quantization, we successfully achieved efficient adaptation in terms of both space and time complexity during LLM fine-tuning and inference. These findings open the possibilities for other methods that fiddle with the architecture of the model to improve the transfer learning for sequence labeling tasks.

\paragraph{RQ3: How to utilize decoder-only LLMs for improved sequence labeling performance in a parameter-efficient manner under a unified adaptation framework?} With this research question, we set out to design a fine-tuning framework that enables decoders to leverage their ICL capabilities during both fine-tuning and inference fully. We proposed this framework in Chapter \ref{ch:method3}. Given that decoder-only LLMs demonstrate remarkable ICL capabilities as their scale increases, the framework emphasizes efficient use of demonstrations in training and inference prompts. We unveiled that decoder-only LLMs are best utilized for sequence labeling with multiple demonstrations in context when guided to focus on responses rather than the entire training prompt. This was achieved by modifying the loss function to mask out all parts of the training prompt except the responses. Compared to using decoders as encoders by removing CM from all layers, we demonstrated that employing decoders as text generators yields superior performance. Encoding task knowledge into a PEFT module ensured parameter efficiency, while constrained decoding encouraged the models to generate higher-quality responses. Decoders are most effective as text generators when combined with their ICL capabilities, both of which are realized within the proposed response-oriented adaptation framework. Experiments with instruction removal revealed that, while instructions during training are beneficial, they are not crucial for decoder-only LLMs to solve sequence labeling tasks effectively.

\section{Implications}

As briefly discussed in Chapter \ref{ch:background}, there are numerous ways to adapt PLMs to various NLP tasks. However, adapting them to sequence labeling tasks remains a promising and underexplored research direction. Our results show that many sequence labeling tasks, such as NER, aspect-based sentiment analysis, text chunking, SRL, ABAM, ETD, ETC, and slot labeling, can be successfully addressed by intervening during the adaptation phase of transfer learning for PLMs. Pre-trained encoder-only models have proved effective for sequence labeling tasks. Therefore, we proposed introducing an external signal into the multi-task model architecture of encoder-only PLMs. However, given the research community's shift from scaling encoder models to focusing on decoder-based LLMs, this thesis emphasized improving sequence labeling tasks using decoder architectures. Our contribution in this regard is twofold: (1) a \textit{layer group unmasking} method that transforms decoder models into efficient encoders and (2) a response-oriented framework that enables decoder models to leverage multiple in-context examples during supervised fine-tuning and inference.

\paragraph{Implications of Introducing an External Signal.}
The successful integration of an external signal, such as a domain-independent OIE system, into the ETD task suggests that similar improvements can be achieved for other span detection tasks. Multi-task learning architectures (cf.~Section \ref{sec:external_signal_model}) and transfer training regimes (cf.~Section \ref{sec:external_signal_exp}), particularly those involving the proposed \textit{implicit} model architecture and the \textit{sequential} transfer learning paradigms, can serve as effective mediums for transfer. This approach is potent when a strong auxiliary task---either target domain-specific or domain-universal---can provide a relevant signal. Finding other auxiliary tasks, similar to relation extraction, could further drive the progress of transfer learning for sequence labeling tasks, diminishing annotation costs. Additionally, reducing the pre-train--fine-tune discrepancy through domain adaptation methods has proven effective (cf.~Subsection \ref{subsec:auxiliary_mlm}), even in scenarios where a small amount of annotated data from the target domain is available.

\paragraph{Implications of Architectural Modifications.}
Simple architectural interventions, such as removing the CM from a subset of decoder blocks during fine-tuning (cf.~Section \ref{sec:cm_removal_method}), are crucial for advancing the field in compute-constrained scenarios. We demonstrated that CM removal from decoder block groups is highly task- and model-dependent (cf.~Section \ref{sec:cm_removal_results}). This suggests the existence of a more fine-grained unmasking configuration per model and dataset that achieves optimal performance. The proposed \textit{layer group unmasking} method can also be interpreted as a regularization technique, where particular layers permit the model to leverage the right-side context, while others aggravate the challenge by restricting access, encouraging the model to solve tasks even with the imposed restrictions. Exploring additional, more targeted regularization methods could further enhance performance during the adaptation phase of sequence labeling tasks. Finally, since most researchers cannot compete in pre-training language models from scratch due to the immense computational resources and data required, architectural interventions during the adaptation phase remain a promising and accessible research direction. By focusing on the nuances of the target task and delving into the model's architecture, researchers can achieve optimal transfer to target sequence labeling tasks and maximize performance gains.

\paragraph{Implications of Modifying the Optimization Scheme.}
Teaching the model to leverage in-context examples through the modification of CLM optimization showed promising results. While ICL can bring substantial gains to sequence labeling tasks, these improvements often plateau due to decoder models' challenges in utilizing long contexts. By teaching models to adhere to specific responses during adaptation and inference (cf.~Section \ref{sec:framework}), we can unlock their potential to handle longer contexts more effectively. Our proposed SIFT framework modifies the autoregressive loss function to emphasize response tokens during SFT, demonstrating that LLMs can leverage longer contexts during the adaptation phase of transfer learning when the optimization scheme is carefully adjusted. This, in turn, improves the ICL capabilities of the model. Further improvements could be realized through more advanced yet straightforward response-oriented optimization schemes. For instance, it may be advantageous to adjust the model's weights using vanilla CLM during the early stages of training, transition to the restricted MRC CLM strategy in the middle stages, and ultimately rely solely on SRC CLM in the final phase of training.

\paragraph{Summary of Implications.}
As the NLP community continues to scale up the pre-training of language models, the focus has shifted toward improving the adaptation phase of transfer learning. Investigating the adaptation phase is arguably less resource-intensive, more practical, and offers higher controllability than pre-training language models and leveraging them for inference purposes. Therefore, advances in adaptation methods of transfer learning play a pivotal role. Through a series of experiments, we have demonstrated that significant improvements can be achieved by focusing on the adaptation phase of transfer learning. Our work provides evidence that each of the three proposed approaches---leveraging external OIE signals, modifying model architectures with layer-wise CM removal, and refining the optimization scheme through supervised in-context fine-tuning---can efficiently and effectively enhance the performance of PLMs on target sequence labeling tasks.

\section{Limitations}

We outline the limitations of our experiments and results by categorizing them into three parts, each corresponding to the specific set of experiments they are associated with.

\paragraph{Limitations of the Experimental Setup for Multi-task Model Leveraging External Signal.}
The available computing resources limited our experiments with the ETD task (cf.~Section \ref{sec:external_signal_exp} and Section \ref{sec:external_signal_results}). For reliability, in our experiments, we report performance scores averaged over three runs (differing in random seeds). Similarly, we sampled the few-shot examples five times. Averaging over larger samples would make the results even more reliable. Furthermore, the results of few-shot experiments can sometimes turn out to be misleading due to the high variance of the sample of examples. Fixing the learning rate and some other hyperparameters across experiments may have resulted in suboptimal adaptation to the event trigger detection task in both source and target domains. Moreover, all experiments were done only with RoBERTa-base; using a different suitable PLM might yield further insights. Finally, our experiments were limited to datasets in the English language; further insights may be gained by extending to cross-lingual ETD domain transfer, more transfer directions, and datasets.

\paragraph{Limitations of the Layer Group Unmasking Method.}
In our experiments for the \textit{layer group unmasking} method (cf.~Section \ref{sec:cm_removal_method} and Section \ref{sec:cm_removal_results}), we ensure reliability by averaging performance scores over five runs with different random seeds. Increasing the sample size for averaging would enhance reliability further. However, maintaining fixed learning rates and other hyperparameters across experiments might have led to suboptimal adaptation for sequence labeling tasks. Exploring additional unmasking configurations could provide valuable insights and potential improvements. While numerous open-weight LLMs are available, we focus solely on Llama2 and Mistral, both with 7B parameters. Enhancing prompt templates used for instruction tuning could boost overall instruction tuning performance. Given more computing resources, experimenting with larger models would be feasible. Additionally, our experiments were limited to English-language datasets. Extending the analysis to sequence labeling tasks in other languages and incorporating more diverse datasets could yield further insights. Finally, the main limitation of our proposed method is the fact that one needs to try out all the possible unmasking configurations to find the best one on the validation set. Trying out all the possible configurations requires extensive fine-tuning. However, this can perhaps be sidestepped using an appropriate heuristic, by leveraging representational similarity analysis or transferability estimation techniques, which is much more compute-efficient than fine-tuning the model with every possible unmasking configuration.

\paragraph{Limitations of the Experimental Setup for Supervised In-context Fine-tuning Framework.}
Finally, our experiments with SIFT and ICL were highly resource-intensive (cf.~Section \ref{sec:sift_results}), and experiments needed to be carefully engineered to adhere to the imposed limitations of constrained computing. For reliability, in our experiments, we report performance scores averaged over four runs (differing in random seeds). We average all ICL experiments only these four times; averaging over more random samples of demonstrations in the context would yield a better estimate of the actual performance and make the results even more reliable. Constrained by the computing resources, we were forced to subsample original subsets for the SRL task. Fixing the learning rate and some other hyperparameters across experiments may have resulted in suboptimal adaptation to sequence labeling tasks of NER, ABAM, slot labeling, and SRL. Moreover, we experiment only with smaller open-weight PLMs with seven and eight billion parameters. Leveraging larger PLMs might yield further insights. Finally, our experiments were limited to datasets in the English language. Additional insights may be gained by extending to cross-lingual transfer of sequence labeling tasks.
\clearpage{}%
\clearpage{}%
\chapter{Conclusion}
\label{ch:conclusion}

Adapting pre-trained language models (PLMs) to sequence labeling tasks is a promising and impactful research direction. Improvements in sequence labeling tasks can initiate a cascade of positive effects, enhancing downstream applications that depend on the outputs of sequence labeling models. Modern encoder- and decoder-only architectures, each with unique pre-training mechanisms, introduce specific challenges requiring careful investigation, individualized approaches, and tailored solutions. In this thesis, we addressed three key limitations of PLMs in the context of sequence labeling tasks, proposing targeted modifications and interventions during the adaptation phase of transfer learning to overcome observed difficulties.

First, we observed that encoder-only PLMs face challenges with domain transfer, particularly for the event trigger detection (ETD) task---a critical component of the information extraction (IE) pipeline. ETD is the foundation for subsequent tasks in the event extraction pipeline, making its robustness crucial for prolonging positive effects down the pipeline. To mitigate negative domain transfer, we introduced a novel multi-task model architecture that leverages outputs from a domain-independent open information extraction (OIE) system. Our approach showcased improved transfer performance in zero-shot and few-shot scenarios by capitalizing on the similarity between triggers and OIE relations and coupling relations with triggers through the multi-task model architecture and transfer training regimes.

Second, we identified a fundamental limitation of decoder-only autoregressive large language models (LLMs) in sequence labeling tasks: the inability to utilize right-side context due to the causal masking mechanism. To enable these models to function as effective text encoders, we proposed a method to remove the causal mask (CM) in a subset of decoder layers, allowing layer-wise bidirectional information flow. This simple yet effective modification significantly improved performance for decoder-only LLMs over strong encoder baselines and instruction-tuned decoders, demonstrating the potential of directly adapting autoregressive models to sequence labeling tasks.

Third, we developed a novel framework for supervised in-context fine-tuning  (SIFT) of autoregressive decoder-only LLMs, explicitly designed for sequence labeling tasks. Our approach put emphasis on the limitations of causal language modeling (CLM) loss function in shaping the model's responses during training with multiple demonstrations in the training prompts. Hence, we proposed modifications of the CLM loss to effectively utilize varying numbers of demonstration examples by masking out everything except response tokens in the training prompt. We compared our framework with existing in-context learning (ICL) methods, supervised fine-tuning (SFT) methods, and decoders with CM removed, highlighting differences, analyzing the effect of instruction on performance, and providing practical recommendations on when and how to use each approach.

The proposed model, method, and framework were developed on the solid foundations of PLM adaptation techniques. Enhancing transfer learning for sequence labeling tasks with modern PLMs requires innovative, effective, and efficient approaches attuned to the model, the task at task, and the available computing resources. Given that pre-training LLMs from scratch remains inaccessible to many researchers, we believe that advancements in the adaptation phase of transfer learning will continue to be a focal point for NLP researchers in the foreseeable future.

For future work, many promising avenues remain open for exploration. Regarding the introduction of an external signal into a multi-task model, we recommend extending the approach to other span detection tasks involving predicate-argument structures, as they could similarly benefit from the domain-agnostic OIE system. Additionally, leveraging subject and object information from rule-based OIE extractors could enhance tasks such as named entity recognition (NER) or other sequence labeling tasks where subjects and objects play a central role in conveying meaning. Finally, many other span detection tasks could leverage their unique auxiliary objectives for improved transfer learning.

For the transformation of decoders into better encoders, future efforts could focus on improving the efficiency of the \textit{layer group unmasking} method. Identifying the optimal unmasking configuration is resource-intensive, as it requires extensive fine-tuning for evaluation. To address this, we suggest exploring transferability estimation methods, such as \textit{LogME} \citep{you2021logme} or \textit{H-score} \citep{8803726}, which could serve as proxy metrics to predict the best unmasking configuration without the need for exhaustive fine-tuning. To aid in better unmasking performance prediction, we can also reach for methods relying on the Fisher information, similar to the \textit{FIT} method proposed in \citep{zandonati2023fit}. Additionally, performance estimation for CM removal from a layer could be improved using representation similarity analysis methods such as centered kernel alignment \citep{pmlr-v97-kornblith19a}. These approaches might reduce the number of unmasking configurations that need to be evaluated to find the best one and significantly reduce computational overhead, streamlining the adaptation process.

Finally, for improved SIFT and ICL results, we propose enhancing the demonstration sampling procedure within our SIFT framework. Currently, demonstrations for the context are sampled randomly, which may not always yield the best performance. We could maximize performance gains and further boost the framework's effectiveness by integrating advanced sampling methods that strategically select the most informative examples for the query from the available pool. This method could be incorporated during inference, as demonstrated in previous work \citep{rubin-etal-2022-learning,ye2023compositional,10707886}, and throughout the fine-tuning process. Furthermore, since decoders struggle with utilizing long contexts, we could enhance SIFT methods by integrating approaches for disentangling latent shifts, such as encoding demonstrations into a vector \citep{10.5555/3692070.3693379} or employing an adapter module \citep{jukic2024disentangling}. These techniques could mitigate or even eliminate the adverse effects caused by the inability of LLMs to process lengthy contexts effectively.\clearpage{}%
\backmatter

\addcontentsline{toc}{chapter}{Bibliography}
\bibliographystyle{IEEEtranFER} 
\bibliography{bibliography, anthology}

\begin{thebibliography}{100}
\providecommand{\url}[1]{#1}
\csname url@samestyle\endcsname
\providecommand{\newblock}{\relax}
\providecommand{\bibinfo}[2]{#2}
\providecommand{\BIBentrySTDinterwordspacing}{\spaceskip=0pt\relax}
\providecommand{\BIBentryALTinterwordstretchfactor}{4}
\providecommand{\BIBentryALTinterwordspacing}{\spaceskip=\fontdimen2\font plus
\BIBentryALTinterwordstretchfactor\fontdimen3\font minus
  \fontdimen4\font\relax}
\providecommand{\BIBforeignlanguage}[2]{{%
\expandafter\ifx\csname l@#1\endcsname\relax
\typeout{** WARNING: IEEEtran.bst: No hyphenation pattern has been}%
\typeout{** loaded for the language `#1'. Using the pattern for}%
\typeout{** the default language instead.}%
\else
\language=\csname l@#1\endcsname
\fi
#2}}
\providecommand{\BIBdecl}{\relax}
\BIBdecl

\bibitem{tjong-kim-sang-de-meulder-2003-introduction}
\BIBentryALTinterwordspacing
Tjong Kim~Sang, E.~F.,  De~Meulder, F., ``Introduction to the {C}o{NLL}-2003
  shared task: Language-independent named entity recognition'', in Proceedings
  of the Seventh Conference on Natural Language Learning at {HLT}-{NAACL} 2003,
  2003, pages 142--147,  \url{https://aclanthology.org/W03-0419/}
\BIBentrySTDinterwordspacing

\bibitem{tjong-kim-sang-buchholz-2000-introduction}
\BIBentryALTinterwordspacing
Tjong Kim~Sang, E.~F.,  Buchholz, S., ``Introduction to the {C}o{NLL}-2000
  shared task chunking'', in Fourth Conference on Computational Natural
  Language Learning and the Second Learning Language in Logic Workshop, 2000,
  \url{https://aclanthology.org/W00-0726/}
\BIBentrySTDinterwordspacing

\bibitem{10.1162/089120102760275983}
\BIBentryALTinterwordspacing
Gildea, D.,  Jurafsky, D., ``Automatic labeling of semantic roles'',
  Computational Linguistics, Vol.~28, No.~3, 09 2002, pages 245-288,
  \url{https://doi.org/10.1162/089120102760275983}
\BIBentrySTDinterwordspacing

\bibitem{hoffart-etal-2011-robust}
\BIBentryALTinterwordspacing
Hoffart, J., Yosef, M.~A., Bordino, I., F{\"u}rstenau, H., Pinkal, M., Spaniol,
  M., Taneva, B., Thater, S.,  Weikum, G., ``Robust disambiguation of named
  entities in text'', in Proceedings of the 2011 Conference on Empirical
  Methods in Natural Language Processing, Barzilay, R.,  Johnson, M., (ur.).
  Edinburgh, Scotland, UK.: Association for Computational Linguistics, Jul.
  2011, pages 782--792,  \url{https://aclanthology.org/D11-1072/}
\BIBentrySTDinterwordspacing

\bibitem{dukic2024closed}
Duki{\'c}, D., Do{\v{s}}ilovi{\'c}, F.~K., Plu{\v{s}}{\v{c}}ec, D.,
  {\v{S}}najder, J., ``Closed-domain event extraction for hard news event
  monitoring: {A} systematic study'', PeerJ Computer Science, Vol.~10, 2024,
  page e2355.

\bibitem{dukic2024takelab}
Duki{\'c}, D., Petri{\v{c}}evi{\'c}, M., {\'C}urkovi{\'c}, S.,  {\v{S}}najder,
  J., ``Takelab {Retriever}: {AI}-driven search engine for articles from
  croatian news outlets'', arXiv preprint arXiv:2411.19718, 2024.

\bibitem{vaswani2017attention}
Vaswani, A., Shazeer, N., Parmar, N., Uszkoreit, J., Jones, L., Gomez, A.~N.,
  Kaiser, {\L}.,  Polosukhin, I., ``Attention is all you need'', in Advances in
  neural information processing systems, Vol.~30, 2017.

\bibitem{radford2018improving}
Radford, A., Narasimhan, K., Salimans, T.,  Sutskever, I., ``Improving language
  understanding by generative pre-training'', 2018.

\bibitem{devlin-etal-2019-bert}
\BIBentryALTinterwordspacing
Devlin, J., Chang, M.-W., Lee, K.,  Toutanova, K., ``{BERT}: Pre-training of
  deep bidirectional transformers for language understanding'', in Proceedings
  of the 2019 Conference of the North {A}merican Chapter of the Association for
  Computational Linguistics: Human Language Technologies, Volume 1 (Long and
  Short Papers), Burstein, J., Doran, C.,  Solorio, T., (ur.). Minneapolis,
  Minnesota: Association for Computational Linguistics, Jun. 2019, pages
  4171--4186,  \url{https://aclanthology.org/N19-1423/}
\BIBentrySTDinterwordspacing

\bibitem{crawshaw2020multi}
Crawshaw, M., ``Multi-task learning with deep neural networks: A survey'',
  arXiv preprint arXiv:2009.09796, 2020.

\bibitem{houlsby2019parameter}
Houlsby, N., Giurgiu, A., Jastrzebski, S., Morrone, B., De~Laroussilhe, Q.,
  Gesmundo, A., Attariyan, M.,  Gelly, S., ``Parameter-efficient transfer
  learning for {NLP}'', in International Conference on Machine Learning. PMLR,
  2019, pages 2790--2799.

\bibitem{distrib}
Harris, Z.~S., ``Distributional structure'', WORD, Vol.~10, 1954, pages
  146--162.

\bibitem{bengio2000neural}
Bengio, Y., Ducharme, R.,  Vincent, P., ``A neural probabilistic language
  model'', in Advances in neural information processing systems, Vol.~13, 2000.

\bibitem{mikolov2013efficient}
Mikolov, T., ``Efficient estimation of word representations in vector space'',
  in International Conference on Learning Representations, 2013.

\bibitem{allen2019analogies}
Allen, C.,  Hospedales, T., ``Analogies explained: Towards understanding word
  embeddings'', in International Conference on Machine Learning. PMLR, 2019,
  pages 223--231.

\bibitem{drozd-etal-2016-word}
\BIBentryALTinterwordspacing
Drozd, A., Gladkova, A.,  Matsuoka, S., ``Word embeddings, analogies, and
  machine learning: Beyond king - man + woman = queen'', in Proceedings of
  {COLING} 2016, the 26th International Conference on Computational
  Linguistics: Technical Papers, Matsumoto, Y.,  Prasad, R., (ur.). Osaka,
  Japan: The COLING 2016 Organizing Committee, Dec. 2016, pages 3519--3530,
  \url{https://aclanthology.org/C16-1332/}
\BIBentrySTDinterwordspacing

\bibitem{ELMAN1990179}
\BIBentryALTinterwordspacing
Elman, J.~L., ``Finding structure in time'', Cognitive Science, Vol.~14, No.~2,
  1990, pages 179-211,
  \url{https://www.sciencedirect.com/science/article/pii/036402139090002E}
\BIBentrySTDinterwordspacing

\bibitem{cho-etal-2014-learning}
\BIBentryALTinterwordspacing
Cho, K., van Merri{\"e}nboer, B., Gulcehre, C., Bahdanau, D., Bougares, F.,
  Schwenk, H.,  Bengio, Y., ``Learning phrase representations using {RNN}
  encoder{--}decoder for statistical machine translation'', in Proceedings of
  the 2014 Conference on Empirical Methods in Natural Language Processing
  ({EMNLP}), Moschitti, A., Pang, B.,  Daelemans, W., (ur.). Doha, Qatar:
  Association for Computational Linguistics, Oct. 2014, pages 1724--1734,
  \url{https://aclanthology.org/D14-1179/}
\BIBentrySTDinterwordspacing

\bibitem{10.1162/neco.1997.9.8.1735}
\BIBentryALTinterwordspacing
Hochreiter, S.,  Schmidhuber, J., ``Long short-term memory'', Neural
  Computation, Vol.~9, No.~8, 11 1997, pages 1735-1780,
  \url{https://doi.org/10.1162/neco.1997.9.8.1735}
\BIBentrySTDinterwordspacing

\bibitem{bahdanau2014neural}
Bahdanau, D., ``Neural machine translation by jointly learning to align and
  translate'', in International Conference on Learning Representations, 2015.

\bibitem{raffel2020exploring}
Raffel, C., Shazeer, N., Roberts, A., Lee, K., Narang, S., Matena, M., Zhou,
  Y., Li, W.,  Liu, P.~J., ``Exploring the limits of transfer learning with a
  unified text-to-text transformer'', The Journal of Machine Learning Research,
  Vol.~21, No.~1, 2020, pages 5485--5551.

\bibitem{clark2020electra}
\BIBentryALTinterwordspacing
Clark, K., Luong, M.-T., Le, Q.~V.,  Manning, C.~D., ``{ELECTRA}: Pre-training
  text encoders as discriminators rather than generators'', in International
  Conference on Learning Representations, 2020,
  \url{https://openreview.net/forum?id=r1xMH1BtvB}
\BIBentrySTDinterwordspacing

\bibitem{yang2019xlnet}
Yang, Z., Dai, Z., Yang, Y., Carbonell, J., Salakhutdinov, R.~R.,  Le, Q.~V.,
  ``{XLNet}: Generalized autoregressive pretraining for language
  understanding'', in Advances in neural information processing systems,
  Vol.~32, 2019.

\bibitem{liu2019roberta}
Liu, Y., Ott, M., Goyal, N., Du, J., Joshi, M., Chen, D., Levy, O., Lewis, M.,
  Zettlemoyer, L.,  Stoyanov, V., ``{RoBERTa}: A robustly optimized {BERT}
  pretraining approach'', arXiv preprint arXiv:1907.11692, 2019.

\bibitem{meng2023representation}
\BIBentryALTinterwordspacing
Meng, Y., Krishnan, J., Wang, S., Wang, Q., Mao, Y., Fang, H., Ghazvininejad,
  M., Han, J.,  Zettlemoyer, L., ``Representation deficiency in masked language
  modeling'', in The Twelfth International Conference on Learning
  Representations, 2024,  \url{https://openreview.net/forum?id=b3l0piOrGU}
\BIBentrySTDinterwordspacing

\bibitem{khanehzar-etal-2019-modeling}
\BIBentryALTinterwordspacing
Khanehzar, S., Turpin, A.,  Mikolajczak, G., ``Modeling political framing
  across policy issues and contexts'', in Proceedings of the 17th Annual
  Workshop of the Australasian Language Technology Association, Mistica, M.,
  Piccardi, M.,  MacKinlay, A., (ur.). Sydney, Australia: Australasian Language
  Technology Association, 4--6 Dec. 2019, pages 61--66,
  \url{https://aclanthology.org/U19-1009/}
\BIBentrySTDinterwordspacing

\bibitem{shoeybi2019megatron}
Shoeybi, M., Patwary, M., Puri, R., LeGresley, P., Casper, J.,  Catanzaro, B.,
  ``{Megatron-LM}: {Training} multi-billion parameter language models using
  model parallelism'', arXiv preprint arXiv:1909.08053, 2019.

\bibitem{radford2019language}
Radford, A., Wu, J., Child, R., Luan, D., Amodei, D., Sutskever, I.
  \emph{et~al.}, ``Language models are unsupervised multitask learners'',
  OpenAI blog, Vol.~1, No.~8, 2019, page~9.

\bibitem{wei2021finetuned}
\BIBentryALTinterwordspacing
Wei, J., Bosma, M., Zhao, V., Guu, K., Yu, A.~W., Lester, B., Du, N., Dai,
  A.~M.,  Le, Q.~V., ``Finetuned language models are zero-shot learners'', in
  International Conference on Learning Representations, 2022,
  \url{https://openreview.net/forum?id=gEZrGCozdqR}
\BIBentrySTDinterwordspacing

\bibitem{brown2020language}
Brown, T., Mann, B., Ryder, N., Subbiah, M., Kaplan, J.~D., Dhariwal, P.,
  Neelakantan, A., Shyam, P., Sastry, G., Askell, A. \emph{et~al.}, ``Language
  models are few-shot learners'', in Advances in neural information processing
  systems, Vol.~33, 2020, pages 1877--1901.

\bibitem{liu2023pre}
Liu, P., Yuan, W., Fu, J., Jiang, Z., Hayashi, H.,  Neubig, G., ``Pre-train,
  prompt, and predict: A systematic survey of prompting methods in natural
  language processing'', ACM Computing Surveys, Vol.~55, No.~9, 2023, pages
  1--35.

\bibitem{NEURIPS2022_9d560961}
\BIBentryALTinterwordspacing
Wei, J., Wang, X., Schuurmans, D., Bosma, M., ichter, b., Xia, F., Chi, E., Le,
  Q.~V.,  Zhou, D., ``Chain-of-thought prompting elicits reasoning in large
  language models'', in Advances in Neural Information Processing Systems,
  Koyejo, S., Mohamed, S., Agarwal, A., Belgrave, D., Cho, K.,  Oh, A., (ur.),
  Vol.~35. Curran Associates, Inc., 2022, pages 24\,824--24\,837,
  \url{https://proceedings.neurips.cc/paper_files/paper/2022/file/9d5609613524ecf4f15af0f7b31abca4-Paper-Conference.pdf}
\BIBentrySTDinterwordspacing

\bibitem{chowdhery2023palm}
Chowdhery, A., Narang, S., Devlin, J., Bosma, M., Mishra, G., Roberts, A.,
  Barham, P., Chung, H.~W., Sutton, C., Gehrmann, S., Schuh, P., Shi, K.,
  Tsvyashchenko, S., Maynez, J., Rao, A., Barnes, P., Tay, Y., Shazeer, N.,
  Prabhakaran, V., Reif, E., Du, N., Hutchinson, B., Pope, R., Bradbury, J.,
  Austin, J., Isard, M., Gur-Ari, G., Yin, P., Duke, T., Levskaya, A.,
  Ghemawat, S., Dev, S., Michalewski, H., Garcia, X., Misra, V., Robinson, K.,
  Fedus, L., Zhou, D., Ippolito, D., Luan, D., Lim, H., Zoph, B., Spiridonov,
  A., Sepassi, R., Dohan, D., Agrawal, S., Omernick, M., Dai, A.~M., Pillai,
  T.~S., Pellat, M., Lewkowycz, A., Moreira, E., Child, R., Polozov, O., Lee,
  K., Zhou, Z., Wang, X., Saeta, B., Diaz, M., Firat, O., Catasta, M., Wei, J.,
  Meier-Hellstern, K., Eck, D., Dean, J., Petrov, S.,  Fiedel, N., ``{PaLM}:
  scaling language modeling with pathways'', J. Mach. Learn. Res., Vol.~24,
  No.~1, Jan. 2023.

\bibitem{mishra-etal-2022-cross}
\BIBentryALTinterwordspacing
Mishra, S., Khashabi, D., Baral, C.,  Hajishirzi, H., ``Cross-task
  generalization via natural language crowdsourcing instructions'', in
  Proceedings of the 60th Annual Meeting of the Association for Computational
  Linguistics (Volume 1: Long Papers), Muresan, S., Nakov, P.,  Villavicencio,
  A., (ur.). Dublin, Ireland: Association for Computational Linguistics, May
  2022, pages 3470--3487,  \url{https://aclanthology.org/2022.acl-long.244/}
\BIBentrySTDinterwordspacing

\bibitem{scaling_lms_src}
\BIBentryALTinterwordspacing
He, K., Mao, R., Lin, Q., Ruan, Y., Lan, X., Feng, M.,  Cambria, E., ``A survey
  of large language models for healthcare: from data, technology, and
  applications to accountability and ethics'', Information Fusion, Vol. 118,
  2025, page 102963,
  \url{https://www.sciencedirect.com/science/article/pii/S1566253525000363}
\BIBentrySTDinterwordspacing

\bibitem{ruder-etal-2019-transfer}
\BIBentryALTinterwordspacing
Ruder, S., Peters, M.~E., Swayamdipta, S.,  Wolf, T., ``Transfer learning in
  natural language processing'', in Proceedings of the 2019 Conference of the
  North {A}merican Chapter of the Association for Computational Linguistics:
  Tutorials, Sarkar, A.,  Strube, M., (ur.). Minneapolis, Minnesota:
  Association for Computational Linguistics, Jun. 2019, pages 15--18,
  \url{https://aclanthology.org/N19-5004/}
\BIBentrySTDinterwordspacing

\bibitem{ruder2019neural}
Ruder, S., ``Neural transfer learning for natural language processing'',
  Doctoral thesis, NUI Galway, 2019.

\bibitem{gururangan-etal-2020-dont}
\BIBentryALTinterwordspacing
Gururangan, S., Marasovi{\'c}, A., Swayamdipta, S., Lo, K., Beltagy, I.,
  Downey, D.,  Smith, N.~A., ``Don`t stop pretraining: Adapt language models to
  domains and tasks'', in Proceedings of the 58th Annual Meeting of the
  Association for Computational Linguistics, Jurafsky, D., Chai, J., Schluter,
  N.,  Tetreault, J., (ur.). Online: Association for Computational Linguistics,
  Jul. 2020, pages 8342--8360,
  \url{https://aclanthology.org/2020.acl-main.740/}
\BIBentrySTDinterwordspacing

\bibitem{ruder2019survey}
Ruder, S., Vuli{\'c}, I.,  S{\o}gaard, A., ``A survey of cross-lingual word
  embedding models'', Journal of Artificial Intelligence Research, Vol.~65,
  2019, pages 569--631.

\bibitem{zhang-etal-2023-survey}
\BIBentryALTinterwordspacing
Zhang, Z., Yu, W., Yu, M., Guo, Z.,  Jiang, M., ``A survey of multi-task
  learning in natural language processing: Regarding task relatedness and
  training methods'', in Proceedings of the 17th Conference of the European
  Chapter of the Association for Computational Linguistics, Vlachos, A.,
  Augenstein, I., (ur.). Dubrovnik, Croatia: Association for Computational
  Linguistics, May 2023, pages 943--956,
  \url{https://aclanthology.org/2023.eacl-main.66/}
\BIBentrySTDinterwordspacing

\bibitem{fu2024data}
Fu, Y., Panda, R., Niu, X., Yue, X., Hajishirzi, H., Kim, Y.,  Peng, H., ``Data
  engineering for scaling language models to 128k context'', in Proceedings of
  the 41st International Conference on Machine Learning, ser. ICML'24.
  JMLR.org, 2024.

\bibitem{wang2019characterizing}
Wang, Z., Dai, Z., P{\'o}czos, B.,  Carbonell, J., ``Characterizing and
  avoiding negative transfer'', in Proceedings of the IEEE/CVF conference on
  computer vision and pattern recognition, 2019, pages 11\,293--11\,302.

\bibitem{ngo-trung-etal-2021-unsupervised}
\BIBentryALTinterwordspacing
Ngo~Trung, N., Phung, D.,  Nguyen, T.~H., ``Unsupervised domain adaptation for
  event detection using domain-specific adapters'', in Findings of the
  Association for Computational Linguistics: ACL-IJCNLP 2021, Zong, C., Xia,
  F., Li, W.,  Navigli, R., (ur.). Online: Association for Computational
  Linguistics, Aug. 2021, pages 4015--4025,
  \url{https://aclanthology.org/2021.findings-acl.351/}
\BIBentrySTDinterwordspacing

\bibitem{meftah-etal-2021-hidden}
\BIBentryALTinterwordspacing
Meftah, S., Semmar, N., Tamaazousti, Y., Essafi, H.,  Sadat, F., ``On the
  hidden negative transfer in sequential transfer learning for domain
  adaptation from news to tweets'', in Proceedings of the Second Workshop on
  Domain Adaptation for NLP, Ben-David, E., Cohen, S., McDonald, R., Plank, B.,
  Reichart, R., Rotman, G.,  Ziser, Y., (ur.). Kyiv, Ukraine: Association for
  Computational Linguistics, Apr. 2021, pages 140--145,
  \url{https://aclanthology.org/2021.adaptnlp-1.14/}
\BIBentrySTDinterwordspacing

\bibitem{9938381}
Zhang, W., Deng, L., Zhang, L.,  Wu, D., ``A survey on negative transfer'',
  IEEE/CAA Journal of Automatica Sinica, Vol.~10, No.~2, 2023, pages 305-329.

\bibitem{dukic-etal-2024-leveraging}
\BIBentryALTinterwordspacing
Duki{\'c}, D., Gashteovski, K., Glava{\v{s}}, G.,  Snajder, J., ``Leveraging
  open information extraction for more robust domain transfer of event trigger
  detection'', in Findings of the Association for Computational Linguistics:
  EACL 2024, Graham, Y.,  Purver, M., (ur.). St. Julian{'}s, Malta: Association
  for Computational Linguistics, Mar. 2024, pages 1197--1213,
  \url{https://aclanthology.org/2024.findings-eacl.80/}
\BIBentrySTDinterwordspacing

\bibitem{hung-etal-2022-ds}
\BIBentryALTinterwordspacing
Hung, C.-C., Lauscher, A., Ponzetto, S.,  Glava{\v{s}}, G., ``{DS}-{TOD}:
  Efficient domain specialization for task-oriented dialog'', in Findings of
  the Association for Computational Linguistics: ACL 2022, Muresan, S., Nakov,
  P.,  Villavicencio, A., (ur.). Dublin, Ireland: Association for Computational
  Linguistics, May 2022, pages 891--904,
  \url{https://aclanthology.org/2022.findings-acl.72/}
\BIBentrySTDinterwordspacing

\bibitem{baric-etal-2023-target}
\BIBentryALTinterwordspacing
Bari{\'c}, A., Majer, L., Duki{\'c}, D., Grbe{\v{s}}a-zenzerovi{\'c}, M.,
  Snajder, J., ``Target two birds with one {ST}o{N}e: Entity-level sentiment
  and tone analysis in {C}roatian news headlines'', in Proceedings of the 9th
  Workshop on Slavic Natural Language Processing 2023 (SlavicNLP 2023),
  Piskorski, J., Marci{\'n}czuk, M., Nakov, P., Ogrodniczuk, M., Pollak, S.,
  P{\v{r}}ib{\'a}{\v{n}}, P., Rybak, P., Steinberger, J.,  Yangarber, R.,
  (ur.). Dubrovnik, Croatia: Association for Computational Linguistics, May
  2023, pages 78--85,  \url{https://aclanthology.org/2023.bsnlp-1.10/}
\BIBentrySTDinterwordspacing

\bibitem{wang-etal-2018-glue}
\BIBentryALTinterwordspacing
Wang, A., Singh, A., Michael, J., Hill, F., Levy, O.,  Bowman, S., ``{GLUE}: A
  multi-task benchmark and analysis platform for natural language
  understanding'', in Proceedings of the 2018 {EMNLP} Workshop {B}lackbox{NLP}:
  Analyzing and Interpreting Neural Networks for {NLP}, Linzen, T.,
  Chrupa{\l}a, G.,  Alishahi, A., (ur.). Brussels, Belgium: Association for
  Computational Linguistics, Nov. 2018, pages 353--355,
  \url{https://aclanthology.org/W18-5446/}
\BIBentrySTDinterwordspacing

\bibitem{jm3}
\BIBentryALTinterwordspacing
Jurafsky, D.,  Martin, J.~H., Speech and Language Processing: An Introduction
  to Natural Language Processing, Computational Linguistics, and Speech
  Recognition with Language Models, 3rd~ed. Draft, 2025, online manuscript
  released January 12, 2025,  \url{https://web.stanford.edu/~jurafsky/slp3/}
\BIBentrySTDinterwordspacing

\bibitem{clark-etal-2018-semi}
\BIBentryALTinterwordspacing
Clark, K., Luong, M.-T., Manning, C.~D.,  Le, Q., ``Semi-supervised sequence
  modeling with cross-view training'', in Proceedings of the 2018 Conference on
  Empirical Methods in Natural Language Processing, Riloff, E., Chiang, D.,
  Hockenmaier, J.,  Tsujii, J., (ur.). Brussels, Belgium: Association for
  Computational Linguistics, Oct.-Nov. 2018, pages 1914--1925,
  \url{https://aclanthology.org/D18-1217/}
\BIBentrySTDinterwordspacing

\bibitem{xie2020unsupervised}
Xie, Q., Dai, Z., Hovy, E., Luong, T.,  Le, Q., ``Unsupervised data
  augmentation for consistency training'', in Advances in neural information
  processing systems, Vol.~33, 2020, pages 6256--6268.

\bibitem{repetal2024electras}
\BIBentryALTinterwordspacing
Rep, I., Duki{\'c}, D.,  {\v{S}}najder, J., ``Are {ELECTRA}{'}s sentence
  embeddings beyond repair? {The} case of semantic textual similarity'', in
  Findings of the Association for Computational Linguistics: EMNLP 2024,
  Al-Onaizan, Y., Bansal, M.,  Chen, Y.-N., (ur.). Miami, Florida, USA:
  Association for Computational Linguistics, Nov. 2024, pages 9159--9169,
  \url{https://aclanthology.org/2024.findings-emnlp.535/}
\BIBentrySTDinterwordspacing

\bibitem{schmidt-etal-2022-dont}
\BIBentryALTinterwordspacing
Schmidt, F.~D., Vuli{\'c}, I.,  Glava{\v{s}}, G., ``Don`t stop fine-tuning: On
  training regimes for few-shot cross-lingual transfer with multilingual
  language models'', in Proceedings of the 2022 Conference on Empirical Methods
  in Natural Language Processing, Goldberg, Y., Kozareva, Z.,  Zhang, Y.,
  (ur.). Abu Dhabi, United Arab Emirates: Association for Computational
  Linguistics, Dec. 2022, pages 10\,725--10\,742,
  \url{https://aclanthology.org/2022.emnlp-main.736/}
\BIBentrySTDinterwordspacing

\bibitem{dukic-snajder-2024-looking}
\BIBentryALTinterwordspacing
Duki{\'c}, D.,  Snajder, J., ``Looking right is sometimes right: Investigating
  the capabilities of decoder-only {LLM}s for sequence labeling'', in Findings
  of the Association for Computational Linguistics: ACL 2024, Ku, L.-W.,
  Martins, A.,  Srikumar, V., (ur.). Bangkok, Thailand: Association for
  Computational Linguistics, Aug. 2024, pages 14\,168--14\,181,
  \url{https://aclanthology.org/2024.findings-acl.843/}
\BIBentrySTDinterwordspacing

\bibitem{ramachandran-etal-2017-unsupervised}
\BIBentryALTinterwordspacing
Ramachandran, P., Liu, P.,  Le, Q., ``Unsupervised pretraining for sequence to
  sequence learning'', in Proceedings of the 2017 Conference on Empirical
  Methods in Natural Language Processing, Palmer, M., Hwa, R.,  Riedel, S.,
  (ur.). Copenhagen, Denmark: Association for Computational Linguistics, Sep.
  2017, pages 383--391,  \url{https://aclanthology.org/D17-1039/}
\BIBentrySTDinterwordspacing

\bibitem{howard-ruder-2018-universal}
\BIBentryALTinterwordspacing
Howard, J.,  Ruder, S., ``Universal language model fine-tuning for text
  classification'', in Proceedings of the 56th Annual Meeting of the
  Association for Computational Linguistics (Volume 1: Long Papers), Gurevych,
  I.,  Miyao, Y., (ur.). Melbourne, Australia: Association for Computational
  Linguistics, Jul. 2018, pages 328--339,
  \url{https://aclanthology.org/P18-1031/}
\BIBentrySTDinterwordspacing

\bibitem{loshchilov2016sgdr}
Loshchilov, I.,  Hutter, F., ``{SGDR}: Stochastic gradient descent with warm
  restarts'', in International Conference on Learning Representations, 2017.

\bibitem{liu2019variance}
Liu, L., Jiang, H., He, P., Chen, W., Liu, X., Gao, J.,  Han, J., ``On the
  variance of the adaptive learning rate and beyond'', in International
  Conference on Learning Representations, 2020.

\bibitem{touvron2023llama}
Touvron, H., Martin, L., Stone, K., Albert, P., Almahairi, A., Babaei, Y.,
  Bashlykov, N., Batra, S., Bhargava, P., Bhosale, S., Bikel, D., Blecher, L.,
  Ferrer, C.~C., Chen, M., Cucurull, G., Esiobu, D., Fernandes, J., Fu, J., Fu,
  W., Fuller, B., Gao, C., Goswami, V., Goyal, N., Hartshorn, A., Hosseini, S.,
  Hou, R., Inan, H., Kardas, M., Kerkez, V., Khabsa, M., Kloumann, I., Korenev,
  A., Koura, P.~S., Lachaux, M.-A., Lavril, T., Lee, J., Liskovich, D., Lu, Y.,
  Mao, Y., Martinet, X., Mihaylov, T., Mishra, P., Molybog, I., Nie, Y.,
  Poulton, A., Reizenstein, J., Rungta, R., Saladi, K., Schelten, A., Silva,
  R., Smith, E.~M., Subramanian, R., Tan, X.~E., Tang, B., Taylor, R.,
  Williams, A., Kuan, J.~X., Xu, P., Yan, Z., Zarov, I., Zhang, Y., Fan, A.,
  Kambadur, M., Narang, S., Rodriguez, A., Stojnic, R., Edunov, S.,  Scialom,
  T., ``Llama 2: Open foundation and fine-tuned chat models'', arXiv preprint
  arXiv:2307.09288, 2023.

\bibitem{dubey2024llama}
Grattafiori, A., Dubey, A., Jauhri, A., Pandey, A., Kadian, A., Al-Dahle, A.,
  Letman, A., Mathur, A., Schelten, A., Vaughan, A., Yang, A., Fan, A., Goyal,
  A., Hartshorn, A., Yang, A., Mitra, A., Sravankumar, A., Korenev, A.,
  Hinsvark, A., Rao, A., Zhang, A., Rodriguez, A., Gregerson, A., Spataru, A.,
  Roziere, B., Biron, B., Tang, B., Chern, B., Caucheteux, C., Nayak, C., Bi,
  C., Marra, C., McConnell, C., Keller, C., Touret, C., Wu, C., Wong, C.,
  Ferrer, C.~C., Nikolaidis, C., Allonsius, D., Song, D., Pintz, D., Livshits,
  D., Wyatt, D., Esiobu, D., Choudhary, D., Mahajan, D., Garcia-Olano, D.,
  Perino, D., Hupkes, D., Lakomkin, E., AlBadawy, E., Lobanova, E., Dinan, E.,
  Smith, E.~M., Radenovic, F., Guzmán, F., Zhang, F., Synnaeve, G., Lee, G.,
  Anderson, G.~L., Thattai, G., Nail, G., Mialon, G., Pang, G., Cucurell, G.,
  Nguyen, H., Korevaar, H., Xu, H., Touvron, H., Zarov, I., Ibarra, I.~A.,
  Kloumann, I., Misra, I., Evtimov, I., Zhang, J., Copet, J., Lee, J., Geffert,
  J., Vranes, J., Park, J., Mahadeokar, J., Shah, J., van~der Linde, J.,
  Billock, J., Hong, J., Lee, J., Fu, J., Chi, J., Huang, J., Liu, J., Wang,
  J., Yu, J., Bitton, J., Spisak, J., Park, J., Rocca, J., Johnstun, J., Saxe,
  J., Jia, J., Alwala, K.~V., Prasad, K., Upasani, K., Plawiak, K., Li, K.,
  Heafield, K., Stone, K., El-Arini, K., Iyer, K., Malik, K., Chiu, K., Bhalla,
  K., Lakhotia, K., Rantala-Yeary, L., van~der Maaten, L., Chen, L., Tan, L.,
  Jenkins, L., Martin, L., Madaan, L., Malo, L., Blecher, L., Landzaat, L.,
  de~Oliveira, L., Muzzi, M., Pasupuleti, M., Singh, M., Paluri, M., Kardas,
  M., Tsimpoukelli, M., Oldham, M., Rita, M., Pavlova, M., Kambadur, M., Lewis,
  M., Si, M., Singh, M.~K., Hassan, M., Goyal, N., Torabi, N., Bashlykov, N.,
  Bogoychev, N., Chatterji, N., Zhang, N., Duchenne, O., Çelebi, O., Alrassy,
  P., Zhang, P., Li, P., Vasic, P., Weng, P., Bhargava, P., Dubal, P.,
  Krishnan, P., Koura, P.~S., Xu, P., He, Q., Dong, Q., Srinivasan, R.,
  Ganapathy, R., Calderer, R., Cabral, R.~S., Stojnic, R., Raileanu, R.,
  Maheswari, R., Girdhar, R., Patel, R., Sauvestre, R., Polidoro, R., Sumbaly,
  R., Taylor, R., Silva, R., Hou, R., Wang, R., Hosseini, S., Chennabasappa,
  S., Singh, S., Bell, S., Kim, S.~S., Edunov, S., Nie, S., Narang, S.,
  Raparthy, S., Shen, S., Wan, S., Bhosale, S., Zhang, S., Vandenhende, S.,
  Batra, S., Whitman, S., Sootla, S., Collot, S., Gururangan, S., Borodinsky,
  S., Herman, T., Fowler, T., Sheasha, T., Georgiou, T., Scialom, T.,
  Speckbacher, T., Mihaylov, T., Xiao, T., Karn, U., Goswami, V., Gupta, V.,
  Ramanathan, V., Kerkez, V., Gonguet, V., Do, V., Vogeti, V., Albiero, V.,
  Petrovic, V., Chu, W., Xiong, W., Fu, W., Meers, W., Martinet, X., Wang, X.,
  Wang, X., Tan, X.~E., Xia, X., Xie, X., Jia, X., Wang, X., Goldschlag, Y.,
  Gaur, Y., Babaei, Y., Wen, Y., Song, Y., Zhang, Y., Li, Y., Mao, Y., Coudert,
  Z.~D., Yan, Z., Chen, Z., Papakipos, Z., Singh, A., Srivastava, A., Jain, A.,
  Kelsey, A., Shajnfeld, A., Gangidi, A., Victoria, A., Goldstand, A., Menon,
  A., Sharma, A., Boesenberg, A., Baevski, A., Feinstein, A., Kallet, A.,
  Sangani, A., Teo, A., Yunus, A., Lupu, A., Alvarado, A., Caples, A., Gu, A.,
  Ho, A., Poulton, A., Ryan, A., Ramchandani, A., Dong, A., Franco, A., Goyal,
  A., Saraf, A., Chowdhury, A., Gabriel, A., Bharambe, A., Eisenman, A.,
  Yazdan, A., James, B., Maurer, B., Leonhardi, B., Huang, B., Loyd, B., Paola,
  B.~D., Paranjape, B., Liu, B., Wu, B., Ni, B., Hancock, B., Wasti, B.,
  Spence, B., Stojkovic, B., Gamido, B., Montalvo, B., Parker, C., Burton, C.,
  Mejia, C., Liu, C., Wang, C., Kim, C., Zhou, C., Hu, C., Chu, C.-H., Cai, C.,
  Tindal, C., Feichtenhofer, C., Gao, C., Civin, D., Beaty, D., Kreymer, D.,
  Li, D., Adkins, D., Xu, D., Testuggine, D., David, D., Parikh, D., Liskovich,
  D., Foss, D., Wang, D., Le, D., Holland, D., Dowling, E., Jamil, E.,
  Montgomery, E., Presani, E., Hahn, E., Wood, E., Le, E.-T., Brinkman, E.,
  Arcaute, E., Dunbar, E., Smothers, E., Sun, F., Kreuk, F., Tian, F.,
  Kokkinos, F., Ozgenel, F., Caggioni, F., Kanayet, F., Seide, F., Florez,
  G.~M., Schwarz, G., Badeer, G., Swee, G., Halpern, G., Herman, G., Sizov, G.,
  Guangyi, Zhang, Lakshminarayanan, G., Inan, H., Shojanazeri, H., Zou, H.,
  Wang, H., Zha, H., Habeeb, H., Rudolph, H., Suk, H., Aspegren, H., Goldman,
  H., Zhan, H., Damlaj, I., Molybog, I., Tufanov, I., Leontiadis, I., Veliche,
  I.-E., Gat, I., Weissman, J., Geboski, J., Kohli, J., Lam, J., Asher, J.,
  Gaya, J.-B., Marcus, J., Tang, J., Chan, J., Zhen, J., Reizenstein, J.,
  Teboul, J., Zhong, J., Jin, J., Yang, J., Cummings, J., Carvill, J., Shepard,
  J., McPhie, J., Torres, J., Ginsburg, J., Wang, J., Wu, K., U, K.~H., Saxena,
  K., Khandelwal, K., Zand, K., Matosich, K., Veeraraghavan, K., Michelena, K.,
  Li, K., Jagadeesh, K., Huang, K., Chawla, K., Huang, K., Chen, L., Garg, L.,
  A, L., Silva, L., Bell, L., Zhang, L., Guo, L., Yu, L., Moshkovich, L.,
  Wehrstedt, L., Khabsa, M., Avalani, M., Bhatt, M., Mankus, M., Hasson, M.,
  Lennie, M., Reso, M., Groshev, M., Naumov, M., Lathi, M., Keneally, M., Liu,
  M., Seltzer, M.~L., Valko, M., Restrepo, M., Patel, M., Vyatskov, M.,
  Samvelyan, M., Clark, M., Macey, M., Wang, M., Hermoso, M.~J., Metanat, M.,
  Rastegari, M., Bansal, M., Santhanam, N., Parks, N., White, N., Bawa, N.,
  Singhal, N., Egebo, N., Usunier, N., Mehta, N., Laptev, N.~P., Dong, N.,
  Cheng, N., Chernoguz, O., Hart, O., Salpekar, O., Kalinli, O., Kent, P.,
  Parekh, P., Saab, P., Balaji, P., Rittner, P., Bontrager, P., Roux, P.,
  Dollar, P., Zvyagina, P., Ratanchandani, P., Yuvraj, P., Liang, Q., Alao, R.,
  Rodriguez, R., Ayub, R., Murthy, R., Nayani, R., Mitra, R., Parthasarathy,
  R., Li, R., Hogan, R., Battey, R., Wang, R., Howes, R., Rinott, R., Mehta,
  S., Siby, S., Bondu, S.~J., Datta, S., Chugh, S., Hunt, S., Dhillon, S.,
  Sidorov, S., Pan, S., Mahajan, S., Verma, S., Yamamoto, S., Ramaswamy, S.,
  Lindsay, S., Lindsay, S., Feng, S., Lin, S., Zha, S.~C., Patil, S., Shankar,
  S., Zhang, S., Zhang, S., Wang, S., Agarwal, S., Sajuyigbe, S., Chintala, S.,
  Max, S., Chen, S., Kehoe, S., Satterfield, S., Govindaprasad, S., Gupta, S.,
  Deng, S., Cho, S., Virk, S., Subramanian, S., Choudhury, S., Goldman, S.,
  Remez, T., Glaser, T., Best, T., Koehler, T., Robinson, T., Li, T., Zhang,
  T., Matthews, T., Chou, T., Shaked, T., Vontimitta, V., Ajayi, V., Montanez,
  V., Mohan, V., Kumar, V.~S., Mangla, V., Ionescu, V., Poenaru, V.,
  Mihailescu, V.~T., Ivanov, V., Li, W., Wang, W., Jiang, W., Bouaziz, W.,
  Constable, W., Tang, X., Wu, X., Wang, X., Wu, X., Gao, X., Kleinman, Y.,
  Chen, Y., Hu, Y., Jia, Y., Qi, Y., Li, Y., Zhang, Y., Zhang, Y., Adi, Y.,
  Nam, Y., Yu, Wang, Zhao, Y., Hao, Y., Qian, Y., Li, Y., He, Y., Rait, Z.,
  DeVito, Z., Rosnbrick, Z., Wen, Z., Yang, Z., Zhao, Z.,  Ma, Z., ``The
  {Llama} 3 herd of models'', arXiv preprint arXiv:2407.21783, 2024.

\bibitem{wiese-etal-2017-neural}
\BIBentryALTinterwordspacing
Wiese, G., Weissenborn, D.,  Neves, M., ``Neural domain adaptation for
  biomedical question answering'', in Proceedings of the 21st Conference on
  Computational Natural Language Learning ({C}o{NLL} 2017), Levy, R.,  Specia,
  L., (ur.). Vancouver, Canada: Association for Computational Linguistics, Aug.
  2017, pages 281--289,  \url{https://aclanthology.org/K17-1029/}
\BIBentrySTDinterwordspacing

\bibitem{tong-etal-2020-improving}
\BIBentryALTinterwordspacing
Tong, M., Xu, B., Wang, S., Cao, Y., Hou, L., Li, J.,  Xie, J., ``Improving
  event detection via open-domain trigger knowledge'', in Proceedings of the
  58th Annual Meeting of the Association for Computational Linguistics,
  Jurafsky, D., Chai, J., Schluter, N.,  Tetreault, J., (ur.). Online:
  Association for Computational Linguistics, Jul. 2020, pages 5887--5897,
  \url{https://aclanthology.org/2020.acl-main.522/}
\BIBentrySTDinterwordspacing

\bibitem{sow2025dynamic}
\BIBentryALTinterwordspacing
Sow, D., Woisetschl{\"a}ger, H., Bulusu, S., Wang, S., Jacobsen, H.~A.,  Liang,
  Y., ``Dynamic loss-based sample reweighting for improved large language model
  pretraining'', in The Thirteenth International Conference on Learning
  Representations, 2025,  \url{https://openreview.net/forum?id=gU4ZgQNsOC}
\BIBentrySTDinterwordspacing

\bibitem{huerta-enochian-ko-2024-instruction}
\BIBentryALTinterwordspacing
Huerta-Enochian, M.,  Ko, S.~Y., ``Instruction fine-tuning: Does prompt loss
  matter?'', in Proceedings of the 2024 Conference on Empirical Methods in
  Natural Language Processing, Al-Onaizan, Y., Bansal, M.,  Chen, Y.-N., (ur.).
  Miami, Florida, USA: Association for Computational Linguistics, Nov. 2024,
  pages 22\,771--22\,795,  \url{https://aclanthology.org/2024.emnlp-main.1267/}
\BIBentrySTDinterwordspacing

\bibitem{an2024response}
An, S.,  Kim, H., ``Response tuning: Aligning large language models without
  instruction'', arXiv preprint arXiv:2410.02465, 2024.

\bibitem{9260074}
Dukić, D., Keča, D.,  Stipić, D., ``Are you human? {Detecting} bots on
  twitter using {BERT}'', in 2020 IEEE 7th International Conference on Data
  Science and Advanced Analytics (DSAA), 2020, pages 631-636.

\bibitem{peters-etal-2018-deep}
\BIBentryALTinterwordspacing
Peters, M.~E., Neumann, M., Iyyer, M., Gardner, M., Clark, C., Lee, K.,
  Zettlemoyer, L., ``Deep contextualized word representations'', in Proceedings
  of the 2018 Conference of the North {A}merican Chapter of the Association for
  Computational Linguistics: Human Language Technologies, Volume 1 (Long
  Papers), Walker, M., Ji, H.,  Stent, A., (ur.). New Orleans, Louisiana:
  Association for Computational Linguistics, Jun. 2018, pages 2227--2237,
  \url{https://aclanthology.org/N18-1202/}
\BIBentrySTDinterwordspacing

\bibitem{peters-etal-2019-tune-new}
\BIBentryALTinterwordspacing
Peters, M.~E., Ruder, S.,  Smith, N.~A., ``To tune or not to tune? {Adapting}
  pretrained representations to diverse tasks'', in Proceedings of the 4th
  Workshop on Representation Learning for NLP (RepL4NLP-2019), Augenstein, I.,
  Gella, S., Ruder, S., Kann, K., Can, B., Welbl, J., Conneau, A., Ren, X.,
  Rei, M., (ur.). Florence, Italy: Association for Computational Linguistics,
  Aug. 2019, pages 7--14,  \url{https://aclanthology.org/W19-4302/}
\BIBentrySTDinterwordspacing

\bibitem{lialin2023scaling}
Lialin, V., Deshpande, V.,  Rumshisky, A., ``Scaling down to scale up: A guide
  to parameter-efficient fine-tuning'', arXiv preprint arXiv:2303.15647, 2023.

\bibitem{MCCLOSKEY1989109}
\BIBentryALTinterwordspacing
McCloskey, M.,  Cohen, N.~J., ``Catastrophic interference in connectionist
  networks: The sequential learning problem'', Psychology of Learning and
  Motivation, Vol.~24, 1989, pages 109-165,
  \url{https://www.sciencedirect.com/science/article/pii/S0079742108605368}
\BIBentrySTDinterwordspacing

\bibitem{ben-zaken-etal-2022-bitfit}
\BIBentryALTinterwordspacing
Ben~Zaken, E., Goldberg, Y.,  Ravfogel, S., ``{B}it{F}it: Simple
  parameter-efficient fine-tuning for transformer-based masked
  language-models'', in Proceedings of the 60th Annual Meeting of the
  Association for Computational Linguistics (Volume 2: Short Papers), Muresan,
  S., Nakov, P.,  Villavicencio, A., (ur.). Dublin, Ireland: Association for
  Computational Linguistics, May 2022, pages 1--9,
  \url{https://aclanthology.org/2022.acl-short.1/}
\BIBentrySTDinterwordspacing

\bibitem{hu2021lora}
\BIBentryALTinterwordspacing
Hu, E.~J., yelong shen, Wallis, P., Allen-Zhu, Z., Li, Y., Wang, S., Wang, L.,
  Chen, W., ``Lo{RA}: Low-rank adaptation of large language models'', in
  International Conference on Learning Representations, 2022,
  \url{https://openreview.net/forum?id=nZeVKeeFYf9}
\BIBentrySTDinterwordspacing

\bibitem{dettmers2023qlora}
\BIBentryALTinterwordspacing
Dettmers, T., Pagnoni, A., Holtzman, A.,  Zettlemoyer, L., ``{QL}o{RA}:
  Efficient finetuning of quantized {LLM}s'', in Thirty-seventh Conference on
  Neural Information Processing Systems, 2023,
  \url{https://openreview.net/forum?id=OUIFPHEgJU}
\BIBentrySTDinterwordspacing

\bibitem{chen2016training}
Chen, T., Xu, B., Zhang, C.,  Guestrin, C., ``Training deep nets with sublinear
  memory cost'', arXiv preprint arXiv:1604.06174, 2016.

\bibitem{Piskorski2013}
\BIBentryALTinterwordspacing
Piskorski, J.,  Yangarber, R., Information Extraction: Past, Present and
  Future. Berlin, Heidelberg: Springer Berlin Heidelberg, 2013, pages 23--49,
  \url{https://doi.org/10.1007/978-3-642-28569-1_2}
\BIBentrySTDinterwordspacing

\bibitem{fischer-etal-2024-concept}
\BIBentryALTinterwordspacing
Fischer, T., Schneider, F., Geislinger, R., Helfer, F., Koch, G.,  Biemann, C.,
  ``Concept over time analysis: Unveiling temporal patterns for qualitative
  data analysis'', in Proceedings of the 2024 Conference of the North American
  Chapter of the Association for Computational Linguistics: Human Language
  Technologies (Volume 3: System Demonstrations), Chang, K.-W., Lee, A.,
  Rajani, N., (ur.). Mexico City, Mexico: Association for Computational
  Linguistics, Jun. 2024, pages 148--157,
  \url{https://aclanthology.org/2024.naacl-demo.15/}
\BIBentrySTDinterwordspacing

\bibitem{acled}
\BIBentryALTinterwordspacing
Piskorski, J., Haneczok, J.,  Jacquet, G., ``New benchmark corpus and models
  for fine-grained event classification: To {BERT} or not to {BERT}?'', in
  Proceedings of the 28th International Conference on Computational
  Linguistics. Barcelona, Spain (Online): International Committee on
  Computational Linguistics, Dec. 2020, pages 6663--6678,
  \url{https://aclanthology.org/2020.coling-main.584}
\BIBentrySTDinterwordspacing

\bibitem{iob2}
Ratnaparkhi, A.,  Marcus, M.~P., ``Maximum entropy models for natural language
  ambiguity resolution'', Doctoral thesis, University of Pennsylvania, USA,
  1998, aAI9840230.

\bibitem{he2020survey}
He, Z., Wang, Z., Wei, W., Feng, S., Mao, X.,  Jiang, S., ``A survey on recent
  advances in sequence labeling from deep learning models'', arXiv preprint
  arXiv:2011.06727, 2020.

\bibitem{ahn-2006-stages}
\BIBentryALTinterwordspacing
Ahn, D., ``The stages of event extraction'', in Proceedings of the Workshop on
  Annotating and Reasoning about Time and Events, Boguraev, B., Mu{\~n}oz, R.,
  Pustejovsky, J., (ur.). Sydney, Australia: Association for Computational
  Linguistics, Jul. 2006, pages 1--8,  \url{https://aclanthology.org/W06-0901/}
\BIBentrySTDinterwordspacing

\bibitem{lafferty2001conditional}
Lafferty, J., McCallum, A., Pereira, F. \emph{et~al.}, ``Conditional random
  fields: Probabilistic models for segmenting and labeling sequence data'', in
  ICML, Vol.~1, No.~2. Williamstown, MA, 2001, page~3.

\bibitem{akbik-etal-2018-contextual}
\BIBentryALTinterwordspacing
Akbik, A., Blythe, D.,  Vollgraf, R., ``Contextual string embeddings for
  sequence labeling'', in Proceedings of the 27th International Conference on
  Computational Linguistics, Bender, E.~M., Derczynski, L.,  Isabelle, P.,
  (ur.). Santa Fe, New Mexico, USA: Association for Computational Linguistics,
  Aug. 2018, pages 1638--1649,  \url{https://aclanthology.org/C18-1139/}
\BIBentrySTDinterwordspacing

\bibitem{fei-etal-2021-better-new}
\BIBentryALTinterwordspacing
Fei, H., Wu, S., Ren, Y., Li, F.,  Ji, D., ``Better combine them together!
  {Integrating} syntactic constituency and dependency representations for
  semantic role labeling'', in Findings of the Association for Computational
  Linguistics: ACL-IJCNLP 2021, Zong, C., Xia, F., Li, W.,  Navigli, R., (ur.).
  Online: Association for Computational Linguistics, Aug. 2021, pages 549--559,
   \url{https://aclanthology.org/2021.findings-acl.49/}
\BIBentrySTDinterwordspacing

\bibitem{wang2022instructionner}
Wang, L., Li, R., Yan, Y., Yan, Y., Wang, S., Wu, W.,  Xu, W.,
  ``Instruction{NER}: A multi-task instruction-based generative framework for
  few-shot {NER}'', arXiv preprint arXiv:2203.03903, 2022.

\bibitem{wang2023instructuie}
Wang, X., Zhou, W., Zu, C., Xia, H., Chen, T., Zhang, Y., Zheng, R., Ye, J.,
  Zhang, Q., Gui, T. \emph{et~al.}, ``Instruct{UIE}: Multi-task instruction
  tuning for unified information extraction'', arXiv preprint arXiv:2304.08085,
  2023.

\bibitem{zhang2023instruction}
Zhang, S., Dong, L., Li, X., Zhang, S., Sun, X., Wang, S., Li, J., Hu, R.,
  Zhang, T., Wu, F. \emph{et~al.}, ``Instruction tuning for large language
  models: A survey'', arXiv preprint arXiv:2308.10792, 2023.

\bibitem{spacy}
Honnibal, M., Montani, I., Van~Landeghem, S.,  Boyd, A., ``{spaCy}:
  {Industrial}-strength natural language processing in {Python}'', 2020.

\bibitem{doddington-etal-2004-automatic}
\BIBentryALTinterwordspacing
Doddington, G., Mitchell, A., Przybocki, M., Ramshaw, L., Strassel, S.,
  Weischedel, R., ``The automatic content extraction ({ACE}) program {--}
  tasks, data, and evaluation'', in Proceedings of the Fourth International
  Conference on Language Resources and Evaluation ({LREC}`04), Lino, M.~T.,
  Xavier, M.~F., Ferreira, F., Costa, R.,  Silva, R., (ur.). Lisbon, Portugal:
  European Language Resources Association (ELRA), May 2004,
  \url{https://aclanthology.org/L04-1011/}
\BIBentrySTDinterwordspacing

\bibitem{guidelines}
Consortium, L.~D. \emph{et~al.}, ``{ACE} (automatic content extraction)
  {English} annotation guidelines for events version 5.4. 3 2005.07. 01'',
  2005.

\bibitem{zhang-etal-2021-eventke-event}
\BIBentryALTinterwordspacing
Zhang, Z., Wang, H., Zhao, H., Tong, H.,  Ji, H., ``{E}vent{KE}: Event-enhanced
  knowledge graph embedding'', in Findings of the Association for Computational
  Linguistics: EMNLP 2021, Moens, M.-F., Huang, X., Specia, L.,  Yih, S. W.-t.,
  (ur.). Punta Cana, Dominican Republic: Association for Computational
  Linguistics, Nov. 2021, pages 1389--1400,
  \url{https://aclanthology.org/2021.findings-emnlp.120/}
\BIBentrySTDinterwordspacing

\bibitem{glavas-snajder-2013-event}
\BIBentryALTinterwordspacing
Glava{\v{s}}, G.,  {\v{S}}najder, J., ``Event-centered information retrieval
  using kernels on event graphs'', in Proceedings of {T}ext{G}raphs-8
  Graph-based Methods for Natural Language Processing, Kozareva, Z., Matveeva,
  I., Melli, G.,  Nastase, V., (ur.). Seattle, Washington, USA: Association for
  Computational Linguistics, Oct. 2013, pages 1--5,
  \url{https://aclanthology.org/W13-5001/}
\BIBentrySTDinterwordspacing

\bibitem{zhang-etal-2023-enhancing}
\BIBentryALTinterwordspacing
Zhang, Z., Elfardy, H., Dreyer, M., Small, K., Ji, H.,  Bansal, M., ``Enhancing
  multi-document summarization with cross-document graph-based information
  extraction'', in Proceedings of the 17th Conference of the European Chapter
  of the Association for Computational Linguistics, Vlachos, A.,  Augenstein,
  I., (ur.). Dubrovnik, Croatia: Association for Computational Linguistics, May
  2023, pages 1696--1707,  \url{https://aclanthology.org/2023.eacl-main.124/}
\BIBentrySTDinterwordspacing

\bibitem{tang-etal-2022-affective}
\BIBentryALTinterwordspacing
Tang, S., Chai, H., Yao, Z., Ding, Y., Gao, C., Fang, B.,  Liao, Q.,
  ``Affective knowledge enhanced multiple-graph fusion networks for
  aspect-based sentiment analysis'', in Proceedings of the 2022 Conference on
  Empirical Methods in Natural Language Processing, Goldberg, Y., Kozareva, Z.,
   Zhang, Y., (ur.). Abu Dhabi, United Arab Emirates: Association for
  Computational Linguistics, Dec. 2022, pages 5352--5362,
  \url{https://aclanthology.org/2022.emnlp-main.359/}
\BIBentrySTDinterwordspacing

\bibitem{xiang2019survey}
Xiang, W.,  Wang, B., ``A survey of event extraction from text'', IEEE Access,
  Vol.~7, 2019, pages 173\,111--173\,137.

\bibitem{wang-etal-2020-maven}
\BIBentryALTinterwordspacing
Wang, X., Wang, Z., Han, X., Jiang, W., Han, R., Liu, Z., Li, J., Li, P., Lin,
  Y.,  Zhou, J., ``{MAVEN}: {A} {M}assive {G}eneral {D}omain {E}vent
  {D}etection {D}ataset'', in Proceedings of the 2020 Conference on Empirical
  Methods in Natural Language Processing (EMNLP), Webber, B., Cohn, T., He, Y.,
   Liu, Y., (ur.). Online: Association for Computational Linguistics, Nov.
  2020, pages 1652--1671,  \url{https://aclanthology.org/2020.emnlp-main.129/}
\BIBentrySTDinterwordspacing

\bibitem{maisonnave2022detecting}
Maisonnave, M., Delbianco, F., Tohm{\'e}, F., Maguitman, A.,  Milios, E.,
  ``Detecting ongoing events using contextual word and sentence embeddings'',
  Expert Systems with Applications, Vol. 209, 2022, page 118257.

\bibitem{glavas_snajder_2015}
Glavaš, G.,  Šnajder, J., ``Construction and evaluation of event graphs'',
  Natural Language Engineering, Vol.~21, No.~4, 2015, page 607–652.

\bibitem{fan-etal-2019-using}
\BIBentryALTinterwordspacing
Fan, A., Gardent, C., Braud, C.,  Bordes, A., ``Using local knowledge graph
  construction to scale {S}eq2{S}eq models to multi-document inputs'', in
  Proceedings of the 2019 Conference on Empirical Methods in Natural Language
  Processing and the 9th International Joint Conference on Natural Language
  Processing (EMNLP-IJCNLP), Inui, K., Jiang, J., Ng, V.,  Wan, X., (ur.). Hong
  Kong, China: Association for Computational Linguistics, Nov. 2019, pages
  4186--4196,  \url{https://aclanthology.org/D19-1428/}
\BIBentrySTDinterwordspacing

\bibitem{ribeiro-etal-2022-factgraph}
\BIBentryALTinterwordspacing
Ribeiro, L. F.~R., Liu, M., Gurevych, I., Dreyer, M.,  Bansal, M.,
  ``{F}act{G}raph: Evaluating factuality in summarization with semantic graph
  representations'', in Proceedings of the 2022 Conference of the North
  American Chapter of the Association for Computational Linguistics: Human
  Language Technologies, Carpuat, M., de~Marneffe, M.-C.,  Meza~Ruiz, I.~V.,
  (ur.). Seattle, United States: Association for Computational Linguistics,
  Jul. 2022, pages 3238--3253,
  \url{https://aclanthology.org/2022.naacl-main.236/}
\BIBentrySTDinterwordspacing

\bibitem{yan2018assertion}
Yan, Z., Tang, D., Duan, N., Liu, S., Wang, W., Jiang, D., Zhou, M.,  Li, Z.,
  ``Assertion-based {QA} with question-aware open information extraction'', in
  Proceedings of the AAAI Conference on Artificial Intelligence, Vol.~32, 2018.

\bibitem{nagumothu-etal-2022-pie}
\BIBentryALTinterwordspacing
Nagumothu, D., Ofoghi, B., Huang, G.,  Eklund, P., ``{PIE}-{QG}: Paraphrased
  information extraction for unsupervised question generation from small
  corpora'', in Proceedings of the 26th Conference on Computational Natural
  Language Learning (CoNLL), Fokkens, A.,  Srikumar, V., (ur.). Abu Dhabi,
  United Arab Emirates (Hybrid): Association for Computational Linguistics,
  Dec. 2022, pages 350--359,  \url{https://aclanthology.org/2022.conll-1.24/}
\BIBentrySTDinterwordspacing

\bibitem{montella-etal-2020-denoising}
\BIBentryALTinterwordspacing
Montella, S., Fabre, B., Urvoy, T., Heinecke, J.,  Rojas-Barahona, L.,
  ``Denoising pre-training and data augmentation strategies for enhanced {RDF}
  verbalization with transformers'', in Proceedings of the 3rd International
  Workshop on Natural Language Generation from the Semantic Web (WebNLG+),
  Castro~Ferreira, T., Gardent, C., Ilinykh, N., van~der Lee, C., Mille, S.,
  Moussallem, D.,  Shimorina, A., (ur.). Dublin, Ireland (Virtual): Association
  for Computational Linguistics, 12 2020, pages 89--99,
  \url{https://aclanthology.org/2020.webnlg-1.9/}
\BIBentrySTDinterwordspacing

\bibitem{chen-etal-2023-led}
\BIBentryALTinterwordspacing
Chen, Y.-P., Yen, A.-Z., Huang, H.-H., Nakayama, H.,  Chen, H.-H., ``{LED}: A
  dataset for life event extraction from dialogs'', in Findings of the
  Association for Computational Linguistics: EACL 2023, Vlachos, A.,
  Augenstein, I., (ur.). Dubrovnik, Croatia: Association for Computational
  Linguistics, May 2023, pages 384--398,
  \url{https://aclanthology.org/2023.findings-eacl.29/}
\BIBentrySTDinterwordspacing

\bibitem{balasubramanian-etal-2013-generating}
\BIBentryALTinterwordspacing
Balasubramanian, N., Soderland, S., Mausam,  Etzioni, O., ``Generating coherent
  event schemas at scale'', in Proceedings of the 2013 Conference on Empirical
  Methods in Natural Language Processing, Yarowsky, D., Baldwin, T., Korhonen,
  A., Livescu, K.,  Bethard, S., (ur.). Seattle, Washington, USA: Association
  for Computational Linguistics, Oct. 2013, pages 1721--1731,
  \url{https://aclanthology.org/D13-1178/}
\BIBentrySTDinterwordspacing

\bibitem{pratapa-etal-2021-cross}
\BIBentryALTinterwordspacing
Pratapa, A., Liu, Z., Hasegawa, K., Li, L., Yamakawa, Y., Zhang, S.,  Mitamura,
  T., ``Cross-document event identity via dense annotation'', in Proceedings of
  the 25th Conference on Computational Natural Language Learning, Bisazza, A.,
  Abend, O., (ur.). Online: Association for Computational Linguistics, Nov.
  2021, pages 496--517,  \url{https://aclanthology.org/2021.conll-1.39/}
\BIBentrySTDinterwordspacing

\bibitem{banko2007open}
Banko, M., Cafarella, M.~J., Soderland, S., Broadhead, M.,  Etzioni, O., ``Open
  information extraction for the web'', in International Joint Conference on
  Artificial Intelligence, Vol.~7, 2007, pages 2670--2676.

\bibitem{biomedoie}
\BIBentryALTinterwordspacing
Wang, X., Zhang, Y., Li, Q., Chen, Y.,  Han, J., ``Open information extraction
  with meta-pattern discovery in biomedical literature'', in Proceedings of the
  2018 ACM International Conference on Bioinformatics, Computational Biology,
  and Health Informatics, ser. BCB '18. New York, NY, USA: Association for
  Computing Machinery, 2018, page 291–300,
  \url{https://doi.org/10.1145/3233547.3233594}
\BIBentrySTDinterwordspacing

\bibitem{sun-etal-2018-logician}
\BIBentryALTinterwordspacing
Sun, M., Li, X.,  Li, P., ``Logician and orator: Learning from the duality
  between language and knowledge in open domain'', in Proceedings of the 2018
  Conference on Empirical Methods in Natural Language Processing, Riloff, E.,
  Chiang, D., Hockenmaier, J.,  Tsujii, J., (ur.). Brussels, Belgium:
  Association for Computational Linguistics, Oct.-Nov. 2018, pages 2119--2130,
  \url{https://aclanthology.org/D18-1236/}
\BIBentrySTDinterwordspacing

\bibitem{gashteovski2019opiec}
\BIBentryALTinterwordspacing
Gashteovski, K., Wanner, S., Hertling, S., Broscheit, S.,  Gemulla, R.,
  ``{OPIEC}: An open information extraction corpus'', in Automated Knowledge
  Base Construction (AKBC), 2019,
  \url{https://openreview.net/forum?id=HJxeGb5pTm}
\BIBentrySTDinterwordspacing

\bibitem{kolluru-etal-2020-openie6}
\BIBentryALTinterwordspacing
Kolluru, K., Adlakha, V., Aggarwal, S., Mausam,  Chakrabarti, S.,
  ``{O}pen{IE}6: {I}terative {G}rid {L}abeling and {C}oordination {A}nalysis
  for {O}pen {I}nformation {E}xtraction'', in Proceedings of the 2020
  Conference on Empirical Methods in Natural Language Processing (EMNLP),
  Webber, B., Cohn, T., He, Y.,  Liu, Y., (ur.). Online: Association for
  Computational Linguistics, Nov. 2020, pages 3748--3761,
  \url{https://aclanthology.org/2020.emnlp-main.306/}
\BIBentrySTDinterwordspacing

\bibitem{kotnis-etal-2022-milie}
\BIBentryALTinterwordspacing
Kotnis, B., Gashteovski, K., Rubio, D., Shaker, A., Rodriguez-Tembras, V.,
  Takamoto, M., Niepert, M.,  Lawrence, C., ``{MILIE}: Modular {\&} iterative
  multilingual open information extraction'', in Proceedings of the 60th Annual
  Meeting of the Association for Computational Linguistics (Volume 1: Long
  Papers), Muresan, S., Nakov, P.,  Villavicencio, A., (ur.). Dublin, Ireland:
  Association for Computational Linguistics, May 2022, pages 6939--6950,
  \url{https://aclanthology.org/2022.acl-long.478/}
\BIBentrySTDinterwordspacing

\bibitem{angeli-etal-2015-leveraging}
\BIBentryALTinterwordspacing
Angeli, G., Johnson~Premkumar, M.~J.,  Manning, C.~D., ``Leveraging linguistic
  structure for open domain information extraction'', in Proceedings of the
  53rd Annual Meeting of the Association for Computational Linguistics and the
  7th International Joint Conference on Natural Language Processing (Volume 1:
  Long Papers), Zong, C.,  Strube, M., (ur.). Beijing, China: Association for
  Computational Linguistics, Jul. 2015, pages 344--354,
  \url{https://aclanthology.org/P15-1034/}
\BIBentrySTDinterwordspacing

\bibitem{gashteovski-etal-2017-minie}
\BIBentryALTinterwordspacing
Gashteovski, K., Gemulla, R.,  del Corro, L., ``{M}in{IE}: Minimizing facts in
  open information extraction'', in Proceedings of the 2017 Conference on
  Empirical Methods in Natural Language Processing, Palmer, M., Hwa, R.,
  Riedel, S., (ur.). Copenhagen, Denmark: Association for Computational
  Linguistics, Sep. 2017, pages 2630--2640,
  \url{https://aclanthology.org/D17-1278/}
\BIBentrySTDinterwordspacing

\bibitem{lauscher2019minscie}
Lauscher, A., Song, Y.,  Gashteovski, K., ``{MinScIE}: {Citation}-centered open
  information extraction'', in 2019 ACM/IEEE Joint Conference on Digital
  Libraries (JCDL). IEEE, 2019, pages 386--387.

\bibitem{gashteovski-etal-2022-benchie}
\BIBentryALTinterwordspacing
Gashteovski, K., Yu, M., Kotnis, B., Lawrence, C., Niepert, M.,  Glava{\v{s}},
  G., ``{B}ench{IE}: A framework for multi-faceted fact-based open information
  extraction evaluation'', in Proceedings of the 60th Annual Meeting of the
  Association for Computational Linguistics (Volume 1: Long Papers), Muresan,
  S., Nakov, P.,  Villavicencio, A., (ur.). Dublin, Ireland: Association for
  Computational Linguistics, May 2022, pages 4472--4490,
  \url{https://aclanthology.org/2022.acl-long.307/}
\BIBentrySTDinterwordspacing

\bibitem{broscheit-etal-2020-predict-new}
\BIBentryALTinterwordspacing
Broscheit, S., Gashteovski, K., Wang, Y.,  Gemulla, R., ``Can we predict new
  facts with open knowledge graph embeddings? {A} benchmark for open link
  prediction'', in Proceedings of the 58th Annual Meeting of the Association
  for Computational Linguistics, Jurafsky, D., Chai, J., Schluter, N.,
  Tetreault, J., (ur.). Online: Association for Computational Linguistics, Jul.
  2020, pages 2296--2308,  \url{https://aclanthology.org/2020.acl-main.209/}
\BIBentrySTDinterwordspacing

\bibitem{9039685}
Li, J., Sun, A., Han, J.,  Li, C., ``A survey on deep learning for named entity
  recognition'', IEEE Transactions on Knowledge and Data Engineering, Vol.~34,
  No.~1, 2022, pages 50-70.

\bibitem{nadeau2007survey}
Nadeau, D.,  Sekine, S., ``A survey of named entity recognition and
  classification'', Lingvisticae Investigationes, Vol.~30, No.~1, 2007, pages
  3--26.

\bibitem{pradhan-etal-2013-towards}
\BIBentryALTinterwordspacing
Pradhan, S., Moschitti, A., Xue, N., Ng, H.~T., Bj{\"o}rkelund, A., Uryupina,
  O., Zhang, Y.,  Zhong, Z., ``Towards robust linguistic analysis using
  {O}nto{N}otes'', in Proceedings of the Seventeenth Conference on
  Computational Natural Language Learning, Hockenmaier, J.,  Riedel, S., (ur.).
  Sofia, Bulgaria: Association for Computational Linguistics, Aug. 2013, pages
  143--152,  \url{https://aclanthology.org/W13-3516/}
\BIBentrySTDinterwordspacing

\bibitem{lhoest-etal-2021-datasets}
\BIBentryALTinterwordspacing
Lhoest, Q., Villanova~del Moral, A., Jernite, Y., Thakur, A., von Platen, P.,
  Patil, S., Chaumond, J., Drame, M., Plu, J., Tunstall, L., Davison, J.,
  {\v{S}}a{\v{s}}ko, M., Chhablani, G., Malik, B., Brandeis, S., Le~Scao, T.,
  Sanh, V., Xu, C., Patry, N., McMillan-Major, A., Schmid, P., Gugger, S.,
  Delangue, C., Matussi{\`e}re, T., Debut, L., Bekman, S., Cistac, P.,
  Goehringer, T., Mustar, V., Lagunas, F., Rush, A.,  Wolf, T., ``Datasets: A
  community library for natural language processing'', in Proceedings of the
  2021 Conference on Empirical Methods in Natural Language Processing: System
  Demonstrations, Adel, H.,  Shi, S., (ur.). Online and Punta Cana, Dominican
  Republic: Association for Computational Linguistics, Nov. 2021, pages
  175--184,  \url{https://aclanthology.org/2021.emnlp-demo.21/}
\BIBentrySTDinterwordspacing

\bibitem{9996141}
Zhang, W., Li, X., Deng, Y., Bing, L.,  Lam, W., ``A survey on aspect-based
  sentiment analysis: Tasks, methods, and challenges'', IEEE Transactions on
  Knowledge and Data Engineering, Vol.~35, No.~11, 2023, pages 11\,019-11\,038.

\bibitem{pontiki-etal-2014-semeval}
\BIBentryALTinterwordspacing
Pontiki, M., Galanis, D., Pavlopoulos, J., Papageorgiou, H., Androutsopoulos,
  I.,  Manandhar, S., ``{S}em{E}val-2014 task 4: Aspect based sentiment
  analysis'', in Proceedings of the 8th International Workshop on Semantic
  Evaluation ({S}em{E}val 2014), Nakov, P.,  Zesch, T., (ur.). Dublin, Ireland:
  Association for Computational Linguistics, Aug. 2014, pages 27--35,
  \url{https://aclanthology.org/S14-2004/}
\BIBentrySTDinterwordspacing

\bibitem{wang-etal-2021-automated}
\BIBentryALTinterwordspacing
Wang, X., Jiang, Y., Bach, N., Wang, T., Huang, Z., Huang, F.,  Tu, K.,
  ``Automated concatenation of embeddings for structured prediction'', in
  Proceedings of the 59th Annual Meeting of the Association for Computational
  Linguistics and the 11th International Joint Conference on Natural Language
  Processing (Volume 1: Long Papers), Zong, C., Xia, F., Li, W.,  Navigli, R.,
  (ur.). Online: Association for Computational Linguistics, Aug. 2021, pages
  2643--2660,  \url{https://aclanthology.org/2021.acl-long.206/}
\BIBentrySTDinterwordspacing

\bibitem{trautmann-2020-aspect}
\BIBentryALTinterwordspacing
Trautmann, D., ``Aspect-based argument mining'', in Proceedings of the 7th
  Workshop on Argument Mining, Cabrio, E.,  Villata, S., (ur.). Online:
  Association for Computational Linguistics, Dec. 2020, pages 41--52,
  \url{https://aclanthology.org/2020.argmining-1.5/}
\BIBentrySTDinterwordspacing

\bibitem{ruckdeschel-wiedemann-2022-boundary}
\BIBentryALTinterwordspacing
Ruckdeschel, M.,  Wiedemann, G., ``Boundary detection and categorization of
  argument aspects via supervised learning'', in Proceedings of the 9th
  Workshop on Argument Mining, Lapesa, G., Schneider, J., Jo, Y.,  Saha, S.,
  (ur.). Online and in Gyeongju, Republic of Korea: International Conference on
  Computational Linguistics, Oct. 2022, pages 126--136,
  \url{https://aclanthology.org/2022.argmining-1.12/}
\BIBentrySTDinterwordspacing

\bibitem{gupta-etal-2019-simple}
\BIBentryALTinterwordspacing
Gupta, A., Hewitt, J.,  Kirchhoff, K., ``Simple, fast, accurate intent
  classification and slot labeling for goal-oriented dialogue systems'', in
  Proceedings of the 20th Annual SIGdial Meeting on Discourse and Dialogue,
  Nakamura, S., Gasic, M., Zukerman, I., Skantze, G., Nakano, M., Papangelis,
  A., Ultes, S.,  Yoshino, K., (ur.). Stockholm, Sweden: Association for
  Computational Linguistics, Sep. 2019, pages 46--55,
  \url{https://aclanthology.org/W19-5906/}
\BIBentrySTDinterwordspacing

\bibitem{razumovskaia-etal-2022-natural}
\BIBentryALTinterwordspacing
Razumovskaia, E., Glava{\v{s}}, G., Majewska, O., Ponti, E.,  Vuli{\'c}, I.,
  ``Natural language processing for multilingual task-oriented dialogue'', in
  Proceedings of the 60th Annual Meeting of the Association for Computational
  Linguistics: Tutorial Abstracts, Benotti, L., Okazaki, N., Scherrer, Y.,
  Zampieri, M., (ur.). Dublin, Ireland: Association for Computational
  Linguistics, May 2022, pages 44--50,
  \url{https://aclanthology.org/2022.acl-tutorials.8/}
\BIBentrySTDinterwordspacing

\bibitem{casanueva-etal-2022-nlu}
\BIBentryALTinterwordspacing
Casanueva, I., Vuli{\'c}, I., Spithourakis, G.,  Budzianowski, P., ``{NLU}++: A
  multi-label, slot-rich, generalisable dataset for natural language
  understanding in task-oriented dialogue'', in Findings of the Association for
  Computational Linguistics: NAACL 2022, Carpuat, M., de~Marneffe, M.-C.,
  Meza~Ruiz, I.~V., (ur.). Seattle, United States: Association for
  Computational Linguistics, Jul. 2022, pages 1998--2013,
  \url{https://aclanthology.org/2022.findings-naacl.154/}
\BIBentrySTDinterwordspacing

\bibitem{mehta-etal-2024-promptly}
\BIBentryALTinterwordspacing
Mehta, M., Pyatkin, V.,  Srikumar, V., ``Promptly predicting structures: The
  return of inference'', in Proceedings of the 2024 Conference of the North
  American Chapter of the Association for Computational Linguistics: Human
  Language Technologies (Volume 1: Long Papers), Duh, K., Gomez, H.,  Bethard,
  S., (ur.). Mexico City, Mexico: Association for Computational Linguistics,
  Jun. 2024, pages 112--130,  \url{https://aclanthology.org/2024.naacl-long.7/}
\BIBentrySTDinterwordspacing

\bibitem{seqeval}
\BIBentryALTinterwordspacing
Nakayama, H., ``{seqeval}: A {Python} framework for sequence labeling
  evaluation'',  \url{https://github.com/chakki-works/seqeval} Software
  available from https://github.com/chakki-works/seqeval. 2018.
\BIBentrySTDinterwordspacing

\bibitem{pustejovsky2005specification}
Pustejovsky, J., Ingria, R., Sauri, R., Casta{\~n}o, J.~M., Littman, J.,
  Gaizauskas, R.~J., Setzer, A., Katz, G.,  Mani, I., ``The specification
  language {TimeML}.'', 2005.

\bibitem{shaw2009lode}
Shaw, R., Troncy, R.,  Hardman, L., ``{LODE}: {Linking} open descriptions of
  events.'', ASWC, Vol.~9, 2009, pages 153--167.

\bibitem{cybulska2014guidelines}
Cybulska, A.,  Vossen, P., ``Guidelines for {ECB+} annotation of events and
  their coreference'', 2014.

\bibitem{song-etal-2015-light}
\BIBentryALTinterwordspacing
Song, Z., Bies, A., Strassel, S., Riese, T., Mott, J., Ellis, J., Wright, J.,
  Kulick, S., Ryant, N.,  Ma, X., ``From light to rich {ERE}: Annotation of
  entities, relations, and events'', in Proceedings of the 3rd Workshop on
  {EVENTS}: Definition, Detection, Coreference, and Representation, Hovy, E.,
  Mitamura, T.,  Palmer, M., (ur.). Denver, Colorado: Association for
  Computational Linguistics, Jun. 2015, pages 89--98,
  \url{https://aclanthology.org/W15-0812/}
\BIBentrySTDinterwordspacing

\bibitem{liu-etal-2016-leveraging}
\BIBentryALTinterwordspacing
Liu, S., Chen, Y., He, S., Liu, K.,  Zhao, J., ``Leveraging {F}rame{N}et to
  improve automatic event detection'', in Proceedings of the 54th Annual
  Meeting of the Association for Computational Linguistics (Volume 1: Long
  Papers), Erk, K.,  Smith, N.~A., (ur.). Berlin, Germany: Association for
  Computational Linguistics, Aug. 2016, pages 2134--2143,
  \url{https://aclanthology.org/P16-1201/}
\BIBentrySTDinterwordspacing

\bibitem{deng-etal-2022-title2event}
\BIBentryALTinterwordspacing
Deng, H., Zhang, Y., Zhang, Y., Ying, W., Yu, C., Gao, J., Wang, W., Bai, X.,
  Yang, N., Ma, J., Chen, X.,  Zhou, T., ``{T}itle2{E}vent: Benchmarking open
  event extraction with a large-scale {C}hinese title dataset'', in Proceedings
  of the 2022 Conference on Empirical Methods in Natural Language Processing,
  Goldberg, Y., Kozareva, Z.,  Zhang, Y., (ur.). Abu Dhabi, United Arab
  Emirates: Association for Computational Linguistics, Dec. 2022, pages
  6511--6524,  \url{https://aclanthology.org/2022.emnlp-main.437/}
\BIBentrySTDinterwordspacing

\bibitem{nguyen-grishman-2015-event}
\BIBentryALTinterwordspacing
Nguyen, T.~H.,  Grishman, R., ``Event detection and domain adaptation with
  convolutional neural networks'', in Proceedings of the 53rd Annual Meeting of
  the Association for Computational Linguistics and the 7th International Joint
  Conference on Natural Language Processing (Volume 2: Short Papers), Zong, C.,
   Strube, M., (ur.). Beijing, China: Association for Computational
  Linguistics, Jul. 2015, pages 365--371,
  \url{https://aclanthology.org/P15-2060/}
\BIBentrySTDinterwordspacing

\bibitem{liu-etal-2017-exploiting}
\BIBentryALTinterwordspacing
Liu, S., Chen, Y., Liu, K.,  Zhao, J., ``Exploiting argument information to
  improve event detection via supervised attention mechanisms'', in Proceedings
  of the 55th Annual Meeting of the Association for Computational Linguistics
  (Volume 1: Long Papers), Barzilay, R.,  Kan, M.-Y., (ur.). Vancouver, Canada:
  Association for Computational Linguistics, Jul. 2017, pages 1789--1798,
  \url{https://aclanthology.org/P17-1164/}
\BIBentrySTDinterwordspacing

\bibitem{ji-etal-2019-exploiting}
\BIBentryALTinterwordspacing
Ji, Y., Lin, Y., Gao, J.,  Wan, H., ``Exploiting the entity type sequence to
  benefit event detection'', in Proceedings of the 23rd Conference on
  Computational Natural Language Learning (CoNLL), Bansal, M.,  Villavicencio,
  A., (ur.). Hong Kong, China: Association for Computational Linguistics, Nov.
  2019, pages 613--623,  \url{https://aclanthology.org/K19-1057/}
\BIBentrySTDinterwordspacing

\bibitem{wolf-etal-2020-transformers}
\BIBentryALTinterwordspacing
Wolf, T., Debut, L., Sanh, V., Chaumond, J., Delangue, C., Moi, A., Cistac, P.,
  Rault, T., Louf, R., Funtowicz, M., Davison, J., Shleifer, S., von Platen,
  P., Ma, C., Jernite, Y., Plu, J., Xu, C., Le~Scao, T., Gugger, S., Drame, M.,
  Lhoest, Q.,  Rush, A., ``Transformers: State-of-the-art natural language
  processing'', in Proceedings of the 2020 Conference on Empirical Methods in
  Natural Language Processing: System Demonstrations, Liu, Q.,  Schlangen, D.,
  (ur.). Online: Association for Computational Linguistics, Oct. 2020, pages
  38--45,  \url{https://aclanthology.org/2020.emnlp-demos.6/}
\BIBentrySTDinterwordspacing

\bibitem{kingma2014adam}
Kingma, D.~P.,  Ba, J., ``Adam: {A} method for stochastic optimization'', in
  International Conference on Learning Representations, 2015.

\bibitem{lauscher-etal-2020-zero}
\BIBentryALTinterwordspacing
Lauscher, A., Ravishankar, V., Vuli{\'c}, I.,  Glava{\v{s}}, G., ``From zero to
  hero: {O}n the limitations of zero-shot language transfer with multilingual
  {T}ransformers'', in Proceedings of the 2020 Conference on Empirical Methods
  in Natural Language Processing (EMNLP), Webber, B., Cohn, T., He, Y.,  Liu,
  Y., (ur.). Online: Association for Computational Linguistics, Nov. 2020,
  pages 4483--4499,  \url{https://aclanthology.org/2020.emnlp-main.363/}
\BIBentrySTDinterwordspacing

\bibitem{artetxe-etal-2022-role}
\BIBentryALTinterwordspacing
Artetxe, M., Du, J., Goyal, N., Zettlemoyer, L.,  Stoyanov, V., ``On the role
  of bidirectionality in language model pre-training'', in Findings of the
  Association for Computational Linguistics: EMNLP 2022, Goldberg, Y.,
  Kozareva, Z.,  Zhang, Y., (ur.). Abu Dhabi, United Arab Emirates: Association
  for Computational Linguistics, Dec. 2022, pages 3973--3985,
  \url{https://aclanthology.org/2022.findings-emnlp.293/}
\BIBentrySTDinterwordspacing

\bibitem{han2023information}
Han, R., Peng, T., Yang, C., Wang, B., Liu, L.,  Wan, X., ``Is information
  extraction solved by {ChatGPT}? an analysis of performance, evaluation
  criteria, robustness and errors'', arXiv preprint arXiv:2305.14450, 2023.

\bibitem{wang2023gpt}
Wang, S., Sun, X., Li, X., Ouyang, R., Wu, F., Zhang, T., Li, J.,  Wang, G.,
  ``{GPT-NER}: Named entity recognition via large language models'', arXiv
  preprint arXiv:2304.10428, 2023.

\bibitem{jiang2023mistral}
Jiang, A.~Q., Sablayrolles, A., Mensch, A., Bamford, C., Chaplot, D.~S., de~las
  Casas, D., Bressand, F., Lengyel, G., Lample, G., Saulnier, L., Lavaud,
  L.~R., Lachaux, M.-A., Stock, P., Scao, T.~L., Lavril, T., Wang, T., Lacroix,
  T.,  Sayed, W.~E., ``Mistral {7B}'', arXiv preprint arXiv:2310.06825, 2023.

\bibitem{team2024gemma}
Team, G., Mesnard, T., Hardin, C., Dadashi, R., Bhupatiraju, S., Pathak, S.,
  Sifre, L., Rivière, M., Kale, M.~S., Love, J., Tafti, P., Hussenot, L.,
  Sessa, P.~G., Chowdhery, A., Roberts, A., Barua, A., Botev, A., Castro-Ros,
  A., Slone, A., Héliou, A., Tacchetti, A., Bulanova, A., Paterson, A., Tsai,
  B., Shahriari, B., Lan, C.~L., Choquette-Choo, C.~A., Crepy, C., Cer, D.,
  Ippolito, D., Reid, D., Buchatskaya, E., Ni, E., Noland, E., Yan, G., Tucker,
  G., Muraru, G.-C., Rozhdestvenskiy, G., Michalewski, H., Tenney, I.,
  Grishchenko, I., Austin, J., Keeling, J., Labanowski, J., Lespiau, J.-B.,
  Stanway, J., Brennan, J., Chen, J., Ferret, J., Chiu, J., Mao-Jones, J., Lee,
  K., Yu, K., Millican, K., Sjoesund, L.~L., Lee, L., Dixon, L., Reid, M.,
  Mikuła, M., Wirth, M., Sharman, M., Chinaev, N., Thain, N., Bachem, O.,
  Chang, O., Wahltinez, O., Bailey, P., Michel, P., Yotov, P., Chaabouni, R.,
  Comanescu, R., Jana, R., Anil, R., McIlroy, R., Liu, R., Mullins, R., Smith,
  S.~L., Borgeaud, S., Girgin, S., Douglas, S., Pandya, S., Shakeri, S., De,
  S., Klimenko, T., Hennigan, T., Feinberg, V., Stokowiec, W., hui Chen, Y.,
  Ahmed, Z., Gong, Z., Warkentin, T., Peran, L., Giang, M., Farabet, C.,
  Vinyals, O., Dean, J., Kavukcuoglu, K., Hassabis, D., Ghahramani, Z., Eck,
  D., Barral, J., Pereira, F., Collins, E., Joulin, A., Fiedel, N., Senter, E.,
  Andreev, A.,  Kenealy, K., ``Gemma: Open models based on gemini research and
  technology'', arXiv preprint arXiv:2403.08295, 2024.

\bibitem{li2023label}
Li, Z., Li, X., Liu, Y., Xie, H., Li, J., Wang, F.-l., Li, Q.,  Zhong, X.,
  ``Label supervised {Llama} finetuning'', arXiv preprint arXiv:2310.01208,
  2023.

\bibitem{behnamghader2024llm2vec}
\BIBentryALTinterwordspacing
BehnamGhader, P., Adlakha, V., Mosbach, M., Bahdanau, D., Chapados, N.,  Reddy,
  S., ``{LLM}2{Vec}: Large language models are secretly powerful text
  encoders'', in First Conference on Language Modeling, 2024,
  \url{https://openreview.net/forum?id=IW1PR7vEBf}
\BIBentrySTDinterwordspacing

\bibitem{lee2024nv}
\BIBentryALTinterwordspacing
Lee, C., Roy, R., Xu, M., Raiman, J., Shoeybi, M., Catanzaro, B.,  Ping, W.,
  ``{NV}-{Embed}: Improved techniques for training {LLM}s as generalist
  embedding models'', in The Thirteenth International Conference on Learning
  Representations, 2025,  \url{https://openreview.net/forum?id=lgsyLSsDRe}
\BIBentrySTDinterwordspacing

\bibitem{HUANG2025112907}
\BIBentryALTinterwordspacing
Huang, Z., Huang, X., Wu, A., Wang, X.,  Cheng, G., ``Transforming decoder-only
  models into encoder-only models with improved understanding capabilities'',
  Knowledge-Based Systems, Vol. 309, 2025, page 112907,
  \url{https://www.sciencedirect.com/science/article/pii/S0950705124015417}
\BIBentrySTDinterwordspacing

\bibitem{loshchilov2017decoupled}
\BIBentryALTinterwordspacing
Loshchilov, I.,  Hutter, F., ``Decoupled weight decay regularization'', in
  International Conference on Learning Representations, 2019,
  \url{https://openreview.net/forum?id=Bkg6RiCqY7}
\BIBentrySTDinterwordspacing

\bibitem{zhu2015aligning}
Zhu, Y., Kiros, R., Zemel, R., Salakhutdinov, R., Urtasun, R., Torralba, A.,
  Fidler, S., ``Aligning books and movies: Towards story-like visual
  explanations by watching movies and reading books'', in Proceedings of the
  IEEE international conference on computer vision, 2015, pages 19--27.

\bibitem{wu-etal-2020-similarity}
\BIBentryALTinterwordspacing
Wu, J., Belinkov, Y., Sajjad, H., Durrani, N., Dalvi, F.,  Glass, J.,
  ``Similarity analysis of contextual word representation models'', in
  Proceedings of the 58th Annual Meeting of the Association for Computational
  Linguistics, Jurafsky, D., Chai, J., Schluter, N.,  Tetreault, J., (ur.).
  Online: Association for Computational Linguistics, Jul. 2020, pages
  4638--4655,  \url{https://aclanthology.org/2020.acl-main.422/}
\BIBentrySTDinterwordspacing

\bibitem{scaria-etal-2024-instructabsa}
\BIBentryALTinterwordspacing
Scaria, K., Gupta, H., Goyal, S., Sawant, S., Mishra, S.,  Baral, C.,
  ``{I}nstruct{ABSA}: Instruction learning for aspect based sentiment
  analysis'', in Proceedings of the 2024 Conference of the North American
  Chapter of the Association for Computational Linguistics: Human Language
  Technologies (Volume 2: Short Papers), Duh, K., Gomez, H.,  Bethard, S.,
  (ur.). Mexico City, Mexico: Association for Computational Linguistics, Jun.
  2024, pages 720--736,  \url{https://aclanthology.org/2024.naacl-short.63/}
\BIBentrySTDinterwordspacing

\bibitem{liu-etal-2022-autoregressive}
\BIBentryALTinterwordspacing
Liu, T., Jiang, Y.~E., Monath, N., Cotterell, R.,  Sachan, M., ``Autoregressive
  structured prediction with language models'', in Findings of the Association
  for Computational Linguistics: EMNLP 2022, Goldberg, Y., Kozareva, Z.,
  Zhang, Y., (ur.). Abu Dhabi, United Arab Emirates: Association for
  Computational Linguistics, Dec. 2022, pages 993--1005,
  \url{https://aclanthology.org/2022.findings-emnlp.70/}
\BIBentrySTDinterwordspacing

\bibitem{yang-li-2024-modeling}
\BIBentryALTinterwordspacing
Yang, H.,  Li, K., ``Modeling aspect sentiment coherency via local sentiment
  aggregation'', in Findings of the Association for Computational Linguistics:
  EACL 2024, Graham, Y.,  Purver, M., (ur.). St. Julian{'}s, Malta: Association
  for Computational Linguistics, Mar. 2024, pages 182--195,
  \url{https://aclanthology.org/2024.findings-eacl.13/}
\BIBentrySTDinterwordspacing

\bibitem{he2020deberta}
\BIBentryALTinterwordspacing
He, P., Liu, X., Gao, J.,  Chen, W., ``De{BERT}a: Decoding-enhanced {BERT} with
  disentangled attention'', in International Conference on Learning
  Representations, 2021,  \url{https://openreview.net/forum?id=XPZIaotutsD}
\BIBentrySTDinterwordspacing

\bibitem{wang-etal-2022-deepstruct}
\BIBentryALTinterwordspacing
Wang, C., Liu, X., Chen, Z., Hong, H., Tang, J.,  Song, D., ``{D}eep{S}truct:
  Pretraining of language models for structure prediction'', in Findings of the
  Association for Computational Linguistics: ACL 2022, Muresan, S., Nakov, P.,
  Villavicencio, A., (ur.). Dublin, Ireland: Association for Computational
  Linguistics, May 2022, pages 803--823,
  \url{https://aclanthology.org/2022.findings-acl.67/}
\BIBentrySTDinterwordspacing

\bibitem{wies2023learnability}
Wies, N., Levine, Y.,  Shashua, A., ``The learnability of in-context
  learning'', in Advances in Neural Information Processing Systems, Vol.~36,
  2023, pages 36\,637--36\,651.

\bibitem{mosbach2023few}
\BIBentryALTinterwordspacing
Mosbach, M., Pimentel, T., Ravfogel, S., Klakow, D.,  Elazar, Y., ``Few-shot
  fine-tuning vs. in-context learning: A fair comparison and evaluation'', in
  Findings of the Association for Computational Linguistics: ACL 2023, Rogers,
  A., Boyd-Graber, J.,  Okazaki, N., (ur.). Toronto, Canada: Association for
  Computational Linguistics, Jul. 2023, pages 12\,284--12\,314,
  \url{https://aclanthology.org/2023.findings-acl.779/}
\BIBentrySTDinterwordspacing

\bibitem{agarwal2024many}
Agarwal, R., Singh, A., Zhang, L., Bohnet, B., Rosias, L., Chan, S., Zhang, B.,
  Anand, A., Abbas, Z., Nova, A. \emph{et~al.}, ``Many-shot in-context
  learning'', in Advances in Neural Information Processing Systems, Vol.~37,
  2024, pages 76\,930--76\,966.

\bibitem{bertsch-etal-2025-context}
\BIBentryALTinterwordspacing
Bertsch, A., Ivgi, M., Xiao, E., Alon, U., Berant, J., Gormley, M.~R.,  Neubig,
  G., ``In-context learning with long-context models: An in-depth
  exploration'', in Proceedings of the 2025 Conference of the Nations of the
  Americas Chapter of the Association for Computational Linguistics: Human
  Language Technologies (Volume 1: Long Papers), Chiruzzo, L., Ritter, A.,
  Wang, L., (ur.). Albuquerque, New Mexico: Association for Computational
  Linguistics, Apr. 2025, pages 12\,119--12\,149,
  \url{https://aclanthology.org/2025.naacl-long.605/}
\BIBentrySTDinterwordspacing

\bibitem{hewitt2024instruction}
Hewitt, J., Liu, N.~F., Liang, P.,  Manning, C.~D., ``Instruction following
  without instruction tuning'', arXiv preprint arXiv:2409.14254, 2024.

\bibitem{min-etal-2022-rethinking}
\BIBentryALTinterwordspacing
Min, S., Lyu, X., Holtzman, A., Artetxe, M., Lewis, M., Hajishirzi, H.,
  Zettlemoyer, L., ``Rethinking the role of demonstrations: What makes
  in-context learning work?'', in Proceedings of the 2022 Conference on
  Empirical Methods in Natural Language Processing, Goldberg, Y., Kozareva, Z.,
   Zhang, Y., (ur.). Abu Dhabi, United Arab Emirates: Association for
  Computational Linguistics, Dec. 2022, pages 11\,048--11\,064,
  \url{https://aclanthology.org/2022.emnlp-main.759/}
\BIBentrySTDinterwordspacing

\bibitem{lampinen2024broader}
Lampinen, A.~K., Chan, S.~C., Singh, A.~K.,  Shanahan, M., ``The broader
  spectrum of in-context learning'', arXiv preprint arXiv:2412.03782, 2024.

\bibitem{duan-etal-2024-exploring}
\BIBentryALTinterwordspacing
Duan, H., Tang, Y., Yang, Y., Abbasi, A.,  Tam, K.~Y., ``Exploring the
  relationship between in-context learning and instruction tuning'', in
  Findings of the Association for Computational Linguistics: EMNLP 2024,
  Al-Onaizan, Y., Bansal, M.,  Chen, Y.-N., (ur.). Miami, Florida, USA:
  Association for Computational Linguistics, Nov. 2024, pages 3197--3210,
  \url{https://aclanthology.org/2024.findings-emnlp.182/}
\BIBentrySTDinterwordspacing

\bibitem{mosbach-etal-2023-shot}
\BIBentryALTinterwordspacing
Mosbach, M., Pimentel, T., Ravfogel, S., Klakow, D.,  Elazar, Y., ``Few-shot
  fine-tuning vs. in-context learning: A fair comparison and evaluation'', in
  Findings of the Association for Computational Linguistics: ACL 2023, Rogers,
  A., Boyd-Graber, J.,  Okazaki, N., (ur.). Toronto, Canada: Association for
  Computational Linguistics, Jul. 2023, pages 12\,284--12\,314,
  \url{https://aclanthology.org/2023.findings-acl.779/}
\BIBentrySTDinterwordspacing

\bibitem{yin-etal-2024-deeper}
\BIBentryALTinterwordspacing
Yin, Q., He, X., Leong, C.~T., Wang, F., Yan, Y., Shen, X.,  Zhang, Q.,
  ``Deeper insights without updates: The power of in-context learning over
  fine-tuning'', in Findings of the Association for Computational Linguistics:
  EMNLP 2024, Al-Onaizan, Y., Bansal, M.,  Chen, Y.-N., (ur.). Miami, Florida,
  USA: Association for Computational Linguistics, Nov. 2024, pages 4138--4151,
  \url{https://aclanthology.org/2024.findings-emnlp.239/}
\BIBentrySTDinterwordspacing

\bibitem{zhuang2024vector}
\BIBentryALTinterwordspacing
Zhuang, Y., Singh, C., Liu, L., Shang, J.,  Gao, J., ``Vector-{ICL}: In-context
  learning with continuous vector representations'', in The Thirteenth
  International Conference on Learning Representations, 2025,
  \url{https://openreview.net/forum?id=xing7dDGh3}
\BIBentrySTDinterwordspacing

\bibitem{min-etal-2022-metaicl}
\BIBentryALTinterwordspacing
Min, S., Lewis, M., Zettlemoyer, L.,  Hajishirzi, H., ``{M}eta{ICL}: Learning
  to learn in context'', in Proceedings of the 2022 Conference of the North
  American Chapter of the Association for Computational Linguistics: Human
  Language Technologies, Carpuat, M., de~Marneffe, M.-C.,  Meza~Ruiz, I.~V.,
  (ur.). Seattle, United States: Association for Computational Linguistics,
  Jul. 2022, pages 2791--2809,
  \url{https://aclanthology.org/2022.naacl-main.201/}
\BIBentrySTDinterwordspacing

\bibitem{chen-etal-2022-meta}
\BIBentryALTinterwordspacing
Chen, Y., Zhong, R., Zha, S., Karypis, G.,  He, H., ``Meta-learning via
  language model in-context tuning'', in Proceedings of the 60th Annual Meeting
  of the Association for Computational Linguistics (Volume 1: Long Papers),
  Muresan, S., Nakov, P.,  Villavicencio, A., (ur.). Dublin, Ireland:
  Association for Computational Linguistics, May 2022, pages 719--730,
  \url{https://aclanthology.org/2022.acl-long.53/}
\BIBentrySTDinterwordspacing

\bibitem{chen-etal-2023-learning}
\BIBentryALTinterwordspacing
Chen, J., Lu, Y., Lin, H., Lou, J., Jia, W., Dai, D., Wu, H., Cao, B., Han, X.,
   Sun, L., ``Learning in-context learning for named entity recognition'', in
  Proceedings of the 61st Annual Meeting of the Association for Computational
  Linguistics (Volume 1: Long Papers), Rogers, A., Boyd-Graber, J.,  Okazaki,
  N., (ur.). Toronto, Canada: Association for Computational Linguistics, Jul.
  2023, pages 13\,661--13\,675,
  \url{https://aclanthology.org/2023.acl-long.764/}
\BIBentrySTDinterwordspacing

\bibitem{li2023task}
Li, C., Zhou, H., Glava{\v{s}}, G., Korhonen, A.,  Vuli{\'c}, I., ``On task
  performance and model calibration with supervised and self-ensembled
  in-context learning'', arXiv preprint arXiv:2312.13772, 2023.

\bibitem{razumovskaia2024analyzing}
Razumovskaia, E., Vuli{\'c}, I.,  Korhonen, A., ``Analyzing and adapting large
  language models for few-shot multilingual {NLU}: Are we there yet?'', arXiv
  preprint arXiv:2403.01929, 2024.

\bibitem{NIPS2017_d5e2c0ad}
\BIBentryALTinterwordspacing
Christiano, P.~F., Leike, J., Brown, T., Martic, M., Legg, S.,  Amodei, D.,
  ``Deep reinforcement learning from human preferences'', in Advances in Neural
  Information Processing Systems, Guyon, I., Luxburg, U.~V., Bengio, S.,
  Wallach, H., Fergus, R., Vishwanathan, S.,  Garnett, R., (ur.), Vol.~30.
  Curran Associates, Inc., 2017,
  \url{https://proceedings.neurips.cc/paper_files/paper/2017/file/d5e2c0adad503c91f91df240d0cd4e49-Paper.pdf}
\BIBentrySTDinterwordspacing

\bibitem{NEURIPS2023_a85b405e}
\BIBentryALTinterwordspacing
Rafailov, R., Sharma, A., Mitchell, E., Manning, C.~D., Ermon, S.,  Finn, C.,
  ``Direct preference optimization: Your language model is secretly a reward
  model'', in Advances in Neural Information Processing Systems, Oh, A.,
  Naumann, T., Globerson, A., Saenko, K., Hardt, M.,  Levine, S., (ur.),
  Vol.~36. Curran Associates, Inc., 2023, pages 53\,728--53\,741,
  \url{https://proceedings.neurips.cc/paper_files/paper/2023/file/a85b405ed65c6477a4fe8302b5e06ce7-Paper-Conference.pdf}
\BIBentrySTDinterwordspacing

\bibitem{vonwerra2022trl}
von Werra, L., Belkada, Y., Tunstall, L., Beeching, E., Thrush, T., Lambert,
  N., Huang, S., Rasul, K.,  Gallouédec, Q., ``{TRL}: Transformer
  reinforcement learning'', \url{https://github.com/huggingface/trl}, 2020.

\bibitem{razumovskaia-etal-2024-sqatin}
\BIBentryALTinterwordspacing
Razumovskaia, E., Glava{\v{s}}, G., Korhonen, A.,  Vuli{\'c}, I., ``{SQATIN}:
  Supervised instruction tuning meets question answering for improved dialogue
  {NLU}'', in Proceedings of the 2024 Conference of the North American Chapter
  of the Association for Computational Linguistics: Human Language Technologies
  (Volume 1: Long Papers), Duh, K., Gomez, H.,  Bethard, S., (ur.). Mexico
  City, Mexico: Association for Computational Linguistics, Jun. 2024, pages
  8195--8211,  \url{https://aclanthology.org/2024.naacl-long.453/}
\BIBentrySTDinterwordspacing

\bibitem{10.1162/tacl_a_00638}
\BIBentryALTinterwordspacing
Liu, N.~F., Lin, K., Hewitt, J., Paranjape, A., Bevilacqua, M., Petroni, F.,
  Liang, P., ``Lost in the middle: How language models use long contexts'',
  Transactions of the Association for Computational Linguistics, Vol.~12, 02
  2024, pages 157-173,  \url{https://doi.org/10.1162/tacl\_a\_00638}
\BIBentrySTDinterwordspacing

\bibitem{willard2023efficient}
Willard, B.~T.,  Louf, R., ``Efficient guided generation for large language
  models'', arXiv preprint arXiv:2307.09702, 2023.

\bibitem{kwon2023efficient}
Kwon, W., Li, Z., Zhuang, S., Sheng, Y., Zheng, L., Yu, C.~H., Gonzalez, J.~E.,
  Zhang, H.,  Stoica, I., ``Efficient memory management for large language
  model serving with paged attention'', in Proceedings of the ACM SIGOPS 29th
  Symposium on Operating Systems Principles, 2023.

\bibitem{you2021logme}
You, K., Liu, Y., Wang, J.,  Long, M., ``Logme: Practical assessment of
  pre-trained models for transfer learning'', in International Conference on
  Machine Learning. PMLR, 2021, pages 12\,133--12\,143.

\bibitem{8803726}
Bao, Y., Li, Y., Huang, S.-L., Zhang, L., Zheng, L., Zamir, A.,  Guibas, L.,
  ``An information-theoretic approach to transferability in task transfer
  learning'', in 2019 IEEE International Conference on Image Processing (ICIP),
  2019, pages 2309-2313.

\bibitem{zandonati2023fit}
\BIBentryALTinterwordspacing
Zandonati, B., Pol, A.~A., Pierini, M., Sirkin, O.,  Kopetz, T., ``{FIT}: A
  metric for model sensitivity'', in The Eleventh International Conference on
  Learning Representations, 2023,
  \url{https://openreview.net/forum?id=PDG4-Y3aboN}
\BIBentrySTDinterwordspacing

\bibitem{pmlr-v97-kornblith19a}
\BIBentryALTinterwordspacing
Kornblith, S., Norouzi, M., Lee, H.,  Hinton, G., ``Similarity of neural
  network representations revisited'', in Proceedings of the 36th International
  Conference on Machine Learning, ser. Proceedings of Machine Learning
  Research, Chaudhuri, K.,  Salakhutdinov, R., (ur.), Vol.~97. PMLR, 09--15 Jun
  2019, pages 3519--3529,
  \url{https://proceedings.mlr.press/v97/kornblith19a.html}
\BIBentrySTDinterwordspacing

\bibitem{rubin-etal-2022-learning}
\BIBentryALTinterwordspacing
Rubin, O., Herzig, J.,  Berant, J., ``Learning to retrieve prompts for
  in-context learning'', in Proceedings of the 2022 Conference of the North
  American Chapter of the Association for Computational Linguistics: Human
  Language Technologies, Carpuat, M., de~Marneffe, M.-C.,  Meza~Ruiz, I.~V.,
  (ur.). Seattle, United States: Association for Computational Linguistics,
  Jul. 2022, pages 2655--2671,
  \url{https://aclanthology.org/2022.naacl-main.191/}
\BIBentrySTDinterwordspacing

\bibitem{ye2023compositional}
Ye, J., Wu, Z., Feng, J., Yu, T.,  Kong, L., ``Compositional exemplars for
  in-context learning'', in International Conference on Machine Learning. PMLR,
  2023, pages 39\,818--39\,833.

\bibitem{10707886}
Imazato, K.,  Shimada, K., ``Automatic few-shot selection on in-context
  learning for aspect term extraction'', in 2024 16th IIAI International
  Congress on Advanced Applied Informatics (IIAI-AAI), 2024, pages 15-20.

\bibitem{10.5555/3692070.3693379}
Liu, S., Ye, H., Xing, L.,  Zou, J., ``In-context vectors: {Making} in context
  learning more effective and controllable through latent space steering'', in
  Proceedings of the 41st International Conference on Machine Learning, ser.
  ICML'24. JMLR.org, 2024.

\bibitem{jukic2024disentangling}
Juki{\'c}, J.,  {\v{S}}najder, J., ``Disentangling latent shifts of in-context
  learning through self-training'', arXiv preprint arXiv:2410.01508, 2024.

\end{thebibliography}

\listoffigures
\cleardoublepage %
\listoftables
\cleardoublepage %

\clearpage{}%
\renewcommand{\leftmark}{Biography}
\chapter*{Biography}
\addcontentsline{toc}{chapter}{Biography}
\setcounter{secnumdepth}{0}

David Dukić was born on October 2, 1997, in Osijek, Croatia. He completed his bachelor studies in computing in July 2019 at the Faculty of Electrical Engineering and Computing, University of Zagreb, where he also completed his master studies in computing in July 2021, specializing in computer science, earning the academic title of Master of Engineering in Computing. During his master studies, he received the Rector's Award for individual scientific work and won Croatia's largest student data science competition (LUMEN Data Science) two times.

In July 2021, he started working as an early stage researcher at the Department of Electronics, Microelectronics, Computer and Intelligent Systems at the Faculty of Electrical Engineering and Computing, University of Zagreb. Since October 2021, he has been employed as an assistant at the same department. Since the beginning of his master studies, he has been a part of the Text Analysis and Knowledge Engineering Laboratory (TakeLab). Since 2021, he has been the project leader of TakeLab Retriever, a platform for processing large amounts of text data from Croatian news outlets. In 2024, he did a three-month research visit to Julius-Maximilians University of Würzburg. His research interests are primarily in the field of natural language processing, specifically in transfer learning, representation learning, and author profiling.

\section*{List of Publications}

\subsection*{Journal Publications}

\begin{enumerate}
    \item Dukić D, Došilović FK, Pluščec D, Šnajder J. 2024. Closed-domain event extraction for hard news event monitoring: a systematic study. PeerJ Computer Science 10:e2355 https://doi.org/10.7717/peerj-cs.2355.
    \item Bilić, P., Dukić, D., Arambašić, L., Gjurković, M., Šnajder, J., \& Furman, I. (2023). Digital news media as a social resilience proxy: A computational political economy perspective. New Media \& Society, 0(0). https://doi.org/10.1177/14614448231214149.
    \item Dukić, D., \& Sović Kržić, A. (2022). Real-Time Facial Expression Recognition Using Deep Learning with Application in the Active Classroom Environment. Electronics, 11(8), 1240. https://doi.org/10.3390/electronics11081240.
\end{enumerate}

\subsection*{Conference Publications}

\begin{enumerate}
    \item Rep, I., Dukić, D., \& Šnajder, J. (2024, November). Are ELECTRA's Sentence Embeddings Beyond Repair? The Case of Semantic Textual Similarity. In Findings of the Association for Computational Linguistics: EMNLP 2024 (pp. 9159-9169).
    \item Dukić, D., \& Šnajder, J. (2024, August). Looking Right is Sometimes Right: Investigating the Capabilities of Decoder-only LLMs for Sequence Labeling. In Findings of the Association for Computational Linguistics ACL 2024 (pp. 14168-14181).
    \item Dukić, D., Gashteovski, K., Glavaš, G., \& Šnajder, J. (2024, March). Leveraging Open Information Extraction for More Robust Domain Transfer of Event Trigger Detection. In Findings of the Association for Computational Linguistics: EACL 2024 (pp. 1197-1213).
    \item Barić, A., Majer, L., Dukić, D., Grbeša-Zenzerović, M., \& Šnajder, J. (2023, May). Target Two Birds With One SToNe: Entity-Level Sentiment and Tone Analysis in Croatian News Headlines. In Proceedings of the 9th Workshop on Slavic Natural Language Processing 2023 (SlavicNLP 2023) (pp. 78-85).
    \item Dukić, D., \& Kržic, A. S. (2021). Detection of Hate Speech Spreaders with BERT. In CLEF (Working Notes) (pp. 1910-1919).
    \item Dukić, D., Keča, D., \& Stipić, D. (2020, October). Are you human? Detecting bots on Twitter Using BERT. In 2020 IEEE 7th International Conference on Data Science and Advanced Analytics (DSAA) (pp. 631-636). IEEE.
\end{enumerate}
\clearpage{}%
\clearpage{}%
\renewcommand{\leftmark}{Životopis}
\chapter*{Životopis}
\addcontentsline{toc}{chapter}{Životopis}

David Dukić rođen je 2.~listopada 1997.~godine u Osijeku, u Hrvatskoj. Prijediplomski sveučilišni studij računarstva završio je u srpnju 2019.~godine na Fakultetu elektrotehnike i računarstva Sveučilišta u Zagrebu, gdje je u srpnju 2021.~godine završio i diplomski sveučilišni studij računarstva, modul računarska znanost, stekavši akademski naziv magistar inženjer računarstva. Za vrijeme diplomskog studija dobio je Rektorovu nagradu za individualni znanstveni rad te je dvostruki pobjednik najvećeg studentskog natjecanja iz podatkovne znanosti u Hrvatskoj (LUMEN Data Science). 

U srpnju 2021.~godine zaposlio se na radno mjesto mladi istraživač na Zavodu za elektroniku, mikroelektroniku, računalne i inteligentne sustave Fakulteta elektrotehnike i računarstva Sveučilišta u Zagrebu. Od listopada 2021.~godine zaposlen je kao asistent na istom zavodu. Još od početka diplomskog studija, dio je Laboratorija za analizu teksta i inženjerstvo znanja (TakeLab). Od 2021.~godine voditelj je projekta TakeLab Retriever, platforme za obradu velikih količina tekstnih podataka na hrvatskim novinskim web portalima. 2024.~godine bio je u tromjesečnom znanstvenom posjetu Sveučilištu Julius-Maximilians u Würzburgu. Njegovi istraživački interesi su primarno u području obrade prirodnog jezika. Kokretno, bavi se prijenosnim učenjem, učenjem reprezentacija te profiliranjem autora.\clearpage{}%

\end{document}